\documentclass{article}

\usepackage{arxiv}

\usepackage[utf8]{inputenc} 
\usepackage[T1]{fontenc}    
\usepackage{hyperref}       
\usepackage{url}            
\usepackage{booktabs}       
\usepackage{amsfonts}       
\usepackage{nicefrac}       
\usepackage{microtype}      
\usepackage{graphicx}
\usepackage[square,semicolon]{natbib}
\usepackage{doi}
\usepackage{xcolor}

\definecolor{boxgrey}{HTML}{F3F3F3}

\newcommand{\hlbox}[2]{
  \begin{center}
    \fcolorbox{white}{boxgrey}{
      \parbox{0.9\columnwidth}{\noindent \textbf{#1}. \textit{#2}}
    }
  \end{center}
}

\title{User Modeling and User Profiling: \\ A Comprehensive Survey}

\date{February 21, 2024}	

\author{
    \href{https://orcid.org/0000-0002-5506-3020}
    {\includegraphics[scale=0.06]{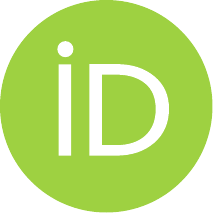}\hspace{1mm}Erasmo Purificato}\thanks{Corresponding author.} \\
	Otto von Guericke University Magdeburg \\
	Magdeburg, Germany \\
    \texttt{erasmo.purificato@acm.org} \\
\And
    \href{https://orcid.org/0000-0002-6053-3015}{\includegraphics[scale=0.06]{orcid.pdf}\hspace{1mm}Ludovico Boratto} \\
	University of Cagliari \\
	Cagliari, Italy \\
    \texttt{ludovico.boratto@acm.org} \\
\And
    \href{https://orcid.org/0000-0003-3621-4118}
    {\includegraphics[scale=0.06]{orcid.pdf}\hspace{1mm}Ernesto William De Luca} \\
	Otto von Guericke University Magdeburg \\
	Magdeburg, Germany \\
    \texttt{ernesto.deluca@ovgu.de} \\
}

\renewcommand{\shorttitle}{User Modeling and User Profiling: A Comprehensive Survey}

\hypersetup{
pdftitle={User Modeling and User Profiling: A Comprehensive Survey},
pdfsubject={User Modeling and User Profiling: A Comprehensive Survey},
pdfauthor={Erasmo Purificato, Ludovico Boratto, Ernesto W. De Luca},
pdfkeywords={User Modeling, User Profiling, User Behavior, User Preferences, User Interests, User Representation},
}

\begin{document}
\maketitle

\begin{abstract}
    The integration of artificial intelligence (AI) into daily life, particularly through information retrieval and recommender systems, has necessitated advanced user modeling and profiling techniques to deliver personalized experiences. These techniques aim to construct accurate user representations based on the rich amounts of data generated through interactions with these systems. This paper presents a comprehensive survey of the current state, evolution, and future directions of user modeling and profiling research. We provide a historical overview, tracing the development from early stereotype models to the latest deep learning techniques, and propose a novel taxonomy that encompasses all active topics in this research area, including recent trends. Our survey highlights the paradigm shifts towards more sophisticated user profiling methods, emphasizing implicit data collection, multi-behavior modeling, and the integration of graph data structures. We also address the critical need for privacy-preserving techniques and the push towards explainability and fairness in user modeling approaches. By examining the definitions of core terminology, we aim to clarify ambiguities and foster a clearer understanding of the field by proposing two novel encyclopedic definitions of the main terms. Furthermore, we explore the application of user modeling in various domains, such as fake news detection, cybersecurity, and personalized education. This survey serves as a comprehensive resource for researchers and practitioners, offering insights into the evolution of user modeling and profiling and guiding the development of more personalized, ethical, and effective AI systems.
\end{abstract}


\section{Introduction}\label{sec:intro}
In the pervasive era of artificial intelligence (AI) systems, the integration of such technologies into daily life is inevitable, whether embraced consciously or not.
Specifically, within the realm of widely adopted tools, information retrieval (IR) and recommender systems (RSs) proficiently supply pertinent information to users in accordance with their information requirements, personality traits, and contextual cues. Within a context where the interaction with such systems yields a voluminous amount of personal data on a daily basis, the imperative to discern individuals' interests, characteristics, and behaviors is fulfilled by \textit{user modeling} and \textit{profiling} techniques~\citep{eke_survey_2019}.
These techniques primarily aim to construct a reliable user representation (i.e., a \textit{user model} or \textit{user profile}) commencing from generated data~\citep{kanoje_user_2015}. 
User modeling and user profiling are pivotal in understanding user behavior and providing personalized experiences. Organizations can gain valuable insights into individual preferences and interests by analyzing user data, such as browsing history, purchase patterns, and social interactions. This, in turn, enables the delivery of tailored content, products, and services,  enhancing user satisfaction and engagement.

In this introduction, we will first illustrate the historical background of user modeling and user profiling research areas. Then, we will present an extensive overview of surveys and literature reviews on user modeling and profiling published over time, followed by a description of the novel contributions presented in this article. Finally, the outline of the paper concludes the introductory section.

\begin{figure}
    \centering
    \includegraphics[width=1\linewidth]{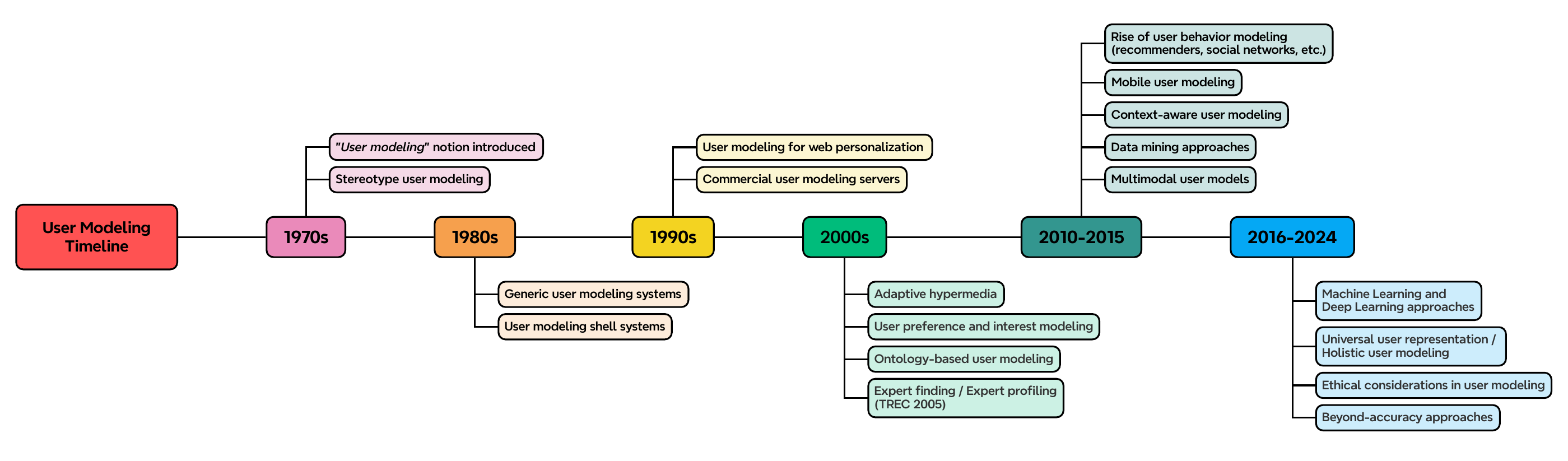}
    \caption{Timeline reporting the major events of the user modeling and profiling history.}
    \label{fig:timeline}
\end{figure}

\subsection{Historical overview}\label{subsec:historical_overview}
Throughout the history of scientific literature on personalization, user modeling, and user profiling fields have constantly witnessed significant advancements, as described by \citet{kobsa_generic_2001}, \citet{cena_real_2019}, and \citet{eke_survey_2019}, among others.

Traditionally, the initial steps in this research area can be dated back to the late 1970s when the notions of \textit{``user model''} and \textit{``user modeling''} were introduced by the contributions of Allen, Cohen, Perrault, and Rich (i.e., \citep{perrault_speech_1978,cohen_elements_1979,rich_user_1979}).
Their pioneering work set the stage for the following decade's research in this field. It led to the creation of numerous application systems that gathered different types of user information with varying adaptive capabilities (e.g., \citep{sleeman_umfe_1985,allgayer_xtra_1989,wahlster_user_1989}).
In these early models, there was no distinct separation between the components used for user modeling and those used for other functions.
Specifically, \citet{perrault_speech_1978} proposed a form of modeling that aimed at understanding the beliefs and goals of a user so that they could communicate. Interestingly, dialogues were exploited to understand these goals. The paper's approach to understanding a user's intentions using a model of the speaker's beliefs aligns with user modeling, where a system infers a user's needs and preferences from their interactions. This implies that user modeling systems should tailor their responses based on the user's profile to ensure effective communication.
\citet{cohen_elements_1979} highlighted the need for a nuanced representation of an agent's mental state, including beliefs, goals, and actions, to predict and influence behavior through communication. This approach was crucial for developing the idea of conversant computers capable of recognizing and generating appropriate speech acts in interaction with users.
\citet{rich_user_1979} introduced one of the earliest successful user modeling techniques by exploring the vision of using \textit{stereotypes} in computer systems to create models of individual users. Stereotypes, in this context, are predefined sets of characteristics that are commonly associated with certain types of users. The author discussed how these stereotypes could infer user preferences and behaviors, allowing the system to make personalized recommendations.
The popular system \textit{Grundy} was developed as a result of this research. It acts like a librarian, suggesting novels to users based on the stereotypes assigned to them. Based on user interactions, the system was designed to learn and adapt its stereotypes over time to improve its recommendations.
\textit{Stereotype user modeling} constituted the first attempt to differentiate a user from other users~\citep{schiaffino_intelligent_2009} and inspired several future works (e.g., \citep{ardissono_using_1995,krulwich_lifestyle_1997,zimmerman_exposing_2002}).

Since the late 1980s, it became apparent that the user modeling component needed to be reusable in creating user-adaptive systems. The initial move in this direction involved the creation of \textit{generic user modeling systems}.
Also defined as \textit{user modeling shell systems} by \citet{kobsa_modeling_1990}, a generic user modeling system (GUMS) operates as an independent component within a system during runtime, requiring developers to input it with application-specific user modeling knowledge~\citep{kobsa_generic_2001}.
Introduced by \citet{finin_gums_1986}, the primary GUMS was never utilized with an application system, but it established the framework for the fundamental functionality of future \textit{general} (i.e., \textit{application-independent}) user modeling systems, namely providing selected user modeling services that can be configured during development.
A few instances of these systems are: UMT~\citep{brajnik_shell_1994}, TAGUS~\citep{paiva_tagus_1994}, BGP-MS~\citep{kobsa_user_1994}, Doppelänger~\citep{orwant_heterogeneous_1994}, and the um toolkit~\citep{kay_toolkit_1995}.
User modeling shells were anticipated to support intricate assumptions and sophisticated reasoning about the user, particularly in the domains where these characteristics were identified. Furthermore, they were expected to be adaptable for use in a broad spectrum of other domains.

The significance of web personalization gained increasing recognition within the domain of electronic commerce at the end of the 1990s~\citep{hof_now_1998,allen_internet_1998}.
It involved tailoring product offerings, sales promotions, product news, and ad banners to individual users based on their navigation data, purchase history, and past interactions with the electronic merchant.
From a broader perspective, personalization facilitated the transition from anonymous mass marketing and sales to a more individualized \textit{one-to-one} marketing approach on the Internet~\citep{peppers_one_1993,peppers_enterprise_1997}. In this context, the involvement of user modeling was deemed crucial~\citep{fink_review_2000}.
Dozens of systems for web personalization, each boasting diverse capabilities, were developed and published in that period (e.g., \citep{caglayan_learn_1997,konstan_grouplens_1997,kay_personis_2002,brusilovsky_knowledgetree_2004,kobsa_ldap-based_2006}).

User profiling and user modeling research saw significant advancements in the 2000s, focusing on improving personalization and adaptability in various systems, including \textit{adaptive hypermedia}~\citep{brusilovsky_adaptive_1998,brusilovsky_adaptive_2001}, a research topic emerged at the intersection of hypermedia systems~\citep{kobsa_personalised_2001} and adaptive user interfaces~\citep{langley_user_1999}.
Unlike regular hypermedia, where all users were offered the same set of hyperlinks, adaptive hypermedia tailored what the user was offered based on a model of the user’s goals, preferences, and knowledge. This resulted in providing links or content that were most appropriate to the current user~\citep{de_bra_adaptive_1999}.
An adaptive hypermedia system tracks the user's browsing behavior and can change the link to a different web page or document that is more tailored to the user. It is used in various fields such as educational hypermedia, online information and help systems, and institutional information systems~\citep{brusilovsky_adaptive_2002}.
The growing interest in personalization in online systems, seen in the mid- and end-2000s, led researchers to start exploring ways to model user preferences to provide more tailored and relevant content~\citep{zhang_efficient_2007,berkovsky_mediation_2008}, such as personalized recommendations for e-commerce websites~\citep{braynov_personalization_2004,semeraro_user_2008} or TV shows~\citep{zimmerman_exposing_2002,zimmerman_tv_2004,martinez_whats_2009}.
In the same period, research also focused on creating user profiles that accurately captured interests based on observations of user behavior on the web (e.g., \citep{godoy_user_2005,castellano_similarity-based_2007,li_dynamic_2007,li_three-tier_2009,calegari_ontology-based_2010,gao_personalisation_2010,plumbaum_user_2011}).
The advent of the \textit{semantic web} prompted investigations into representing and modeling user preferences through \textit{ontologies}~\citep{middleton_ontological_2004,mehta_ontologically-enriched_2005,sieg_web_2007,deluca_multilingual_2010,sosnovsky_ontological_2010,elallioui_user_2012}.
These ontologies were employed to semantically organize and connect user profiles, thereby enhancing the comprehension of user preferences and relationships.

A significant turning point in user modeling and profiling research was characterized by the introduction of the \textit{expert finding} and \textit{expert profiling} tasks within the Enterprise Track at the Text REtrieval Conference (TREC) 2005~\citep{craswell_overview_2005}.
This was when the field started gaining considerable attention~\citep{eke_survey_2019}.
The TREC Enterprise Track's acknowledgment of the necessity to delve into and comprehend \textit{expertise retrieval} emphasized the significance of these tasks~\citep{balog_broad_2007}.
This, in turn, paved the way for a more profound investigation and comprehension of user profiling within the realm of expert finding and expert profiling~\citep{pavan_semantic_2015, liang_formal_2016}.
\textit{Expert finding}~\citep{balog_expertise_2012} is about identifying individuals who possess knowledge or expertise in a specific area within a given dataset. The aim is to find the best matches for a user’s specific request for expertise on a topic. For instance, a user might be looking for an expert who has knowledge in a particular field, and the task would involve searching through the dataset to find individuals who have demonstrated expertise in that area. The challenge lies in accurately determining a person’s expertise based on the available data and ranking these individuals based on their relevance to the given topic.
\textit{Expert profiling}~\citep{balog_expertise_2012}, on the other hand, is about creating a comprehensive profile of an individual’s expertise. It involves gathering all relevant information about a person’s skills and knowledge from their documents. This could include their educational background, areas of specialization, projects they have worked on, their contributions to their field, and so on. The aim is to provide a detailed overview of what the person knows and what they are capable of.

The 2010s marked a significant transformation in focus toward more sophisticated user profiling methods, with a growing emphasis on personalization in various digital services, particularly in RSs, where researchers developed advanced algorithms to analyze user behavior and preferences for improved content recommendations~\citep{abel_analyzing_2011,lakiotaki_multicriteria_2011,masthoff_group_2011,konstan_recommender_2012}.
Innovative approaches involved the use of personality-based user adaptation, where automated methods were developed for recognizing personality traits from user behaviors~\citep{gao_improving_2013,gou_knowme_2014,barnett_five-factor_2015} and conversations~\citep{mairesse_towards_2010,ivanov_recognition_2011,wei_beyond_2017}.
A notable surge in academic literature was due to \textit{semantic user modeling} techniques, which concern the creation of computational models to understand and predict user preferences, behaviors, and intentions based on semantic information derived from various data sources~\citep{aroyo_user_2010,bakalov_introspectiveviews_2010,plumbaum_user_2011,cena_semantic_2012,plumbaum_serum_2012,piao_exploring_2016}.
This period also saw the proliferation of mobile devices, which led to research on user modeling in mobile and location-based contexts~\citep{ouanaim_dynamic_2010,kuflik_challenges_2012}.
The integration of location data was explored to provide context-aware strategies and recommendations. \textit{Context-aware user modeling} gained traction as researchers aimed to understand how user preferences and behaviors change in different contexts. This included factors such as location, time, and device, leading to more adaptive and responsive systems~\citep{adomavicius_context-aware_2011,skillen_ontological_2012,verbert_context-aware_2012,said_movie_2013,codina_context-aware_2015}.
Concurrently, the rise of social media platforms spurred an increased interest in understanding and modeling user behavior within these online social environments~\citep{zafarani_connecting_2013,gou_knowme_2014,gilbert_rise_2023}.
Researchers began exploring ways to incorporate social network data into user models, applying social network analysis to understand the influence of social connections on user preferences and behaviors~\citep{vassileva_motivating_2012,zhong_comsoc_2012,zhou_state---art_2012}.
This included integrating social network data and user-generated content to create more accurate and context-aware user profiles~\citep{piao_inferring_2018,kaushal_methods_2019}.

During the same period, the ascent of \textit{big data} drove the investigation of advanced \textit{data mining} techniques for user modeling~\citep{romero_data_2013,doca_data-mining_2014,van_dam_online_2015}.
Large datasets have witnessed the application of \textit{machine learning} (ML) algorithms, encompassing clustering and classification methods, to unveil meaningful patterns and insights into user behavior~\citep{qi_research_2010,zghal_rebai_adaptive_2013,krishnan_dynamic_2017}.
As users began engaging with services across diverse platforms and devices, the emphasis turned towards formulating cross-platform user models~\citep{deng_personalized_2013,mercado_impact_2016,shin_cross-platform_2016,lin_cross_2019}. Researchers explored methods to construct cohesive user profiles that could encompass user preferences and behaviors across different digital environments. The proliferation of varied data types, including text, images, and audio, spurred efforts in developing \textit{multimodal user models}~\citep{saevanee_multi-modal_2012,saevanee_continuous_2015,farseev_harvesting_2015,guo_multi-modal_2018}. These models aimed to attain a more comprehensive understanding of user preferences and behaviors by integrating information from various modalities.
Subsequent advances in ML, particularly \textit{neural networks}, began to deeply influence this research area in the latter half of the 2010s.
From this period onwards, almost all relevant works in this user modeling and profiling focused on the use of deep neural networks for modeling complex user behaviors and enabling more accurate predictions and personalized experiences~\citep{tang_user_2015,zolna_user_2017,ge_personalizing_2018,ni_perceive_2018,an_neural_2019,chen_semi-supervised_2019,hu_graph_2020,wanda_deepprofile_2020,chen_catgcn_2021,fazil_deepsbd_2021,yan_relation-aware_2021,li_novel_2022,yan_interaction-aware_2022,zhao_co-learning_2022,xuan_knowledge_2023}.
\textit{Deep learning}~(DL) models were also adopted to automatically learn hierarchical representations of user preferences from raw data, leading to improvements in recommendation accuracy~\citep{gu_hierarchical_2020,wen_hierarchically_2021,li_hierarchical_2022,wei_hierarchical_2022,xue_factorial_2022}.

As the collection of user data became more pervasive, in recent years, there was a growing awareness of \textit{privacy} concerns, even though security and privacy approaches for user modeling and profiling have already been proposed in the past~\citep{kobsa_privacy_2003,schreck_security_2003}.
Researchers and practitioners began exploring ways to balance the need for personalized services with user privacy, leading to the development of privacy-preserving techniques~\citep{aghasian_scoring_2017,isaak_user_2018,raber_transferring_2022}, such as \textit{federated user modeling}~\citep{wu_hierarchical_2021,chu_mitigating_2022,luo_personalized_2022,liu_federated_2023}.
Furthermore, the increasing need for transparency and interpretability in AI systems has resulted in a widespread surge of scientific contributions focused on \textit{explainable AI} (XAI), including its application in user modeling~\citep{balog_transparent_2019,huang_explainable_2019,guesmi_explaining_2022}.
Efforts were directed towards enhancing the understandability and interpretability of ML models, enabling users to gain a clearer understanding of the construction of user models and the rationale behind recommendations and predictions~\citep{wang_designing_2019,hase_evaluating_2020,xian_ex3_2021,liu_triple_2023}.

In the past few years, ethical concerns related to user modeling, particularly focusing on \textit{bias} and \textit{fairness} issues, have gained significant prominence. Addressing these ethical considerations in user profiling has become a priority, with a commitment to fostering transparency, accountability, and fairness in algorithmic decision-making~\citep{balog_transparent_2019,dai_say_2021,shen_sar-net_2021,purificato_graph_2022,zheng_cbr_2022,abdelrazek_fairup_2023,zhang_fairlisa_2023}.
This transition towards a more human-centric design and inclusive approach to user modeling underscores the importance of recognizing diverse user demographics and preventing biases in the modeling process. Active exploration of methods is underway to ensure that user models maintain representativeness and fairness across different user groups~\citep{purificato_what_2023}.
Moreover, the evolving landscape includes a growing emphasis on fostering effective human-AI collaboration to enhance the ethical and inclusive dimensions of user modeling~\citep{celikok_modeling_2023}.

\subsection{Existing surveys on User Modeling and User Profiling}\label{subsec:existing_surveys}
Over the years, several surveys and literature reviews have been published on user modeling and user profiling.
Plenty of them targeted a specific application domain or feature. Only a few general, comprehensive, and exhaustive reviews and surveys have been presented.

\paragraph{Soft computing and personal information agents (2005)}
\citet{frias-martinez_modeling_2005} investigated the application of \textit{soft computing} methodologies, such as fuzzy logic, neural networks, genetic algorithms, fuzzy clustering, and neuro-fuzzy systems, in user modeling from 1999 to 2004. The research explores the utilization of these techniques either independently or in conjunction with other machine learning methods. Each technique is analyzed for its primary applications, constraints, and prospective avenues for user modeling. Additionally, the study offers guidance on selecting appropriate soft computing techniques based on the specific task implemented by the application.
\citet{godoy_user_2005} provided a survey of user profiling techniques within \textit{personal information agents}, exploring a range of algorithms and methods that are pivotal for personalization and recommender systems. It delves into collaborative filtering, text classification, user modeling, and the application of machine learning techniques such as Naïve Bayes classifiers, genetic algorithms, and artificial neural networks. The effectiveness of these methods is compared, particularly in their ability to model and adapt to user interests and preferences, with specific applications in news filtering and email classification. The paper also examines the use of feature selection, stemming, and various supervised learning approaches in text classification, discussing their pros and cons and comparing different methods like decision trees and rule-based classifiers.

\paragraph{Data mining approaches for adaptive hypermedia systems and their applications (2006 - 2007)}
\citet{frias-martinez_survey_2006} presented a comprehensive review of \textit{data mining} techniques applied to user modeling within the context of adaptive hypermedia systems. It addresses the challenges involved in selecting the most suitable data mining approach and offers guidelines for the design of user models that leverage these techniques. The paper underscores the importance of hybrid systems and the necessity for standardization in the field of user modeling. Each method is thoroughly described, detailing its fundamental algorithms, applications in user modeling, and inherent limitations. The paper provides practical examples and explores the potential of each technique in crafting user models for adaptive systems.
\citet{brusilovsky_user_2007} provided a comprehensive overview of adaptive hypermedia and user modeling, with a focus on \textit{personalized information access} and \textit{adaptive educational systems}. The study discusses various aspects of user modeling, including the representation of user knowledge, interests, goals, background, and individual traits, as well as the context of the user's work. It emphasizes the use of Bayesian Networks (BNs) for creating dynamic and qualitative student models that integrate expert knowledge with learning techniques, allowing for real-time improvements in adaptive web applications.

\paragraph{Online Social Networks and Mobile Social Networks (2009 - 2013)}
\citet{benevenuto_characterizing_2009} published a broad analysis of user behaviors on \textit{online social networks} (OSNs) by examining detailed clickstream data from a social network aggregator website. The study identifies two types of user activities: ``publicly visible'' actions and ``silent'' actions, such as browsing profiles or viewing photos, which are not immediately apparent to others. The authors highlight the dominance of browsing activities, which constitute 92\% of all user actions, and note that including silent interactions significantly increases the observed level of user engagement. The research extends to multiple social networks and finds considerable variation in user behaviors and session lengths across different platforms.
\citet{mezghani_user_2012} delved into the complexities of utilizing \textit{social annotations}, particularly \textit{tags}, for modeling user profiles within social networks. The paper underlines the importance of considering both static and dynamic elements when representing user profiles to ensure they remain up-to-date and reflective of users' evolving interests. In the context of recommendations, the paper discusses the use of vector representations and the FOAF (``Friend of a Friend'') ontology as methods for describing users and their interests.
Remaining in the sphere of social networks, \citet{jin_understanding_2013} provided an in-depth analysis of user behavior modeling methods within OSNs, examining aspects such as social connectivity, interaction patterns, traffic activity, and the dynamics of \textit{mobile social networks} (MSNs). The survey underscores the significance of clickstream data as a tool for understanding these behaviors and acknowledges the challenges associated with its use. A key focus of the paper is on the security and privacy concerns prevalent in OSNs. It reviews existing strategies to counter these threats and identifies areas where further research is needed to develop more effective solutions.

\paragraph{Human-Computer Interaction, multi-application environments and interoperability (2010 - 2011)}
\citet{biswas_brief_2010} presented a brief survey of different user modeling techniques used in \textit{Human-Computer Interaction} (HCI). After a discussion of the historical context of user modeling, the paper delves into various approaches, including the \textit{GOMS} family of models (which stands for ``Goals, Operators, Method and Selection''), cognitive architectures, grammar-based models, and application-specific models. Each modeling technique is examined for its strengths and weaknesses. Additionally, the authors underscore the significance of understanding the variety of user modeling paradigms available to system analysts.
\citet{viviani_survey_2010} analyzed the challenges and current research in user modeling within \textit{multi-application environments}. The study focuses on the concept of user modeling \textit{interoperability}, which is essential for cross-system personalization, and identifies two primary approaches to achieving interoperability: \textit{standardization-based} user modeling, to create common standards for user models that can be used across different systems; \textit{mediation-based} user modeling, to translate and adapt user models between systems that use different representations and formats by means of a mediator. The authors emphasize the importance of developing effective user modeling systems that can operate in a multi-application context. They acknowledge the difficulties in achieving interoperability and suggest that hybrid solutions, which combine elements of both standardization and mediation, may be necessary to overcome these challenges.
Another review of user model interoperability was presented by \citet{carmagnola_user_2011}. The authors explored the motivations behind interoperability and the evolution from centralized to decentralized user model architectures. The study also addresses the challenges associated with this research topic, including privacy concerns, access control, encryption, and the management of conflicting data models and values. Solutions like pseudonymous personalization, perturbation techniques, and scrutable user models are discussed as ways to handle these challenges.

\paragraph{Demographic recommenders, learning environments, and intrusion detection and prevention systems (2016)}
\citet{al-shamri_user_2016} delivered a thorough review of various techniques and approaches utilized in recommender systems, with a focus on demographic data, fuzzy profiling, and similarity measures. The core of the paper is an analysis of different user profiling methods for \textit{demographic recommender systems} (DRS). It highlights the significance of profiling in improving the performance of these systems, noting that certain methods can lead to substantial enhancements. The study's findings are intended to inform the design and implementation of DRS.
\citet{medina-medina_taxonomy_2016} presented a comprehensive examination of user modeling approaches within the context of \textit{learning environments}, particularly focusing on natural language systems, intelligent tutoring systems, web personalization, and adaptive educational hypermedia. The core contribution of the paper is the development of a taxonomy for classifying user models. This taxonomy is based on criteria related to the structure and management of user models, such as initialization, update methods, inference techniques, and the types of information they store.
\citet{peng_user_2016} analyzed the role of user profiling in \textit{intrusion detection and prevention systems} (IDPS). The review delves into the concept of behavioral profiling, which encompasses both system and user behaviors. It discusses the use of behavioral biometrics, such as keystroke dynamics and eye movement patterns, and psychometrics, which involve the analysis of a user's psychological attributes like intelligence, decision-making patterns, and preferences. These profiling methods aim to create a detailed user profile that can be used to detect unauthorized access or malicious activities.

\paragraph{Entity profiling and modeling (2017)}
\citet{barforoush_new_2017} introduced a classification framework designed to evaluate \textit{entity profiling} (EP) systems on the web, summarizing advancements in the field from 2000 to 2015. The survey provides a set of criteria to compare and classify EP systems, aiming to assist researchers in developing or selecting robust systems tailored to their specific needs. EP is defined as the process of collecting and compiling descriptive information about a specific entity, which could be a person, organization, or location. This information is gathered from various web sources and is used to create a comprehensive profile that represents the entity's characteristics, behaviors, or relationships. \textit{Entity modeling} (EM), on the other hand, refers to the creation of abstract representations of entities, often using formal models that define the types of entities and the possible relationships between them. EM is crucial for understanding and organizing the information collected during the profiling process.

\paragraph{Microblogging social networks and updated Human-Computer Interaction models (2018)}
\citet{piao_inferring_2018} presented a comprehensive review of user modeling strategies for inferring user interests in \textit{microblogging social networks}. It aims to provide an overview of the state-of-the-art techniques and methodologies used to construct and evaluate user interest profiles, which are essential for delivering personalized content and recommendations.
Similar to previous works, \citet{sajib_al_seraj_survey_2018} examined user modeling approaches within the field of HCI. The author provided an updated review of various models, such as the GOMS family, cognitive architectures, grammar-based models, and application-specific techniques. Among the discussed future challenges, the paper highlights the significance of user control in intelligent systems and raises concerns about privacy and security in relation to user models.

\paragraph{Wearable, ubiquitous and mobile computing technologies (2019)}
\citet{cena_real_2019} provided an in-depth analysis of the evolution of user modeling, particularly with the advent of \textit{wearable} and \textit{ubiquitous computing technologies}. The authors introduce the concept of \textit{Real World User Model} (RWUM), which integrates data from a person's everyday life, significantly expanding the scope and depth of traditional user models.
\citet{zhao_user_2019} presented a survey on user profiling through the analysis of \textit{smartphone application usage}, including the types of data that can be collected, the methods employed for profiling, and the potential applications and implications of this practice. The paper discusses how personal information such as demographic attributes, personality traits, psychological status, personal interests, and lifestyles can be inferred from smartphone app data. It also examines the use of other data sources, such as online social networks and call detail records (CDRs), for effective user profiling.

\paragraph{Representation learning and review-based user profiling (2020 - 2021)}
\citet{li_survey_2020} published a detailed analysis of the latest developments in user modeling, with a particular focus on \textit{representation learning} techniques. The study categorizes the methods into two main types: \textit{static} and \textit{sequential} representation learning. Static representation learning methods include matrix factorization and deep collaborative filtering, which are used to capture user preferences and item characteristics in a static context. These methods are foundational to many recommender systems and are crucial for understanding user behavior. Sequential representation learning methods, such as recurrent neural networks, are discussed as they pertain to capturing the dynamic nature of user preferences over time. These methods are particularly relevant for applications that require understanding the temporal aspects of user behavior.
\citet{dong_profiling_2021} presented a systematic mapping study on \textit{review-based user profiling} (RBUP), an area of research that utilizes user-generated reviews to create detailed user profiles. The study aims to provide an overview of the current state of the field, its evolution, publication trends, application areas, and co-authorship patterns. The research process for RBUP involves formulating research questions, conducting systematic literature searches using keywords, "snowball" searching, and applying inclusion and exclusion criteria to select relevant papers. The analysis of the selected papers includes bibliometric analysis, examining the time distribution of publications, identifying paper venues, and exploring co-author networks.

\paragraph{Recommender systems (2022)}
\citet{tenemaza_user_2022} highlighted the critical role of user modeling in the development of \textit{recommender systems}. The work delves into the various characteristics and structures that constitute user models, examining the diverse range of data sources and attributes that are employed to represent and understand users. It also points out a significant gap in the current research landscape: the lack of a \textit{generalized user model}, conveying that existing studies tend to focus on specific components of user modeling rather than adopting a holistic approach that integrates all relevant aspects.

\paragraph{User behavior modeling with Deep Learning, Long Short-Term Memory networks, and Large Language Models (2022 - 2023)}
In the same context of recommender systems, \citet{he_survey_2023} presented a survey of user behavior modeling. The paper discusses the deep network structures and techniques used to capture behavior dependencies, including the handling of long user behavior sequences and the incorporation of different behavior types. It also explores the challenges and advancements in user behavior modeling, such as the use of long-term behavior histories, multi-type behaviors, and side information.
\citet{sudhakar_web_2022} offered a thorough study on the application of \textit{Long Short-Term Memory} (LSTM) networks for generating and analyzing web user profiles. The primary objective is to create detailed and personalized user profiles that can enhance the customization of applications. The research encompasses several stages, including data collection, preprocessing, feature extraction, and the classification process, all of which are critical for the effective implementation of LSTM in user profiling. A significant portion of the paper is dedicated to comparing the LSTM-based method with existing user profiling techniques, concluding by underscoring the potential of LSTM networks in revolutionizing the field of user profiling and providing a pathway for more personalized and effective web experiences.
\citet{tan_user_2023} provided an overview of the use of \textit{Large Language Models} (LLMs) in the field of user modeling and recommender systems. The article highlights the potential of LLMs in understanding user behavior and generating personalized recommendations. It also delves into the challenges associated with LLM-based User Modeling, such as detecting suspicious behavior like fraud and misinformation and addressing privacy and security concerns. It emphasizes the need for comprehensive benchmarks, evaluation criteria, trustworthy user modeling, fairness, and efficient domain adaptation.

\paragraph{General comprehensive surveys (2009 - 2022)}
\citet{schiaffino_intelligent_2009} discussed the key components of user profiles adopted in that period for intelligent agents, adaptive systems, intelligent tutoring systems, recommender systems, e-commerce applications, and knowledge management systems. The paper also discusses several intelligent user profiling techniques, including stereotypes, Bayesian networks, association rules, and case-based reasoning.
\citet{cufoglu_user_2014} examined the advantages and disadvantages of existing user profiling methods and explored the potential of these methods for enhancing service personalization. A key part of the paper is the analysis of classification and clustering algorithms, which are pivotal in the construction of user profiles. The author conducts simulations using real-world user profile data to evaluate the performance of various algorithms in this context.
The survey presented by \citet{kanoje_user_2015} delves into the significance of user profiling and examines the progression, methodologies, and practical applications across various sectors, including academic literature recommendations, e-tourism, energy conservation, and employment matchmaking. The article outlines a range of user profiling techniques, such as implicit profiling, perceptual preference questionnaires, social profiling, weakly supervised extraction, ontological approaches, and the use of implicit feedback.
\citet{farid_user_2018} explored the essential information required for constructing diverse user models, the methods employed for collecting this information, the representation and maintenance of the user model, and, ultimately, the utilization of the user model to provide personalized services. A classification schema for user profiling research is proposed, and the paper also looks at how user profiling is applied in areas like personalized web search, recommender systems, adaptive learning, visualization, and personalized services in online social networks.
The most recent general and comprehensive survey on user modeling and user profiling has been provided by \citet{eke_survey_2019}. This research paper presents an extensive review that focuses on the methodologies, models, and processes involved in constructing accurate user profiles for service personalization. The authors aim to fill a gap in the existing literature by examining user profiling from the perspectives of data acquisition, feature extraction, modeling techniques, and performance evaluation. The paper discusses both static and dynamic user profiles, the process of user modeling, and the construction and updating of user profiles. It delves into the methods of user information collection, data preprocessing, and feature extraction, providing examples of studies that have utilized these techniques in user profiling. Various approaches to user profiling are explored, including collaborative and hybrid filtering, statistical models, neighborhood-based techniques, machine learning, user ontology, and filtering methods.
For the sake of completeness, in chronological order, it is worth mentioning two other articles, published, however, with slightly different purposes than to provide a complete overview of user modeling and profiling features.
\citet{abri_classification_2021} proposed a classification of the major dimensions of user models. The paper briefly explores the user profiling process, detailing the methods for collecting user information, such as browsing history, search queries, and user feedback. It examines learning techniques for constructing user profiles, including rule induction and predictive statistical models.
Recently, \citet{tchuente_user_2022} conducted a bibliometric analysis of the field of user modeling and user profiling within information systems, utilizing a dataset of 52\,027 publications. The study aims to map out the landscape of research in this area by identifying key authors, publication sources, institutions, and countries, as well as examining their collaborative efforts. The article also analyzes the main research topics and their respective subtopics, outlining the evolution of the field and identifying current trends. It further discusses potential future research directions and highlights existing gaps in the literature. This provides valuable insights for academics, researchers, and industry professionals interested in user modeling and profiling.

\subsection{Scope and novel contributions of the survey}\label{subsec:contributions}
The scientific literature has long devoted considerable attention to the research areas of user modeling and user profiling, as illustrated in the previous sections.
Across the years, tons of papers and numerous surveys on these topics have been published.
Our survey stems primarily from a number of key issues that emerged from the study of the literature: (1)~the absence of an article including a wide historical overview and general taxonomy; (2)~the ambiguous use of basic core terminology (i.e., ``user modeling'', ``user model'', ``user profiling'', and ``user profile''); and (3)~the lack of a comprehensive and up-to-date overview of this research area since 2019, to fill the gap due to the publication date of the most recent exhaustive survey (i.e., \citep{eke_survey_2019}).

Specifically, the novel contributions presented by our survey are listed and briefly described below, with the indication of the section including the related content:
\begin{itemize}
    \item We provide a complete outline of the large, long-standing, and ever-growing research fields of user modeling and user profiling (Section~\ref{sec:intro}), by offering a detailed historical overview (Section~\ref{subsec:historical_overview}) and an inspection of the literature reviews published across the decades (Section~\ref{subsec:existing_surveys}).
    After retracing the most significant milestones of the literature, starting from the \textit{stereotype user modeling}, introduced in 1979, to the recent studies on \textit{beyond-accuracy} perspectives, a formal taxonomy is proposed (Section~\ref{sec:taxonomy}), taking into account all the currently active topics in the research area, including the trends that emerged in the last few years.
    
    \item We examine the definitions associated with each key term in this research domain (i.e., ``user modeling'', ``user model'', ``user profiling'', and ``user profile''), aiming to eliminate ambiguity and confusion in their usage and proposing new, encyclopedic, and easily understandable definitions (Section~\ref{sec:terminology}).
    
    \item We present in-depth the paradigm shifts that have occurred in recent years, especially due to technological evolution, as well as the current research directions and novel trends in the field (Section~\ref{sec:paradigm_shift}).
    In particular, we illustrate and discuss the advances in the following research topics:
    \begin{itemize}
        \item \textit{Implicit vs. explicit user profiling} (Section~\ref{subsec:implicit_explicit}): in recent years, studies have adopted almost solely implicit (or hybrid) approaches for data collection. True explicit profiling (i.e., user data retrieved from surveys or questionnaires) is replaced by a sort of \textit{pseudo-explicit} profiling, where the explicit information is taken from public data already shared by the users for different purposes (e.g., social network accounts).
        
        \item \textit{User preferences and interests} (Section~\ref{subsec:preferences_interests}): the rise and daily use of digital platforms, such as e-commerce services and RSs of all kinds, have inevitably led to a steady increase in information about users' preferences and interests. This phenomenon is obviously reflected in the context of user modeling and profiling research.
        
        \item \textit{User behavior modeling} (Section~\ref{subsec:user_behavior_modeling}): the examination of user behaviors has evolved over time, bringing forth novel concepts, such as \textit{micro} and \textit{macro behavior modeling} (i.e., respectively, the immediate actions that a user takes reflecting short-term preferences, and large-scale actions that reflect a user's long-term commitment), \textit{multi-behavior modeling} (i.e., the practice of integrating diverse forms of user interactions with items, as opposed to depending on a single type), \textit{sequential behavior modeling} (i.e., considering the temporal sequences of behaviors influencing user interests), \textit{hierarchical user profiling} (i.e., a technique used in personalized recommender systems, particularly in e-commerce, to model users' real-time interests at different levels of granularity), and \textit{mobile user profiling} (i.e., the effort of discerning users' interests and behavioral patterns from their mobile activities).
        
        \item \textit{User representation} (Section~\ref{subsec:user_representation}): several works have underlined the absence of a generalized user model, suggesting that current research tends only to concentrate on particular facets of user modeling. This led to the advent and diffusion of the concepts of \textit{universal user representation} and \textit{holistic user modeling}.
        
        \item \textit{Evaluation} (Section~\ref{subsec:evaluation}): currently, there are two main lines for evaluating the performance of a standalone user profiling approach, i.e., assess the proposed model or method based on the effectiveness of a \textit{classification task} at predicting a user's personal characteristics, or generate \textit{simulated data} to aid in minimizing the volume of gathered user data while maintaining profiling efficiency and safeguarding the privacy and confidentiality of users' personal information.
        
        \item \textit{Graph data structures} (Section~\ref{subsec:graph_data_structures}): as in many other domains, also in user modeling and profiling, extensive attention has been given to exploring and adopting \textit{graph structures} (including \textit{knowledge graphs}), particularly in the context of online social media.
        
        \item \textit{Deep learning} (Section~\ref{subsec:deep_learning}): the advent of \textit{deep neural network-based models} has ushered in a transformative era across various domains, and user modeling and profiling have not been exceptions. We discuss contributions that employed different deep learning architectures and models, i.e., \textit{differentiable user models}, \textit{attention mechanism}, \textit{graph neural networks}, \textit{convolutional neural networks}, \textit{autoencoders}, \textit{recurrent neural networks}, \textit{long-short term memory}, and \textit{transformers}.
        
        \item \textit{Beyond-accuracy perspectives} (Section~\ref{subsec:beyond_accuracy_perspectives}): as for deep learning architectures, the adoption of \textit{``beyond-accuracy''} techniques signifies a pivotal transition in every domain, including user modeling and profiling. These approaches go beyond predictive precision, prioritizing values like \textit{privacy}, \textit{fairness}, and \textit{explainability}. In user modeling, it ensures accurate predictions while safeguarding user privacy, addressing biases, and promoting transparency. 

        \item \textit{Application domains} (Section~\ref{subsec:application_domains}): user modeling and profiling are finding novel applications in critical domains, addressing contemporary challenges. In the realm of \textit{fake news detection}, these techniques analyze user behavior and content interactions to identify and combat misinformation. In \textit{social networks}, even if not an unexplored domain, user profiling enhances community engagement and content recommender systems by understanding individual preferences. \textit{Cybersecurity} benefits from user modeling through anomaly detection, providing a proactive approach to identifying potential threats. In \textit{Massive Open Online Course} (MOOC) platforms, user modeling tailors educational content, ensuring a personalized learning experience for each student.
    \end{itemize}

    \item We discuss the findings of our survey, underlining the concepts that hold over time in the field, along with the emerging future research directions (Section~\ref{sec:discussion}).
\end{itemize}

\section{Analysis of the Terminology}\label{sec:terminology}
In this section, we examine the interpretations tied to each significant term related to the research fields studied, with the intention of removing any uncertainty or misinterpretation in their use.
Furthermore, we put forth a novel, general, and easily comprehensible set of definitions.
Particularly, after selecting specific contributions where precise characterizations were provided, we methodically offer a series of definitions for the terms \textit{``user profile''}, \textit{``user model''}, \textit{``user profiling''}, and \textit{``user modeling''}.
Our consideration encompasses both widely accepted and recently formulated definitions, with a focus on capturing the most contemporary meaning of each term in question.

\textbf{\textit{User profile}}, the first term to get a formal and precise definition together with \textit{user model}, has been described as:
\begin{itemize}
    \item a representation of the preferences of any individual user; roughly, it is a structured representation of the user's needs through which a retrieval system should, e.g., act upon one or more goals based on that profile and autonomously, pursuing the goals posed by the user~\citep{amato_user_1999}.

    \item the narration of a user’s behavior, interests, characteristics, and preferences obtained through interviews and questionnaires or dynamically with the aid of machine learning algorithms and data mining techniques~\citep{godoy_user_2005}.

    \item the application of ontology for the systematic representation of the user’s interest; it enables the conceptual representation of the knowledge that constitutes user preference and context~\citep{calegari_ontology-based_2010}.

    \item the information that offers insight into a user’s need and predicts his future intention; this information depends on similarities, trace handling, and prediction through ML~\citep{alaoui_building_2015}.

    \item the procedure for gathering information on the user’s interest; the system utilizes such information to tailor services and improve the user’s satisfaction~\citep{kanoje_user_2015}.

    \item a set of information that describes a user; it consists of demographic information, such as the user’s name, age, country, and level of education, which represents user preferences or interests in either a single or group of users~\citep{ouaftouh_user_2015}.

    \item a pattern that consists of user behavioral tendencies and preferences; the user profile knowledge acquired provides an idea of the user’s behavior knowledge and can predict his/her intentions~\citep{lashkari_survey_2019}.

    \item a collection of user interests, characteristics, behaviors, and preferences, and also a system for collecting, organizing, and guessing user information~\citep{sudhakar_web_2022}.
\end{itemize}

\textbf{\textit{User model}} has been defined as:
\begin{itemize}
    \item a representation of information about an individual user that is essential for an adaptive system to provide the adaptation effect~\citep{brusilovski_adaptive_2007}.
    
    \item user’s information for effective adaptation in an adaptive system; for instance, it facilitates prioritizing an adaptive selection of important/relevant items for users when searching for relevant data~\citep{gao_personalisation_2010}.

    \item a data structure that is used to capture specific characteristics about an individual user~\citep{piao_inferring_2018}.

    \item a representation of the knowledge and preferences of users; it is not a mandatory part of the software, but it helps to get the system to serve the user better. Any information stored about the user or usage pattern is not a user model unless it can be used to get some explicit assumption about the user~\citep{sajib_al_seraj_survey_2018}.

    \item a summary of the user’s interests, characteristics, behaviors, or preferences~\citep{tchuente_user_2022}.
\end{itemize}

Moving to the operations, the definitions provided for \textbf{\textit{user profiling}} depict it as:
\begin{itemize}
    \item the process of acquiring, extracting, and representing the features of users~\citep{zhou_state---art_2012}.

    \item the process to produce an accurate user information representation, usually stored in the user profile~\citep{de_campos_using_2014}.

    \item a means of determining the user’s interest data that is built upon the knowledge of the user and the accurate system’s retrieval of user satisfaction~\citep{kanoje_user_2015}.

    \item the task of inferring user personality traits from user-generated data~\citep{chen_semi-supervised_2019}.

    \item the process of inferring an individual’s interests, personality traits, or behaviors from generated data to create an efficient user representation, i.e., a user model, which is exploited by adaptive and personalized systems~\citep{eke_survey_2019}.

    \item the efforts of extracting a user’s interest and behavioral patterns from users’ activities~\citep{wang_adversarial_2019}.

    \item the process of automatically converting user information into a predefined and interpretable format that reflects the most important aspects of the user’s profile, which are useful for further decision-making in practical applications~\citep{vo_integrated_2021}.
\end{itemize}

\textbf{\textit{User modeling}} has been defined as:
\begin{itemize}
    \item the process of gathering information about a user’s interests, constructing, maintaining, and using user profiles~\citep{farid_user_2018}.

    \item the practice of capturing the latent interests of the user and deriving the adaptive representation for each user~\citep{ren_lifelong_2019}.

    \item the process of building up and modifying a conceptual understanding of the user. Its task is to learn a latent representation for each user, with the help of items, item features, and/or user-item response matrix, with applications to response prediction, recommendation, and others~\citep{li_survey_2020}.

    \item the process of capturing, recording, and managing user needs and interests by creating a user profile~\citep{abri_classification_2021}.

    \item the process of obtaining the user profile, which is a conceptual understanding of the user for personalized recommender systems. The key idea is to learn the representation for each user by leveraging the user’s interacted items or the features of the items, and the obtained representations are used for a wide range of applications such as response prediction and recommendation~\citep{kim_task_2023}.
\end{itemize}

Originally, another term was also used in research papers (not for long time), i.e., \textbf{\textit{user profile modeling}}, which has been described as:
\begin{itemize}
    \item the process that constitutes the methodology for building a user profile; it requires two steps to describe: ``what'' has to be represented, and ``how'' this information is effectively represented~\citep{amato_user_1999}.
\end{itemize}

Upon examining the provided definitions, it becomes apparent that in the literature, particularly when considering articles from the last decade, the terms ``user model'' and ``user profile'' (along with ``user modeling'', ``user profiling'', and ``user profile modeling'') exhibit conceptually overlapping descriptions. Consequently, we can assert definitively that these two terms can be utilized interchangeably, and within the scientific literature reviewed, they can be regarded as synonymous. Hereinafter, in our survey, we will use just one term directly, without explicitly specifying the other.
In this scenario, bringing together the peculiarities of the provided characterization, we propose two novel, comprehensive, encyclopedic definitions:
\begin{itemize}
    \item \textit{A \textbf{user model} (or \textbf{user profile}) is a structured representation of an individual user's preferences, needs, behaviors, and demographic details to personalize system interactions. It is derived from direct user feedback or inferred through machine learning and data mining techniques. It supports the predictions of future user intentions and the refinement of systems response to enhance user satisfaction. User models are often instrumental in optimizing the relevance and efficiency of adaptive systems, ensuring that user interactions are aligned with individual needs and preferences.}

    \item \textit{\textbf{User modeling} (or \textbf{user profiling}) is the process of acquiring, extracting, and representing user features and personal characteristics to build accurate user models (or user profiles). It encompasses inferring personality traits and behaviors from user-generated data. This dynamic practice includes automatically converting user information into interpretable formats, capturing latent interests, and learning conceptual user representations. Essentially, user modeling constitutes the methodology for building and modifying user models, determining ``what'' to represent and ``how'' to effectively represent this information for adaptive and personalized systems.}
\end{itemize}

\section{Paradigm Shifts and New Trends}\label{sec:paradigm_shift}
As explained in Section~\ref{subsec:contributions}, one of the driving factors behind the creation of this survey is the absence of a thorough and contemporary literature review on the subject of user modeling.
The most recent general survey in this research field is dated back to 2019 (i.e.,~\citep{eke_survey_2019}), and considers scientific contributions up until December 2018\footnote[1]{The paper by \citet{lashkari_survey_2019} cited within the survey is dated 2019, but it was officially published on September 22, 2018, as reported on the publisher website (\url{https://journals.riverpublishers.com/index.php/JCSANDM/article/view/5321}).}.
To be accurate, a few articles on the subject were published after that date, but they cannot be regarded as complete surveys. They offer a bibliometric analysis~\citep{tchuente_user_2022} and a concise, non-exhaustive categorization of user model dimensions~\citep{abri_classification_2021}.

In this section, we offer a thorough exploration of the recent paradigm shifts, particularly influenced by technological advancements, along with the current research orientations and emerging trends in the field.

\subsection{Implicit and explicit user profiling}\label{subsec:implicit_explicit}
Traditionally, early profiling methods focused solely on examining static attributes.
\textit{Explicit user profiling} (also referred to as \textit{static profiling} or \textit{factual profiling}) required direct input from the user, such as filling out a questionnaire, completing an online form, providing ratings, or explicitly stating preferences (e.g., \citep{raghu_dynamic_2001,poo_hybrid_2003,zigoris_bayesian_2006,brusilovski_adaptive_2007}).
Relying exclusively on explicit profiling quickly became problematic as users were reluctant to disclose their information due to privacy concerns, or found the form-filling process to be cumbersome and avoided it. Consequently, the accuracy of this type of profiling diminished over time~\citep{kanoje_user_2015,kasper_user_2017}.

In this context, although the two approaches were used simultaneously even in the past years, contemporary systems shifted their view by placing greater emphasis on \textit{implicit user profiling} (also known as \textit{behavioral} or \textit{adaptive profiling})~\citep{kanoje_user_2015,eke_survey_2019}, which entails the passive collection and analysis of dynamic user data, such as observing user behavior, interactions, and preferences without requiring direct user input (e.g., \citep{kulkarni_user_2019,el-ansari_improved_2020,qian_learning_2021,ma_one_2021,gu_self-supervised_2021,bedi_framework_2022,han_multi-aggregator_2022,yan_interaction-aware_2022}).

However, at the same time, the use of static data for user profiling has turned into gathering explicit information derived from public data that users have previously shared for different purposes, such as the creation of social network accounts (e.g.,~\citep{shu_understanding_2018,shu_role_2019}) or the usage of travel platforms (e.g.,~\citep{zhang_daily_2020}).
For this emerged category, we introduce the term \textbf{\textit{pseudo-explicit user profiling}}.

\subsection{User preferences and interests}\label{subsec:preferences_interests}
In the realm of user modeling, the evolution in exploring user preferences and interests has closely mirrored the advancements in implicit and explicit profiling.
Initially, user preferences and interests have been modeled using explicit and direct feedback (e.g.,~\citep{amatriain_rate_2009,lakiotaki_multicriteria_2011,kellner_i-know_2012,fu_why_2013}).

In recent times, the increasing prevalence and everyday utilization of digital platforms, including e-commerce services and various recommender systems, as well as the hesitancy of users in providing direct feedback, led to an abundance of user-generated data, such as social interactions and opinionated text content, and to a subsequent growing emphasis of research studies on capturing user interests and preferences hidden in users' historical behaviors (e.g.,~\citep{guo_user_2018,cami_user_2019,kulkarni_user_2019,logesh_efficient_2019,majumder_generating_2019,greco_emotional_2020,hassan_learning_2021,kostric_soliciting_2021,lu_standing_2021,olaleke_dynamic_2021,vo_integrated_2021,yilma_personalisation_2021,zhang_learning_2021,curmei_towards_2022,fan_modeling_2022,gomez_bruballa_learning_2022,ma_nest_2022,ariannezhad_personalized_2023,sguerra_ex2vec_2023,wang_improving_2023}).

Also worth mentioning, in the area of preference modeling, is the concomitant increase in specific studies pertaining to \textit{short-} and \textit{long-term preference modeling} (e.g.,~\citep{an_neural_2019,guo_attentive_2019,wu_long-_2019,yu_adaptive_2019,hu_graph_2020,sun_where_2020,fazelnia_variational_2022,liu_gnn-based_2023}).

\subsection{User behavioral modeling}\label{subsec:user_behavior_modeling}
The examination of user behaviors has significantly evolved to include a variety of sophisticated modeling techniques that provide a deeper understanding of user behavior in various contexts.

\paragraph{Micro and macro behavior modeling}
\textit{Micro behavior modeling} refers to the analysis of immediate actions taken by a user, which reflect their short-term preferences. These actions might include clicks, views, or interactions with specific components on a webpage or app.
\textit{Macro behavior modeling}, on the other hand, looks at large-scale actions that indicate a user's long-term commitment or patterns, such as purchase history or subscription renewals.
Examples of contributions on these topics are provided by \citet{gu_hierarchical_2020} and \citet{wen_hierarchically_2021}. 

\paragraph{Multi-behavior modeling}
 This approach integrates various forms of user interactions with items, rather than relying on a single type of interaction. \textit{Multi-behavior modeling} acknowledges that users often engage with platforms in multiple ways (such as clicking, favoriting, reviewing, and purchasing), and each of these behaviors can provide valuable insights into their preferences and intentions, and improve the performance of personalized recommender systems (e.g.,~\citep{jin_multi-behavior_2020,xia_knowledge-enhanced_2021,xia_graph_2021,zhang_leaving_2022,cheng_multi-behavior_2023,cho_dynamic_2023,xuan_knowledge_2023}).

\paragraph{Sequential behavior modeling}
\textit{Sequential behavior modeling} takes into account the order and timing of user actions, recognizing that the sequence of behaviors can influence a user's interests. This temporal aspect is crucial for services like online shopping, news feeds, and advertising, where the sequence of user interactions can reveal evolving preferences and help in predicting future actions (e.g., \citep{ren_lifelong_2019,yuan_parameter-efficient_2020,bian_contrastive_2021,cao_sampling_2022,chen_enhancing_2022}).

\paragraph{Hierarchical user profiling}
\textit{Hierarchical user profiling}, integral to personalized recommender systems in e-commerce, models users' real-time interests at various levels of granularity. This approach enables a comprehensive understanding of both immediate and long-term user preferences, crucial for accurate recommendations. Examining user behaviors hierarchically provides nuanced insights, particularly beneficial for predicting conversion rates in e-commerce platforms. This method enhances the precision of recommendations by delving into user preferences and behaviors, contributing to a tailored and effective user experience in the dynamic e-commerce landscape.
Examples of works presenting hierarchical user profiling approaches can be found in \citet{gu_hierarchical_2020}, \citet{wen_hierarchically_2021}, \citet{li_hierarchical_2022}, \citet{wei_hierarchical_2022}, and \citet{xue_factorial_2022}.

\paragraph{Mobile user profiling}
\textit{Mobile user profiling} involves discerning users' interests and behavioral patterns based on their activities on mobile devices. Given the ubiquity of smartphones and the wealth of data they generate, mobile user profiling is increasingly important for delivering personalized content and services that align with users' on-the-go lifestyles.
Contributions in this topic are provided by \citet{bhogi_user_2019}, \citet{wang_adversarial_2019,wang_incremental_2020,wang_reinforced_2021}, and \citet{zhao_co-learning_2022}.

\subsection{User representation}\label{subsec:user_representation}
Different studies have highlighted the scarcity of studies on generalized user model representation (e.g.,~\citep{tenemaza_user_2022}), pointing out that current research tends to focus on specific aspects of user modeling rather than a holistic approach. This observation has led to the development and widespread adoption of concepts such as \textit{universal user representation} and \textit{holistic user modeling}.
The shift towards these inclusive modeling approaches allows researchers and practitioners to better understand and cater to the complex and multifaceted nature of users in digital environments.

\paragraph{Universal user representation}
\textit{Universal user representation} is an emerging concept that aims to create a unified and generalized profile of a user by encapsulating a broad spectrum of user behaviors and preferences without bias towards any specific task. This approach is designed to be applicable across various domains and applications, providing a more complete understanding of users that can be leveraged for multiple purposes, such as user profiling and preference prediction (e.g., \citep{ni_perceive_2018,yuan_parameter-efficient_2020,gu_exploiting_2021,yuan_one_2021,kim_task_2023}).

\paragraph{Holistic user modeling}
\textit{Holistic user modeling} takes this idea further by integrating diverse and heterogeneous personal data sources, such as social networks and personal devices, to construct a comprehensive representation of the user. This model seeks to capture the full range of user characteristics and behaviors, providing a more nuanced and complete picture of the user that can be used to personalize experiences and interactions across different platforms and services (e.g., \citep{gong_when_2018,musto_framework_2018,musto_towards_queryable_2020,musto_myrror_2020,musto_towards_2020,musto_myrrorbot_2021,gong_jnet_2020}).

\subsection{Evaluation}\label{subsec:evaluation}
There are currently two primary strategies for evaluating the performance of standalone user profiling methods.

The first approach involves assessing the proposed model or method based on its effectiveness in a \textit{classification task}, which aims to predict a user's personal characteristics. This method often involves using machine learning techniques to classify user profiles as genuine or not (e.g., \citep{costanzo_towards_2019,chen_catgcn_2021,dai_say_2021,yan_relation-aware_2021}).

The second approach involves generating \textit{simulated data} to help reduce the amount of user data collected while maintaining the efficiency of profiling and protecting the privacy and confidentiality of users' personal information (e.g., \citep{zerhoudi_evaluating_2022,balog_user_2023}).

\subsection{Graph data structures}\label{subsec:graph_data_structures}
In the field of user modeling, much like in many other domains, there has been a significant focus on the exploration and implementation of \textit{graph structures}, including \textit{knowledge graphs}. This is particularly evident in the context of online social media.

\textit{Graph structures} are powerful data representations that capture relationships among data objects, making them ubiquitous in real-world applications. In the context of user modeling, graph structures can be used to represent and analyze user behavior, preferences, and interactions (e.g., \citep{guo_user_2018,wang_adversarial_2019,wanda_deepprofile_2020,wang_reinforced_2021,chen_global_2022,guan_personalized_2022,yang_modeling_2022,liu_triple_2023,yang_going_2023}).

\textit{Knowledge graphs}, a specific type of graph structure, have been recognized for their ability to effectively represent complex information, thereby gaining the attention of both academia and industry. They accumulate and convey knowledge of the real world, making them particularly useful in the context of social media, where they can be used to analyze critical information from people's activities and posts. For example, knowledge graphs can be used to recommend accurate content that interests users and to connect users with persons of interest.
Examples of contributions leveraging knowledge graphs for user modeling can be found in \citet{huang_explainable_2019}, \citet{wang_incremental_2020}, \citet{anelli_sparse_2021}, \citet{wang_enhancing_2021}, and \citet{xuan_knowledge_2023}.

\subsection{Deep learning}\label{subsec:deep_learning}
The advent of deep neural network-based models has sparked a revolutionary shift across various fields, including user modeling. These models have significantly contributed to advancements in this field, enabling more accurate and comprehensive user profiling and prediction of user behavior.
Different deep learning approaches and architectures have been employed in user modeling. In particular, \textit{differentiable user models} (e.g.,~\citep{hamalainen_differentiable_2023}), \textit{attention mechanism} (e.g.,~\citep{wang_user_2020,wang_attention-based_2022,fazil_deepsbd_2021,qi_trilateral_2021,qi_news_2022,chu_mitigating_2022}), \textit{graph neural networks} (e.g.,~\citep{chen_semi-supervised_2019,chen_catgcn_2021,dai_say_2021,wu_user-as-graph_2021,yan_relation-aware_2021,yan_interaction-aware_2022,han_multi-aggregator_2022,zhao_co-learning_2022,cheng_multi-behavior_2023}), \textit{convolutional neural networks} (e.g.,~\citep{wang_adversarial_2019,wanda_deepprofile_2020,fazil_deepsbd_2021,qi_news_2022}), \textit{autoencoders} (e.g.,~\citep{abu_sulayman_user_2019,wang_adversarial_2019,fazelnia_variational_2022,liu_federated_2023}), \textit{recurrent neural networks} (e.g.,~\citep{ge_personalizing_2018,ni_perceive_2018,yu_adaptive_2019,gu_hierarchical_2020,chu_mitigating_2022,li_novel_2022}), \textit{long-short term memory} (e.g.,~\citep{zolna_user_2017,ma_pstie_2020,fazil_deepsbd_2021,sahoo_multiple_2021,nkambou_learning_2023}), and \textit{transformers} (e.g.,~\citep{gu_partner_2021,kota_understanding_2021,zhu_contrastive_2021,avny_brosh_bruce_2022,wu_userbert_2022,zheng_perd_2022}).

\subsection{Beyond-accuracy perspectives}\label{subsec:beyond_accuracy_perspectives}
Similar to the transformation observed in deep learning, the incorporation of advanced techniques extending beyond mere accuracy marks a significant shift in various domains, with a particular impact on user modeling and profiling. These approaches transcend the conventional pursuit of predictive precision and instead prioritize fundamental values such as \textit{privacy}, \textit{fairness}, and \textit{explainability}.
In our domain, these methods ensure not only precise predictions but also the protection of user privacy (especially through \textit{federated learning} approaches, e.g.,~\citep{wu_hierarchical_2021,chu_mitigating_2022,luo_personalized_2022,liu_federated_2023,zhang_dual_2023}), the detection and mitigation of biases (e.g.,~\citep{dai_say_2021,shen_sar-net_2021,purificato_graph_2022,zheng_cbr_2022,abdelrazek_fairup_2023,purificato_what_2023,zhang_fairlisa_2023}), and the promotion of transparency and interpretability (e.g.,~\citep{balog_transparent_2019,huang_explainable_2019,hase_evaluating_2020,xian_ex3_2021,de_pauw_who_2022,guesmi_explaining_2022,minn_interpretable_2022,ding_interpretable_2023}).

\subsection{Application domains}\label{subsec:application_domains}
User modeling is being innovatively applied in various important research fields to tackle contemporary challenges.

In the context of \textit{fake news detection}, user modeling techniques are employed to scrutinize user behavior and their interactions with content. This analysis aids in the identification and mitigation of misinformation, thereby enhancing the credibility of the information ecosystem. Examples can be found in \citet{shu_understanding_2018,shu_role_2019}, \citet{monti_fake_2019}, \citet{sahoo_multiple_2021}, and \citet{allein_preventing_2023}. 

\textit{Social networks}, although their domain is not entirely unexplored for user modeling, greatly benefit from it by comprehending individual preferences for enriching community engagement and refining content recommender systems (e.g.,~\citep{kaushal_methods_2019,gilbert_rise_2023}), or detecting fake profiles and bots (e.g.,~\citep{wanda_deepprofile_2020,fazil_deepsbd_2021}).

In the field of \textit{cybersecurity}, user modeling contributes significantly through anomaly detection. It offers a proactive strategy to spot potential threats by identifying unusual patterns in user behavior, thereby bolstering security measures (e.g.,~\citep{lashkari_survey_2019,kwon_user_2021}).

Lastly, in the education sector, particularly on \textit{massive open online course} (MOOC) platforms, user modeling is used to customize educational content. It ensures that each student receives a learning experience tailored to their unique needs and learning style, thereby enhancing the effectiveness of online education. Examples are seen in \citet{sunar_modelling_2020}, \citet{sanchez-gordon_model_2021}, and \citet{jin_mooc_2023}.

\begin{figure}
    \centering
    \includegraphics[width=\linewidth]{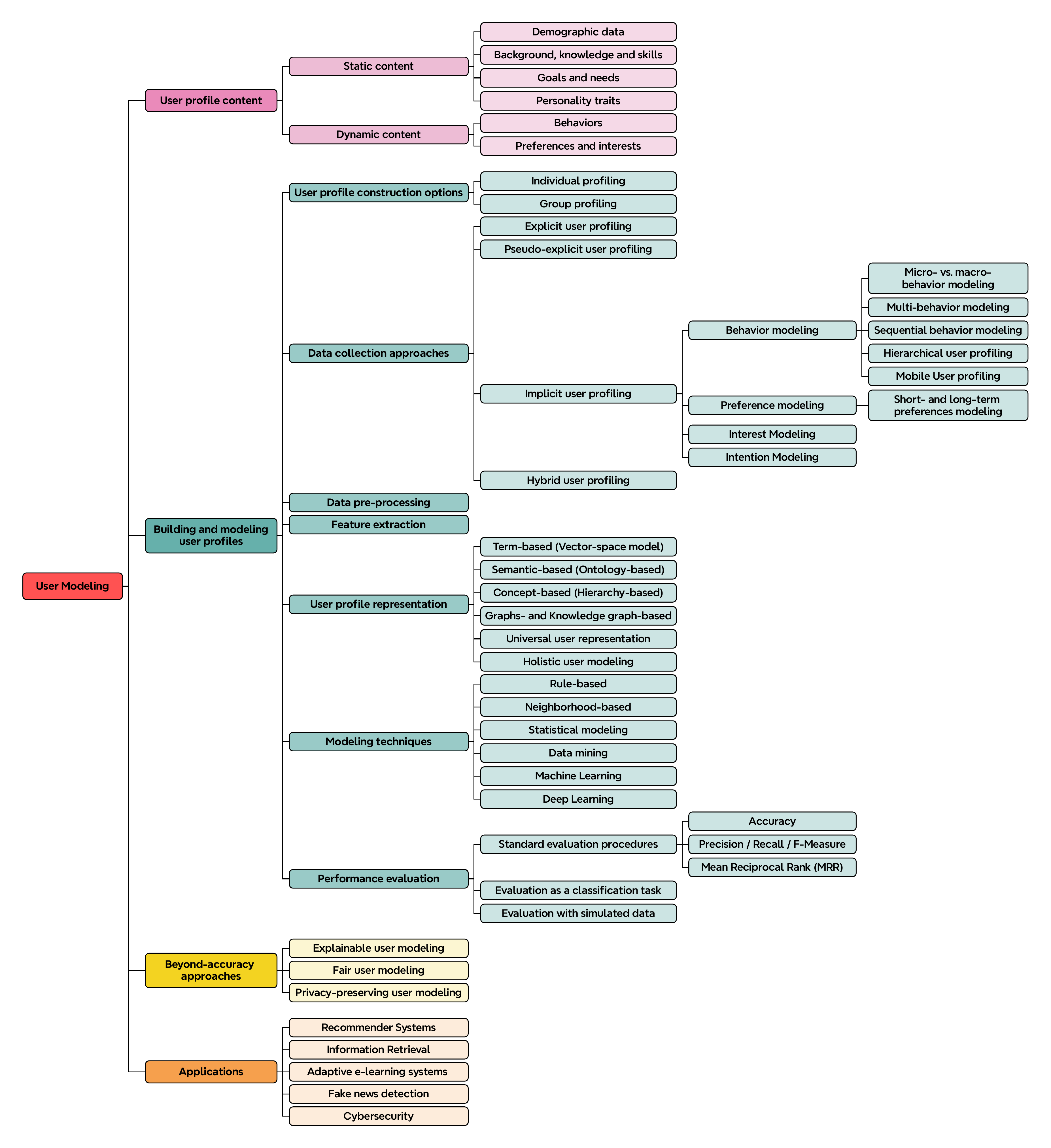}
    \caption{Taxonomy of the reviewed literature and trends for user modeling. The \textit{Modeling techniques} tree is detailed in Figure~\ref{fig:modeling_techniques}.}
    \label{fig:taxonomy}
\end{figure}

\section{Current Taxonomy}\label{sec:taxonomy}
The taxonomy presented in this section arises from the necessity to offer a renewed and contemporary reference in user modeling research.
This aims to benefit present researchers and serve as a valuable resource for individuals who will engage in the study of this subject in the upcoming times.
The categorization proposed in our survey takes inspiration from the similar contributions published over the years and described in Section~\ref{subsec:existing_surveys} (especially \citep{schiaffino_intelligent_2009,kanoje_user_2015,farid_user_2018,eke_survey_2019}), to make sure to maintain continuity in the literature.
Figure~\ref{fig:taxonomy} shows the formal taxonomy we put forth. It considers all presently active topics within the research area and incorporates the trends that have surfaced in recent years.

\subsection{User profile content}\label{subsec:profile_content}
User profiles typically hold details about an individual in a specific system or application, incorporating diverse content to shape and personalize the user experience. This content usually falls into two categories: \textit{static}, which provides foundational information, and \textit{dynamic}, offering real-time updates and interactivity. This distinction highlights how content behaves and its frequency of change, creating a complete user representation.

\subsubsection{Static content}\label{subsubsec:static_content}
\textit{Static content} refers to information that does not change frequently or automatically~\citep{poo_hybrid_2003}. It is consistent across user sessions and does not adapt in real-time to user interactions~\citep{eke_survey_2019}. Static content is typically set by the user during account creation or through profile settings and remains the same until the user decides to update it manually~\citep{schiaffino_intelligent_2009}.
Examples of static content in a user profile include: \textit{demographic data}, \textit{background}, \textit{knowledge}, \textit{skills}, \textit{goals}, \textit{needs}, and \textit{personality traits}.

\paragraph{Demographic data}
The demographic characteristics of a user include basic features such as name, country, gender, age, native language, education, family members, and more.
\citet{dong_user_2017} presented a study on user demographic and profile modeling within the context of large-scale mobile communication networks. It focuses on the prediction of demographic attributes such as gender and age by analyzing users' online behaviors, including browsing, gaming, and search activities, as well as their social decisions and communication patterns.
\citet{solomon_predict_2018} proposed an approach for predicting demographic attributes of cellphone users by analyzing their GPS data. The authors focus on extracting mobility profiles and location traces to infer characteristics such as age, gender, marital status, and academic affiliation.
\citet{zhang_daily_2020} introduced a method for simulating the daily electric vehicle (EV) charging load profiles by incorporating the demographics and social characteristics of vehicle users, such as gender, age, and education level. The study emphasizes that these user attributes significantly influence the magnitude and peak times of EV charging loads.
\citet{rozen_predicting_2021} presented an approach to predicting the demographic characteristics of online users by analyzing their comments on news articles.

\paragraph{Background, knowledge, and skills}
Background data, containing information about a user's educational, professional, and personal background, including their field of study, work experience, and cultural background, is crucial for creating wide user profiles.
Knowledge data refers to a user's expertise, domain knowledge, or specific knowledge related to a system or platform, and may include details about their educational qualifications, certifications, or areas of expertise.
Additionally, skills data pertains to a user's competencies and abilities, covering both technical and soft skills such as communication, problem-solving, and leadership abilities.
This diverse set of background, knowledge, and skills data is instrumental in personalizing user experiences, enabling adaptive interactions, and informing the development of tailored training and educational applications (e.g., \citep{brusilovsky_user_2007,schiaffino_intelligent_2009,guo_user_2018,li_survey_2020}).
To provide some specific and recent cases, \citet{minn_interpretable_2022} presented a method for student modeling called Interpretable Knowledge Tracing (IKT). This approach is designed to predict student performance by leveraging three meaningful latent features: individual skill mastery, ability profile, and problem difficulty. The motivation behind IKT is to address the shortcomings of existing knowledge tracing models, such as Bayesian Knowledge Tracing (BKT), which struggle with capturing learning transfer across different skills.
\citet{nkambou_learning_2023} developed an advanced learner model within an Intelligent Tutoring System (ITS) aimed at enhancing logical reasoning skills. This learner model is a composite of several components, and it is designed to provide a personalized learning experience by adapting to the individual needs of each student.

\paragraph{Goals and needs}
Recognizing a user's goals is a fundamental aspect of user modeling, focusing on understanding the specific objectives users intend to achieve within the application they are utilizing. On the other hand, comprehending information needs is centered around the essential requirement of obtaining pertinent information while, for instance, navigating the web.
Modeling users' goals and needs has been a practice applied in every period of the history of user profiling research.
\citet{horvitz_lumiere_1998} presented an in-depth exploration of Bayesian user modeling with the primary goal of inferring the goals and needs of users by analyzing their backgrounds, actions, and queries, thereby providing tailored assistance.
The paper by \citet{zhou_inferring_2003} discusses the development and evaluation of a causal model designed to infer user emotions during interactions with an educational game. The model utilizes nodes and links to deduce user goals based on personality traits and interaction patterns.
\citet{barua_modelling_2014} focused on the development and evaluation of a user model designed to assist individuals in setting and achieving personal long-term health-related goals. Drawing from psychological theories, the authors created a Goal Model representation and a user interface to facilitate goal setting and self-reflection.
Recently, a few studies explored user needs as a dynamic feature. For instance, \citet{ma_nest_2022} designed and implemented the NEST (Need Evolution Simulation Testbed) framework to model the \textit{dynamic} nature of user preferences and needs, particularly during extraordinary events that cause significant shifts in user behavior.

\paragraph{Personality traits}
In the context of user modeling, personality (or individual) traits are defined as stable and enduring internal characteristics that change only over a long period of time, and can be considered static. These traits, which include learning and cognitive styles, reflect people's characteristic patterns of thoughts, feelings, behaviors, and habits, implying consistency and stability. These features define a user as an individual and are used to create the user model. This allows for different users interacting with the same system to have unique experiences based on their individual traits.
The most widely used system of traits in psychology is the Five-Factor Model~\citep{mccrae_introduction_1992}, and even though many works in user modeling have referred to them (e.g.,~\citep{barnett_five-factor_2015}), contributions like that presented by \citet{cena_incorporating_2022} studied the impact of different specific traits (i.e. need for recognition, locus of control, mindset and self-efficacy) on the development of user models.
In general, predicting users' personality traits has seen applications in various contexts.
\citet{gao_improving_2013} presented a method for deducing personality traits from social media content, specifically targeting the Chinese language environment. By employing machine learning techniques, the researchers developed a model capable of predicting personality traits with a reasonable degree of accuracy. The findings indicate that there is a significant correlation between the content posted on social media and the personality dimensions of the users.
\citet{kim_personality_2013} explored the impact of personality traits on the effectiveness of e-learning systems. It presents the idea that understanding and integrating personality traits into the design of e-learning platforms can lead to more personalized and efficient learning experiences.
\citet{gou_knowme_2014} examined the idea of deriving personality traits from social media activity, specifically Twitter, and analyzed user attitudes towards sharing these traits in professional settings. The study aimed to determine the feasibility of automatically extracting personality traits from social media and to understand the factors influencing users' willingness to share these traits at work.
\citet{berkovsky_detecting_2019} introduced a framework aimed at user modeling by detecting personality traits through physiological responses to external stimuli. The study specifically explores the use of affective image and video stimuli alongside eye-tracking data to model user personality.
\citet{shen_user_2021} presented a user profiling system designed to infer gender and personality traits from non-linguistic audio data during conversations. The system leverages conversational features such as turn-taking and interruption patterns to identify user attributes effectively.

\subsubsection{Dynamic content}\label{subsubsec:dynamic_content}
\textit{Dynamic content} refers to information that frequently and automatically changes in response to user actions or system updates. This type of content is generated in real-time or on a regular basis, often driven by algorithms that analyze user behavior, preferences, and interactions with the system~\citep{farid_user_2018}. A dynamic profile is autogenerated by the system, leading to changes in user attributes and contents over time~\citep{eke_survey_2019}.
Despite the recent surge of studies also considering the dynamicity of static attributes, as mentioned in the previous section, common dynamic features in user profiles are: \textit{behaviors}, \textit{interests}, and \textit{preferences}.

\paragraph{Behaviors}
User \textit{behaviors} refer to the actions that users take while interacting with a platform. This can include things like the pages they visit, the links they click on, the amount of time they spend on certain tasks, and the frequency of their visits~\citep{farid_user_2018,eke_survey_2019}. By analyzing these behaviors, platforms can gain insights into what users are interested in and how they use the services~\citep{kanoje_user_2015}. Frequent actions on websites or social networks are exploited to infer a user's intentions (e.g.,~\citep{zhong_comsoc_2012,jiamthapthaksin_user_2017}).
Traditional methods rely on user behavior and labeled data for modeling users, such as \citet{covington_deep_2016}, who proposed a YouTubeNet model for video recommendation to model users by leveraging their watched videos and search tokens.
Other similar examples can be found in \citet{zhou_deep_2018} and \citet{ouyang_representation_2019}.
These techniques may not be optimal when labeled data are scarce. To address this limitation, \citet{wu_ptum_2020} introduced a pre-trained user model (PTUM) approach, inspired by the success of pre-trained language models in natural language processing tasks, to capture the relatedness between past and future behaviors.
However, for modern systems in general, the domains where user behaviors are most studied are \textit{e-commerce} (e.g.,~\citep{gu_hierarchical_2020,gu_self-supervised_2021,fan_modeling_2022,ma_caen_2022}) and general \textit{recommender systems} (e.g., \citep{jin_multi-behavior_2020,shen_sar-net_2021,zhao_ameir_2021,qian_fwseqblock_2022,xuan_knowledge_2023}).

\paragraph{Preferences and interests}
\textit{Preferences} and \textit{interests} of users are fundamental aspects of personalization~\citep{farid_user_2018}. They are related concepts, but they refer to different aspects of an individual's inclinations or likes.
Preference is more about choosing or favoring one option over another, often based on user personal tastes or judgments (e.g., \citep{jiamthapthaksin_user_2017,cami_user_2019,kostric_soliciting_2021,zhang_learning_2021,yang_modeling_2022,zheng_perd_2022}).
To provide some specific examples, \citet{kellner_i-know_2012} developed a tool to assess individual perceptual preferences in the context of information processing, knowledge acquisition, and learning. The primary goal of their contribution is to facilitate user profiling by identifying cognitive preferences, which can then be used to tailor information presentation to enhance user experience.
\citet{guo_multi-modal_2018} presented a personalized product search model that leverages both visual and textual information to capture user preferences.
\citet{majumder_generating_2019} introduced a method for personalized recipe generation, which aims to assist users who have incomplete knowledge about ingredients for specific dishes. The core of this study is the adoption of historical user preferences to create tailored recipes based on partial input specifications.
\citet{curmei_towards_2022} proposed a methodology that integrates psychological principles into the development of dynamic user preference models. The focus is on capturing the nuances of user behavior by formalizing three classic psychological effects.

On the other hand, interest is about the level of curiosity or engagement a user has in a particular subject or activity, which may or may not translate into making choices or decisions (examples of general approaches employing user interests, see \citet{piao_inferring_2018}, \citet{zhou_deep_2018}, and \citet{wang_enhancing_2021}).
\citet{wang_modeling_2022} presented an approach for enhancing user profiling in the context of personalized point-of-interest recommendations by leveraging a knowledge graph with temporal information to capture the dynamic and evolving preferences of users with respect to various locations.
\citet{xuan_knowledge_2023} introduced a framework designed to enhance user modeling in recommender systems by leveraging multi-behavior information. This approach acknowledges that different behaviors can reflect varying levels of user preference and intent, providing a richer and more nuanced understanding of user interests.

Dynamic profile that considers time may distinguish between \textit{short-term} and \textit{long-term interests}. In user modeling research, short-term interests refer to immediate preferences, subject to quick changes. In contrast, long-term interests are enduring preferences that remain relatively stable over an extended period, offering insights into consistent user likes and engagement areas.
Early approaches are dated to mid-2000s (e.g.,~\citep{diaz_adaptive_2004,li_dynamic_2007,liu_user_2007}), but numerous significant contributions have been published in recent years (e.g.,~\citep{an_neural_2019,guo_attentive_2019,wu_long-_2019,yu_adaptive_2019,xu_rethinking_2022,liu_gnn-based_2023}).
\citet{hu_graph_2020} implemented a graph-based model designed to capture both long-term and short-term user interests to enhance the accuracy of news recommendations. The model achieves this by constructing a heterogeneous graph that interlinks users, news articles, and topics, thereby encapsulating the complex relationships within the data.
\citet{sun_where_2020} introduced an approach for the next point of interest (POI) recommendation by incorporating both long-term and short-term user preferences, as well as considering the spatial and temporal contexts of user visits.
\citet{fazelnia_variational_2022} proposed a user modeling model designed for music recommender systems that is distinctive in its ability to capture both stable, long-term user interests and dynamic, short-term preferences. The model differentiates between ``slow features'', which represent historical interactions indicative of enduring interests, and ``fast features'', which capture recent interactions reflecting immediate preferences.

\vspace{0.1cm}
\hlbox{Take-home messages from user profile content}{
\begin{itemize}
    \item \textbf{Combination of static and dynamic content}: User profiles blend static content (like demographics and skills, which are stable) with dynamic content (such as behaviors and preferences, which evolve over time) to create a full picture of the user.
    \item \textbf{Detailed user data for personalization}: In-depth static data covering a user's background, skills, and goals are key for personalized experiences, especially in educational and adaptive systems.
    \item \textbf{Adaptive profiles with dynamic content}: Dynamic content, reflecting real-time user behaviors and preferences, allows systems to adapt and personalize experiences, catering to both immediate and long-term user interests.
\end{itemize}
}

\subsection{Building and modeling user profiles}\label{subsec:building_modeling_profiles}
The process of building and modeling a user profile, which aims to capture, record, and manage user needs, interests, and any kind of information described in Section~\ref{subsec:profile_content}, incorporates several elements, from profile construction and data collection techniques to their evaluation.
In this section, we discuss every component of the process.

\subsubsection{User profile construction options}\label{subsubsec:profile_construction}
A crucial facet of user modeling revolves around profile construction, which can be classified into \textit{individual} and \textit{group profile modeling} (or simply \textit{individual/group profiling}).
Individual profiling relies on information linked to a single user, such as enclosing aspects like demographic details, and it refers to the standard and almost the entirety of the studies published in the research area; 
On the other hand, group profiling involves a collective of users who share common interests, goals, or preferences. In group profiling, a partially complete profile is generated by overlapping information across users.

\paragraph{Group profile modeling}
\citet{masthoff_group_2004} investigated the complexities of group modeling in the context of adaptive television viewing, where the goal is to accommodate the preferences of multiple viewers simultaneously. The research delves into how individuals within a group make collective decisions about what to watch and how satisfied they are with those decisions. Experiments conducted as part of the study reveal that individuals consider factors like fairness, avoiding individual misery, and overall group approval when making selections.
Generally, the contributions on group modeling are strictly related to the topic of \textit{group recommender systems}.
\citet{masthoff_group_2011} provided an in-depth analysis of group recommender systems, focusing on the methodologies for aggregating individual preferences to cater to a collective audience. The article emphasizes the importance of understanding and modeling the affective states of users to enhance the recommendation process for groups.
\citet{boratto_modeling_2014} focused on the evaluation of various group modeling strategies within group recommenders. The paper underlines the importance of accurately combining individual preferences to generate group recommendations and highlights the influence of group size and the diversity of individual preferences on recommendation accuracy.
The same authors implemented the Automatic Recommendation Technologies (ART) framework~\citep{boratto_art_2015}, which employs clustering algorithms to automatically detect groups of users based on their preferences. The study identifies the Additive Utilitarian strategy as the best approach for modeling group preferences.
More lately, \citet{logesh_efficient_2019} introduced an intelligent travel recommender system designed to generate personalized recommendations for both individual and group users. Central to this system is a user profiling module that leverages a broad set of user-specific data. This data includes social, contextual, behavioral, geographical, and categorical information.
\citet{zhou_group_2021} presented a group-based personalized search model that leverages both search behavior and friend networks to enhance search results for users, particularly those with sparse historical data. Group profiles are created by identifying friend circles through the user's social connections and shared search behaviors.

\subsubsection{Data collection approaches}\label{subsubsec:data_collection_approaches}
A central phase of user modeling involves \textit{collecting data} about the user, which can be obtained either through the user's input or automatically collected by an intelligent agent. This information gathering about a specific user serves, first and foremost, as the starting point for any user modeling technique.
We categorize the data collection approaches for user profiling into: \textit{explicit}, \textit{pseudo-explicit}, \textit{implicit}, and \textit{hybrid}.

\paragraph{Explicit user profiling}\label{par:explicit_profiling}
\textit{Explicit profiling} approaches, also known as \textit{static} or \textit{factual profiling}, rely on manual techniques that require user intervention through the completion of online forms, the fulfillment of questionnaires, or the explicit provision of ratings and preferences. These techniques were originally adopted by the earlier contributions (examples are described in \citet{raghu_dynamic_2001}, \citet{poo_hybrid_2003}, \citet{brusilovski_adaptive_2007} and \citet{schiaffino_intelligent_2009}).
To offer specific instances, \citet{kern_shaping_2008} investigated user attitudes towards explicit profile generation for targeted advertising on public electronic displays. The study reveals a preference among users for explicit systems that grant them the ability to view and edit their profiles, despite the additional effort required. It also indicates that users are inclined to accept a less convenient system if it offers them greater control over their personal data.
\citet{pannu_explicit_2011} addressed the problem of low-quality web search results by proposing a system that enhances search outcomes through explicit profiling. The system employs the Vector Space Model (VSM) to construct user profiles that reflect individual preferences and interests. These profiles are then used to filter web documents by assessing the similarity between the user profile and the document content.

Another side of this area considers research relying only on analyzing user's predictable and static characteristics. As illustrated in Section~\ref{subsubsec:static_content}, static content includes users' demographic information, background, knowledge, skills, goals, needs, and personality traits.
Generally, a static profiling contribution analyzes and studies different static features at the same time (e.g.,~\citep{gao_improving_2013,kim_personality_2013,gou_knowme_2014}).
\citet{zhou_inferring_2003} developed a model for recognizing student emotions in real-time, which employs the OCC cognitive theory of emotions (named after its developers, Ortony, Clore, and Collins~\citep{ortony_cognitive_1988}) and is structured using Dynamic Decision Networks. The study conducted to refine the model looked at student interaction patterns and goals, finding correlations between personality traits and goals.
More recently, \citet{dong_user_2017} proposed a method called CoupledMFG (Multiple Label Factor Graph Model) for demographic prediction in coupled networks, which are characterized by users who have limited historical data within the network (i.e., a cold start problem). This model captures the correlations between users' communication behaviors and demographic profiles, as well as the interrelations among different demographic attributes. It utilizes both nonstructural attribute features and structural features to infer demographic information.
\citet{fernandez-lanvin_dimension_2018} provided a comprehensive examination of user attitudes and behaviors in the realm of e-commerce, delving into various facets such as user adaptation, success factors for online entrepreneurship, personality modeling, and demographic-specific interaction strategies. A key focus of the study is the impact of age and gender on web interaction performance. The findings revealed that both age and gender play a significant role in influencing user performance on e-commerce sites. Notably, the study observed consistent performance across different interaction tasks among individuals, suggesting the feasibility of developing a system that could automatically classify users based on their interaction behaviors.
\citet{guo_user_2018} presented an in-depth discussion on the evolution of user models and introduced the Cyber-I model, an innovative approach aimed at creating a comprehensive digital representation of an individual. The Cyber-I model is designed to encapsulate the personality traits, characteristics, and behaviors of users in the digital realm, striving to closely mirror the real individual.
\citet{rozen_predicting_2021} introduced ProfBERT, a model that leverages a BERT-based framework combined with attention networks to generate user representations from comments on news articles. This model utilizes both user-generated comments and the summaries of the articles they comment on to create detailed user profiles. The attributes predicted include gender, location type, and mobile device usage.

\paragraph{Pseudo-explicit user profiling}\label{par:pseudo_explicit_profiling}
The practice of utilizing static data for user profiling has evolved in recent years. This information is extracted from public data that users have willingly shared for various reasons, such as setting up social networking accounts or using travel platforms. We propose the term \textit{pseudo-explicit user profiling} for this newly emerged category.
To elaborate, pseudo-explicit user profiling refers to a method where explicit user information is gathered not directly from the user for the purpose of profiling, but indirectly from data that the user has shared publicly for other purposes. This could include data shared on social media platforms, travel booking sites, online forums, and more. This method offers a new approach to user profiling, leveraging the wealth of data users generate as they interact with various platforms on the internet. It is called “pseudo-explicit” because while the information is explicit and voluntarily provided by the users, it was not shared with the intention of being used for profiling.
The papers by \citet{shu_understanding_2018,shu_role_2019} focused on the role of user profiles in the context of fake news detection on social media, by exploiting explicit information extracted directly from metadata returned by querying social network APIs.
\citet{zhang_daily_2020} demonstrated that the daily EV charging load profiles vary with different demographic and social attributes by presenting a refined EV charging load simulation method considering people's demographics and social characteristics (e.g., gender, age, education level) retrieved from the US National Household Travel Survey.

\paragraph{Implicit user profiling}\label{par:implicit_profiling}
Exclusively depending on explicit profiling became a challenge as users hesitated to share their information due to privacy concerns or found the process of filling out forms to be unmanageable, leading to avoidance.
Modern systems have changed their perspective by giving more importance to \textit{implicit user profiling}, also known as \textit{behavioral}, \textit{dynamic}, or \textit{adaptive profiling}.
This involves the passive collection and analysis of dynamic user data, including observing user behavior, interactions, and preferences, without necessitating direct input from the user (e.g.,~\citep{kasper_user_2017,pujahari_item_2022,qi_fum_2022,li_stan_2023}).
Implicit profiling approaches involve several aspects, including \textit{behavior}, \textit{preference}, \textit{interest}, and \textit{intention modeling}.

\textbf{\textit{Behavior modeling}} is about observing and analyzing users' actions, interactions, and patterns while they engage with a system or application. This could include tracking the pages they visit, the time spent on different tasks, and the sequences of actions performed (e.g.,~\citep{zhong_comsoc_2012,covington_deep_2016,zhou_deep_2018,wu_ptum_2020,morshed_fahid_progression_2021,zhao_ameir_2021,agarwal_modeling_2022,fan_modeling_2022,qian_fwseqblock_2022,he_survey_2023}).
\citet{wang_calendar_2020} presented a deep learning architecture to predict user attributes by learning from spatiotemporal behavior.
\citet{gu_self-supervised_2021} introduced a framework designed to enhance ranking models in e-commerce by leveraging user behaviors and utilizing a two-stage approach: pre-training on users' spontaneous behaviors and fine-tuning on implicit feedback across multiple e-commerce scenarios.
\citet{zhu_contrastive_2021} designed a system to enhance document ranking by optimizing the representation of user behavior sequences through contrastive learning. The system focuses on capturing the nuances of user behavior sequences to improve document ranking.
Several methods introduced approaches for constructing user profiles which rely on the implicit collection of user browsing data to generate precise profiles, whose accuracy can be improved by integrating various sources of browsing data and by differentiating between significant and insignificant concepts (e.g.,~\citep{kulkarni_user_2019,el-ansari_improved_2020,gu_self-supervised_2021,bedi_framework_2022,han_multi-aggregator_2022,yan_interaction-aware_2022}).
Different contributions exploited the user's dialogue history from conversational platforms to build the profiles (e.g.,~\citep{ma_one_2021,qian_learning_2021,li_user-centric_2022}).

The analysis of behavior profiling has advanced considerably in the last few years, incorporating a range of refined modeling techniques that offer a more profound insight into user behavior across diverse contexts:
\begin{itemize}
    \item \textit{\textbf{Micro}} and \textit{\textbf{macro behavior modeling}} refers to the study of different levels of user interactions and activities within an online platform, particularly in the context of e-commerce and recommender systems, reflecting, respectively, short-term and long-term user preferences or interests.
    \citet{gu_hierarchical_2020} developed a deep learning framework to capture user actions and dynamically model their interests at different temporal levels.
    \citet{wen_hierarchically_2021} introduced a deep neural recommendation model designed to address two critical challenges in the field: sample selection bias and data sparsity. They achieved this by incorporating both micro and macro user behaviors into a unified framework, which allows for a more comprehensive understanding of user interactions.

    \item \textit{\textbf{Multi-behavior modeling}} integrates diverse user interactions with items, moving beyond dependence on a single type of interaction (e.g.,~\citep{jin_multi-behavior_2020,xia_knowledge-enhanced_2021,xia_graph_2021,zhang_leaving_2022,cheng_multi-behavior_2023}).
    To cite specific recent cases, \citet{cho_dynamic_2023} introduced two user modeling approaches, DyMuS and DyMuS+, designed to address the inherent challenges in multi-behavior data, such as data imbalance, heterogeneity, and the need for personalized recommendations.
    \citet{xuan_knowledge_2023} proposed the Knowledge Enhancement Multi-Behavior Contrastive Learning Recommendation (KMCLR) framework, aiming to augment user modeling in recommender systems through the utilization of multi-behavior information. This strategy recognizes that various behaviors can indicate different degrees of user preference and intent, thereby offering a more comprehensive and nuanced comprehension of user interests.

    \item \textit{\textbf{Sequential behavior modeling}} considers the order and timing of user actions, acknowledging that the sequence of behaviors can impact a user's interests. This temporal aspect holds significance in many application domains, as the sequence of user interactions can unveil evolving preferences and contribute to predicting future actions (e.g.,~\citep{ren_lifelong_2019,yuan_parameter-efficient_2020,cao_sampling_2022}).
    To provide explicit samples, \citet{bian_contrastive_2021} presented a Contrastive Curriculum Learning (CCL) framework designed to enhance the modeling of sequential user behaviors and produce more effective user behavior representations.
    \citet{chen_enhancing_2022} introduced the Auto-Session-Encoder (ASE), a novel model designed to improve the modeling of user behavior sequences in session search. The conducted experimental studies revealed that the proposed model benefits more from predicting future sequences and clicked documents rather than recovering historical ones.

    \item \textit{\textbf{Hierarchical user profiling}} is relevant to personalized e-commerce recommendations and models real-time interests at varying levels. Strictly related to the concepts of \textit{micro} and \textit{macro behavior modeling}, this yields nuanced insights for accurate predictions, enhancing precision and tailoring user experiences in the dynamic e-commerce landscape.
    \citet{gu_hierarchical_2020} engineered a Hierarchical User Profiling (HUP) framework to capture and model users' real-time interests at varying levels of granularity, acknowledging the hierarchical structure inherent in product categories and user interactions.
    \citet{wen_hierarchically_2021} introduced a deep neural recommendation model, designed to enhance conversion rate prediction by hierarchically modeling both micro and macro behaviors. By considering both levels of behavior, the model can better understand and anticipate user actions, leading to more effective recommendations and higher conversion rates.
    In the same research territory, \citet{li_hierarchical_2022} designed a framework for enhancing cross-domain click-through rate prediction by incorporating a hierarchical user behavior modeling approach that leverages an element-wise behavior transfer layer and a user representation layer.
    \citet{wei_hierarchical_2022} and \citet{xue_factorial_2022} employed Graph Neural Networks to hierarchically analyze multi-level user intents and item representations to improve the accuracy of recommendations.

    \item \textit{\textbf{Mobile user profiling}} identifies user interests and behavior patterns from their mobile device activities. With smartphones being widespread and generating substantial data, this profiling is vital for delivering personalized content and services that suit users' dynamic lifestyles.
    \citet{bhogi_user_2019} presented an approach to user profiling for mobile phone users, particularly focusing on the challenge of creating accurate profiles without access to ground truth data. The core of the proposed approach is the identification of seed features that are indicative of certain user attributes. These features are derived from user behavior and are used to train a positive unlabeled learning model.
    \citet{wang_adversarial_2019,wang_incremental_2020,wang_reinforced_2021} are particularly active in this area. They proposed several innovative deep learning approaches for studying the unique characteristics and preferences of mobile users.
    \citet{zhao_co-learning_2022} introduced a graph-based model for improving mobile user profiling by leveraging app text data. Traditional methods struggle with the sparse semantics and limited context of this kind of data, but the proposed model addresses these issues by constructing two heterogeneous graphs and applying a selective-scale attention mechanism to better capture semantic information.
\end{itemize}

\textbf{\textit{Preference modeling}} involves deducing user preferences based on their past interactions and choices. By analyzing the historical data of user preferences, systems can make predictions about what a user might prefer in the future. This is one of the historical areas exploited in user modeling research (e.g.,~\citep{amatriain_rate_2009,lakiotaki_multicriteria_2011,fu_why_2013}), and studies are currently applied in various contexts, such as recommendations, personalized user interfaces, or targeted advertising.
To illustrate with particular examples, \citet{zhao_preference_2019} focused on modeling user-system interactions within recommender systems by implementing a two-stage framework and a specific neural model. The model is specifically designed to better capture user behavior and preferences by considering the temporal dynamics of user interactions.
\citet{wu_user_2020} developed a model to capture a more holistic view of user preferences by not only considering which articles are clicked but also how users interact with them in terms of reading time. The experimental results demonstrated that incorporating reading satisfaction into user modeling significantly enhances the quality of news recommendations.
\citet{kostric_soliciting_2021} presented a method for improving preference elicitation in conversational recommender systems. The core idea is to generate implicit questions that can be used to understand user preferences more effectively.
\citet{gomez_bruballa_learning_2022} introduced a recommender system specifically designed for an image marketplace, with the primary goal of learning and catering to users' preferred visual styles.
\citet{yang_modeling_2022} proposed a user preference modeling approach in the context of online recruitment, specifically addressing the challenge of person-job fit.

\textbf{\textit{Interest modeling}} focuses instead on identifying and understanding the topics, subjects, or content that capture a user's attention. This is often inferred from the content they engage with, the keywords they search for, or the types of products they explore. By building a model of user interests, systems can provide more relevant and tailored recommendations, content suggestions, or targeted information.
\citet{cami_user_2019} introduced a Bayesian non-parametric approach to capture and adapt to the evolving preferences and interests of users.
\citet{yilma_personalisation_2021} developed a recommendation and guidance method that is sensitive to the interests and needs of multiple stakeholders, including visitors, curators, and those responsible for crowd management. The novelty of the proposed method lies in its ability to balance user interests with other factors, such as the popularity of Points of Interest (POIs) and the objectives set by the curators.

Other relevant contributions for preference and interest modeling can be found in \citet{jiamthapthaksin_user_2017}, \citet{logesh_efficient_2019}, \citet{majumder_generating_2019}, \citet{olaleke_dynamic_2021}, \citet{zhang_learning_2021}, \citet{curmei_towards_2022}, \citet{fan_modeling_2022}, and \citet{zheng_perd_2022}.

Noteworthy to mention is the simultaneous focused research on both \textit{\textbf{short-}} and \textit{\textbf{long-term preference}} (or \textit{interest}) \textit{\textbf{modeling}} (e.g.,~\citep{an_neural_2019,hu_graph_2020,sun_where_2020,zhou_encoding_2020,fazelnia_variational_2022,liu_gnn-based_2023}).
\citet{guo_attentive_2019} introduced a neural network model named ALSTP (Attentive Long- and Short-Term Preference) designed for personalized product search. The model aims to enhance search accuracy by incorporating both the long-term and short-term preferences of users, along with the current search query.
\citet{wu_long-_2019} developed a model to recommend the next point of interest (POI) to users. The model is designed to incorporate both the long-term preferences and the short-term behaviors of users, as well as contextual factors like POI categories and check-in times.
\citet{yu_adaptive_2019} implemented a recommender that adeptly captures both short-term and long-term user preferences. The approach is tailored to address the dynamic nature of user behavior, which is characterized by changing time intervals and latent intents.
\citet{xu_rethinking_2022} presented a recommendation engine designed by Pinterest to address the dual temporal aspects of user behavior: long-term interests and short-term intentions. The engine incorporates user embeddings that are specifically optimized to predict long-term future actions, as well as sequences of real-time actions to capture immediate user intent.

\textbf{\textit{Intention modeling}} covers an important part of the behavior modeling research. It involves predicting or understanding the user's goals, objectives, or intentions based on their behavior and interactions. By analyzing the sequence of actions and choices made by a user, systems can infer what the user is likely to pursue or achieve. This aspect is valuable in providing anticipatory and proactive support, aligning system responses with the user's intended outcomes. Intention modeling contributes to a more dynamic and responsive user experience, as the system can adapt to user goals in real-time, offering relevant assistance or suggestions.
\citet{lugo_modeling_2021} presented an approach to user search task modeling, whose key aspect is its focus on user intent modeling, which is achieved by analyzing clicked URLs from a large-scale query-clicked document collection.
\citet{deng_improving_2022} introduced a personalized search model that leverages a dual-feedback network to enhance the understanding of user search intentions. The model is designed to improve personalized search by incorporating multi-granular user feedback, both positive and negative, to accurately model the user's current search intention.
\citet{li_intent-aware_2023} proposed an approach that defines user intent as a probability distribution of item categories and behaviors. This allows the system to understand the user's multiple objectives and preferences more accurately. The model utilizes user intents as guidance for fusing user preferences across different behavior objectives, rather than learning user preference for each intent separately.

\paragraph{Hybrid user profiling}\label{par:hybrid_profiling}
This user profiling approach, which combines both implicit and explicit user profiling methods, integrates the benefits of both strategies. It considers static characteristics and retrieves behavioral information about a user, enhancing profiling efficiency.
Examples can be found in \citet{poo_hybrid_2003}, \cite{stanescu_hybrid_2013}, \citet{luo_hybrid_2014}, and \citet{liu_modeling_2015}.
To supply detailed instances, \citet{du_folksonomy-based_2016} addressed the challenge of personalized resource search within collaborative tagging systems by introducing a novel hybrid user profiling model. The model is designed to enhance the personalization of search results by incorporating both tags and ratings, thereby capturing a user's preferences and aversions more accurately.
In \citet{natarajan_recommending_2016}, the hybrid user profiles are constructed using a combination of click-through analysis, user tweet analysis, and user follower analysis. This multifaceted approach to profile building allows the system to capture a more comprehensive view of user preferences.
Lately, \citet{logesh_efficient_2019} and \citet{nkambou_learning_2023} provided hybrid approaches leveraging data encompassing contextual, behavioral, geographical, and categorical information, which is instrumental in understanding the user's preferences and activities, particularly in online social networks and recommender systems.

\subsubsection{Data preprocessing}\label{subsubsec:data_preprocessing}
A significant portion of profile data obtained, especially dynamic content from social media and recommendation engines, is often incorrect. Consequently, there is a necessity to cleanse the acquired datasets to ensure that the extracted features for profile modeling yield improved performance results. Many researchers adopt diverse \textit{data preprocessing} techniques in their studies on user profiling to ready their data for the subsequent analysis phase.
For instance, \citet{tang_combination_2010} employed a Hidden Markov Model approach to efficiently purify data, improving the user profile system. This method successfully addressed the issue of disambiguation in collecting user profile information. Basic pre-processing techniques, including tokenization, stop word removal, and tagging, were applied. Tokenization involves segmenting textual data into tokens using a tokenizer, with each token being assigned a tag.
Recently, \citet{ali_integrated_2020} provided a thorough examination of the data preprocessing steps necessary for effective web usage mining, which is the process of extracting useful information from web usage data to model user behavior. The paper outlines the various sources of web data, such as click-stream analysis and web server log files, and emphasizes the importance of preprocessing in the overall behavior user modeling process.

\subsubsection{Feature extraction}\label{subsubsec:feature_extraction}
\textit{Feature extraction} is another important step in user modeling, involving the retrieval of user profile features from diverse domains. Various procedures have been employed by researchers in this field to extract these features, aiming to enhance modeling performance. Commonly used elements include content features, pattern features, profile features, term features, and user behavioral features.
For instance, \citet{hijikata_implicit_2004} proposed a method for text data extraction based on user mouse behavior features on web platforms.
\citet{tang_combination_2010} focused on authorship profiling, evaluating a set of profile features consisting of six sets of attributes in article publication data, such as publication title, abstract, venue, abstract authors, publication year, and references. These features were extracted from digital libraries (ACM, Springer, and IEEE) using heuristics and represented as a feature vector with the number of occurrences as values.
\citet{meftah_emotion_2012} introduced a method that focuses on the extraction of relevant features from physiological data, such as electromyography (EMG) and respiration signals, which are critical for accurately identifying not only basic but also complex emotions. The feature extraction process is a part of the training module designed to discern patterns within the physiological signals that correspond to different emotional states.
\citet{li_weakly_2014} utilized a supervised learning classifier to extract data from Twitter and train the model. However, this approach solely took into account network information, which proves insufficient for inferring user attributes. 
\citet{ben_hassen_deep_2022} proposed a method to extract latent features from images of items, such as movie posters and clothing, by leveraging pre-trained deep learning models. The extracted features are then subjected to dimensionality reduction to make them more manageable and suitable for the subsequent stages of the recommendation process.

\subsubsection{User profile representation}\label{subsubsec:user_profile_representation}
\textit{User profile representation} refers to the way in which information stored in user profiles is structured and presented through a system or an application. The goal is to create an accurate portrayal of the user, enabling the system to make informed decisions and predictions about the user's interactions and preferences.
We discuss several representation variants: \textit{term-based}, \textit{semantic-network}, \textit{concept-based}, \textit{graph-based}, \textit{universal}, and \textit{holistic}.

\paragraph{Term-based (Vector-space model)}
The prevalent representation of users' interests is through a keyword-based or vector-based user model, comprising a set of terms weighted by vectors of keywords. These vectors, utilized to expand user queries, can be developed using different methods like Boolean, Term Frequency (TF), or Term Frequency-Inverse Document Frequency (TF-IDF)~\citep{shen_implicit_2005}.
User interest keywords, extracted from visited documents during browsing, can be represented by a single vector encompassing all interests or multiple vectors reflecting interests in various domains. The effectiveness of this model relies on the degree of generalization in the vectors~\citep{gauch_user_2007}. However, drawbacks include polysemy and potential misinterpretation of user interests.
An interesting article was presented by \citet{hu_user_2017}, who introduced an extraction algorithm that combines Word2Vec and TF-IDF to address shortcomings in the vector representation method.

\paragraph{Semantic-based (Ontology-based)}
To tackle the challenges of polysemy and synonymy in keyword-based profiles, a solution involves representing profiles using weighted concepts in a \textit{semantic network}. However, this introduces complexities in system construction, requiring an existing mapping between words of interest and concepts, such as WordNet ontology, a learning system method, or manual intervention.
\citet{sieg_web_2007} explored the concept of personalizing web search results through the use of ontological user profiles. These profiles are built by assigning interest scores to concepts within a domain ontology.
\citet{skillen_ontological_2012} introduced an ontological user profile model designed to enhance adaptive, context-aware applications in mobile environments. The model emphasizes the importance of dynamic user attributes and contains classes that represent user preferences, education, health profiles, capabilities, interests, and contextual information. The goal of the model is to facilitate personalized services that can adapt to users as they transition through different environments.
In the context of recommender systems, \citet{rimitha_ontologies_2018} proposed an ontology that encapsulates various aspects of the job recommendation domain, such as location, education, salary ranges, job description, and job experience. The ontology is designed to provide a structured and standardized representation of user preferences and job characteristics, which includes defining classes and the relationships between them. For instance, it includes relations like ``Has preferred salary'' and ``Has preferred location'', which are crucial for matching users with suitable job opportunities.
In the same domain, \citet{olufisayo_dahunsi_ontology-based_2021} presented a proposal for an advanced user profiling methodology within the context of apparel recommender systems. The proposed methodology is centered around the development of an ontology-based knowledge base designed to enhance the personalization of apparel recommendations by mapping user features to garment and ensemble features, utilizing domain expert knowledge and predefined style rules.

\paragraph{Concept-based (Hierarchy-based)}
Concept-based profiles and semantic network-based profiling share the common approach of representing information through nodes of concepts and the relationships between them. Different methods are employed to assign weights indicating the user's level of interest in each topic.
\citet{nanas_building_2003} utilized various techniques to categorize user interests within a hierarchical structure, employing a knowledge base for this purpose. The researchers applied a library classification system (LCS) to identify user interest hierarchies, while the edge weights were calculated using the TF-IDF mechanism.
\citet{malinowski_hierarchies_2006} introduced a conceptual framework designed to enhance data warehouses and Online Analytical Processing (OLAP) systems. The model builds upon the Entity-Relationship (ER) model, incorporating specific constructs to represent dimensions, hierarchies, and fact relationships, and provides a graphical notation for these elements.
Recently, \citet{wu_hierarchical_2019} developed a hierarchical user and item representation model that leverages a three-tier attention mechanism to enhance recommender systems. This model is designed to extract and learn representations from user and item reviews at multiple levels of granularity, including words, sentences, and reviews themselves.

\paragraph{Graph- and Knowledge graph-based}
Graph-based representation in user modeling involves utilizing graphs or networks to represent and analyze information about users. This approach is particularly effective in capturing complex relationships, dependencies, and interactions among various elements in a user's profile. The representation is typically structured with nodes representing entities or concepts and edges denoting relationships or connections between them.
In the last few years, works leveraging graph-structures increased.
\citet{wang_adversarial_2019} introduced an adversarial substructured learning framework tailored for mobile user profiling. The approach hinges on the construction of user activity graphs that encapsulate the unique characteristics and preferences of mobile users. The central challenge addressed is the learning of deep representations from these graphs to accurately profile users
\citet{chen_global_2022} presented a novel approach named Global and Personalized Graphs for Heterogeneous Sequential Recommendation (GPG4HSR) that enhances user profiling and modeling in the context of heterogeneous sequential recommender systems. The proposed model incorporates two distinct graph layers to improve user profiling. The first is a global graph layer that captures the transitions between different behaviors across all users, providing a macro-level view of user interactions. The second is a personalized graph layer that models the interaction sequences of individual users, taking into account the unique intentions and preferences of each user by focusing on the contextually relevant nodes adjacent to their interactions.
\citet{guan_personalized_2022} implemented a novel approach designed to enhance the personalization of fashion compatibility recommendations by leveraging user preference modeling. The proposed model employs a heterogeneous graph that integrates users, items, and attribute entities, along with their interrelations, to create a comprehensive representation of the fashion domain.
Other relevant contributions in the area can be seen in \citet{guo_user_2018}, \citet{wanda_deepprofile_2020}, \citet{wang_incremental_2020}, \citet{yang_modeling_2022}, and \citet{yang_going_2023}.

\textit{Knowledge graphs} are essential in user modeling, providing a robust framework to organize and represent complex relationships. In user modeling, they efficiently capture users' preferences and behaviors, enabling personalized experiences and enhancing recommender systems. By mapping connections between users and their interests, knowledge graphs facilitate a nuanced understanding of individual profiles, allowing systems to adapt dynamically to users' unique needs and preferences.
\citet{wang_incremental_2020} presented an innovative approach to mobile user profiling by proposing an integrated reinforcement learning framework that incorporates spatial knowledge graphs.
\citet{anelli_sparse_2021} introduced a knowledge-aware recommender system that emphasizes user modeling through a sparse factorization approach. By analyzing historical data and item attributes, the model constructs a detailed model of user-item interactions, which is essential for understanding user preferences and tailoring recommendations.
Instances of similar works are seen in \citet{huang_explainable_2019}, \citet{wang_adversarial_2019}, \citet{wang_enhancing_2021}, and \citet{xuan_knowledge_2023}.

\paragraph{Universal user representation}
It aims to create a generalized user representation that encapsulates the essential characteristics and behaviors of a user. This representation is constructed to be versatile and applicable across a variety of real-world applications, without the need for specific adjustments for each task. Universal user representation emphasizes adaptability and versatility to accommodate a wide range of users and their diverse needs, seeking to create a model that can be applied across different contexts and user groups.
For instance, \citet{ni_perceive_2018} developed a deep learning framework designed to create universal user representations for a variety of e-commerce personalization tasks. The proposed architecture is capable of simultaneously performing multiple personalization tasks, such as click-through rate (CTR) prediction, learning to rank (L2R), price preference prediction, fashion icon following prediction, and shop preference prediction. The model is trained on a substantial offline dataset from Taobao, one of the largest e-commerce platforms.
\citet{yuan_parameter-efficient_2020} introduced a transfer learning architecture designed to create universal user representations for recommender systems, which consists of two stages: the first involves pre-training on large-scale datasets to capture universal user behaviors, and the second involves fine-tuning with supervised labels to adapt the pre-trained representations to several different tasks.
\citet{gu_exploiting_2021} presented an approach for learning universal user representations from unlabeled behavior data. The primary objective is to generate user representations that encapsulate a wealth of information, enabling their application across a variety of downstream tasks, such as predicting user preferences and inferring user profiles.
\citet{yuan_one_2021} developed a method of user modeling that embodies the ``One Person, One Model, One World'' paradigm. This paradigm is designed to create a universal user representation that can continually learn across different tasks without succumbing to catastrophic forgetting, which is the loss of previously acquired knowledge when new information is learned.
An innovative technique in this area has been proposed by \citet{kim_task_2023}, where they introduced a continual user representation learning method. Their main idea is to use task embeddings to generate task-specific soft masks, allowing model parameters to be updated throughout the training sequence and capturing the relationship between tasks. Additionally, a knowledge retention module with a pseudo-labeling strategy is introduced to alleviate catastrophic forgetting.

\paragraph{Holistic user modeling}
Although it can be related to \textit{universal user representation} as both concepts aim to create generalized user models, \textit{holistic user modeling} is about building a detailed and comprehensive model that takes into account all relevant aspects of a user's interaction with a system. This includes not only the user's actions but also their expectations, preferences, and limitations. It is more focused on the integration of various data sources to create a complete picture of the user.
Specific examples can be found in \citet{gong_when_2018} and \citet{gong_jnet_2020}. In the former work, they designed a probabilistic generative model for holistic user behavior modeling in social media. It integrates opinionated content modeling with social network structure modeling to capture the consistency and heterogeneity of user behavior across various user-generated data modalities.
In the latter, the authors proposed a solution that learns user representations in social networks from both network structure and text content by capturing the dependency between these two modalities in the latent topic space. This approach uses statistical topic models to handle the unstructured nature of the text and embeds both users and topics in a low-dimensional space to capture their mutual dependency. The user's affinity to a topic is characterized by their proximity to the topic's embedding, which is used to generate text documents, while the affinity between users is modeled by the proximity between their embeddings, which is used to generate social network connections.
Another robust line of research in this area has been conducted by \citeauthor{musto_framework_2018}, from 2018 and 2021.
In \citet{musto_framework_2018}, the authors discussed the development of a holistic user profile framework that integrates a user's digital footprints from both social networks and personal devices to create a comprehensive digital identity. This framework collects data from platforms such as Twitter and Facebook, as well as from devices like Android smartphones and FitBit wristbands. The collected data undergoes processing and enrichment to form a unique user model that reflects the individual's behavior and interests.
\citet{musto_towards_2020} presented a knowledge-aware food recommender system designed to provide personalized recipe suggestions based on user characteristics and health-related constraints. The system incorporates a holistic user model to re-rank recipes and was assessed through a web-based experiment. The findings indicate that holistic user profiles significantly influence user preferences.
\citet{musto_towards_queryable_2020,musto_myrror_2020,musto_myrrorbot_2021} discussed the implementation of Myrror and MyrrorBot.
Myrror is a platform designed to create a holistic user model by integrating data from social networks, smartphones, and wearable devices. The platform's goal is to alleviate the problem of information overload and offer personalized services based on the user's profile.
MyrrorBot is a conversational agent developed to enable users to interact with and query their user models using natural language and is built on the Myrror platform. The core components of MyrrorBot include the Intent Recognizer module, which interprets user requests, and the Generator module, which formulates responses in natural language. The system's conversational interface aims to enhance traditional web interfaces, allowing for a more intuitive and efficient way for users to access and inspect the information contained in their profiles.

\begin{figure}
    \centering
    \includegraphics[width=\linewidth]{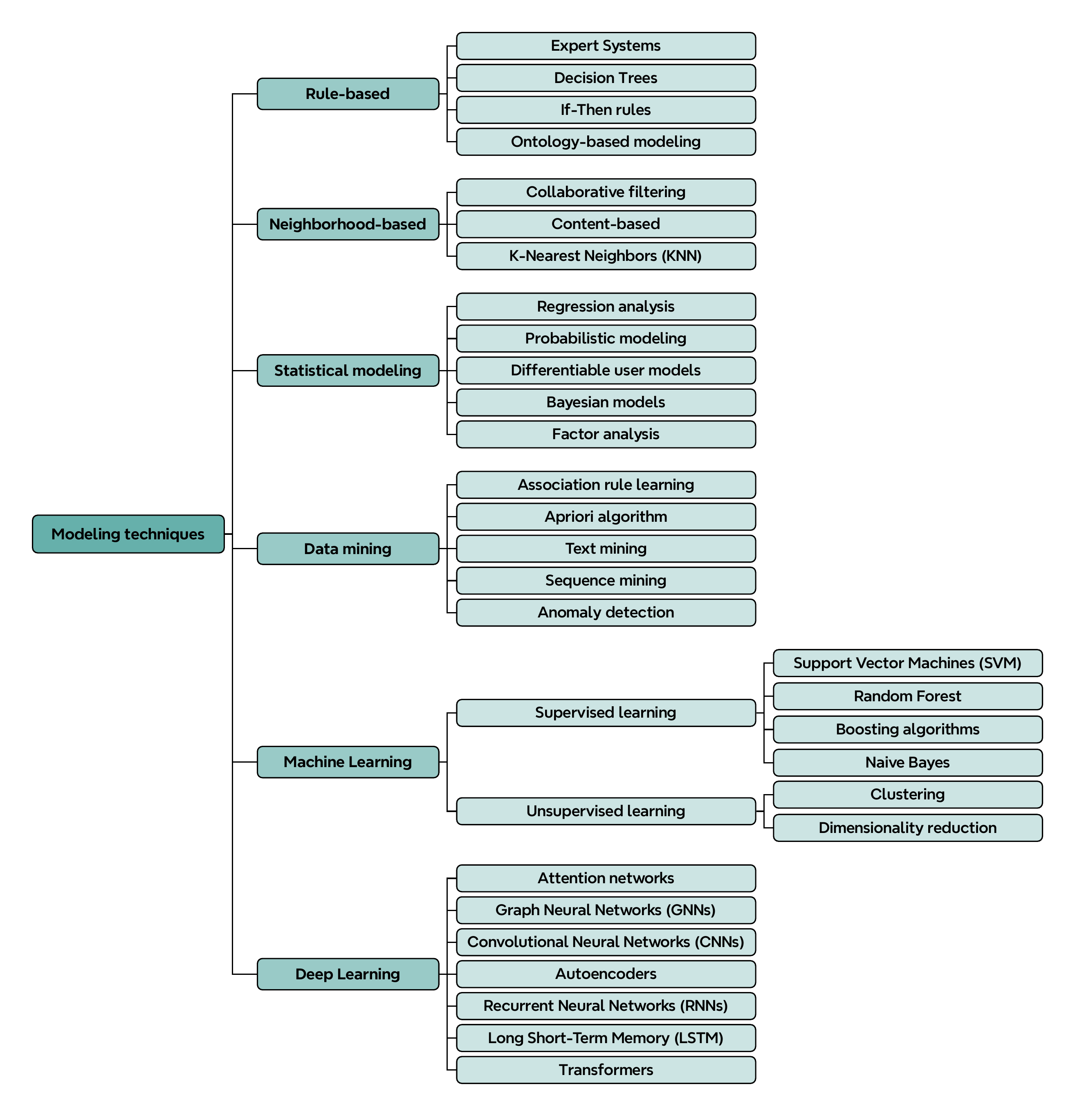}
    \caption{Specific taxonomy of the \textit{Modeling techniques}.}
    \label{fig:modeling_techniques}
\end{figure}

\subsubsection{Modeling techniques}
Once the information is organized, the next step is to select suitable techniques with varying characteristics to build a user profile. This phase is essentially about the creation of a computational model, which is constructed using the extracted features and is capable of predicting the needs or preferences of the user.
In the proposed categorization, whose specific taxonomy is displayed in Figure~\ref{fig:modeling_techniques}, we present in detail recent relevant works for each topic.
It is also important to note that some techniques may span multiple categories depending on their specific implementation and use case. We examine each method within the category in which it is most discussed in the literature in the context of user modeling.

\paragraph{Rule-based}
\textit{Rule-based user modeling} involves creating a set of predefined rules or conditions that determine user behavior or preferences. These rules are typically designed based on expert knowledge or domain-specific guidelines (e.g., \citep{cheung_chiu_using_1998,adomavicius_expert-driven_2001,razmerita_ontology-based_2003,brut_rule-based_2009,aleven_rule-based_2010}).
Rule-based approaches are classified into:
\begin{itemize}
    \item \textbf{\textit{Expert systems}} refer to computer programs that emulate the decision-making ability of a human expert in a specific domain. They use a knowledge base of rules to draw inferences and make decisions.
    For example, \citet{rizzo_modeling_2016} explored the development of a rule-based expert system designed to model and infer mental workload in users, with the aim of enhancing user modeling techniques. The expert system is constructed by eliciting knowledge from domain experts and translating this knowledge into computational rules. These rules are then applied to model user's mental workloads, using heuristics to combine them into a numerical index. The system accounts for various attributes that contribute to the mental workloads, such as mental demand, temporal demand, effort, performance, and frustration.

    \item \textbf{\textit{If-Then rules}} are a form of declarative knowledge representation where specific conditions (\textit{if}) are defined to trigger certain actions or conclusions (\textit{then}). In the context of user modeling, these rules are used to make decisions or predictions based on observed user behavior or input features.
    \citet{sola_rule-based_2021} introduced a rule-based recommender system specifically designed to assist users in the task of business process modeling. In the context of user modeling, this approach is significant as it adapts to the user's input and the structure of the process model at hand. By analyzing a repository of BPMN 2.0 models, the system learns the common sequences and dependencies of activities, which reflects the collective expertise and practices encoded in these models.

    \item \textbf{\textit{Decision trees}} involve creating a tree-like model where each node represents a decision based on input features. They are used for classification and regression tasks, making decisions by traversing the branches of the tree.
    To cite a specific work, \citet{sarker_behavdt_2020} proposed the Behavioral Decision Tree (BehavDT) model, which is designed to predict smartphone user behavior by considering multi-dimensional contexts. Unlike traditional decision tree models, BehavDT emphasizes behavior-oriented generalization and individualized preference-oriented decision-making, which enhances its predictive accuracy.
    In a recent article, \citet{streeb_task-based_2022} provided a comprehensive survey of visualization techniques for decision trees and rule-based classifiers, with a focus on user modeling and the tasks associated with the development and application of these models.

    \item \textbf{\textit{Ontology-based modeling}} concerns using ontologies, which define concepts and relationships in a domain, to represent and reason about user preferences, characteristics, and behavior. It utilizes ontological rules for making inferences about users.
    \citet{rimitha_ontologies_2018} presented a method for enhancing personalized job RSs through the development of ontology-based user profiles. It underscores the significance of user profiles in delivering tailored job suggestions to individuals by accurately reflecting user preferences and experiences.
    \citet{chari_explanation_2020} introduced the Explanation Ontology, a comprehensive model specifically crafted to encapsulate the various facets of explanations within user-centered AI systems. The model is presented as a valuable tool for system designers, offering a structured approach to user modeling with ontology in the context of AI explanations. It emphasizes the importance of user-centric design in AI and the ongoing efforts to update and refine the ontology to support the dynamic landscape of user needs and system capabilities. 
\end{itemize}

\paragraph{Neighborhood-based}
\textit{Neighborhood-based user modeling} deals with the idea that users who are similar to each other in certain aspects are likely to have similar preferences. It involves analyzing the behavior or characteristics of a user in relation to their ``neighborhood'' or a group of similar users (e.g., \citep{konstan_grouplens_1997,kim_collaborative_2011,meftah_emotion_2012,subramaniyaswamy_adaptive_2017,cami_user_2019,sanchez_building_2019,javed_review_2021}).
We identify the following groups:
\begin{itemize}
    \item \textbf{\textit{Collaborative filtering}} relies on user-item interactions and recommendations from a group of users with similar tastes. It can be \textit{user-based}, \textit{item-based}, or a combination of both.
    \citet{ma_nest_2022} investigated the effects of rapid, population-scale concept drifts, such as those induced by pandemic-like events, on collaborative filtering models within RSs. The core of the study is the introduction of the NEST framework, which simulates the evolution of user needs and behaviors in response to non-stationary external events.
    \citet{wang_attention-based_2022} introduced a recommender system framework known as Time-aware Attention-based Deep Collaborative Filtering (TADCF), whose core is its ability to model dynamic user preferences over time.

    \item \textbf{\textit{Content-based filtering}}\footnote{\textit{Content-based filtering} is placed under the ``\textit{Neighborhood-based}'' category to highlight its similarity to \textit{collaborative filtering} in terms of creating a neighborhood of similar items. In content-based filtering, this neighborhood is formed based on the content features of items and the user's historical interactions. The placement acknowledges that, in a sense, content-based filtering creates a ``content-based neighborhood'' for recommendations. However, it is important to note that content-based filtering is often considered a separate and distinct approach due to its emphasis on item characteristics and individual user profiles.} recommends items based on the user's past preferences and the characteristics of the items. It considers the content features of items and user profiles.
    To offer a concrete example, \citet{pujahari_item_2022} presented a wide approach to enhancing content-based user modeling in recommender systems. For user profile generation, the paper introduces an ensemble classifier that operates within an iterative learning framework. This classifier is designed to incrementally learn and update user preferences, allowing the system to adapt to user interactions and feedback over time. The iterative update rules ensure that the user profiles evolve, capturing the dynamic nature of user interests.

    \item \textbf{\textit{K-Nearest Neighbors}} (KNN) is a technique that classifies or recommends based on the majority class or average of the k-nearest data points in feature space.
    For instance, \citet{nagaraj_university_2022} developed a recommender system aimed at aiding undergraduate and graduate students in selecting the most suitable colleges based on their individual profiles. The system employs user modeling through the KNN algorithm to calculate weighted scores for college recommendations. By analyzing factors such as SAT scores, tuition fees, and academic records, the KNN algorithm helps to match students with institutions where they have the highest likelihood of acceptance and success.
\end{itemize}

\paragraph{Statistical modeling}
\textit{Statistical modeling} for user behavior refers to the use of statistical methods to analyze and model patterns in user data, along with understanding relationships and making predictions (e.g., \citep{horvitz_lumiere_1998,zukerman_predictive_2001,zigoris_bayesian_2006,zhang_efficient_2007,harvey_bayesian_2011,seroussi_personalised_2011,benmakrelouf_towards_2015,tomeo_exploiting_2015,vu_temporal_2015,yadav_detection_2015,chen_joint_2019}).
This category includes:
\begin{itemize}
    \item \textbf{\textit{Regression analysis}} involves modeling the relationship between a dependent variable and one or more independent variables. It is used to understand and predict the impact of changes in the independent variables on the dependent variable.
    To cite a relevant paper, \citet{gao_collaborative_2017} introduced an item recommendation model named CDUE, which stands for Collaborative Dynamic User profile Evolution. This model is designed to predict a user's future interests by incorporating a regression approach that accounts for the temporal dynamics of both user interests and item topics. Specifically, the model employs a dynamic sparse topic regression to track how item topics evolve over time, ensuring that the recommendations remain relevant as the content changes.

    \item \textbf{\textit{Probabilistic modeling}} concerns the use of probability distributions to represent uncertainty and variability in user behavior or preferences. This approach allows for the incorporation of statistical methods, likelihoods, and estimation techniques to capture the inherent uncertainty in modeling user characteristics based on observed data. Probabilistic modeling encompasses a variety of statistical techniques that utilize probability theory to model and understand complex relationships within user data.
    To illustrate with particular examples, \citet{hadoux_learning_2018} introduced a framework for probabilistic user modeling in the context of persuasive argumentation. The core of the framework is the use of beta distributions to represent the uncertainty in user beliefs, which is a departure from traditional models that use sharp, deterministic values.
    An intriguing approach has been proposed in \citet{vardasbi_when_2020} and \citet{oosterhuis_robust_2021}. A key aspect of the papers is the use of probabilistic user models for simulating user interactions in the experimental setup. These models are crucial for generating data that reflects the biases present in real-world scenarios. By employing probabilistic user models, the authors are able to create a more realistic testing environment for their proposed affine estimator.

    \item \textbf{\textit{Differentiable user models}} are computational constructs that enable efficient probabilistic user modeling by providing a differentiable approximation of cognitive behavior simulators, which are often non-differentiable and computationally prohibitive for practical applications. These models are designed to be compatible with modern machine learning frameworks, often implemented using neural networks, and allow for real-time applications even with advanced cognitive models that lack a closed-form likelihood.
    The most important work on this topic has been written by \citet{hamalainen_differentiable_2023}, where the authors introduced this innovative approach to cognitive user modeling by employing differentiable surrogates, which significantly enhance computational efficiency and enable real-time applications. The differentiable surrogates are trained offline and are designed to be utilized online, effectively addressing the computational challenges that have traditionally limited the use of probabilistic user modeling based on cognitive models. The differentiable user models are shown to overcome the limitations of existing likelihood-free inference methods, providing a solution that is both computationally viable for online applications and maintains user modeling accuracy.

    \item \textbf{\textit{Bayesian models}} specifically adhere to the principles of Bayesian probability theory to update beliefs about user preferences or behavior based on new evidence. They provide a framework for incorporating prior knowledge and updating it with observed data.
    \citet{shen_user_2021} developed a user profiling system for inferring gender and personality traits from non-linguistic audio data during conversations. The core component of the system is a voice activity detection (VAD) mechanism and a feature extraction process that focuses on conversational dynamics. The VAD component utilizes a Bayesian algorithm to distinguish between speech and non-speech segments within the audio data, which is crucial for accurately capturing the conversational features relevant to user profiling.

    \item \textbf{\textit{Factor analysis}} is a statistical method used to identify \textit{latent} (i.e., unobservable) \textit{factors} that explain patterns of correlations within observed variables. It is often applied to uncover underlying dimensions in user behavior or preferences. An interesting work has been presented by \citet{yuan_generalized_2020}. The paper introduces a Generalized and Fast-converging Non-negative Latent Factor (GFNLF) model designed to predict user preferences in recommender systems. The GFNLF model aims to overcome the limitations of slow convergence and restricted representational capabilities observed in traditional latent factor models. 
\end{itemize}

\paragraph{Data mining}
\textit{Data mining} refers to the process of discovering patterns, associations, and trends within large datasets. Techniques like association rule mining and clustering are used to extract valuable information and insights from user-related data (e.g., \citep{eirinaki_web_2003,pierrakos_web_2003,mobasher_data_2007,xie_large-scale_2009,angeline_association_2013,erlandsson_finding_2016,burgos_data_2018,sarker_mining_2018,abu_sulayman_user_2019,shazad_finding_2019}).
The topics pertaining to data mining are:
\begin{itemize}
    \item \textbf{\textit{Association rule learning}} is a technique designed to unveil meaningful relationships, patterns, or associations among variables (in our context, user behaviors or preferences) within extensive datasets. This method seeks to identify rules expressing correlations or co-occurrences between different user actions or characteristics.
    To provide some samples, \citet{si_association_2019} presented an innovative use of association rules to uncover and analyze the interests of users on social networks, with a particular focus on Twitter and LinkedIn. For Twitter, the authors propose a method that reorders a user's interests based on association rules derived from user behavior. On LinkedIn, the researchers collected profile data to identify the distribution of user interests. They applied association rules to this dataset and discovered a substantial number of correlations between various human interests.
    \citet{agouti_graph-based_2022} proposed an innovative algorithm named Diffusion-Graph-Based Influence Maximization (DGIM), which leverages association rule mining to model the spread of influence and to detect relationships between users concerning specific topics. 

    \item \textbf{\textit{Apriori algorithm}} is specifically tailored association rule learning in user modeling and is employed to identify frequent item sets and generate association rules. Following the ``\textit{apriori}'' principle, it iteratively refines rules based on user behavior, helping to reveal patterns and associations in interactions or preferences.
    \citet{singh_optimized_2021} developed an optimized collaborative filtering algorithm for recommenders that utilizes the apriori algorithm to enhance user profile creation. The paper details how the apriori algorithm is employed to generate detailed user profiles by analyzing users' interactions with items. These profiles are then used to modify the similarity measure for finding similar neighbors, which is a critical step in collaborative filtering algorithms.

    \item \textbf{\textit{Text mining}} involves extracting meaningful information and patterns from unstructured text data. It includes techniques such as natural language processing, sentiment analysis, and text clustering to understand and model user preferences expressed in textual form.
    \citet{greco_emotional_2020} explored the use of Emotional Text Mining (ETM) as a tool for user profiling in the context of brand management, with a specific focus on analyzing Twitter messages about a sportswear brand. The paper demonstrates how ETM can be leveraged for user profiling in brand management, providing theoretical and managerial insights. It showcases the method's ability to simplify the analysis of large textual datasets and its contribution to the field through the extraction of actionable information for enhancing customer understanding and brand strategy.
    \citet{heidari_deep_2020} presented, in the same domain of social networks, an approach to detecting social bots on Twitter by focusing on user profiling extracted from tweet text.

    \item \textbf{\textit{Sequence mining}} focuses on discovering patterns or relationships within sequential data. It is commonly used to identify sequences of events or actions in a user's interactions over time, providing insights into behavior patterns.
    An interesting approach has been proposed by \citet{gurbanov_action_2017}. They developed a hybrid recommender system that integrates sequence mining with collaborative filtering to enhance the prediction of user actions on an online career-oriented social networking platform. The focus of the system is to utilize the sequential patterns of user interactions, particularly with job postings, to forecast specific actions such as replying to a job posting. The sequence mining component is crucial in the hybrid model as it is responsible for predicting the probability of the next user action based on the observed sequence of interactions.

    \item \textbf{\textit{Anomaly detection}}, in user modeling, aims to identify instances that deviate significantly from the expected or normal behavior. It involves detecting unusual patterns, outliers, or anomalies in user data that may indicate fraudulent activity, errors, or novel user behavior.
    To supply detailed instances, \citet{bohmer_mining_2020} presented an extended study on anomaly detection for temporal data in business processes, with a focus on user modeling. The authors propose a novel approach that leverages association rule mining to detect anomalies in process runtime behavior, aiming to differentiate between benign and malign anomalies, minimize false positives, and elucidate the root causes of anomalies to users.
    \citet{sharma_user_2020} introduced a procedure for anomaly detection in user behavior analytics by employing an unsupervised method. The primary objective is to identify insider threats by analyzing patterns in user session activities.
\end{itemize}

\paragraph{Machine learning}
\textit{Machine learning} (ML) refers to a set of computational techniques that empower systems to automatically learn patterns, relationships, and predictions from user data and behaviors. Generally, ML methods are divided into two groups:
\begin{itemize}
    \item \textbf{\textit{Supervised learning}} involves training algorithms on labeled datasets to predict or classify user behavior based on provided examples, enabling the system to make accurate predictions for new user interactions (e.g.,~\citep{cufoglu_comparative_2008,tang_combination_2010,santra_classification_2012,raghuram_efficient_2016}).
    Specific applications in user modeling make use of the following supervised learning algorithms:
    \begin{itemize}
        \item \textbf{\textit{Support Vector Machines}} (SVM) are models that can be used for classification and regression tasks. They find a hyperplane that best separates classes or predicts a target variable.
        \citet{ahmad_spam_2021} introduced a method for identifying spam tweets on Twitter by leveraging an SVM classifier, with a particular focus on analyzing user interactions. In this study, it is applied to differentiate between spam and non-spam tweets using a variety of features, including those derived from user interactions.

        \item \textbf{\textit{Random Forest}} is an ensemble learning method that constructs a multitude of decision trees during training and outputs the class that is the mode of the classes from individual trees.
        \citet{chen_monitoring_2021} presented a study that evaluates the effectiveness of the random forest algorithm in identifying high-risk users within public opinion discussions on digital platforms. In particular, is used to classify users based on their risk of contributing to negative public opinion.

        \item \textbf{\textit{Boosting algorithms}} are a set of ensemble learning techniques that combine the predictions of multiple weak models to create a robust and accurate predictive model. These algorithms sequentially train weak learners, with each subsequent model focusing on the mistakes of its predecessors. Popular boosting algorithms include \textit{AdaBoost} and \textit{Gradient Boosting}, which are applied to capture complex relationships in user data.
        An extensive study on the application of boosting algorithms for user profiling, specifically for predicting student performance in higher education, has been presented by \citet{hamim_student_2022}. To achieve this, the analysis compares different research studies that utilize boosting algorithms for student profile modeling. The paper also explores the impact of student behaviors and characteristics on their academic performance. It proposes a student profile model that leverages educational data mining techniques to predict, classify, and adaptively support students' learning processes. This model can also be used for e-RSs that tailor educational content to individual student needs.

        \item \textbf{\textit{Na{\"i}ve Bayes}} is a probabilistic algorithm based on Bayes' theorem. It assumes that features are conditionally independent, making it efficient and effective for classification tasks. A valuable research paper in the student profiling domain has also been provided for this specific algorithm. Specifically, \citet{tripathi_naive_2019} presented a study on student profile modeling, with a particular focus on predicting student performance. The core of the study is the application of the Naïve Bayes classifier, which is compared against the existing Support Vector Machine (SVM) classifier. The results indicate that the Naïve Bayes classifier outperforms the SVM in terms of both accuracy and execution time, making it a more efficient and effective tool for predicting student performance.
    \end{itemize}

    \item \textbf{\textit{Unsupervised learning}} harnesses input data devoid of explicit labels to uncover inherent patterns or structures within user behavior, providing valuable insights into the underlying data structures without the need for predefined categories or outcomes (e.g.,~\citep{castellano_similarity-based_2007,boratto_modeling_2014,benmakrelouf_towards_2015,van_dam_online_2015,canigueral_flexibility_2021,yang_local_2021}.
    Algorithms belonging to this group are:
    \begin{itemize}
        \item \textbf{\textit{Cluster analysis}} (or simply \textit{clustering}) groups data points based on similarities, creating clusters or groups of similar instances. Common algorithms include \textit{k-means clustering} and \textit{hierarchical clustering}. Clustering helps identify user segments with similar behavior or preferences.
        To provide concrete examples, \citet{ouaftouh_social_2019} proposed a clustering-based approach to enhance social recommender systems. The core of the research is the application of partitional clustering algorithms, with a focus on the K-means algorithm, to group user profiles into clusters. These clusters are intended to represent users with similar interests and preferences, which can then be used to provide more accurate recommendations within an e-commerce environment.
        
        \item \textbf{\textit{Dimensionality reduction}} techniques reduce the number of features in the dataset while retaining its essential information. Reducing dimensionality aids in visualizing user data patterns and improving computational efficiency. \textit{Principal Component Analysis} (PCA) is the most popular method of this type.
        \citet{bi_anomaly_2016} introduced an anomaly detection model to scrutinize user behavior within database systems and web browsing environments. PCA analyzes user behavior by calculating covariance matrices and extracting significant behavior features. This statistical approach simplifies the complexity of user behavior data, allowing the model to distinguish between normal and abnormal patterns effectively.
    \end{itemize}
\end{itemize}

\paragraph{Deep learning}
\textit{Deep learning} (DL) involves using \textit{deep neural networks} (i.e., neural networks with multiple layers) to capture complex patterns and representations in user data. It excels at handling large and unstructured user data, making them suitable for tasks such as user behavior prediction and personalized recommendations. As observed in various fields, plenty of articles have been published in user modeling employing several DL techniques and architectures (e.g.,~\citep{elkahky_multi-view_2015,an_neural_2019,ren_lifelong_2019,heidari_deep_2020,zhou_enhancing_2020,wang_reinforced_2021,wen_hierarchically_2021,zhao_ameir_2021,zhou_equivariant_2023}).
A description of these methods and examples of contributions for each of them are provided below. It is important to highlight that a single approach can include more than one category. We consider these cases only under one topic.
\begin{itemize}
    \item \textbf{\textit{Attention networks}} (or \textit{mechanisms}) allocate varying degrees of importance to different parts of the input sequence. This enables the model to focus on specific aspects of user behavior, enhancing its ability to capture and understand intricate patterns in the data (e.g.,~\citep{fazil_deepsbd_2021,chu_mitigating_2022,qi_news_2022}).
    To detail specific works, \citet{qi_trilateral_2021} introduced TRISAN (Trilateral Spatiotemporal Attention Network), an attention-based neural model designed for user behavior modeling in the context of location-based search. TRISAN is unique in its ability to incorporate temporal relatedness into the analysis of user behavior, particularly in how it relates to the geographic proximity of items and user requests. This is achieved through a specialized fusion mechanism that considers not only physical distance but also semantic similarity to model geographic closeness.
    \citet{wang_user_2020} and \citet{wang_modeling_2022} combines \textit{graph neural networks} and \textit{attention mechanisms}.
    In the former article, the authors presented UIL-HGAN, a model for user identity linkage across social networks, emphasizing the enhancement of user modeling. UIL-HGAN stands for User Identity Linkage using a Heterogeneous Graph Attention Network, a method that integrates user profiles, user-generated content, and structural information within a heterogeneous social network graph.
    In the latter, an approach to enhance user profiling in the context of personalized point-of-interest (POI) recommendations is proposed. The paper introduces the Spatial-Temporal Graph Convolutional Attention Network (STGCAN), a model that leverages a knowledge graph with temporal information to capture the dynamic and evolving preferences of users with respect to various locations.

    \item \textbf{\textit{Graph Neural Networks}} (GNNs) are designed to handle graph-structured data, representing relationships between entities. They are proficient in capturing the complex interdependencies within user networks or interaction graphs, providing a comprehensive view of user relationships and preferences (e.g.,~\citep{chen_semi-supervised_2019,hu_graph_2020,wang_calendar_2020,wang_user_2020,chen_catgcn_2021,dai_say_2021,wu_user_2020,xia_knowledge-enhanced_2021,xia_graph_2021,yan_relation-aware_2021,luo_personalized_2022,wang_modeling_2022,zhao_co-learning_2022}).
    To provide specific instances, \citet{agarwal_modeling_2022} introduced SEINE (Spam Detection using Interaction Networks), a graph neural network model tailored for identifying spam users on web-scale social media platforms. SEINE's primary innovation lies in its sophisticated user behavior and interaction modeling, which is achieved through the construction of a user-entity graph. This graph is not merely a collection of nodes and edges but a rich, heterogeneous network that captures the multifaceted interactions between users and various entities such as posts, comments, and likes.
    \citet{han_multi-aggregator_2022} developed the Multi-Aggregator Time-Warping Heterogeneous Graph Neural Network (MTHGNN), a model for personalized micro-video recommendation, with a strong emphasis on user modeling. The MTHGNN is designed to capture the dynamic and multi-faceted nature of user preferences in the context of micro-video content, such as that found on platforms like TikTok.
    \citet{yan_interaction-aware_2022} presented a framework named Interaction-aware Hypergraph Neural Networks (IHNN) specifically designed for enhancing user profiling on e-commerce and social media platforms. IHNN employs hypergraph and meta-path based graphs to effectively model user interactions and attributes. The framework is constructed with a multi-view hypergraph approach and utilizes convolutional operations to aggregate high-order user information comprehensively. It also incorporates a semi-supervised learning approach for predicting user profiles. This method is particularly useful for tasks such as predicting user characteristics like age and gender.
    \citet{cheng_multi-behavior_2023} introduced a model named Multi-Behavior Recommendation with Cascading Graph Convolution Networks (MB-CGCN), designed to enhance recommendations by leveraging multiple user behaviors. The architecture is structured around a sequence of Graph Convolutional Network (GCN) blocks, each representing a different user behavior in a behavior chain. The key innovation of MB-CGCN lies in its ability to capture the dependencies between various behaviors to improve the accuracy of recommendations.

    \item \textbf{\textit{Convolutional Neural Networks}} (CNNs) leverage convolutional layers to learn hierarchical features and spatial relationships within sequential or tabular data. While traditionally used in image processing, CNNs can be adapted to understand complex patterns in user interactions, identifying important temporal and contextual features for effective representation learning in user behavior (e.g.,~\citep{karatzoglou_convolutional_2018,wang_adversarial_2019,wanda_deepprofile_2020,cura_driver_2021,mekruksavanich_convolutional_2021}).
    Other contributions are, for example, \citet{yuan_one_2021}, who introduced Conure, a continual learning framework that utilizes a Temporal Convolutional Network (TCN) architecture to represent users. It is specifically engineered to handle sequential tasks and is adept at transferring knowledge from one task to another while retaining important information from past tasks.
    \citet{qi_news_2022} presented an approach to personalized news recommendation through a Candidate-aware User Modeling (CAUM) framework. By using a candidate-aware self-attention network, a candidate-aware convolutional neural network, and a candidate-aware attention network, the CAUM framework is designed to capture both global and short-term user interests that are specifically relevant to the candidate news items. This convolutional neural network is adapted to process user behavior data with a focus on candidate news, helping to capture local and sequential patterns that are indicative of user interests.

    \item \textbf{\textit{Autoencoders}} are neural network architectures designed for unsupervised learning. They aim to reconstruct input data, and in the context of user modeling, autoencoders can be employed for tasks such as feature learning and data compression to uncover latent representations in user behavior (e.g.,~\citep{rajashekar_smart_2016,abu_sulayman_user_2019,wang_adversarial_2019,pan_learning_2020,sharma_user_2020}).
    Noteworthy specific publications include the work by \citet{deng_user_2022}, who introduced a user behavior analysis model that leverages stacked autoencoders for dimensionality reduction and feature selection in the context of a complex power grid environment. The model is designed to optimize resource coordination and planning by analyzing user behavior characteristics.
    \citet{fazelnia_variational_2022} developed a user modeling approach named FS-VAE, which employs a variational autoencoder (VAE) framework that leverages user-item interaction data to learn effective user representations in the music recommender systems domain.
    \citet{liu_federated_2023} proposed Hierarchical Personalized Federated Learning (HPFL), a model designed to tackle a number of heterogeneity challenges. In its ``augmented'' version (i.e., AHPFL), the authors incorporate augmented mechanisms to filter out low-quality information and integrate high-quality information, such as singular value decomposition (SVD) and autoencoders, to improve the effectiveness of user models.

    \item \textbf{\textit{Recurrent Neural Networks}} (RNNs) are tailored for sequential data, capturing dependencies over time. Their cyclic connections make them adept at understanding the temporal aspects of user behavior, making them suitable for tasks involving sequential interactions (e.g.,~\citep{carvalho_exploiting_2017,ishitaki_application_2017,ni_perceive_2018,tripathi_reinforcement_2018,yu_adaptive_2019,gu_hierarchical_2020}).
    Interesting works can be found, for instance, in \citet{donkers_sequential_2017}, where the authors proposed an approach to sequential recommender systems by incorporating user-specific information into RNNs. The paper introduces a user-based RNN model that personalizes recommendations by considering the unique preferences and behavioral patterns of individual users. To achieve this, they extend traditional RNNs with a gated architecture using Gated Recurrent Units (GRUs), which facilitates the integration of user data into the sequential learning process.
    \citet{ge_personalizing_2018} designed a framework to personalize search engine results by leveraging a hierarchical recurrent neural network (HRNN) with a query-aware attention mechanism. The framework's objective is to utilize sequential data from users' past interactions to construct dynamic user profiles, thereby enhancing the personalization of search results.
    \citet{chu_mitigating_2022} presented a methodology for predicting student performance in online learning environments, with a focus on enhancing user modeling across diverse demographic groups such as race and gender. Utilizing RNNs, specifically an attention-based Gated Recurrent Unit (GRU) with a self-attention mechanism, the model predicts course passing likelihood based on students' activity sequences.
    \citet{li_novel_2022} implemented a user profiling procedure that combines semantic behavior modeling with RNNs by proposing a methodology that enriches user behavior data with semantic information to enhance the accuracy of user profiling.

    \item \textbf{\textit{Long Short Term Memory}} (LSTM) represents a specialized form of recurrent neural network. LSTMs address the vanishing gradient problem, enabling effective modeling of long-term dependencies in sequential user data, leading to an improved understanding of user preferences over time (e.g.,~\citep{zhu_what_2017,zolna_user_2017,singh_user_2019,sharma_user_2020,cura_driver_2021,fazil_deepsbd_2021}).
    To cite specific relevant contributions, \citet{gu_hierarchical_2020} developed a hierarchical user profiling framework, whose core is the integration of Pyramid RNNs with a specialized component known as Behavior-LSTM. The Behavior-LSTM layer is particularly crucial as it models the temporal sequence of user behaviors, providing a dynamic and evolving representation of user interests.
    Another valuable LSTM-based framework, named Time Information Enhanced Personalized Search (PSTIE), has been proposed by \citet{ma_pstie_2020}. The PSTIE framework utilizes time-aware LSTM architectures to model the evolution of user interests over time. This allows the model to calculate both short-term and long-term query intent and document interest. The framework also includes techniques for re-ranking by combining these interest representations.
    \citet{sahoo_multiple_2021} presented a system for detecting fake news on Facebook by leveraging user profile features alongside news content features. The approach involves using LSTM to effectively process and learn from the user profile features, which are indicative of the credibility of the news shared.
    \citet{nkambou_learning_2023}

    \item \textbf{\textit{Transformers}} utilize attention mechanisms to capture contextual information bidirectionally. They succeed in natural language understanding, providing rich representations of user-generated text data for various applications in user modeling (e.g.,~\citep{huertas-garcia_profiling_2021,zhu_contrastive_2021}).
    A specific case is provided by \citet{avny_brosh_bruce_2022}, who introduced a neural network-based model, BRUCE, which focuses on personalized bundle recommendations. The main ability of BRUCE is modeling user preferences and the latent relationships between items within a bundle using a Transformer-based architecture. This architecture employs self-attention mechanisms to create contextualized item embeddings that reflect both the individual items' characteristics and their interactions within the bundle.
    The most important transformer-based model is BERT, which stands for Bidirectional Encoder Representations from Transformers (examples of BERT-based models in user modeling can be seen in \citet{gu_partner_2021} and \citet{kota_understanding_2021}).
    To provide some detailed instances, \citet{wu_userbert_2022} presented UserBERT, an approach for pre-training user models on unlabeled user behavior data. This method employs contrastive self-supervision to capture universal user information, which is crucial for enhancing personalization in various applications. UserBERT introduces two key self-supervision tasks: Masked Behavior Prediction, involving predicting user behaviors that have been intentionally masked in the input data, and Behavior Sequence Matching, required to determine whether two sequences of user behavior are from the same user, promoting the model's ability to discern patterns and similarities in user activities. To further refine the pre-training process, UserBERT incorporates a medium-hard negative sampling method. This technique selects negative samples that are neither too easy nor too difficult for the model to discriminate, providing an optimal challenge for learning.
    \citet{zheng_perd_2022} developed the Personalized Emoji Recommendation with Dynamic user preference (PERD) model. The core of PERD is a text encoder that utilizes a BERT model to capture the semantic representations of tweets. This allows the system to understand the context and content of user tweets effectively. Alongside the text encoder, the model incorporates a personalized attention mechanism. This mechanism is crucial as it sifts through a user's historical tweets to identify and emphasize those that are most informative for determining the user's preferences.
\end{itemize}

\subsubsection{Performance evaluation}\label{subsubsec:evaluation}
The evaluation phase is a critical step in any modeling problem as it allows for the assessment of a model's performance and its suitability for the given task. This phase is particularly important in user profiling, where standard evaluation measures are used to assess the performance of developed approaches. Common techniques for model evaluation include performance metrics like \textit{accuracy}, \textit{precision}, \textit{recall}, \textit{F-measure} (also known as \textit{F1 score}), \textit{mean reciprocal rank} (MRR), \textit{cosine similarity}, and \textit{Jaccard similarity} \citep{kaur_authcom_2018,costanzo_towards_2019,eke_survey_2019}.

Presently, there exist two principal approaches for assessing the efficacy of independent user profiling methods: (1)~relating them to \textit{classification tasks} and (2) adopting ad-hoc generated \textit{simulated data}.

\paragraph{Evaluation as a classification task}
This strategy adheres to conventional evaluation protocols and entails appraising the proposed model or method through its proficiency in a classification task. Typically, this method employs machine learning techniques to classify user profiles, discerning between authentic and non-authentic ones. By subjecting the model to this classification task, it becomes possible to gauge its effectiveness in accurately identifying and categorizing users based on their personal attributes. This process is essential in understanding the model's performance and reliability in the context of user profiling.
Clear recent demonstrations of user modeling methods assessed as a classification task can be found, for instance, in \citet{chen_catgcn_2021}, \citet{dai_say_2021}, \citet{yan_relation-aware_2021}, and \citet{purificato_graph_2022}.

\paragraph{Evaluation with simulated data}
The generation of simulated data is a means to curtail the volume of actual user data collected. This approach is designed not only to streamline the efficiency of profiling but also to uphold the privacy and confidentiality of users' personal information. By creating synthetic datasets that emulate the characteristics and patterns found in genuine user data, practitioners can significantly diminish the reliance on actual user information. This reduction in the dependence on real data serves a dual purpose: it mitigates potential privacy concerns by limiting the exposure of sensitive information, while concurrently ensuring that the profiling process remains effective and accurate.
A comprehensive methodology for simulating and evaluating user interactions within search sessions has been presented by \citet{zerhoudi_evaluating_2022}. The authors introduce the use of first-order Markov models and contextual Markov models as tools to replicate user search behavior. The realism of these simulated interactions is assessed using two main methods: Kolmogorov-Smirnov test and classification-based evaluation. The findings indicate that the simulated sessions closely resemble the real user log sessions, suggesting that the simulated interactions are representative of actual user behavior.
In \citet{keurulainen_amortised_2023}, the authors proposed an ``amortised experimental design'' for user models. The paper explores the use of Bayesian optimal experimental design and reinforcement learning to create a user model described as a simulation-based reinforcement learning model.

\vspace{0.1cm}
\hlbox{Take-home messages from building and modeling user profiles}{
\begin{itemize}
    \item \textbf{Variety of user modeling methodologies}: A range of methods for creating user profiles is explored, including rule-based modeling that uses predefined rules, ontology-based modeling for structured knowledge representation, and collaborative filtering for leveraging user community data to predict individual preferences.
    \item \textbf{Advanced analytical techniques in user modeling}: Highlights the use of statistical modeling, machine learning, and deep learning techniques. These methods are crucial for analyzing complex user data, identifying patterns, and making accurate predictions about user behavior, along with preferences, interests, and intentions.
    \item \textbf{Importance of the evaluation phase}: Emphasizes the essential role of the evaluation phase in user modeling. This phase assesses the performance and appropriateness of models, using standard measures to ensure their reliability and effectiveness in representing and predicting user behavior.
\end{itemize}
}

\subsection{Beyond-accuracy approaches}
The integration of advanced techniques surpassing mere accuracy represents a substantial paradigm shift across diverse domains, notably influencing user modeling and profiling. These methodologies go beyond the traditional emphasis on predictive precision, placing greater importance on core principles that also consider the ethical dimensions of user modeling, focusing on issues like privacy, transparency, and the responsible use of data. This broader view encourages the development of models that not only predict user actions accurately but also contribute positively to the user's overall interaction with a system, fostering a more holistic and user-centric approach to modeling.
Within this context, we identify relevant contributions belonging to three topics: \textit{explainable user modeling}, \textit{fair user modeling}, and \textit{privacy-preserving user modeling} (specifically referring to methods leveraging \textit{federated learning} method).

\subsubsection{Explainable user modeling}
\citet{balog_transparent_2019} introduced a set-based recommendation technique that emphasizes the importance of user models that are transparent, scrutable, and explainable. Transparency is achieved by providing users with insights into how their preferences are understood by the system and how the recommendation process works. Scrutability is incorporated by allowing users to directly and meaningfully revise their model. Explainability is provided to the level of user preferences, rather than just explaining why a given item was recommended, as commonly done in related works.
\citet{de_pauw_who_2022} proposed a recommender system, named TEASER, which is particularly geared towards enhancing the user experience by providing recommendations that are not only relevant but also accompanied by detailed explanations. The transparency and interpretability of TEASER stem from its ability to articulate the reasoning behind its suggestions.
\citet{guesmi_explaining_2022} focused on how transparency, user control, and personal characteristics influence the adoption of recommendation agents. The research introduces the Recommendation and Interest Modeling Application (RIMA), a tool designed to enhance transparency by providing explanations of user interest models at varying levels of detail. The study assesses the effectiveness of different explanation levels within the system and how they are perceived by users with diverse personal characteristics.
Other examples of works addressing explainability in user modeling can be seen in \citet{huang_explainable_2019}, \citet{wang_designing_2019}, \citet{chari_explanation_2020}, \citet{hase_evaluating_2020}, \citet{xian_ex3_2021}, \citet{minn_interpretable_2022}, and \citet{ding_interpretable_2023}.

\subsubsection{Fair user modeling}
\citet{dai_say_2021} designed FairGNN, a framework to ensure fairness in GNN-based user profiling (evaluated as a node classification task). The framework employs adversarial debiasing, where the adversary's goal is to ensure that the sensitive attributes cannot be accurately predicted from the node representations, thereby encouraging the GNN to learn fair representations. The theoretical analysis provided in the paper supports the effectiveness of this approach, showing that FairGNN can guarantee fair representations under specific conditions. To further promote fairness, a regularizer is added to the GNN classifier, and a covariance constraint is introduced to control the correlation between the sensitive attributes and the learned representations. These measures help to ensure that the predictions made by the GNN are not unfairly influenced by sensitive attributes.
Within the same context of GNN-based user profiling models, \citet{purificato_graph_2022} conducted a thorough investigation into the fairness of these models. The study specifically focuses on behavioral user profiling, which is a critical aspect of personalized services in various online platforms. The core of the research revolves around evaluating the fairness of these models in terms of how they might discriminate against certain user groups. To quantify fairness, the paper employs several disparate impact and disparate mistreatment metrics, utilizing two real-world datasets from e-commerce platforms in their experiments.
This work is followed up by the implementation of a unified framework, i.e., FairUP~\citep{abdelrazek_fairup_2023}, that allows researchers and practitioners to interact with state-of-the-art GNN-based models for user profiling and evaluate fairness on several graph data sources.
\citet{purificato_what_2023} addressed the topic of algorithmic fairness within the realm of user profiling. It identifies a gap in current fairness analysis, which is predominantly applied to binary classification problems and relies on the absolute difference in fairness metrics. This approach, the authors argue, can result in evaluations that do not accurately reflect the fairness of a system. To illustrate their point, the authors present a case study on the use of GNN-based models for user profiling. They discuss the limitations of current practices and the potential ethical consequences that arise from these limitations. The paper aims to spark conversation within the academic community about these issues and to encourage more comprehensive and nuanced analyses of fairness in algorithmic systems.
Another interesting framework for enhancing fairness in user modeling is proposed by \citet{zhang_fairlisa_2023}. The framework employs an adversarial learning approach that includes a filter module and a discriminator module to minimize the mutual information between user representations and sensitive attributes. FairLISA operates by combining data with both known and unknown sensitive attributes. It consists of two main components: a filter that aims to remove sensitive information from user representations and a discriminator that attempts to detect any remaining sensitive information.
Further contributions in this domain are \citet{shen_sar-net_2021}, \citet{chu_mitigating_2022}, and \citet{zheng_cbr_2022}.

\subsubsection{Privacy-preserving user modeling}
The most relevant recent works about privacy-preserving in user modeling employ \textit{federated learning} approaches.  
\citet{chu_mitigating_2022} introduced a methodology for predicting student performance in online learning environments. The core of the procedure is a personalized federated learning framework that allows for the creation of individualized models for student subgroups, derived from a global model that aggregates data across all students.
\citet{liu_federated_2023} proposed a federated learning technique for user modeling in environments characterized by inconsistent clients. The proposed approach, Hierarchical Personalized Federated Learning (HPFL), is designed to tackle the challenges of statistical heterogeneity, privacy heterogeneity, model heterogeneity, and quality heterogeneity that are prevalent in such settings.
\citet{zhang_dual_2023} presented the Personalized Federated Recommendation (PFedRec) framework, a system that leverages federated learning to create user profile-specific recommendation models. PFedRec is designed to be deployed on smart devices, enabling a decentralized and privacy-preserving method for capturing user preferences. The proposed federated learning approach is centered around a dual personalization mechanism that allows for fine-grained personalization at both the user and item levels. This is a significant departure from traditional federated learning methods, which often focus on a single model that is shared across all users. Instead, PFedRec formulates a unified federated optimization framework that learns individual lightweight models tailored to each user's unique preferences.
Privacy-preserving user modeling techniques have also been presented by \citet{wu_hierarchical_2021} and \citet{luo_personalized_2022}.

\vspace{0.1cm}
\hlbox{Take-home messages from beyond-accuracy approaches}{
\begin{itemize}
    \item \textbf{Ethical advancements in user modeling}: A shift towards advanced user modeling techniques emphasizes ethical principles, focusing on privacy, transparency, and responsible data use. This approach aims to enhance both the accuracy and the overall user-system interaction.
    \item \textbf{Three key beyond-accuracy approaches}:
    \begin{itemize}
        \item \textbf{Explainable user modeling} emphasizes the importance transparency and user understanding in modeling processes;
        \item \textbf{Fair user modeling} focuses on creating unbiased models that prevent discrimination against specific user groups;
        \item \textbf{Privacy-preserving user modeling} focuses on safeguarding privacy through techniques such as federated learning.
    \end{itemize}
\end{itemize}
}

\subsection{Applications}\label{subsec:applications}
User modeling and profiling techniques play a crucial role in various domains and applications, with different data being extracted and utilized based on the purpose and collection methods. User models, historically integral only to recommender systems and any kind of personalized platforms, are now widely employed across various applications.

\subsubsection{Recommender systems}
A recommender system, belonging to the category of information filtering systems, aims to predict item ratings or user preferences. The key components of a personalized recommender system include the efficient understanding of the user and recommending items aligned with their interests.
Recommender systems always benefited from user profiling methods~\citep{zimmerman_exposing_2002,middleton_ontological_2004,liu_user_2007,zhang_efficient_2007,berkovsky_mediation_2008,adomavicius_context-aware_2011,lakiotaki_multicriteria_2011,konstan_recommender_2012,liu_modeling_2015}.
Significant modern systems in this field naturally leverage the power of DL architectures (e.g.,~\citep{an_neural_2019,yu_adaptive_2019,jin_multi-behavior_2020,pan_learning_2020,xia_knowledge-enhanced_2021,han_multi-aggregator_2022,qi_news_2022,wang_attention-based_2022,xue_factorial_2022,cheng_multi-behavior_2023}
To provide specific examples, \citet{wu_hierarchical_2019} developed a user representation model that employs a neural architecture to enhance recommendations from reviews. The model operates by first using a sentence-level CNN to capture contextual information within sentences. It then applies a sentence-level attention network to weigh the importance of different sentences.
\citet{hu_graph_2020} implemented a news recommender system that adopts a graph convolutional network for modeling long-term user interests. In contrast, short-term user interests are captured using an attention-based LSTM model.
Another GNN-based recommender system, named Hierarchical User Intent Graph Network, has been provided by \citet{wei_hierarchical_2022}. The core idea is to leverage multi-level user intents and item representations to improve the accuracy of recommendations. The framework is composed of three main components: intra-level aggregation, inter-level aggregation, and interaction prediction.
The Core Attribute Evolution Network (CAEN), developed by \citet{ma_caen_2022}, incorporates a user behavior modeling method known as UB-GRUA (User Behavior with Gated Recurrent Unit Attention), which is part of the Personalized Attention Layer (PAL). This method is designed to capture and interpret user interactions with items, particularly under the influence of specific attributes. By employing a two-stage hierarchical attention network, CAEN can discern the significance of different attributes to individual users, leading to more accurate and personalized item recommendations.

Application domains covered include, among others, \textit{e-commerce} (e.g.,~\citep{lin_cross_2019,gu_hierarchical_2020,ma_caen_2022}), \textit{job recommendation} (e.g.,~\citep{rimitha_ontologies_2018,kota_understanding_2021}), and \textit{news recommendation} (e.g.,~\citep{natarajan_recommending_2016,an_neural_2019,wu_user-as-graph_2021,yang_going_2023}

\subsubsection{Information retrieval}
In the information retrieval context, user modeling research focuses particularly on personalized search, which is a process that prioritizes a user's interests when searching through a set of documents or web pages. This is achieved by creating a user profile based on the sites the user has searched for, thereby identifying their interests. The accuracy of the system is enhanced by predicting the user's interests more accurately than a typical search. In these personalized search systems, the search results are ranked not only based on the query information but also according to the user's interests. This approach ensures a more tailored and relevant search experience for the user.
To cite specific cases, \citet{zhou_enhancing_2020} proposed a personalized search model that aims to improve the re-finding behavior of users when they search for previously encountered information. The model leverages memory networks to capture two distinct types of re-finding behavior: query-based and document-based. It utilizes Recurrent Neural Networks (RNNs) to process sequential data and generate refined vectors that represent user intent over multiple sessions.
\citet{zhou_group_2021} introduced a group-based model that employs both search behavior and friend networks to enhance search results for users, particularly those with sparse historical data. By grouping users into friend circles and utilizing neural networks, the model creates a more accurate semantic representation of user interests, leading to significant improvements in search personalization.
\citet{deng_improving_2022} leveraged a dual-feedback network to enhance the understanding of user search intentions. The model is designed to improve personalized search by incorporating multi-granular user feedback, both positive and negative, to accurately model the user's current search intention.
Other works can be seen in \citet{du_folksonomy-based_2016}, \citet{ma_pstie_2020}, and \citet{zhou_pssl_2021}.

\subsubsection{Adaptive e-learning systems}
An adaptive e-learning system in the context of user modeling refers to an educational technology platform that leverages user modeling techniques to dynamically tailor the learning experience to the individual needs, preferences, and abilities of each learner. In such a system, the emphasis is on creating and updating a personalized user model that captures relevant information about the learner, allowing the platform to adapt various aspects of the e-learning process. Works in this area have always been proposed   (e.g.,~\citep{brusilovsky_knowledgetree_2004,brut_rule-based_2009,li_three-tier_2009,kim_personality_2013}).
Recently, \citet{kulkarni_user_2019} addressed the issue of delivering relevant and high-quality content to users within e-learning systems, aiming to minimize the time users spend searching for information. User profiles are created by combining explicit information provided by the user with implicit data gathered from their interactions with the system, resulting in a dynamic and adaptive profile.

Along with ethical principles, there has also been an increasing recognition of the imperative for accessibility in digital platforms, driven by a growing awareness of the diverse needs of users and a commitment to ensuring inclusive and equitable access to information and services. In this scenario, techniques were also developed to generate user profiles that are tailored to the specific accessibility needs of users with disabilities (e.g.,~\citep{sanchez-gordon_model_2021}).

The last few years have witnessed a significant surge in the popularity of \textit{Massive Open Online Courses} (MOOCs) within the realm of e-learning platforms, with the COVID-19 pandemic further accelerating their adoption as individuals sought flexible and accessible online learning options.
\citet{sunar_modelling_2020} investigated the role of social engagement in MOOCs and its correlation with course completion rates. The research specifically examines user behavior aiming to model and understand the patterns of social interactions among participants. The authors introduce the concept of ``behavior chains'' to model the social engagement of learners. The study reveals that learners who engage more deeply in peer interactions and contribute consistently over the duration of the course are more likely to complete it.
\citet{jin_mooc_2023} focused, instead, on the development of a predictive model to identify students at risk of dropping out of MOOCs by analyzing their learning behavior data. The model employs Support Vector Regression (SVR) as the core predictive algorithm due to its suitability for the task. To enhance the performance of the SVR, the paper introduces an Improved Quantum-behaved Particle Swarm Optimization (IQPSO) algorithm for optimizing its parameters.

\subsubsection{Fake news detection}
The innovative application of user modeling has seen a notable rise in the development of advanced approaches for detecting fake news, underscoring the growing importance of leveraging personalized user profiles to enhance the accuracy and effectiveness of misinformation detection in digital environments.
In the user modeling context, the majority of work analyzes social media platforms and social networks (e.g.,~\citep{shu_understanding_2018,monti_fake_2019,shu_role_2019}). A specific example is the paper by \citet{sahoo_multiple_2021}, where the authors presented a framework for detecting fake news on Facebook by leveraging user profile features alongside news content features. The system employs a combination of machine learning and deep learning techniques to analyze the behavior of Facebook accounts. By collecting data through a crawler and constructing datasets for classification, the study utilizes user profile information as a critical component in identifying fake news.
Concerning ethical implications in AI systems, particularly the risks of mislabeling news and unjust profiling, \citet{allein_preventing_2023} proposed an ethical fake news detection approach that incorporates user insights without relying on direct profiling, addressing the challenge of profiling-dependent fake news detection. The proposed algorithm is designed to leverage the social context of Twitter users during the training phase but excludes user information during the evaluation of news article veracity, thus avoiding decision-making based on profiling. The study employs a multimodal learning algorithm that uses a cross-modal loss function inspired by social sciences to analyze the relationship between news articles, their creators, and the audience.

\subsubsection{Cybersecurity}
The innovative application of user modeling has witnessed a surge in the integration of cybersecurity approaches, reflecting a heightened awareness of the crucial role personalized user profiles play in enhancing digital security and safeguarding against evolving cyber threats.
\citet{lashkari_survey_2019} presented a broad examination of user profiling as a critical component in cybersecurity, specifically for anomaly detection. The authors review existing user profiling models, highlighting their strengths and weaknesses, and propose a novel user profiling model designed to enhance the detection of suspicious activities in cyberspace.
\citet{addae_exploring_2019} investigated the factors that influence users' cybersecurity behaviors and the adoption of personalized adaptive cybersecurity measures. The study integrates insights from behavioral science, machine learning, and theoretical models such as the Technology Acceptance Model (TAM) and Protection Motivation Theory (PMT) to understand and predict user behavior in non-corporate environments.

\vspace{0.1cm}
\hlbox{Take-home messages from applications}{
\begin{itemize}
    \item \textbf{Wide-ranging applications}: User modeling and profiling techniques are crucial in a variety of domains. While initially integral to recommender systems and personalized platforms, these techniques are now employed in a broader range of applications, including adaptive e-learning systems, fake news detection, and cybersecurity.
    \item \textbf{Adaptation to diverse data and purposes}: These techniques involve extracting and utilizing different types of data based on the specific purpose and method of collection in each application.
\end{itemize}
}

\section{Discussion and Future Research Directions}\label{sec:discussion}
In this survey, we embarked on a comprehensive exploration of the evolving landscape of \textit{user modeling} and \textit{user profiling}.
Our journey commenced with a thorough historical overview, tracing the roots of user modeling to provide a contextual foundation for understanding its evolution.
Subsequently, we conducted an in-depth examination of existing surveys, critically analyzing their contributions and consolidating diverse perspectives to glean insights into the overarching developments.
The analysis of terminology emerged as a pivotal undertaking, revealing the semantic intricacies that permeate user modeling and user profiling literature. By dissecting and categorizing the terminological landscape, we aimed to foster a clearer understanding of the field's conceptual foundations, facilitating a more coherent discourse among researchers and practitioners alike.
As a culmination of this effort, we have ventured to contribute to the field by providing two novel definitions that encapsulate the essence of ``user model'' (or ``user profile'') and ``user modeling'' (or ``user profiling'').
A pivotal focus on paradigm shifts and new trends unveiled the transformative forces shaping user modeling methodologies. This section not only scrutinized historical transitions but also projected forward, anticipating emerging trends that could redefine the landscape.
Concluding our exploration, we present the current taxonomy of user modeling, synthesizing the diverse strands of research into a cohesive framework. This taxonomy serves as a guidepost for researchers, offering a structured lens through which to comprehend the evolving facets of user-centric modeling.

Despite the paradigm shifts that have characterized the evolution of user modeling and profiling, certain core topics have exhibited remarkable resilience, persisting as focal points throughout the discipline's history.
The continual examination of users' preferences, interests, and needs serves as a foundational element, exemplifying an enduring pursuit to comprehend the complexities of human behavior across diverse interactive contexts.
Statistical approaches, a traditional cornerstone of user modeling research, have weathered the transformative waves of technological advancements, highlighting their robustness in deciphering patterns within user data.
Moreover, the faithful applications of user modeling in recommender systems and personalized, adaptive interfaces underscore the enduring relevance of tailoring technological interactions to individual user characteristics.
As we navigate the currents of change, these enduring themes serve as anchors, guiding and grounding user modeling research, ensuring continuity of inquiry even in the face of evolving methodologies and technological landscapes.

With a future-oriented outlook, the principles of \textit{Human-Centered AI} (HCAI)~\citep{shneiderman_human_2022} and \textit{Responsible AI}~\citep{dignum_responsible_2019}, as well as the specific regulations for the development of trustworthy AI systems, such as the \textit{EU~AI~Act}\footnote{\url{https://www.europarl.europa.eu/RegData/etudes/BRIE/2021/698792/EPRS_BRI(2021)698792_EN.pdf}}, are gaining widespread attention for their potential to shape the future of technology in ways that serve human needs more effectively. These principles emphasize designing AI systems that support human autonomy, enhance human performance, and ensure that technology serves to empower people rather than replace them.
Building on these foundational concepts, research in user modeling will probably undergo a significant transformation, highlighting the importance of developing systems that are not only technologically advanced but also deeply aligned with human values and needs.
We can thus identify the following emerging directions:

\paragraph{Human-AI Collaboration}
The significance of human-AI collaboration is increasingly recognized as a fundamental aspect of the future of research in user modeling, highlighting the transformative potential of melding human intuition with AI's analytical prowess. Developing computational tools that act as interactive assistants exemplifies this collaborative approach, where AI systems serve as partners in the modeling process, aiding human modelers even in the absence of fully specified outcomes. This symbiotic relationship enables a more adaptable and responsive modeling process, attuned to the dynamic needs of users~\citep{celikok_modeling_2023}.
Among recent contributions, the work of \citet{gao_joint_2020} proposes an explainable AI framework aimed at achieving human-like communication in human-robot collaborations. By constructing a hierarchical mind model of the human user, their approach allows for the generation of explanations based on online Bayesian inference of the user's mental state, significantly enhancing collaboration performance and user perception of the robot.
Similarly, \citet{banerjee_system_2023} address the challenge of AI bots lacking personalization and understanding of user intent by developing a system where a human support agent collaborates with an AI agent to answer customer queries, showcasing the potential of human-AI collaboration to enhance user experience.
These examples underscore the critical role of human-AI collaboration in enhancing the adaptability, effectiveness, and ethical grounding of user modeling systems. By fostering a collaborative dynamic between humans and AI, the field of user modeling is set to advance towards developing technologies that are not only sophisticated and efficient but also deeply aligned with human values and needs.

\paragraph{Advanced User Models}
Focusing on the development of user models that regard the user as an active decision-maker engaged in a two-agent interaction with an AI collaborator highlights the need for understanding not only the user's goals but also their cognitive computational capacity~\citep{celikok_modeling_2023}.
This perspective is critical, as emphasized by \citet{nasrudin_systematic_2023}, who underscore the importance of models that accurately reflect the complex nature of human behavior by considering user goals, cognitive capacities, and perceptions, thereby impacting the sustainability of applications.
Additional contributions to this field include the work of \citet{jegou_computational_2018}, who propose a computational model to facilitate turn-taking behaviors in user-agent interactions, illustrating the significance of continuous decision-making based on both the agent's goals and the user's intentions.
Similarly, \citet{humann_modeling_2023} present a graphical user interface tool that optimizes multi-robot, multi-operator systems according to user preferences, serving as a testament to the importance of tailoring computational models to accommodate diverse user needs.
These references collectively underscore the evolving complexity and necessity of advanced user models that not only acknowledge the user as a key decision-maker but also adapt to their unique cognitive landscapes, thereby paving the way for more personalized and effective human-AI collaborations.

\paragraph{Ethical and Equitable User Modeling}
In line with Human-Centered AI (HCAI) principles, the ethical dimensions of user modeling, including data sovereignty and privacy, hold significant importance.
\citet{shadbolt_architectures_2020} discusses methodologies for ensuring data privacy and fostering individual autonomy within digital ecosystems, emphasizing a critical aspect of responsible AI development. This focus on ethical considerations is essential in the development and deployment of user modeling systems, ensuring that they align with broader societal values and individual rights.
Further exploration into the ethical dimensions of user modeling reveals a growing body of research dedicated to addressing these concerns. For instance, work by \citet{shahar_ethical_2021} on advanced user models in shared medical decision-making showcases the potential to address human cognitive limitations while ensuring ethical decision-making, highlighting the balance between AI assistance and human autonomy.
By focusing on privacy, autonomy, and transparency, the field can advance toward developing technologies that are advanced and efficient, as well as ethically grounded and aligned with human values and societal norms.

\paragraph{Integration of Cognitive Sciences}
Combining insights from cognitive sciences, psychology, and human-computer interaction, an interdisciplinary approach to user modeling is essential for developing more advanced and accurate user models~\citep{celikok_modeling_2023}.
This integration offers significant potential to enhance our understanding of user behavior, informing the development of systems that can adapt to human actions in a nuanced and personalized manner.
\citet{streicher_cognitive_2024}'s application of a Bayesian cognitive state modeling approach to adaptive educational serious games showcases how cognitive modeling can dynamically adapt in serious gaming environments, illustrating the impact of cognitive insights on user model effectiveness.
\citet{madsen_analytic_2019} discuss the use of agent-based models in cognitive sciences for predicting behaviors in dynamic, adaptive, and heterogeneous agents, highlighting the utility of agent-based models in scaling cognitive models within social networks and validating model predictions. This work serves as a crucial link between individual and socially oriented models.
Incorporating cognitive principles and methodologies allows researchers and developers to create user models that are not only more aligned with human thought processes and behaviors but also capable of fostering more engaging and effective human-machine interactions. This interdisciplinary approach is foundational for creating systems that truly understand and adapt to user needs in a personalized manner.

\paragraph{AI Systems Understanding User Goals}
Creating AI systems capable of understanding and adapting to user goals is pivotal for the advancement of responsive and supportive technologies~\citep{celikok_modeling_2023}.
This notion is reinforced by \citet{zeng_how_2021}'s exploration into human-centered intelligent gaming systems, which utilize machine learning algorithms to enhance player experiences by personalizing gameplay and understanding player motivations, underscores the significance of adapting AI systems to user goals. This approach not only enhances user engagement but also fosters the development of human-like characters and adaptive recommender systems, marking a step forward in creating more immersive and personalized interactive environments.
Moreover, the study by \citet{viros-i-martin_improving_2022} on AI assistants that adapt to designers' learning goals during design space exploration illustrates the potential benefits of AI systems that adjust to user preferences and objectives. Their research highlights the importance of AI systems in improving users' understanding of complex tasks, even though it may affect task performance due to changes in user interaction patterns.
Collectively, these insights underline the critical role of AI systems designed to understand and adapt to user goals. Such systems not only promise to revolutionize how we interact with technology but also pave the way for more personalized, intuitive, and effective human-machine collaborations.

\paragraph{Cross-Modal User Interaction}
Cross-modal user interaction has seen notable advancements, particularly in enhancing user experiences by bridging diverse modalities.
The development of a method for user-generalized cross-modal retrieval is a prime example of this progress, providing significant improvements in user experience across different interaction modes~\citep{ma_model_2022}. This advancement emphasizes the evolving landscape of user modeling research, where understanding and catering to the multifaceted needs and preferences of users through diverse sensory modalities are becoming increasingly important.
In the educational domain, implementing a cross-modal UX course for industrial design students, incorporating auditory and haptic techniques, aims to create inclusive and accessible user experiences. This initiative highlights the potential of cross-modal approaches in fostering a more comprehensive understanding of user interaction within user modeling research~\citep{temor_cross-modal_2022}.
Moreover, the exploration of ``cued-gaze'' with voice agents by \citet{jaber_cross-modal_2022} reveals that integrating visual cues with auditory instructions through cross-modality repair is more effective than speech reformulation alone. This finding emphasizes the importance of multimodal interactions in advancing user modeling, suggesting that user experiences can be significantly enriched by understanding and leveraging the interplay between different sensory modalities.
These advancements in Cross-Modal User Interaction directly contribute to the field of user modeling by offering new insights into how users interact with and perceive multimodal systems. By integrating diverse modalities, researchers can develop more nuanced and comprehensive models of user behavior, preferences, and needs. This opens new avenues for designing more accessible, intuitive, and engaging human-computer interfaces, accentuating the symbiotic relationship between cross-modal interaction and user modeling research.

\vspace{0.1cm}
\hlbox{Consideration}{These novel research directions in user modeling, underpinned by the principles of Human-Centered AI and Responsible AI, represent a significant shift towards developing technologies that are not only advanced but also deeply aligned with human values, needs, and ethical considerations. By focusing on these areas, the field of user modeling is poised to contribute to the creation of more personalized, efficient, and intuitive interactions between humans and technology, ultimately fostering a more humane and sustainable future.}

\bibliographystyle{plainnat}
\bibliography{references}

\begin{thebibliography}{435}
\providecommand{\natexlab}[1]{#1}
\providecommand{\url}[1]{\texttt{#1}}
\expandafter\ifx\csname urlstyle\endcsname\relax
  \providecommand{\doi}[1]{doi: #1}\else
  \providecommand{\doi}{doi: \begingroup \urlstyle{rm}\Url}\fi

\bibitem[Abdelrazek et~al.(2023)Abdelrazek, Purificato, Boratto, and
  De~Luca]{abdelrazek_fairup_2023}
Mohamed Abdelrazek, Erasmo Purificato, Ludovico Boratto, and Ernesto~William
  De~Luca.
\newblock {FairUP}: {A} {Framework} for {Fairness} {Analysis} of {Graph}
  {Neural} {Network}-{Based} {User} {Profiling} {Models}.
\newblock In \emph{Proceedings of the 46th {International} {ACM} {SIGIR}
  {Conference} on {Research} and {Development} in {Information} {Retrieval}},
  {SIGIR} '23, pages 3165--3169, New York, NY, USA, 2023. Association for
  Computing Machinery.
\newblock ISBN 978-1-4503-9408-6.
\newblock \doi{10.1145/3539618.3591814}.
\newblock URL \url{https://dl.acm.org/doi/10.1145/3539618.3591814}.

\bibitem[Abel et~al.(2011)Abel, Gao, Houben, and Tao]{abel_analyzing_2011}
Fabian Abel, Qi~Gao, Geert-Jan Houben, and Ke~Tao.
\newblock Analyzing {User} {Modeling} on {Twitter} for {Personalized} {News}
  {Recommendations}.
\newblock In Joseph~A. Konstan, Ricardo Conejo, José~L. Marzo, and Nuria
  Oliver, editors, \emph{User {Modeling}, {Adaption} and {Personalization}},
  Lecture {Notes} in {Computer} {Science}, pages 1--12, Berlin, Heidelberg,
  2011. Springer.
\newblock ISBN 978-3-642-22362-4.
\newblock \doi{10.1007/978-3-642-22362-4_1}.

\bibitem[Abri et~al.(2021)Abri, Abri, and Cetin]{abri_classification_2021}
Sara Abri, Rayan Abri, and Salih Cetin.
\newblock A {Classification} on {Different} {Aspects} of {User} {Modelling} in
  {Personalized} {Web} {Search}.
\newblock In \emph{Proceedings of the 4th {International} {Conference} on
  {Natural} {Language} {Processing} and {Information} {Retrieval}}, {NLPIR}
  '20, pages 194--199, New York, NY, USA, February 2021. Association for
  Computing Machinery.
\newblock ISBN 978-1-4503-7760-7.
\newblock \doi{10.1145/3443279.3443291}.
\newblock URL \url{https://dl.acm.org/doi/10.1145/3443279.3443291}.

\bibitem[Abu~Sulayman and Ouda(2019)]{abu_sulayman_user_2019}
Iman I.~M. Abu~Sulayman and Abdelkader Ouda.
\newblock User {Modeling} via {Anomaly} {Detection} {Techniques} for {User}
  {Authentication}.
\newblock In \emph{2019 {IEEE} 10th {Annual} {Information} {Technology},
  {Electronics} and {Mobile} {Communication} {Conference} ({IEMCON})}, pages
  0169--0176, Vancouver, BC, Canada, October 2019. IEEE.
\newblock ISBN 978-1-72812-530-5.
\newblock \doi{10.1109/IEMCON.2019.8936183}.
\newblock URL \url{https://ieeexplore.ieee.org/document/8936183/}.

\bibitem[Addae et~al.(2019)Addae, Sun, Towey, and
  Radenkovic]{addae_exploring_2019}
Joyce~H. Addae, Xu~Sun, Dave Towey, and Milena Radenkovic.
\newblock Exploring user behavioral data for adaptive cybersecurity.
\newblock \emph{User Modeling and User-Adapted Interaction}, 29\penalty0
  (3):\penalty0 701--750, July 2019.
\newblock ISSN 1573-1391.
\newblock \doi{10.1007/s11257-019-09236-5}.
\newblock URL \url{https://doi.org/10.1007/s11257-019-09236-5}.

\bibitem[Adomavicius and Tuzhilin(2001)]{adomavicius_expert-driven_2001}
Gediminas Adomavicius and Alexander Tuzhilin.
\newblock Expert-{Driven} {Validation} of {Rule}-{Based} {User} {Models} in
  {Personalization} {Applications}.
\newblock \emph{Data Mining and Knowledge Discovery}, 5\penalty0 (1):\penalty0
  33--58, January 2001.
\newblock ISSN 1573-756X.
\newblock \doi{10.1023/A:1009839827683}.
\newblock URL \url{https://doi.org/10.1023/A:1009839827683}.

\bibitem[Adomavicius and Tuzhilin(2011)]{adomavicius_context-aware_2011}
Gediminas Adomavicius and Alexander Tuzhilin.
\newblock Context-{Aware} {Recommender} {Systems}.
\newblock In Francesco Ricci, Lior Rokach, Bracha Shapira, and Paul~B. Kantor,
  editors, \emph{Recommender {Systems} {Handbook}}, pages 217--253. Springer
  US, Boston, MA, 2011.
\newblock ISBN 978-0-387-85820-3.
\newblock \doi{10.1007/978-0-387-85820-3_7}.
\newblock URL \url{https://doi.org/10.1007/978-0-387-85820-3_7}.

\bibitem[Agarwal et~al.(2022)Agarwal, Srivastava, Singh, and
  Rosenberg]{agarwal_modeling_2022}
Prabhat Agarwal, Manisha Srivastava, Vishwakarma Singh, and Charles Rosenberg.
\newblock Modeling {User} {Behavior} {With} {Interaction} {Networks} for {Spam}
  {Detection}.
\newblock In \emph{Proceedings of the 45th {International} {ACM} {SIGIR}
  {Conference} on {Research} and {Development} in {Information} {Retrieval}},
  pages 2437--2442, Madrid Spain, July 2022. ACM.
\newblock ISBN 978-1-4503-8732-3.
\newblock \doi{10.1145/3477495.3531875}.
\newblock URL \url{https://dl.acm.org/doi/10.1145/3477495.3531875}.

\bibitem[Aghasian et~al.(2017)Aghasian, Garg, Gao, Yu, and
  Montgomery]{aghasian_scoring_2017}
Erfan Aghasian, Saurabh Garg, Longxiang Gao, Shui Yu, and James Montgomery.
\newblock Scoring {Users}’ {Privacy} {Disclosure} {Across} {Multiple}
  {Online} {Social} {Networks}.
\newblock \emph{IEEE Access}, 5:\penalty0 13118--13130, 2017.
\newblock ISSN 2169-3536.
\newblock \doi{10.1109/ACCESS.2017.2720187}.
\newblock URL \url{https://ieeexplore.ieee.org/abstract/document/7959160}.

\bibitem[Agouti(2022)]{agouti_graph-based_2022}
Tarik Agouti.
\newblock Graph-based modeling using association rule mining to detect
  influential users in social networks.
\newblock \emph{Expert Systems with Applications}, 202:\penalty0 117436,
  September 2022.
\newblock ISSN 0957-4174.
\newblock \doi{10.1016/j.eswa.2022.117436}.
\newblock URL
  \url{https://www.sciencedirect.com/science/article/pii/S0957417422007722}.

\bibitem[Ahmad et~al.(2021)Ahmad, Rafie, and Ghorabie]{ahmad_spam_2021}
Saleh Beyt~Sheikh Ahmad, Mahnaz Rafie, and Seyed~Mojtaba Ghorabie.
\newblock Spam detection on {Twitter} using a support vector machine and
  users’ features by identifying their interactions.
\newblock \emph{Multimedia Tools and Applications}, 80\penalty0 (8):\penalty0
  11583--11605, March 2021.
\newblock ISSN 1573-7721.
\newblock \doi{10.1007/s11042-020-10405-7}.
\newblock URL \url{https://doi.org/10.1007/s11042-020-10405-7}.

\bibitem[Al-Shamri(2016)]{al-shamri_user_2016}
Mohammad Yahya~H. Al-Shamri.
\newblock User profiling approaches for demographic recommender systems.
\newblock \emph{Knowledge-Based Systems}, 100:\penalty0 175--187, May 2016.
\newblock ISSN 0950-7051.
\newblock \doi{10.1016/j.knosys.2016.03.006}.
\newblock URL
  \url{https://www.sciencedirect.com/science/article/pii/S0950705116001192}.

\bibitem[Alaoui et~al.(2015)Alaoui, Idrissi, and Ajhoun]{alaoui_building_2015}
Sara Alaoui, Younès El Bouzekri~El Idrissi, and Rachida Ajhoun.
\newblock Building {Rich} {User} {Profile} {Based} on {Intentional}
  {Perspective}.
\newblock \emph{Procedia Computer Science}, 73:\penalty0 342--349, 2015.
\newblock ISSN 18770509.
\newblock \doi{10.1016/j.procs.2015.12.002}.
\newblock URL
  \url{https://linkinghub.elsevier.com/retrieve/pii/S1877050915034638}.

\bibitem[Aleven(2010)]{aleven_rule-based_2010}
Vincent Aleven.
\newblock Rule-{Based} {Cognitive} {Modeling} for {Intelligent} {Tutoring}
  {Systems}.
\newblock In Roger Nkambou, Jacqueline Bourdeau, and Riichiro Mizoguchi,
  editors, \emph{Advances in {Intelligent} {Tutoring} {Systems}}, Studies in
  {Computational} {Intelligence}, pages 33--62. Springer, Berlin, Heidelberg,
  2010.
\newblock ISBN 978-3-642-14363-2.
\newblock \doi{10.1007/978-3-642-14363-2_3}.
\newblock URL \url{https://doi.org/10.1007/978-3-642-14363-2_3}.

\bibitem[Ali et~al.(2020)Ali, Gadallah, Hefny, and
  Novikov]{ali_integrated_2020}
Noaman~M. Ali, Ahmed~M. Gadallah, Hesham~A. Hefny, and Boris Novikov.
\newblock An {Integrated} {Framework} for {Web} {Data} {Preprocessing}
  {Towards} {Modeling} {User} {Behavior}.
\newblock In \emph{2020 {International} {Multi}-{Conference} on {Industrial}
  {Engineering} and {Modern} {Technologies} ({FarEastCon})}, pages 1--8,
  October 2020.
\newblock \doi{10.1109/FarEastCon50210.2020.9271467}.
\newblock URL \url{https://ieeexplore.ieee.org/abstract/document/9271467}.

\bibitem[Allein et~al.(2023)Allein, Moens, and
  Perrotta]{allein_preventing_2023}
Liesbeth Allein, Marie-Francine Moens, and Domenico Perrotta.
\newblock Preventing profiling for ethical fake news detection.
\newblock \emph{Information Processing \& Management}, 60\penalty0
  (2):\penalty0 103206, March 2023.
\newblock ISSN 0306-4573.
\newblock \doi{10.1016/j.ipm.2022.103206}.
\newblock URL
  \url{https://www.sciencedirect.com/science/article/pii/S0306457322003077}.

\bibitem[Allen et~al.(1998)Allen, Yaeckel, and Kania]{allen_internet_1998}
Cliff Allen, Beth Yaeckel, and Deborah Kania.
\newblock \emph{Internet world guide to one-to-one web marketing}.
\newblock John Wiley \& Sons, Inc., 1998.

\bibitem[Allgayer et~al.(1989)Allgayer, Harbusch, Kobsa, Reddig, Reithinger,
  and Schmauks]{allgayer_xtra_1989}
Jürgen Allgayer, Karin Harbusch, Alfred Kobsa, Carola Reddig, Norbert
  Reithinger, and Dagmar Schmauks.
\newblock {XTRA}: a natural-language access system to expert systems.
\newblock \emph{International Journal of Man-Machine Studies}, 31\penalty0
  (2):\penalty0 161--195, August 1989.
\newblock ISSN 0020-7373.
\newblock \doi{10.1016/0020-7373(89)90026-6}.
\newblock URL
  \url{https://www.sciencedirect.com/science/article/pii/0020737389900266}.

\bibitem[Amato and Straccia(1999)]{amato_user_1999}
Giuseppe Amato and Umberto Straccia.
\newblock User {Profile} {Modeling} and {Applications} to {Digital}
  {Libraries}.
\newblock In \emph{Research and {Advanced} {Technology} for {Digital}
  {Libraries}}, volume 1696, pages 184--197. Springer Berlin Heidelberg,
  Berlin, Heidelberg, 1999.
\newblock ISBN 978-3-540-66558-8 978-3-540-48155-3.
\newblock \doi{10.1007/3-540-48155-9_13}.
\newblock URL \url{http://link.springer.com/10.1007/3-540-48155-9_13}.

\bibitem[Amatriain et~al.(2009)Amatriain, Pujol, Tintarev, and
  Oliver]{amatriain_rate_2009}
Xavier Amatriain, Josep~M. Pujol, Nava Tintarev, and Nuria Oliver.
\newblock Rate it again: increasing recommendation accuracy by user re-rating.
\newblock In \emph{Proceedings of the third {ACM} conference on {Recommender}
  systems}, {RecSys} '09, pages 173--180, New York, NY, USA, 2009. Association
  for Computing Machinery.
\newblock ISBN 978-1-60558-435-5.
\newblock \doi{10.1145/1639714.1639744}.
\newblock URL \url{https://dl.acm.org/doi/10.1145/1639714.1639744}.

\bibitem[An et~al.(2019)An, Wu, Wu, Zhang, Liu, and Xie]{an_neural_2019}
Mingxiao An, Fangzhao Wu, Chuhan Wu, Kun Zhang, Zheng Liu, and Xing Xie.
\newblock Neural {News} {Recommendation} with {Long}- and {Short}-term {User}
  {Representations}.
\newblock In Anna Korhonen, David Traum, and Lluís Màrquez, editors,
  \emph{Proceedings of the 57th {Annual} {Meeting} of the {Association} for
  {Computational} {Linguistics}}, pages 336--345, Florence, Italy, July 2019.
  Association for Computational Linguistics.
\newblock \doi{10.18653/v1/P19-1033}.
\newblock URL \url{https://aclanthology.org/P19-1033}.

\bibitem[Anelli et~al.(2021)Anelli, Di~Noia, Di~Sciascio, Ferrara, and
  Mancino]{anelli_sparse_2021}
Vito~Walter Anelli, Tommaso Di~Noia, Eugenio Di~Sciascio, Antonio Ferrara, and
  Alberto Carlo~Maria Mancino.
\newblock Sparse {Feature} {Factorization} for {Recommender} {Systems} with
  {Knowledge} {Graphs}.
\newblock In \emph{Fifteenth {ACM} {Conference} on {Recommender} {Systems}},
  pages 154--165, Amsterdam Netherlands, September 2021. ACM.
\newblock ISBN 978-1-4503-8458-2.
\newblock \doi{10.1145/3460231.3474243}.
\newblock URL \url{https://dl.acm.org/doi/10.1145/3460231.3474243}.

\bibitem[Angeline(2013)]{angeline_association_2013}
D.~Magdalene~Delighta Angeline.
\newblock Association {Rule} {Generation} for {Student} {Performance}
  {Analysis} using {Apriori} {Algorithm}.
\newblock \emph{The SIJ Transactions on Computer Science Engineering \& its
  Applications (CSEA)}, 01\penalty0 (01):\penalty0 16--20, April 2013.
\newblock ISSN 23212373, 23212381.
\newblock \doi{10.9756/SIJCSEA/V1I1/01010252}.
\newblock URL
  \url{http://www.thesij.com/TableOfContentArticleDetails.aspx?JournalID=1&IssueID=29}.

\bibitem[Ardissono and Sestero(1995)]{ardissono_using_1995}
Liliana Ardissono and Dario Sestero.
\newblock Using dynamic user models in the recognition of the plans of the
  user.
\newblock \emph{User Modeling and User-Adapted Interaction}, 5\penalty0
  (2):\penalty0 157--190, June 1995.
\newblock ISSN 1573-1391.
\newblock \doi{10.1007/BF01099760}.
\newblock URL \url{https://doi.org/10.1007/BF01099760}.

\bibitem[Ariannezhad et~al.(2023)Ariannezhad, Li, Schelter, and
  De~Rijke]{ariannezhad_personalized_2023}
Mozhdeh Ariannezhad, Ming Li, Sebastian Schelter, and Maarten De~Rijke.
\newblock A {Personalized} {Neighborhood}-based {Model} for {Within}-basket
  {Recommendation} in {Grocery} {Shopping}.
\newblock In \emph{Proceedings of the {Sixteenth} {ACM} {International}
  {Conference} on {Web} {Search} and {Data} {Mining}}, pages 87--95, Singapore
  Singapore, February 2023. ACM.
\newblock ISBN 978-1-4503-9407-9.
\newblock \doi{10.1145/3539597.3570417}.
\newblock URL \url{https://dl.acm.org/doi/10.1145/3539597.3570417}.

\bibitem[Aroyo and Houben(2010)]{aroyo_user_2010}
Lora Aroyo and Geert-Jan Houben.
\newblock User modeling and adaptive {Semantic} {Web}.
\newblock \emph{Semantic Web}, 1\penalty0 (1-2):\penalty0 105--110, January
  2010.
\newblock ISSN 1570-0844.
\newblock \doi{10.3233/SW-2010-0006}.
\newblock URL \url{https://content.iospress.com/articles/semantic-web/sw006}.

\bibitem[Avny~Brosh et~al.(2022)Avny~Brosh, Livne, Sar~Shalom, Shapira, and
  Last]{avny_brosh_bruce_2022}
Tzoof Avny~Brosh, Amit Livne, Oren Sar~Shalom, Bracha Shapira, and Mark Last.
\newblock {BRUCE}: {Bundle} {Recommendation} {Using} {Contextualized} item
  {Embeddings}.
\newblock In \emph{Proceedings of the 16th {ACM} {Conference} on {Recommender}
  {Systems}}, pages 237--245, Seattle WA USA, September 2022. ACM.
\newblock ISBN 978-1-4503-9278-5.
\newblock \doi{10.1145/3523227.3546754}.
\newblock URL \url{https://dl.acm.org/doi/10.1145/3523227.3546754}.

\bibitem[Bakalov et~al.(2010)Bakalov, König-Ries, Nauerz, and
  Welsch]{bakalov_introspectiveviews_2010}
Fedor Bakalov, Birgitta König-Ries, Andreas Nauerz, and Martin Welsch.
\newblock {IntrospectiveViews}: {An} {Interface} for {Scrutinizing} {Semantic}
  {User} {Models}.
\newblock In Paul De~Bra, Alfred Kobsa, and David Chin, editors, \emph{User
  {Modeling}, {Adaptation}, and {Personalization}}, Lecture {Notes} in
  {Computer} {Science}, pages 219--230, Berlin, Heidelberg, 2010. Springer.
\newblock ISBN 978-3-642-13470-8.
\newblock \doi{10.1007/978-3-642-13470-8_21}.

\bibitem[Balog and Zhai(2023)]{balog_user_2023}
Krisztian Balog and ChengXiang Zhai.
\newblock User {Simulation} for {Evaluating} {Information} {Access} {Systems}.
\newblock In \emph{Proceedings of the {Annual} {International} {ACM} {SIGIR}
  {Conference} on {Research} and {Development} in {Information} {Retrieval} in
  the {Asia} {Pacific} {Region}}, {SIGIR}-{AP} '23, pages 302--305, New York,
  NY, USA, November 2023. Association for Computing Machinery.
\newblock ISBN 9798400704086.
\newblock \doi{10.1145/3624918.3629549}.
\newblock URL \url{https://dl.acm.org/doi/10.1145/3624918.3629549}.

\bibitem[Balog et~al.(2007)Balog, Bogers, Azzopardi, de~Rijke, and van~den
  Bosch]{balog_broad_2007}
Krisztian Balog, Toine Bogers, Leif Azzopardi, Maarten de~Rijke, and Antal
  van~den Bosch.
\newblock Broad expertise retrieval in sparse data environments.
\newblock In \emph{Proceedings of the 30th annual international {ACM} {SIGIR}
  conference on {Research} and development in information retrieval}, {SIGIR}
  '07, pages 551--558, New York, NY, USA, 2007. Association for Computing
  Machinery.
\newblock ISBN 978-1-59593-597-7.
\newblock \doi{10.1145/1277741.1277836}.
\newblock URL \url{https://dl.acm.org/doi/10.1145/1277741.1277836}.

\bibitem[Balog et~al.(2012)Balog, Fang, De~Rijke, Serdyukov, Si,
  et~al.]{balog_expertise_2012}
Krisztian Balog, Yi~Fang, Maarten De~Rijke, Pavel Serdyukov, Luo Si, et~al.
\newblock Expertise retrieval.
\newblock \emph{Foundations and Trends{\textregistered} in Information
  Retrieval}, 6\penalty0 (2--3):\penalty0 127--256, 2012.

\bibitem[Balog et~al.(2019)Balog, Radlinski, and
  Arakelyan]{balog_transparent_2019}
Krisztian Balog, Filip Radlinski, and Shushan Arakelyan.
\newblock Transparent, {Scrutable} and {Explainable} {User} {Models} for
  {Personalized} {Recommendation}.
\newblock In \emph{Proceedings of the 42nd {International} {ACM} {SIGIR}
  {Conference} on {Research} and {Development} in {Information} {Retrieval}},
  pages 265--274, Paris France, July 2019. ACM.
\newblock ISBN 978-1-4503-6172-9.
\newblock \doi{10.1145/3331184.3331211}.
\newblock URL \url{https://dl.acm.org/doi/10.1145/3331184.3331211}.

\bibitem[Banerjee et~al.(2023)Banerjee, Poser, Wiethof, Subramanian, Paucar,
  Bittner, and Biemann]{banerjee_system_2023}
Debayan Banerjee, Mathis Poser, Christina Wiethof, Varun~Shankar Subramanian,
  Richard Paucar, Eva A.~C. Bittner, and Chris Biemann.
\newblock A {System} for {Human}-{AI} collaboration for {Online} {Customer}
  {Support}, February 2023.
\newblock URL \url{http://arxiv.org/abs/2301.12158}.
\newblock arXiv:2301.12158 [cs].

\bibitem[Barforoush et~al.(2017)Barforoush, Shirazi, and
  Emami]{barforoush_new_2017}
Ahmad Barforoush, Hossein Shirazi, and Hojjat Emami.
\newblock A {New} {Classification} {Framework} to {Evaluate} the {Entity}
  {Profiling} on the {Web}: {Past}, {Present} and {Future}.
\newblock \emph{ACM Computing Surveys}, 50\penalty0 (3):\penalty0 39:1--39:39,
  June 2017.
\newblock ISSN 0360-0300.
\newblock \doi{10.1145/3066904}.
\newblock URL \url{https://dl.acm.org/doi/10.1145/3066904}.

\bibitem[Barnett et~al.(2015)Barnett, Pearson, Pearson, and
  Kellermanns]{barnett_five-factor_2015}
Tim Barnett, Allison~W Pearson, Rodney Pearson, and Franz~W Kellermanns.
\newblock Five-factor model personality traits as predictors of perceived and
  actual usage of technology.
\newblock \emph{European Journal of Information Systems}, 24\penalty0
  (4):\penalty0 374--390, July 2015.
\newblock ISSN 1476-9344.
\newblock \doi{10.1057/ejis.2014.10}.
\newblock URL \url{https://doi.org/10.1057/ejis.2014.10}.

\bibitem[Barua et~al.(2014)Barua, Kay, Kummerfeld, and
  Paris]{barua_modelling_2014}
Debjanee Barua, Judy Kay, Bob Kummerfeld, and Cécile Paris.
\newblock Modelling {Long} {Term} {Goals}.
\newblock In Vania Dimitrova, Tsvi Kuflik, David Chin, Francesco Ricci, Peter
  Dolog, and Geert-Jan Houben, editors, \emph{User {Modeling}, {Adaptation},
  and {Personalization}}, Lecture {Notes} in {Computer} {Science}, pages 1--12,
  Cham, 2014. Springer International Publishing.
\newblock ISBN 978-3-319-08786-3.
\newblock \doi{10.1007/978-3-319-08786-3_1}.

\bibitem[Bedi et~al.(2022)Bedi, Goyal, Rajawat, Shaw, and
  Ghosh]{bedi_framework_2022}
Pradeep Bedi, S.~B. Goyal, Anand~Singh Rajawat, Rabindra~Nath Shaw, and Ankush
  Ghosh.
\newblock A {Framework} for {Personalizing} {Atypical} {Web} {Search}
  {Sessions} with {Concept}-{Based} {User} {Profiles} {Using} {Selective}
  {Machine} {Learning} {Techniques}.
\newblock In Monica Bianchini, Vincenzo Piuri, Sanjoy Das, and Rabindra~Nath
  Shaw, editors, \emph{Advanced {Computing} and {Intelligent} {Technologies}},
  Lecture {Notes} in {Networks} and {Systems}, pages 279--291, Singapore, 2022.
  Springer.
\newblock ISBN 9789811621642.
\newblock \doi{10.1007/978-981-16-2164-2_23}.

\bibitem[Ben~Hassen et~al.(2022)Ben~Hassen, Ben~Ticha, and
  Chaibi]{ben_hassen_deep_2022}
Aymen Ben~Hassen, Sonia Ben~Ticha, and Anja~Habacha Chaibi.
\newblock Deep {Learning} for {Visual}-{Features} {Extraction} {Based}
  {Personalized} {User} {Modeling}.
\newblock \emph{SN Computer Science}, 3\penalty0 (4):\penalty0 261, April 2022.
\newblock ISSN 2661-8907.
\newblock \doi{10.1007/s42979-022-01131-y}.
\newblock URL \url{https://doi.org/10.1007/s42979-022-01131-y}.

\bibitem[Benevenuto et~al.(2009)Benevenuto, Rodrigues, Cha, and
  Almeida]{benevenuto_characterizing_2009}
Fabrício Benevenuto, Tiago Rodrigues, Meeyoung Cha, and Virgílio Almeida.
\newblock Characterizing user behavior in online social networks.
\newblock In \emph{Proceedings of the 9th {ACM} {SIGCOMM} conference on
  {Internet} measurement}, {IMC} '09, pages 49--62, New York, NY, USA, November
  2009. Association for Computing Machinery.
\newblock ISBN 978-1-60558-771-4.
\newblock \doi{10.1145/1644893.1644900}.
\newblock URL \url{https://dl.acm.org/doi/10.1145/1644893.1644900}.

\bibitem[Benmakrelouf et~al.(2015)Benmakrelouf, Mezghani, and
  Kara]{benmakrelouf_towards_2015}
Souhila Benmakrelouf, Neila Mezghani, and Nadjia Kara.
\newblock Towards the {Identification} of {Players}' {Profiles} {Using}
  {Game}'s {Data} {Analysis} {Based} on {Regression} {Model} and {Clustering}.
\newblock In \emph{Proceedings of the 2015 {IEEE}/{ACM} {International}
  {Conference} on {Advances} in {Social} {Networks} {Analysis} and {Mining}
  2015}, {ASONAM} '15, pages 1403--1410, New York, NY, USA, 2015. Association
  for Computing Machinery.
\newblock ISBN 978-1-4503-3854-7.
\newblock \doi{10.1145/2808797.2809429}.
\newblock URL \url{https://dl.acm.org/doi/10.1145/2808797.2809429}.

\bibitem[Berkovsky et~al.(2008)Berkovsky, Kuflik, and
  Ricci]{berkovsky_mediation_2008}
Shlomo Berkovsky, Tsvi Kuflik, and Francesco Ricci.
\newblock Mediation of user models for enhanced personalization in recommender
  systems.
\newblock \emph{User Modeling and User-Adapted Interaction}, 18\penalty0
  (3):\penalty0 245--286, August 2008.
\newblock ISSN 1573-1391.
\newblock \doi{10.1007/s11257-007-9042-9}.
\newblock URL \url{https://doi.org/10.1007/s11257-007-9042-9}.

\bibitem[Berkovsky et~al.(2019)Berkovsky, Taib, Koprinska, Wang, Zeng, Li, and
  Kleitman]{berkovsky_detecting_2019}
Shlomo Berkovsky, Ronnie Taib, Irena Koprinska, Eileen Wang, Yucheng Zeng,
  Jingjie Li, and Sabina Kleitman.
\newblock Detecting {Personality} {Traits} {Using} {Eye}-{Tracking} {Data}.
\newblock In \emph{Proceedings of the 2019 {CHI} {Conference} on {Human}
  {Factors} in {Computing} {Systems}}, {CHI} '19, pages 1--12, New York, NY,
  USA, May 2019. Association for Computing Machinery.
\newblock ISBN 978-1-4503-5970-2.
\newblock \doi{10.1145/3290605.3300451}.
\newblock URL \url{https://dl.acm.org/doi/10.1145/3290605.3300451}.

\bibitem[Bhogi et~al.(2019)Bhogi, Singla, and Bose]{bhogi_user_2019}
Suleep~Kumar Bhogi, Kushal Singla, and Joy Bose.
\newblock User {Profile} in {Absence} of {Ground} {Truth} for {Mobile} {Users}.
\newblock In \emph{2019 {IEEE} 13th {International} {Conference} on {Semantic}
  {Computing} ({ICSC})}, pages 170--173, January 2019.
\newblock \doi{10.1109/ICOSC.2019.8665580}.
\newblock URL \url{https://ieeexplore.ieee.org/document/8665580}.

\bibitem[Bi et~al.(2016)Bi, Xu, Wang, and Zhou]{bi_anomaly_2016}
Meng Bi, Jian Xu, Mo~Wang, and Fucai Zhou.
\newblock Anomaly detection model of user behavior based on principal component
  analysis.
\newblock \emph{Journal of Ambient Intelligence and Humanized Computing},
  7\penalty0 (4):\penalty0 547--554, August 2016.
\newblock ISSN 1868-5145.
\newblock \doi{10.1007/s12652-015-0341-4}.
\newblock URL \url{https://doi.org/10.1007/s12652-015-0341-4}.

\bibitem[Bian et~al.(2021)Bian, Zhao, Zhou, Cai, He, Yin, and
  Wen]{bian_contrastive_2021}
Shuqing Bian, Wayne~Xin Zhao, Kun Zhou, Jing Cai, Yancheng He, Cunxiang Yin,
  and Ji-Rong Wen.
\newblock Contrastive {Curriculum} {Learning} for {Sequential} {User}
  {Behavior} {Modeling} via {Data} {Augmentation}.
\newblock In \emph{Proceedings of the 30th {ACM} {International} {Conference}
  on {Information} \& {Knowledge} {Management}}, pages 3737--3746, Virtual
  Event Queensland Australia, October 2021. ACM.
\newblock ISBN 978-1-4503-8446-9.
\newblock \doi{10.1145/3459637.3481905}.
\newblock URL \url{https://dl.acm.org/doi/10.1145/3459637.3481905}.

\bibitem[Biswas and Robinson(2010)]{biswas_brief_2010}
Pradipta Biswas and Peter Robinson.
\newblock A brief survey on user modelling in {HCI}.
\newblock In \emph{Proceedings of the {International} {Conference} on
  {Intelligent} {Human} {Computer} {Interaction} ({IHCI})}, 2010.

\bibitem[Boratto and Carta(2014)]{boratto_modeling_2014}
Ludovico Boratto and Salvatore Carta.
\newblock Modeling the {Preferences} of a {Group} of {Users} {Detected} by
  {Clustering}: a {Group} {Recommendation} {Case}-{Study}.
\newblock In \emph{Proceedings of the 4th {International} {Conference} on {Web}
  {Intelligence}, {Mining} and {Semantics} ({WIMS14})}, {WIMS} '14, pages 1--7,
  New York, NY, USA, 2014. Association for Computing Machinery.
\newblock ISBN 978-1-4503-2538-7.
\newblock \doi{10.1145/2611040.2611073}.
\newblock URL \url{https://dl.acm.org/doi/10.1145/2611040.2611073}.

\bibitem[Boratto and Carta(2015)]{boratto_art_2015}
Ludovico Boratto and Salvatore Carta.
\newblock {ART}: group recommendation approaches for automatically detected
  groups.
\newblock \emph{International Journal of Machine Learning and Cybernetics},
  6\penalty0 (6):\penalty0 953--980, December 2015.
\newblock ISSN 1868-808X.
\newblock \doi{10.1007/s13042-015-0371-4}.
\newblock URL \url{https://doi.org/10.1007/s13042-015-0371-4}.

\bibitem[Brajnik and Tasso(1994)]{brajnik_shell_1994}
Giorgio Brajnik and Carlo Tasso.
\newblock A shell for developing non-monotonic user modeling systems.
\newblock \emph{International Journal of Human-Computer Studies}, 40\penalty0
  (1):\penalty0 31--62, January 1994.
\newblock ISSN 1071-5819.
\newblock \doi{10.1006/ijhc.1994.1003}.
\newblock URL
  \url{https://www.sciencedirect.com/science/article/pii/S1071581984710032}.

\bibitem[Braynov(2004)]{braynov_personalization_2004}
Sviatoslav Braynov.
\newblock Personalization and {Customization} {Technologies}.
\newblock In Hossein Bidgoli, editor, \emph{The {Internet} {Encyclopedia}}.
  Wiley, 1 edition, January 2004.
\newblock ISBN 978-0-471-22201-9 978-0-471-48296-3.
\newblock \doi{10.1002/047148296X.tie141}.
\newblock URL
  \url{https://onlinelibrary.wiley.com/doi/10.1002/047148296X.tie141}.

\bibitem[Brusilovski et~al.(2007)Brusilovski, Kobsa, and
  Nejdl]{brusilovski_adaptive_2007}
Peter Brusilovski, Alfred Kobsa, and Wolfgang Nejdl.
\newblock \emph{The {Adaptive} {Web}: {Methods} and {Strategies} of {Web}
  {Personalization}}.
\newblock Springer Science \& Business Media, April 2007.
\newblock ISBN 978-3-540-72078-2.

\bibitem[Brusilovsky(2001)]{brusilovsky_adaptive_2001}
Peter Brusilovsky.
\newblock Adaptive {Hypermedia}.
\newblock \emph{User Modeling and User-Adapted Interaction}, 11\penalty0
  (1):\penalty0 87--110, March 2001.
\newblock ISSN 1573-1391.
\newblock \doi{10.1023/A:1011143116306}.
\newblock URL \url{https://doi.org/10.1023/A:1011143116306}.

\bibitem[Brusilovsky(2004)]{brusilovsky_knowledgetree_2004}
Peter Brusilovsky.
\newblock {KnowledgeTree}: a distributed architecture for adaptive e-learning.
\newblock In \emph{Proceedings of the 13th international {World} {Wide} {Web}
  conference on {Alternate} track papers \& posters - {WWW} {Alt}. '04}, page
  104, New York, NY, USA, 2004. ACM Press.
\newblock ISBN 978-1-58113-912-9.
\newblock \doi{10.1145/1013367.1013386}.
\newblock URL \url{http://portal.acm.org/citation.cfm?doid=1013367.1013386}.

\bibitem[Brusilovsky and Maybury(2002)]{brusilovsky_adaptive_2002}
Peter Brusilovsky and Mark~T Maybury.
\newblock From adaptive hypermedia to the adaptive web.
\newblock \emph{Communications of the ACM}, 45\penalty0 (5):\penalty0 30--33,
  2002.

\bibitem[Brusilovsky and Millán(2007)]{brusilovsky_user_2007}
Peter Brusilovsky and Eva Millán.
\newblock User {Models} for {Adaptive} {Hypermedia}
  and {Adaptive} {Educational} {Systems}.
\newblock In Peter Brusilovsky, Alfred Kobsa, and Wolfgang Nejdl, editors,
  \emph{The {Adaptive} {Web}: {Methods} and {Strategies} of {Web}
  {Personalization}}, Lecture {Notes} in {Computer} {Science}, pages 3--53.
  Springer, Berlin, Heidelberg, 2007.
\newblock ISBN 978-3-540-72079-9.
\newblock URL \url{https://doi.org/10.1007/978-3-540-72079-9_1}.

\bibitem[Brusilovsky et~al.(1998)Brusilovsky, Kobsa, and
  Vassileva]{brusilovsky_adaptive_1998}
Peter Brusilovsky, Alfred Kobsa, and Julita Vassileva, editors.
\newblock \emph{Adaptive {Hypertext} and {Hypermedia}}.
\newblock Springer Netherlands, Dordrecht, 1998.
\newblock ISBN 978-90-481-4944-5 978-94-017-0617-9.
\newblock URL \url{https://link.springer.com/10.1007/978-94-017-0617-9}.

\bibitem[Brut et~al.(2009)Brut, Asandului, and Grigoras]{brut_rule-based_2009}
Mihaela Brut, Laura Asandului, and Gheorghe Grigoras.
\newblock A {Rule}-{Based} {Approach} for {Developing} a
  {Competency}-{Oriented} {User} {Model} for {E}-{Learning} {Systems}.
\newblock In \emph{2009 {Fourth} {International} {Conference} on {Internet} and
  {Web} {Applications} and {Services}}, pages 555--560, May 2009.
\newblock \doi{10.1109/ICIW.2009.90}.

\bibitem[Burgos et~al.(2018)Burgos, Campanario, Peña, Lara, Lizcano, and
  Martínez]{burgos_data_2018}
Concepción Burgos, María~L. Campanario, David de~la Peña, Juan~A. Lara,
  David Lizcano, and María~A. Martínez.
\newblock Data mining for modeling students’ performance: {A} tutoring action
  plan to prevent academic dropout.
\newblock \emph{Computers \& Electrical Engineering}, 66:\penalty0 541--556,
  February 2018.
\newblock ISSN 0045-7906.
\newblock \doi{10.1016/j.compeleceng.2017.03.005}.
\newblock URL
  \url{https://www.sciencedirect.com/science/article/pii/S0045790617305220}.

\bibitem[Böhmer and Rinderle-Ma(2020)]{bohmer_mining_2020}
Kristof Böhmer and Stefanie Rinderle-Ma.
\newblock Mining association rules for anomaly detection in dynamic process
  runtime behavior and explaining the root cause to users.
\newblock \emph{Information Systems}, 90, May 2020.
\newblock ISSN 0306-4379.
\newblock \doi{10.1016/j.is.2019.101438}.
\newblock URL
  \url{https://www.sciencedirect.com/science/article/pii/S0306437919304909}.

\bibitem[Caglayan et~al.(1997)Caglayan, Snorrason, Jacoby, Mazzu, Jones, and
  Kumar]{caglayan_learn_1997}
Alper Caglayan, Magnus Snorrason, Jennifer Jacoby, James Mazzu, Robin Jones,
  and Krishna Kumar.
\newblock Learn sesame a learning agent engine.
\newblock \emph{Applied Artificial Intelligence}, 11\penalty0 (5):\penalty0
  393--412, July 1997.
\newblock ISSN 0883-9514.
\newblock \doi{10.1080/088395197118109}.
\newblock URL \url{https://doi.org/10.1080/088395197118109}.

\bibitem[Calegari and Pasi(2010)]{calegari_ontology-based_2010}
Silvia Calegari and Gabriella Pasi.
\newblock Ontology-{Based} {Information} {Behaviour} to {Improve} {Web}
  {Search}.
\newblock \emph{Future Internet}, 2\penalty0 (4):\penalty0 533--558, December
  2010.
\newblock ISSN 1999-5903.
\newblock \doi{10.3390/fi2040533}.
\newblock URL \url{https://www.mdpi.com/1999-5903/2/4/533}.

\bibitem[Cami et~al.(2019)Cami, Hassanpour, and Mashayekhi]{cami_user_2019}
Bagher~Rahimpour Cami, Hamid Hassanpour, and Hoda Mashayekhi.
\newblock User preferences modeling using dirichlet process mixture model for a
  content-based recommender system.
\newblock \emph{Knowledge-Based Systems}, 163:\penalty0 644--655, January 2019.
\newblock ISSN 09507051.
\newblock \doi{10.1016/j.knosys.2018.09.028}.
\newblock URL
  \url{https://linkinghub.elsevier.com/retrieve/pii/S0950705118304799}.

\bibitem[Cao et~al.(2022)Cao, Zhou, Feng, Huang, Xiao, Chen, and
  Chen]{cao_sampling_2022}
Yue Cao, Xiaojiang Zhou, Jiaqi Feng, Peihao Huang, Yao Xiao, Dayao Chen, and
  Sheng Chen.
\newblock Sampling {Is} {All} {You} {Need} on {Modeling} {Long}-{Term} {User}
  {Behaviors} for {CTR} {Prediction}.
\newblock In \emph{Proceedings of the 31st {ACM} {International} {Conference}
  on {Information} \& {Knowledge} {Management}}, pages 2974--2983, Atlanta GA
  USA, October 2022. ACM.
\newblock ISBN 978-1-4503-9236-5.
\newblock \doi{10.1145/3511808.3557082}.
\newblock URL \url{https://dl.acm.org/doi/10.1145/3511808.3557082}.

\bibitem[Carmagnola et~al.(2011)Carmagnola, Cena, and
  Gena]{carmagnola_user_2011}
Francesca Carmagnola, Federica Cena, and Cristina Gena.
\newblock User model interoperability: a survey.
\newblock \emph{User Modeling and User-Adapted Interaction}, 21\penalty0
  (3):\penalty0 285--331, August 2011.
\newblock ISSN 1573-1391.
\newblock \doi{10.1007/s11257-011-9097-5}.
\newblock URL \url{https://doi.org/10.1007/s11257-011-9097-5}.

\bibitem[Carvalho et~al.(2017)Carvalho, Ferreira, Ferreira, de~Souza, Carvalho,
  Suhara, Pentland, and Pessin]{carvalho_exploiting_2017}
Eduardo Carvalho, Bruno~V. Ferreira, Jair Ferreira, Cleidson de~Souza, Hanna~V.
  Carvalho, Yoshihiko Suhara, Alex~Sandy Pentland, and Gustavo Pessin.
\newblock Exploiting the use of recurrent neural networks for driver behavior
  profiling.
\newblock In \emph{2017 {International} {Joint} {Conference} on {Neural}
  {Networks} ({IJCNN})}, pages 3016--3021, May 2017.
\newblock \doi{10.1109/IJCNN.2017.7966230}.
\newblock URL
  \url{https://ieeexplore.ieee.org/abstract/document/7966230?casa_token=qdBQ6ZJHlKsAAAAA:G2UJiXLEDC-EizefHgvpQZB0Dn6YZnHsYu0V1PrBUQX2Le9n1kjshOrgclteaWIHyZpw7MEgm41P}.
\newblock ISSN: 2161-4407.

\bibitem[Castellano et~al.(2007)Castellano, Fanelli, Mencar, and
  Torsello]{castellano_similarity-based_2007}
Giovanna Castellano, A.~Maria Fanelli, Corrado Mencar, and M.~Alessandra
  Torsello.
\newblock Similarity-{Based} {Fuzzy} {Clustering} for {User} {Profiling}.
\newblock In \emph{2007 {IEEE}/{WIC}/{ACM} {International} {Conferences} on
  {Web} {Intelligence} and {Intelligent} {Agent} {Technology} - {Workshops}},
  pages 75--78, November 2007.
\newblock \doi{10.1109/WI-IATW.2007.32}.

\bibitem[Cañigueral and Meléndez(2021)]{canigueral_flexibility_2021}
Marc Cañigueral and Joaquim Meléndez.
\newblock Flexibility management of electric vehicles based on user profiles:
  {The} {Arnhem} case study.
\newblock \emph{International Journal of Electrical Power \& Energy Systems},
  133:\penalty0 107195, December 2021.
\newblock ISSN 01420615.
\newblock \doi{10.1016/j.ijepes.2021.107195}.
\newblock URL
  \url{https://linkinghub.elsevier.com/retrieve/pii/S0142061521004348}.

\bibitem[Cena et~al.(2012)Cena, Dattolo, De~Luca, Lops, Plumbaum, and
  Vassileva]{cena_semantic_2012}
Federica Cena, Antonina Dattolo, Ernesto~William De~Luca, Pasquale Lops, Till
  Plumbaum, and Julita Vassileva.
\newblock Semantic {Adaptive} {Social} {Web}.
\newblock In Liliana Ardissono and Tsvi Kuflik, editors, \emph{Advances in
  {User} {Modeling}}, Lecture {Notes} in {Computer} {Science}, pages 176--180,
  Berlin, Heidelberg, 2012. Springer.
\newblock ISBN 978-3-642-28509-7.
\newblock \doi{10.1007/978-3-642-28509-7_17}.

\bibitem[Cena et~al.(2019)Cena, Likavec, and Rapp]{cena_real_2019}
Federica Cena, Silvia Likavec, and Amon Rapp.
\newblock Real {World} {User} {Model}: {Evolution} of {User} {Modeling}
  {Triggered} by {Advances} in {Wearable} and {Ubiquitous} {Computing}.
\newblock \emph{Information Systems Frontiers}, 21\penalty0 (5):\penalty0
  1085--1110, October 2019.
\newblock ISSN 1572-9419.
\newblock \doi{10.1007/s10796-017-9818-3}.
\newblock URL \url{https://doi.org/10.1007/s10796-017-9818-3}.

\bibitem[Cena et~al.(2022)Cena, Gena, Mattutino, Mioli, Treccani, Vernero, and
  Zancanaro]{cena_incorporating_2022}
Federica Cena, Cristina Gena, Claudio Mattutino, Michele Mioli, Barbara
  Treccani, Fabiana Vernero, and Massimo Zancanaro.
\newblock Incorporating {Personality} {Traits} in {User} {Modeling} for {EUD}.
\newblock In \emph{Proceedings of the 3rd {International} {Workshop} on
  {Empowering} {People} in {Dealing} with {Internet} of {Things} {Ecosystems}
  co-located with {International} {Conference} on {Advanced} {Visual}
  {Interfaces} ({AVI}) 2022}, Frascati, Italy, June 2022.

\bibitem[Chari et~al.(2020)Chari, Seneviratne, Gruen, Foreman, Das, and
  McGuinness]{chari_explanation_2020}
Shruthi Chari, Oshani Seneviratne, Daniel~M. Gruen, Morgan~A. Foreman, Amar~K.
  Das, and Deborah~L. McGuinness.
\newblock Explanation {Ontology}: {A} {Model} of {Explanations} for
  {User}-{Centered} {AI}.
\newblock In Jeff~Z. Pan, Valentina Tamma, Claudia d’Amato, Krzysztof
  Janowicz, Bo~Fu, Axel Polleres, Oshani Seneviratne, and Lalana Kagal,
  editors, \emph{The {Semantic} {Web} – {ISWC} 2020}, Lecture {Notes} in
  {Computer} {Science}, pages 228--243, Cham, 2020. Springer International
  Publishing.
\newblock ISBN 978-3-030-62466-8.
\newblock \doi{10.1007/978-3-030-62466-8_15}.

\bibitem[Chen et~al.(2022{\natexlab{a}})Chen, Dou, Zhu, Cao, Cheng, and
  Wen]{chen_enhancing_2022}
Haonan Chen, Zhicheng Dou, Yutao Zhu, Zhao Cao, Xiaohua Cheng, and Ji-Rong Wen.
\newblock Enhancing {User} {Behavior} {Sequence} {Modeling} by {Generative}
  {Tasks} for {Session} {Search}.
\newblock In \emph{Proceedings of the 31st {ACM} {International} {Conference}
  on {Information} \& {Knowledge} {Management}}, pages 180--190, Atlanta GA
  USA, October 2022{\natexlab{a}}. ACM.
\newblock ISBN 978-1-4503-9236-5.
\newblock \doi{10.1145/3511808.3557310}.
\newblock URL \url{https://dl.acm.org/doi/10.1145/3511808.3557310}.

\bibitem[Chen et~al.(2019{\natexlab{a}})Chen, Cheng, Yang, Liang, Quan, and
  Li]{chen_joint_2019}
Jing Chen, Long Cheng, Xi~Yang, Jun Liang, Bing Quan, and Shoushan Li.
\newblock Joint {Learning} with both {Classification} and {Regression} {Models}
  for {Age} {Prediction}.
\newblock \emph{Journal of Physics: Conference Series}, 1168\penalty0
  (3):\penalty0 032016, February 2019{\natexlab{a}}.
\newblock ISSN 1742-6596.
\newblock \doi{10.1088/1742-6596/1168/3/032016}.
\newblock URL \url{https://dx.doi.org/10.1088/1742-6596/1168/3/032016}.

\bibitem[Chen et~al.(2021{\natexlab{a}})Chen, Yin, Peng, Rong, Yang, and
  Cong]{chen_monitoring_2021}
Tinggui Chen, Xiaohua Yin, Lijuan Peng, Jingtao Rong, Jianjun Yang, and Guodong
  Cong.
\newblock Monitoring and {Recognizing} {Enterprise} {Public} {Opinion} from
  {High}-{Risk} {Users} {Based} on {User} {Portrait} and {Random} {Forest}
  {Algorithm}.
\newblock \emph{Axioms}, 10\penalty0 (2), June 2021{\natexlab{a}}.
\newblock ISSN 2075-1680.
\newblock \doi{10.3390/axioms10020106}.
\newblock URL \url{https://www.mdpi.com/2075-1680/10/2/106}.
\newblock Number: 2 Publisher: Multidisciplinary Digital Publishing Institute.

\bibitem[Chen et~al.(2019{\natexlab{b}})Chen, Gu, Ren, He, Xie, Guo, Yin, and
  Zhang]{chen_semi-supervised_2019}
Weijian Chen, Yulong Gu, Zhaochun Ren, Xiangnan He, Hongtao Xie, Tong Guo,
  Dawei Yin, and Yongdong Zhang.
\newblock Semi-supervised {User} {Profiling} with {Heterogeneous} {Graph}
  {Attention} {Networks}.
\newblock In \emph{Proceedings of the {Twenty}-{Eighth} {International} {Joint}
  {Conference} on {Artificial} {Intelligence}}, pages 2116--2122, Macao, China,
  August 2019{\natexlab{b}}. International Joint Conferences on Artificial
  Intelligence Organization.
\newblock ISBN 978-0-9992411-4-1.
\newblock \doi{10.24963/ijcai.2019/293}.
\newblock URL \url{https://www.ijcai.org/proceedings/2019/293}.

\bibitem[Chen et~al.(2021{\natexlab{b}})Chen, Feng, Wang, He, Song, Ling, and
  Zhang]{chen_catgcn_2021}
Weijian Chen, Fuli Feng, Qifan Wang, Xiangnan He, Chonggang Song, Guohui Ling,
  and Yongdong Zhang.
\newblock {CatGCN}: {Graph} {Convolutional} {Networks} with {Categorical}
  {Node} {Features}.
\newblock \emph{IEEE Transactions on Knowledge and Data Engineering}, pages
  1--1, 2021{\natexlab{b}}.
\newblock ISSN 1558-2191.
\newblock \doi{10.1109/TKDE.2021.3133013}.

\bibitem[Chen et~al.(2022{\natexlab{b}})Chen, He, Ni, Pan, Chen, and
  Ming]{chen_global_2022}
Weixin Chen, Mingkai He, Yongxin Ni, Weike Pan, Li~Chen, and Zhong Ming.
\newblock Global and {Personalized} {Graphs} for {Heterogeneous} {Sequential}
  {Recommendation} by {Learning} {Behavior} {Transitions} and {User}
  {Intentions}.
\newblock In \emph{Proceedings of the 16th {ACM} {Conference} on {Recommender}
  {Systems}}, pages 268--277, Seattle WA USA, September 2022{\natexlab{b}}.
  ACM.
\newblock ISBN 978-1-4503-9278-5.
\newblock \doi{10.1145/3523227.3546761}.
\newblock URL \url{https://dl.acm.org/doi/10.1145/3523227.3546761}.

\bibitem[Cheng et~al.(2023)Cheng, Han, Liu, Zhu, Gao, and
  Peng]{cheng_multi-behavior_2023}
Zhiyong Cheng, Sai Han, Fan Liu, Lei Zhu, Zan Gao, and Yuxin Peng.
\newblock Multi-{Behavior} {Recommendation} with {Cascading} {Graph}
  {Convolution} {Networks}.
\newblock In \emph{Proceedings of the {ACM} {Web} {Conference} 2023}, {WWW}
  '23, pages 1181--1189, New York, NY, USA, April 2023. Association for
  Computing Machinery.
\newblock ISBN 978-1-4503-9416-1.
\newblock \doi{10.1145/3543507.3583439}.
\newblock URL \url{https://dl.acm.org/doi/10.1145/3543507.3583439}.

\bibitem[Cheung~Chiu and Webb(1998)]{cheung_chiu_using_1998}
Bark Cheung~Chiu and Geoffrey~I. Webb.
\newblock Using {Decision} {Trees} for {Agent} {Modeling}: {Improving}
  {Prediction} {Performance}.
\newblock \emph{User Modeling and User-Adapted Interaction}, 8\penalty0
  (1):\penalty0 131--152, March 1998.
\newblock ISSN 1573-1391.
\newblock \doi{10.1023/A:1008296930163}.
\newblock URL \url{https://doi.org/10.1023/A:1008296930163}.

\bibitem[Cho et~al.(2023)Cho, Hyun, Lim, Cheon, Park, and Yu]{cho_dynamic_2023}
Junsu Cho, Dongmin Hyun, Dong~won Lim, Hyeon~jae Cheon, Hyoung-iel Park, and
  Hwanjo Yu.
\newblock Dynamic {Multi}-{Behavior} {Sequence} {Modeling} for {Next} {Item}
  {Recommendation}.
\newblock \emph{Proceedings of the AAAI Conference on Artificial Intelligence},
  37\penalty0 (4):\penalty0 4199--4207, June 2023.
\newblock ISSN 2374-3468.
\newblock \doi{10.1609/aaai.v37i4.25537}.
\newblock URL \url{https://ojs.aaai.org/index.php/AAAI/article/view/25537}.

\bibitem[Chu et~al.(2022)Chu, Hosseinalipour, Tenorio, Cruz, Douglas, Lan, and
  Brinton]{chu_mitigating_2022}
Yun-Wei Chu, Seyyedali Hosseinalipour, Elizabeth Tenorio, Laura Cruz, Kerrie
  Douglas, Andrew Lan, and Christopher Brinton.
\newblock Mitigating {Biases} in {Student} {Performance} {Prediction} via
  {Attention}-{Based} {Personalized} {Federated} {Learning}.
\newblock In \emph{Proceedings of the 31st {ACM} {International} {Conference}
  on {Information} \& {Knowledge} {Management}}, pages 3033--3042, Atlanta GA
  USA, October 2022. ACM.
\newblock ISBN 978-1-4503-9236-5.
\newblock \doi{10.1145/3511808.3557108}.
\newblock URL \url{https://dl.acm.org/doi/10.1145/3511808.3557108}.

\bibitem[Codina et~al.(2015)Codina, Mena, and Oliva]{codina_context-aware_2015}
Victor Codina, Jose Mena, and Luis Oliva.
\newblock Context-{Aware} {User} {Modeling} {Strategies} for {Journey} {Plan}
  {Recommendation}.
\newblock In Francesco Ricci, Kalina Bontcheva, Owen Conlan, and Séamus
  Lawless, editors, \emph{User {Modeling}, {Adaptation} and {Personalization}},
  Lecture {Notes} in {Computer} {Science}, pages 68--79, Cham, 2015. Springer
  International Publishing.
\newblock ISBN 978-3-319-20267-9.
\newblock \doi{10.1007/978-3-319-20267-9_6}.

\bibitem[Cohen and Perrault(1979)]{cohen_elements_1979}
Philip~R. Cohen and C.~Raymond Perrault.
\newblock Elements of a {Plan}-{Based} {Theory} of {Speech} {Acts}.
\newblock \emph{Cognitive Science}, 3\penalty0 (3):\penalty0 177--212, 1979.
\newblock ISSN 1551-6709.
\newblock \doi{10.1207/s15516709cog0303_1}.
\newblock URL
  \url{https://onlinelibrary.wiley.com/doi/abs/10.1207/s15516709cog0303_1}.

\bibitem[Costanzo et~al.(2019)Costanzo, Deldjoo, Dacrema, Schedl, and
  Cremonesi]{costanzo_towards_2019}
Luca~Luciano Costanzo, Yashar Deldjoo, Maurizio~Ferrari Dacrema, Markus Schedl,
  and Paolo Cremonesi.
\newblock Towards {Evaluating} {User} {Profiling} {Methods} {Based} on
  {Explicit} {Ratings} on {Item} {Features}.
\newblock In \emph{Proceedings of the 6th {Joint} {Workshop} on {Interfaces}
  and {Human} {Decision} {Making} for {Recommender} {Systems} co-located with
  13th {ACM} {Conference} on {Recommender} {Systems}({RecSys} 2019)},
  Copenhagen, Denmark, September 2019.

\bibitem[Covington et~al.(2016)Covington, Adams, and
  Sargin]{covington_deep_2016}
Paul Covington, Jay Adams, and Emre Sargin.
\newblock Deep {Neural} {Networks} for {YouTube} {Recommendations}.
\newblock In \emph{Proceedings of the 10th {ACM} {Conference} on {Recommender}
  {Systems}}, {RecSys} '16, pages 191--198, New York, NY, USA, 2016.
  Association for Computing Machinery.
\newblock ISBN 978-1-4503-4035-9.
\newblock \doi{10.1145/2959100.2959190}.
\newblock URL \url{https://dl.acm.org/doi/10.1145/2959100.2959190}.

\bibitem[Craswell et~al.(2005)Craswell, de~Vries, and
  Soboroff]{craswell_overview_2005}
Nick Craswell, Arjen de~Vries, and Ian Soboroff.
\newblock Overview of the {TREC}-2005 enterprise track.
\newblock In \emph{TREC}, volume~5, pages 1--7, January 2005.

\bibitem[Cufoglu(2014)]{cufoglu_user_2014}
Ayse Cufoglu.
\newblock User {Profiling} - {A} {Short} {Review}.
\newblock \emph{International Journal of Computer Applications}, 108\penalty0
  (3):\penalty0 1--9, December 2014.
\newblock ISSN 09758887.
\newblock \doi{10.5120/18888-0179}.
\newblock URL
  \url{http://research.ijcaonline.org/volume108/number3/pxc3900179.pdf}.

\bibitem[Cufoglu et~al.(2008)Cufoglu, Lohi, and
  Madani]{cufoglu_comparative_2008}
Ayse Cufoglu, Mahi Lohi, and Kambiz Madani.
\newblock A {Comparative} {Study} of {Selected} {Classification} {Accuracy} in
  {User} {Profiling}.
\newblock In \emph{2008 {Seventh} {International} {Conference} on {Machine}
  {Learning} and {Applications}}, pages 787--791, December 2008.
\newblock \doi{10.1109/ICMLA.2008.139}.
\newblock URL \url{https://ieeexplore.ieee.org/abstract/document/4725067}.

\bibitem[Cura et~al.(2021)Cura, Küçük, Ergen, and
  Öksüzoğlu]{cura_driver_2021}
Aslıhan Cura, Haluk Küçük, Erdem Ergen, and İsmail~Burak Öksüzoğlu.
\newblock Driver {Profiling} {Using} {Long} {Short} {Term} {Memory} ({LSTM})
  and {Convolutional} {Neural} {Network} ({CNN}) {Methods}.
\newblock \emph{IEEE Transactions on Intelligent Transportation Systems},
  22\penalty0 (10):\penalty0 6572--6582, October 2021.
\newblock ISSN 1558-0016.
\newblock \doi{10.1109/TITS.2020.2995722}.
\newblock URL
  \url{https://ieeexplore.ieee.org/abstract/document/9110758?casa_token=fkFYuzEvG9AAAAAA:b5kKbGrpRUK8J4vLP3XbWQMLBIvfoHvosQb6ykSetomcyvULuGAPfgtX9U7Zs7Kcisbdxt3-t4Em}.

\bibitem[Curmei et~al.(2022)Curmei, Haupt, Recht, and
  Hadfield-Menell]{curmei_towards_2022}
Mihaela Curmei, Andreas~A. Haupt, Benjamin Recht, and Dylan Hadfield-Menell.
\newblock Towards {Psychologically}-{Grounded} {Dynamic} {Preference} {Models}.
\newblock In \emph{Proceedings of the 16th {ACM} {Conference} on {Recommender}
  {Systems}}, pages 35--48, Seattle WA USA, September 2022. ACM.
\newblock ISBN 978-1-4503-9278-5.
\newblock \doi{10.1145/3523227.3546778}.
\newblock URL \url{https://dl.acm.org/doi/10.1145/3523227.3546778}.

\bibitem[Dai and Wang(2021)]{dai_say_2021}
Enyan Dai and Suhang Wang.
\newblock Say {No} to the {Discrimination}: {Learning} {Fair} {Graph} {Neural}
  {Networks} with {Limited} {Sensitive} {Attribute} {Information}.
\newblock In \emph{Proceedings of the 14th {ACM} {International} {Conference}
  on {Web} {Search} and {Data} {Mining}}, {WSDM} '21, pages 680--688, New York,
  NY, USA, March 2021. Association for Computing Machinery.
\newblock ISBN 978-1-4503-8297-7.
\newblock \doi{10.1145/3437963.3441752}.
\newblock URL \url{https://dl.acm.org/doi/10.1145/3437963.3441752}.

\bibitem[De~Bra et~al.(1999)De~Bra, Brusilovsky, and
  Houben]{de_bra_adaptive_1999}
Paul De~Bra, Peter Brusilovsky, and Geert-Jan Houben.
\newblock Adaptive hypermedia: from systems to framework.
\newblock \emph{ACM Computing Surveys}, 31\penalty0 (4es):\penalty0 12,
  December 1999.
\newblock ISSN 0360-0300, 1557-7341.
\newblock \doi{10.1145/345966.345996}.
\newblock URL \url{https://dl.acm.org/doi/10.1145/345966.345996}.

\bibitem[De~Campos et~al.(2014)De~Campos, Fernandez-Luna, Huete, and
  Vicente-Lopez]{de_campos_using_2014}
Luis~M. De~Campos, Juan~M. Fernandez-Luna, Juan~F. Huete, and Eduardo
  Vicente-Lopez.
\newblock Using {Personalization} to {Improve} {XML} {Retrieval}.
\newblock \emph{IEEE Transactions on Knowledge and Data Engineering},
  26\penalty0 (5):\penalty0 1280--1292, May 2014.
\newblock ISSN 1041-4347.
\newblock \doi{10.1109/TKDE.2013.75}.
\newblock URL \url{http://ieeexplore.ieee.org/document/6514872/}.

\bibitem[De~Luca et~al.(2010)De~Luca, Plumbaum, Kunegis, and
  Albayrak]{deluca_multilingual_2010}
Ernesto~William De~Luca, Till Plumbaum, J{\'e}r{\^o}me Kunegis, and Sahin
  Albayrak.
\newblock Multilingual ontology-based user profile enrichment.
\newblock In \emph{MSW}, pages 41--42, 2010.

\bibitem[De~Pauw et~al.(2022)De~Pauw, Ruymbeek, and Goethals]{de_pauw_who_2022}
Joey De~Pauw, Koen Ruymbeek, and Bart Goethals.
\newblock Who do you think {I} am? {Interactive} {User} {Modelling} with {Item}
  {Metadata}.
\newblock In \emph{Proceedings of the 16th {ACM} {Conference} on {Recommender}
  {Systems}}, pages 640--643, Seattle WA USA, September 2022. ACM.
\newblock ISBN 978-1-4503-9278-5.
\newblock \doi{10.1145/3523227.3551470}.
\newblock URL \url{https://dl.acm.org/doi/10.1145/3523227.3551470}.

\bibitem[Deng et~al.(2022{\natexlab{a}})Deng, Zhou, and
  Dou]{deng_improving_2022}
Chenlong Deng, Yujia Zhou, and Zhicheng Dou.
\newblock Improving {Personalized} {Search} with {Dual}-{Feedback} {Network}.
\newblock In \emph{Proceedings of the {Fifteenth} {ACM} {International}
  {Conference} on {Web} {Search} and {Data} {Mining}}, pages 210--218, Virtual
  Event AZ USA, February 2022{\natexlab{a}}. ACM.
\newblock ISBN 978-1-4503-9132-0.
\newblock \doi{10.1145/3488560.3498447}.
\newblock URL \url{https://dl.acm.org/doi/10.1145/3488560.3498447}.

\bibitem[Deng et~al.(2022{\natexlab{b}})Deng, Cai, Zhang, and
  Wu]{deng_user_2022}
Song Deng, Qingyuan Cai, Zi~Zhang, and Xindong Wu.
\newblock User {Behavior} {Analysis} {Based} on {Stacked} {Autoencoder} and
  {Clustering} in {Complex} {Power} {Grid} {Environment}.
\newblock \emph{IEEE Transactions on Intelligent Transportation Systems},
  23\penalty0 (12):\penalty0 25521--25535, December 2022{\natexlab{b}}.
\newblock ISSN 1558-0016.
\newblock \doi{10.1109/TITS.2021.3076607}.
\newblock URL \url{https://ieeexplore.ieee.org/abstract/document/9430771}.
\newblock Conference Name: IEEE Transactions on Intelligent Transportation
  Systems.

\bibitem[Deng et~al.(2013)Deng, Sang, and Xu]{deng_personalized_2013}
Zhengyu Deng, Jitao Sang, and Changsheng Xu.
\newblock Personalized video recommendation based on cross-platform user
  modeling.
\newblock In \emph{2013 {IEEE} {International} {Conference} on {Multimedia} and
  {Expo} ({ICME})}, pages 1--6, July 2013.
\newblock \doi{10.1109/ICME.2013.6607513}.
\newblock URL
  \url{https://ieeexplore.ieee.org/abstract/document/6607513?casa_token=6o5upYPGcmEAAAAA:LBu07wRx214QnmK_7zMR4-Jh79MA77Pv0U8wH96AgcittffEn1tzHfkU8WNyudM6QwNVNdta6w}.

\bibitem[Dignum(2019)]{dignum_responsible_2019}
Virginia Dignum.
\newblock \emph{Responsible {Artificial} {Intelligence}: {How} to {Develop} and
  {Use} {AI} in a {Responsible} {Way}}.
\newblock Artificial {Intelligence}: {Foundations}, {Theory}, and {Algorithms}.
  Springer International Publishing, Cham, 2019.
\newblock ISBN 978-3-030-30370-9 978-3-030-30371-6.
\newblock \doi{10.1007/978-3-030-30371-6}.
\newblock URL \url{http://link.springer.com/10.1007/978-3-030-30371-6}.

\bibitem[Ding et~al.(2023)Ding, Xie, Hao, Yang, Ge, Zhang, Zhou, and
  Lin]{ding_interpretable_2023}
Rui Ding, Ruobing Xie, Xiaobo Hao, Xiaochun Yang, Kaikai Ge, Xu~Zhang, Jie
  Zhou, and Leyu Lin.
\newblock Interpretable {User} {Retention} {Modeling} in {Recommendation}.
\newblock In \emph{Proceedings of the 17th {ACM} {Conference} on {Recommender}
  {Systems}}, pages 702--708, Singapore Singapore, September 2023. ACM.
\newblock ISBN 9798400702419.
\newblock \doi{10.1145/3604915.3608818}.
\newblock URL \url{https://dl.acm.org/doi/10.1145/3604915.3608818}.

\bibitem[D'Oca and Hong(2014)]{doca_data-mining_2014}
Simona D'Oca and Tianzhen Hong.
\newblock A data-mining approach to discover patterns of window opening and
  closing behavior in offices.
\newblock \emph{Building and Environment}, 82:\penalty0 726--739, December
  2014.
\newblock ISSN 0360-1323.
\newblock \doi{10.1016/j.buildenv.2014.10.021}.
\newblock URL
  \url{https://www.sciencedirect.com/science/article/pii/S0360132314003424}.

\bibitem[Dong et~al.(2021)Dong, Li, Song, and Ding]{dong_profiling_2021}
Xin Dong, Tong Li, Rui Song, and Zhiming Ding.
\newblock Profiling users via their reviews: an extended systematic mapping
  study.
\newblock \emph{Software and Systems Modeling}, 20\penalty0 (1):\penalty0
  49--69, February 2021.
\newblock ISSN 1619-1374.
\newblock \doi{10.1007/s10270-020-00790-w}.
\newblock URL \url{https://doi.org/10.1007/s10270-020-00790-w}.

\bibitem[Dong et~al.(2017)Dong, Chawla, Tang, Yang, and Yang]{dong_user_2017}
Yuxiao Dong, Nitesh~V. Chawla, Jie Tang, Yang Yang, and Yang Yang.
\newblock User {Modeling} on {Demographic} {Attributes} in {Big} {Mobile}
  {Social} {Networks}.
\newblock \emph{ACM Transactions on Information Systems}, 35\penalty0
  (4):\penalty0 35:1--35:33, 2017.
\newblock ISSN 1046-8188.
\newblock \doi{10.1145/3057278}.
\newblock URL \url{https://dl.acm.org/doi/10.1145/3057278}.

\bibitem[Donkers et~al.(2017)Donkers, Loepp, and
  Ziegler]{donkers_sequential_2017}
Tim Donkers, Benedikt Loepp, and Jürgen Ziegler.
\newblock Sequential {User}-based {Recurrent} {Neural} {Network}
  {Recommendations}.
\newblock In \emph{Proceedings of the {Eleventh} {ACM} {Conference} on
  {Recommender} {Systems}}, {RecSys} '17, pages 152--160, New York, NY, USA,
  2017. Association for Computing Machinery.
\newblock ISBN 978-1-4503-4652-8.
\newblock \doi{10.1145/3109859.3109877}.
\newblock URL \url{https://dl.acm.org/doi/10.1145/3109859.3109877}.

\bibitem[Du et~al.(2016)Du, Xie, Cai, Leung, Li, Min, and
  Wang]{du_folksonomy-based_2016}
Qing Du, Haoran Xie, Yi~Cai, Ho-fung Leung, Qing Li, Huaqing Min, and Fu~Lee
  Wang.
\newblock Folksonomy-based personalized search by hybrid user profiles in
  multiple levels.
\newblock \emph{Neurocomputing}, 204:\penalty0 142--152, September 2016.
\newblock ISSN 0925-2312.
\newblock \doi{10.1016/j.neucom.2015.10.135}.
\newblock URL
  \url{https://www.sciencedirect.com/science/article/pii/S0925231216301163}.

\bibitem[Díaz and Gervás(2004)]{diaz_adaptive_2004}
Alberto Díaz and Pablo Gervás.
\newblock Adaptive {User} {Modeling} for {Personalization} of {Web} {Contents}.
\newblock In Paul M.~E. De~Bra and Wolfgang Nejdl, editors, \emph{Adaptive
  {Hypermedia} and {Adaptive} {Web}-{Based} {Systems}}, Lecture {Notes} in
  {Computer} {Science}, pages 65--74, Berlin, Heidelberg, 2004. Springer.
\newblock ISBN 978-3-540-27780-4.
\newblock \doi{10.1007/978-3-540-27780-4_10}.

\bibitem[Eirinaki and Vazirgiannis(2003)]{eirinaki_web_2003}
Magdalini Eirinaki and Michalis Vazirgiannis.
\newblock Web mining for web personalization.
\newblock \emph{ACM Transactions on Internet Technology}, 3\penalty0
  (1):\penalty0 1--27, February 2003.
\newblock ISSN 1533-5399.
\newblock \doi{10.1145/643477.643478}.
\newblock URL \url{https://dl.acm.org/doi/10.1145/643477.643478}.

\bibitem[Eke et~al.(2019)Eke, Norman, Shuib, and Nweke]{eke_survey_2019}
Christopher~Ifeanyi Eke, Azah~Anir Norman, Liyana Shuib, and Henry~Friday
  Nweke.
\newblock A {Survey} of {User} {Profiling}: {State}-of-the-{Art}, {Challenges},
  and {Solutions}.
\newblock \emph{IEEE Access}, 7:\penalty0 144907--144924, 2019.
\newblock ISSN 2169-3536.
\newblock \doi{10.1109/ACCESS.2019.2944243}.

\bibitem[El-Ansari et~al.(2020)El-Ansari, Beni-Hssane, and
  Saadi]{el-ansari_improved_2020}
Anas El-Ansari, Abderrahim Beni-Hssane, and Mostafa Saadi.
\newblock An improved modeling method for profile-based personalized search.
\newblock In \emph{Proceedings of the 3rd {International} {Conference} on
  {Networking}, {Information} {Systems} \& {Security}}, {NISS} '20, pages 1--6,
  New York, NY, USA, 2020. Association for Computing Machinery.
\newblock ISBN 978-1-4503-7634-1.
\newblock \doi{10.1145/3386723.3387874}.
\newblock URL \url{https://dl.acm.org/doi/10.1145/3386723.3387874}.

\bibitem[Elallioui and El~Beqqali(2012)]{elallioui_user_2012}
Youssouf Elallioui and Omar El~Beqqali.
\newblock User profile {Ontology} for the {Personalization} approach.
\newblock \emph{International Journal of Computer Applications}, 41:\penalty0
  31--40, March 2012.
\newblock \doi{10.5120/5531-7577}.

\bibitem[Elkahky et~al.(2015)Elkahky, Song, and He]{elkahky_multi-view_2015}
Ali~Mamdouh Elkahky, Yang Song, and Xiaodong He.
\newblock A {Multi}-{View} {Deep} {Learning} {Approach} for {Cross} {Domain}
  {User} {Modeling} in {Recommendation} {Systems}.
\newblock In \emph{Proceedings of the 24th {International} {Conference} on
  {World} {Wide} {Web}}, {WWW} '15, pages 278--288, Republic and Canton of
  Geneva, CHE, 2015. International World Wide Web Conferences Steering
  Committee.
\newblock ISBN 978-1-4503-3469-3.
\newblock \doi{10.1145/2736277.2741667}.
\newblock URL \url{https://dl.acm.org/doi/10.1145/2736277.2741667}.

\bibitem[Erlandsson et~al.(2016)Erlandsson, Bródka, Borg, and
  Johnson]{erlandsson_finding_2016}
Fredrik Erlandsson, Piotr Bródka, Anton Borg, and Henric Johnson.
\newblock Finding {Influential} {Users} in {Social} {Media} {Using}
  {Association} {Rule} {Learning}.
\newblock \emph{Entropy}, 18\penalty0 (5):\penalty0 164, May 2016.
\newblock ISSN 1099-4300.
\newblock \doi{10.3390/e18050164}.
\newblock URL \url{https://www.mdpi.com/1099-4300/18/5/164}.
\newblock Number: 5 Publisher: Multidisciplinary Digital Publishing Institute.

\bibitem[Fan et~al.(2022)Fan, Ou, Gu, Fu, Li, Bao, Dai, Zeng, Zhuang, and
  Liu]{fan_modeling_2022}
Zhifang Fan, Dan Ou, Yulong Gu, Bairan Fu, Xiang Li, Wentian Bao, Xin-Yu Dai,
  Xiaoyi Zeng, Tao Zhuang, and Qingwen Liu.
\newblock Modeling {Users}' {Contextualized} {Page}-wise {Feedback} for
  {Click}-{Through} {Rate} {Prediction} in {E}-commerce {Search}.
\newblock In \emph{Proceedings of the {Fifteenth} {ACM} {International}
  {Conference} on {Web} {Search} and {Data} {Mining}}, pages 262--270, Virtual
  Event AZ USA, February 2022. ACM.
\newblock ISBN 978-1-4503-9132-0.
\newblock \doi{10.1145/3488560.3498478}.
\newblock URL \url{https://dl.acm.org/doi/10.1145/3488560.3498478}.

\bibitem[Farid et~al.(2018)Farid, Elgohary, Moawad, and
  Roushdy]{farid_user_2018}
Marina Farid, Rania Elgohary, Ibrahim Moawad, and Mohamed Roushdy.
\newblock User {Profiling} {Approaches}, {Modeling}, and {Personalization},
  October 2018.
\newblock URL \url{https://papers.ssrn.com/abstract=3389811}.

\bibitem[Farseev et~al.(2015)Farseev, Nie, Akbari, and
  Chua]{farseev_harvesting_2015}
Aleksandr Farseev, Liqiang Nie, Mohammad Akbari, and Tat-Seng Chua.
\newblock Harvesting {Multiple} {Sources} for {User} {Profile} {Learning}: a
  {Big} {Data} {Study}.
\newblock In \emph{Proceedings of the 5th {ACM} on {International} {Conference}
  on {Multimedia} {Retrieval}}, {ICMR} '15, pages 235--242, New York, NY, USA,
  2015. Association for Computing Machinery.
\newblock ISBN 978-1-4503-3274-3.
\newblock \doi{10.1145/2671188.2749381}.
\newblock URL \url{https://dl.acm.org/doi/10.1145/2671188.2749381}.

\bibitem[Fazelnia et~al.(2022)Fazelnia, Simon, Anderson, Carterette, and
  Lalmas]{fazelnia_variational_2022}
Ghazal Fazelnia, Eric Simon, Ian Anderson, Benjamin Carterette, and Mounia
  Lalmas.
\newblock Variational {User} {Modeling} with {Slow} and {Fast} {Features}.
\newblock In \emph{Proceedings of the {Fifteenth} {ACM} {International}
  {Conference} on {Web} {Search} and {Data} {Mining}}, pages 271--279, Virtual
  Event AZ USA, February 2022. ACM.
\newblock ISBN 978-1-4503-9132-0.
\newblock \doi{10.1145/3488560.3498477}.
\newblock URL \url{https://dl.acm.org/doi/10.1145/3488560.3498477}.

\bibitem[Fazil et~al.(2021)Fazil, Sah, and Abulaish]{fazil_deepsbd_2021}
Mohd Fazil, Amit~Kumar Sah, and Muhammad Abulaish.
\newblock {DeepSBD}: {A} {Deep} {Neural} {Network} {Model} {With} {Attention}
  {Mechanism} for {SocialBot} {Detection}.
\newblock \emph{IEEE Transactions on Information Forensics and Security},
  16:\penalty0 4211--4223, 2021.
\newblock ISSN 1556-6013, 1556-6021.
\newblock \doi{10.1109/TIFS.2021.3102498}.
\newblock URL \url{https://ieeexplore.ieee.org/document/9505695/}.

\bibitem[Fernandez-Lanvin et~al.(2018)Fernandez-Lanvin, Andres-Suarez,
  Gonzalez-Rodriguez, and Pariente-Martinez]{fernandez-lanvin_dimension_2018}
D.~Fernandez-Lanvin, J.~de Andres-Suarez, M.~Gonzalez-Rodriguez, and
  B.~Pariente-Martinez.
\newblock The dimension of age and gender as user model demographic factors for
  automatic personalization in e-commerce sites.
\newblock \emph{Computer Standards \& Interfaces}, 59:\penalty0 1--9, August
  2018.
\newblock ISSN 0920-5489.
\newblock \doi{10.1016/j.csi.2018.02.001}.
\newblock URL
  \url{https://www.sciencedirect.com/science/article/pii/S0920548917303070}.

\bibitem[Finin and Drager(1986)]{finin_gums_1986}
Tim Finin and David Drager.
\newblock {GUMS}: {A} {General} {User} {Modeling} {System}.
\newblock \emph{Proceedings of the workshop on Strategic computing natural
  language of the Human Language Technology Conference}, pages 224--230, May
  1986.
\newblock URL
  \url{https://ebiquity.umbc.edu/paper/abstract/id/325/GUMS-A-General-User-Modeling-System}.

\bibitem[Fink and Kobsa(2000)]{fink_review_2000}
Josef Fink and Alfred Kobsa.
\newblock A {Review} and {Analysis} of {Commercial} {User} {Modeling} {Servers}
  for {Personalization} on the {World} {Wide} {Web}.
\newblock \emph{User Modeling and User-Adapted Interaction}, 10\penalty0
  (2):\penalty0 209--249, June 2000.
\newblock ISSN 1573-1391.
\newblock \doi{10.1023/A:1026597308943}.
\newblock URL \url{https://doi.org/10.1023/A:1026597308943}.

\bibitem[Frias-Martinez et~al.(2005)Frias-Martinez, Magoulas, Chen, and
  Macredie]{frias-martinez_modeling_2005}
E.~Frias-Martinez, G.~Magoulas, S.~Chen, and R.~Macredie.
\newblock Modeling human behavior in user-adaptive systems: {Recent} advances
  using soft computing techniques.
\newblock \emph{Expert Systems with Applications}, 29\penalty0 (2):\penalty0
  320--329, August 2005.
\newblock ISSN 0957-4174.
\newblock \doi{10.1016/j.eswa.2005.04.005}.
\newblock URL
  \url{https://www.sciencedirect.com/science/article/pii/S0957417405000588}.

\bibitem[Frias-Martinez et~al.(2006)Frias-Martinez, Chen, and
  Liu]{frias-martinez_survey_2006}
E.~Frias-Martinez, S.Y. Chen, and X.~Liu.
\newblock Survey of {Data} {Mining} {Approaches} to {User} {Modeling} for
  {Adaptive} {Hypermedia}.
\newblock \emph{IEEE Transactions on Systems, Man and Cybernetics, Part C
  (Applications and Reviews)}, 36\penalty0 (6):\penalty0 734--749, November
  2006.
\newblock ISSN 1094-6977.
\newblock \doi{10.1109/TSMCC.2006.879391}.
\newblock URL \url{http://ieeexplore.ieee.org/document/1715503/}.

\bibitem[Fu et~al.(2013)Fu, Lin, Li, Faloutsos, Hong, and Sadeh]{fu_why_2013}
Bin Fu, Jialiu Lin, Lei Li, Christos Faloutsos, Jason Hong, and Norman Sadeh.
\newblock Why people hate your app: making sense of user feedback in a mobile
  app store.
\newblock In \emph{Proceedings of the 19th {ACM} {SIGKDD} international
  conference on {Knowledge} discovery and data mining}, {KDD} '13, pages
  1276--1284, New York, NY, USA, 2013. Association for Computing Machinery.
\newblock ISBN 978-1-4503-2174-7.
\newblock \doi{10.1145/2487575.2488202}.
\newblock URL \url{https://dl.acm.org/doi/10.1145/2487575.2488202}.

\bibitem[Gao et~al.(2017)Gao, Wu, Zhou, and Hu]{gao_collaborative_2017}
Li~Gao, Jia Wu, Chuan Zhou, and Yue Hu.
\newblock Collaborative {Dynamic} {Sparse} {Topic} {Regression} with {User}
  {Profile} {Evolution} for {Item} {Recommendation}.
\newblock \emph{Proceedings of the AAAI Conference on Artificial Intelligence},
  31\penalty0 (1), February 2017.
\newblock ISSN 2374-3468.
\newblock \doi{10.1609/aaai.v31i1.10726}.
\newblock URL \url{https://ojs.aaai.org/index.php/AAAI/article/view/10726}.

\bibitem[Gao et~al.(2010)Gao, Liu, and Wu]{gao_personalisation_2010}
Min Gao, Kecheng Liu, and Zhongfu Wu.
\newblock Personalisation in web computing and informatics: {Theories},
  techniques, applications, and future research.
\newblock \emph{Information Systems Frontiers}, 12\penalty0 (5):\penalty0
  607--629, November 2010.
\newblock ISSN 1572-9419.
\newblock \doi{10.1007/s10796-009-9199-3}.
\newblock URL \url{https://doi.org/10.1007/s10796-009-9199-3}.

\bibitem[Gao et~al.(2013)Gao, Hao, Bai, Li, Li, and Zhu]{gao_improving_2013}
Rui Gao, Bibo Hao, Shuotian Bai, Lin Li, Ang Li, and Tingshao Zhu.
\newblock Improving user profile with personality traits predicted from social
  media content.
\newblock In \emph{Proceedings of the 7th {ACM} conference on {Recommender}
  systems}, {RecSys} '13, pages 355--358, New York, NY, USA, 2013. Association
  for Computing Machinery.
\newblock ISBN 978-1-4503-2409-0.
\newblock \doi{10.1145/2507157.2507219}.
\newblock URL \url{https://dl.acm.org/doi/10.1145/2507157.2507219}.

\bibitem[Gao et~al.(2020)Gao, Gong, Zhao, Wang, Shu, and Zhu]{gao_joint_2020}
Xiaofeng Gao, Ran Gong, Yizhou Zhao, Shu Wang, Tianmin Shu, and Song-Chun Zhu.
\newblock Joint {Mind} {Modeling} for {Explanation} {Generation} in {Complex}
  {Human}-{Robot} {Collaborative} {Tasks}.
\newblock In \emph{2020 29th {IEEE} {International} {Conference} on {Robot} and
  {Human} {Interactive} {Communication} ({RO}-{MAN})}, pages 1119--1126, August
  2020.
\newblock \doi{10.1109/RO-MAN47096.2020.9223595}.
\newblock URL \url{https://ieeexplore.ieee.org/document/9223595}.
\newblock ISSN: 1944-9437.

\bibitem[Gauch et~al.(2007)Gauch, Speretta, Chandramouli, and
  Micarelli]{gauch_user_2007}
Susan Gauch, Mirco Speretta, Aravind Chandramouli, and Alessandro Micarelli.
\newblock User {Profiles} for {Personalized} {Information} {Access}.
\newblock In Peter Brusilovsky, Alfred Kobsa, and Wolfgang Nejdl, editors,
  \emph{The {Adaptive} {Web}: {Methods} and {Strategies} of {Web}
  {Personalization}}, Lecture {Notes} in {Computer} {Science}, pages 54--89.
  Springer, Berlin, Heidelberg, 2007.
\newblock ISBN 978-3-540-72079-9.
\newblock \doi{10.1007/978-3-540-72079-9_2}.
\newblock URL \url{https://doi.org/10.1007/978-3-540-72079-9_2}.

\bibitem[Ge et~al.(2018)Ge, Dou, Jiang, Nie, and Wen]{ge_personalizing_2018}
Songwei Ge, Zhicheng Dou, Zhengbao Jiang, Jian-Yun Nie, and Ji-Rong Wen.
\newblock Personalizing {Search} {Results} {Using} {Hierarchical} {RNN} with
  {Query}-aware {Attention}.
\newblock In \emph{Proceedings of the 27th {ACM} {International} {Conference}
  on {Information} and {Knowledge} {Management}}, pages 347--356, Torino Italy,
  October 2018. ACM.
\newblock ISBN 978-1-4503-6014-2.
\newblock \doi{10.1145/3269206.3271728}.
\newblock URL \url{https://dl.acm.org/doi/10.1145/3269206.3271728}.

\bibitem[Gilbert et~al.(2023)Gilbert, Hamid, Hashem, Ghani, and
  Boluwatife]{gilbert_rise_2023}
Justin Gilbert, Suraya Hamid, Ibrahim Abaker~Targio Hashem, Norjihan~Abdul
  Ghani, and Fatokun~Faith Boluwatife.
\newblock The rise of user profiling in social media: review, challenges and
  future direction.
\newblock \emph{Social Network Analysis and Mining}, 13\penalty0 (1):\penalty0
  137, October 2023.
\newblock ISSN 1869-5469.
\newblock \doi{10.1007/s13278-023-01146-0}.
\newblock URL \url{https://doi.org/10.1007/s13278-023-01146-0}.

\bibitem[Godoy and Amandi(2005)]{godoy_user_2005}
Daniela Godoy and Analia Amandi.
\newblock User profiling in personal information agents: a survey.
\newblock \emph{The Knowledge Engineering Review}, 20\penalty0 (4):\penalty0
  329--361, December 2005.
\newblock ISSN 1469-8005, 0269-8889.
\newblock \doi{10.1017/S0269888906000397}.
\newblock URL
  \url{https://www.cambridge.org/core/journals/knowledge-engineering-review/article/user-profiling-in-personal-information-agents-a-survey/383527ACC2A6C40746AD7FE48E7E8B1D}.

\bibitem[Gomez~Bruballa et~al.(2022)Gomez~Bruballa, Burnham-King, and
  Sala]{gomez_bruballa_learning_2022}
Raul Gomez~Bruballa, Lauren Burnham-King, and Alessandra Sala.
\newblock Learning {Users}’ {Preferred} {Visual} {Styles} in an {Image}
  {Marketplace}.
\newblock In \emph{Proceedings of the 16th {ACM} {Conference} on {Recommender}
  {Systems}}, pages 466--468, Seattle WA USA, September 2022. ACM.
\newblock ISBN 978-1-4503-9278-5.
\newblock \doi{10.1145/3523227.3547382}.
\newblock URL \url{https://dl.acm.org/doi/10.1145/3523227.3547382}.

\bibitem[Gong and Wang(2018)]{gong_when_2018}
Lin Gong and Hongning Wang.
\newblock When {Sentiment} {Analysis} {Meets} {Social} {Network}: {A}
  {Holistic} {User} {Behavior} {Modeling} in {Opinionated} {Data}.
\newblock In \emph{Proceedings of the 24th {ACM} {SIGKDD} {International}
  {Conference} on {Knowledge} {Discovery} \& {Data} {Mining}}, pages
  1455--1464, London United Kingdom, July 2018. ACM.
\newblock ISBN 978-1-4503-5552-0.
\newblock \doi{10.1145/3219819.3220120}.
\newblock URL \url{https://dl.acm.org/doi/10.1145/3219819.3220120}.

\bibitem[Gong et~al.(2020)Gong, Lin, Song, and Wang]{gong_jnet_2020}
Lin Gong, Lu~Lin, Weihao Song, and Hongning Wang.
\newblock {JNET}: {Learning} {User} {Representations} via {Joint} {Network}
  {Embedding} and {Topic} {Embedding}.
\newblock In \emph{Proceedings of the 13th {International} {Conference} on
  {Web} {Search} and {Data} {Mining}}, {WSDM} '20, pages 205--213, New York,
  NY, USA, 2020. Association for Computing Machinery.
\newblock ISBN 978-1-4503-6822-3.
\newblock \doi{10.1145/3336191.3371770}.
\newblock URL \url{https://dl.acm.org/doi/10.1145/3336191.3371770}.

\bibitem[Gou et~al.(2014)Gou, Zhou, and Yang]{gou_knowme_2014}
Liang Gou, Michelle~X. Zhou, and Huahai Yang.
\newblock {KnowMe} and {ShareMe}: understanding automatically discovered
  personality traits from social media and user sharing preferences.
\newblock In \emph{Proceedings of the {SIGCHI} {Conference} on {Human}
  {Factors} in {Computing} {Systems}}, {CHI} '14, pages 955--964, New York, NY,
  USA, April 2014. Association for Computing Machinery.
\newblock ISBN 978-1-4503-2473-1.
\newblock \doi{10.1145/2556288.2557398}.
\newblock URL \url{https://dl.acm.org/doi/10.1145/2556288.2557398}.

\bibitem[Greco and Polli(2020)]{greco_emotional_2020}
Francesca Greco and Alessandro Polli.
\newblock Emotional {Text} {Mining}: {Customer} profiling in brand management.
\newblock \emph{International Journal of Information Management}, 51:\penalty0
  101934, April 2020.
\newblock ISSN 0268-4012.
\newblock \doi{10.1016/j.ijinfomgt.2019.04.007}.
\newblock URL
  \url{https://www.sciencedirect.com/science/article/pii/S0268401218313598}.

\bibitem[Gu et~al.(2021{\natexlab{a}})Gu, Liu, Ling, Liu, Chen, and
  Zhu]{gu_partner_2021}
Jia-Chen Gu, Hui Liu, Zhen-Hua Ling, Quan Liu, Zhigang Chen, and Xiaodan Zhu.
\newblock Partner {Matters}! {An} {Empirical} {Study} on {Fusing} {Personas}
  for {Personalized} {Response} {Selection} in {Retrieval}-{Based} {Chatbots}.
\newblock In \emph{Proceedings of the 44th {International} {ACM} {SIGIR}
  {Conference} on {Research} and {Development} in {Information} {Retrieval}},
  pages 565--574, Virtual Event Canada, July 2021{\natexlab{a}}. ACM.
\newblock ISBN 978-1-4503-8037-9.
\newblock \doi{10.1145/3404835.3462858}.
\newblock URL \url{https://dl.acm.org/doi/10.1145/3404835.3462858}.

\bibitem[Gu et~al.(2021{\natexlab{b}})Gu, Wang, Sun, Ye, Xu, Chen, and
  Zhang]{gu_exploiting_2021}
Jie Gu, Feng Wang, Qinghui Sun, Zhiquan Ye, Xiaoxiao Xu, Jingmin Chen, and Jun
  Zhang.
\newblock Exploiting {Behavioral} {Consistence} for {Universal} {User}
  {Representation}.
\newblock \emph{Proceedings of the AAAI Conference on Artificial Intelligence},
  35\penalty0 (5):\penalty0 4063--4071, May 2021{\natexlab{b}}.
\newblock ISSN 2374-3468, 2159-5399.
\newblock \doi{10.1609/aaai.v35i5.16527}.
\newblock URL \url{https://ojs.aaai.org/index.php/AAAI/article/view/16527}.

\bibitem[Gu et~al.(2020)Gu, Ding, Wang, and Yin]{gu_hierarchical_2020}
Yulong Gu, Zhuoye Ding, Shuaiqiang Wang, and Dawei Yin.
\newblock Hierarchical {User} {Profiling} for {E}-commerce {Recommender}
  {Systems}.
\newblock In \emph{Proceedings of the 13th {International} {Conference} on
  {Web} {Search} and {Data} {Mining}}, {WSDM} '20, pages 223--231, New York,
  NY, USA, January 2020. Association for Computing Machinery.
\newblock ISBN 978-1-4503-6822-3.
\newblock \doi{10.1145/3336191.3371827}.
\newblock URL \url{https://dl.acm.org/doi/10.1145/3336191.3371827}.

\bibitem[Gu et~al.(2021{\natexlab{c}})Gu, Bao, Ou, Li, Cui, Ma, Huang, Liu, and
  Zeng]{gu_self-supervised_2021}
Yulong Gu, Wentian Bao, Dan Ou, Xiang Li, Baoliang Cui, Biyu Ma, Haikuan Huang,
  Qingwen Liu, and Xiaoyi Zeng.
\newblock Self-{Supervised} {Learning} on {Users}' {Spontaneous} {Behaviors}
  for {Multi}-{Scenario} {Ranking} in {E}-commerce.
\newblock In \emph{Proceedings of the 30th {ACM} {International} {Conference}
  on {Information} \& {Knowledge} {Management}}, pages 3828--3837, Virtual
  Event Queensland Australia, October 2021{\natexlab{c}}. ACM.
\newblock ISBN 978-1-4503-8446-9.
\newblock \doi{10.1145/3459637.3481953}.
\newblock URL \url{https://dl.acm.org/doi/10.1145/3459637.3481953}.

\bibitem[Guan et~al.(2022)Guan, Jiao, Song, Wen, Yeh, and
  Chang]{guan_personalized_2022}
Weili Guan, Fangkai Jiao, Xuemeng Song, Haokun Wen, Chung-Hsing Yeh, and
  Xiaojun Chang.
\newblock Personalized {Fashion} {Compatibility} {Modeling} via
  {Metapath}-guided {Heterogeneous} {Graph} {Learning}.
\newblock In \emph{Proceedings of the 45th {International} {ACM} {SIGIR}
  {Conference} on {Research} and {Development} in {Information} {Retrieval}},
  pages 482--491, Madrid Spain, July 2022. ACM.
\newblock ISBN 978-1-4503-8732-3.
\newblock \doi{10.1145/3477495.3532038}.
\newblock URL \url{https://dl.acm.org/doi/10.1145/3477495.3532038}.

\bibitem[Guesmi et~al.(2022)Guesmi, Chatti, Vorgerd, Ngo, Joarder, Ain, and
  Muslim]{guesmi_explaining_2022}
Mouadh Guesmi, Mohamed~Amine Chatti, Laura Vorgerd, Thao Ngo, Shoeb Joarder,
  Qurat~Ul Ain, and Arham Muslim.
\newblock Explaining {User} {Models} with {Different} {Levels} of {Detail} for
  {Transparent} {Recommendation}: {A} {User} {Study}.
\newblock In \emph{Adjunct {Proceedings} of the 30th {ACM} {Conference} on
  {User} {Modeling}, {Adaptation} and {Personalization}}, {UMAP} '22 {Adjunct},
  pages 175--183, New York, NY, USA, 2022. Association for Computing Machinery.
\newblock ISBN 978-1-4503-9232-7.
\newblock \doi{10.1145/3511047.3537685}.
\newblock URL \url{https://dl.acm.org/doi/10.1145/3511047.3537685}.

\bibitem[Guo et~al.(2018{\natexlab{a}})Guo, Ma, and Wang]{guo_user_2018}
Ao~Guo, Jianhua Ma, and Kevin I-Kai Wang.
\newblock From {User} {Models} to the {Cyber}-{I} {Model}: {Approaches},
  {Progresses} and {Issues}.
\newblock In \emph{2018 {IEEE} 16th {Intl} {Conf} on {Dependable}, {Autonomic}
  and {Secure} {Computing}, 16th {Intl} {Conf} on {Pervasive} {Intelligence}
  and {Computing}, 4th {Intl} {Conf} on {Big} {Data} {Intelligence} and
  {Computing} and {Cyber} {Science} and {Technology}
  {Congress}({DASC}/{PiCom}/{DataCom}/{CyberSciTech})}, pages 33--40, Athens,
  August 2018{\natexlab{a}}. IEEE.
\newblock ISBN 978-1-5386-7518-2.
\newblock \doi{10.1109/DASC/PiCom/DataCom/CyberSciTec.2018.00021}.
\newblock URL \url{https://ieeexplore.ieee.org/document/8511864/}.

\bibitem[Guo et~al.(2018{\natexlab{b}})Guo, Cheng, Nie, Xu, and
  Kankanhalli]{guo_multi-modal_2018}
Yangyang Guo, Zhiyong Cheng, Liqiang Nie, Xin-Shun Xu, and Mohan Kankanhalli.
\newblock Multi-modal {Preference} {Modeling} for {Product} {Search}.
\newblock In \emph{Proceedings of the 26th {ACM} international conference on
  {Multimedia}}, {MM} '18, pages 1865--1873, New York, NY, USA,
  2018{\natexlab{b}}. Association for Computing Machinery.
\newblock ISBN 978-1-4503-5665-7.
\newblock \doi{10.1145/3240508.3240541}.
\newblock URL \url{https://dl.acm.org/doi/10.1145/3240508.3240541}.

\bibitem[Guo et~al.(2019)Guo, Cheng, Nie, Wang, Ma, and
  Kankanhalli]{guo_attentive_2019}
Yangyang Guo, Zhiyong Cheng, Liqiang Nie, Yinglong Wang, Jun Ma, and Mohan
  Kankanhalli.
\newblock Attentive {Long} {Short}-{Term} {Preference} {Modeling} for
  {Personalized} {Product} {Search}.
\newblock \emph{ACM Transactions on Information Systems}, 37\penalty0
  (2):\penalty0 19:1--19:27, 2019.
\newblock ISSN 1046-8188.
\newblock \doi{10.1145/3295822}.
\newblock URL \url{https://dl.acm.org/doi/10.1145/3295822}.

\bibitem[Gurbanov and Ricci(2017)]{gurbanov_action_2017}
Tural Gurbanov and Francesco Ricci.
\newblock Action prediction models for recommender systems based on
  collaborative filtering and sequence mining hybridization.
\newblock In \emph{Proceedings of the {Symposium} on {Applied} {Computing}},
  {SAC} '17, pages 1655--1661, New York, NY, USA, April 2017. Association for
  Computing Machinery.
\newblock ISBN 978-1-4503-4486-9.
\newblock \doi{10.1145/3019612.3019759}.
\newblock URL \url{https://dl.acm.org/doi/10.1145/3019612.3019759}.

\bibitem[Hadoux and Hunter(2018)]{hadoux_learning_2018}
E.~Hadoux and A.~Hunter.
\newblock Learning and {Updating} {User} {Models} for {Subpopulations} in
  {Persuasive} {Argumentation} {Using} {Beta} {Distribution}, July 2018.
\newblock URL \url{https://dl.acm.org/citation.cfm?id=3237865}.

\bibitem[Hamim et~al.(2022)Hamim, Benabbou, and Sael]{hamim_student_2022}
Touria Hamim, Faouzia Benabbou, and Nawal Sael.
\newblock Student {Profile} {Modeling} {Using} {Boosting} {Algorithms}.
\newblock \emph{International Journal of Web-Based Learning and Teaching
  Technologies (IJWLTT)}, 17\penalty0 (5):\penalty0 1--13, September 2022.
\newblock ISSN 1548-1093.
\newblock \doi{10.4018/IJWLTT.20220901.oa4}.
\newblock URL
  \url{https://www.igi-global.com/article/student-profile-modeling-using-boosting-algorithms/www.igi-global.com/article/student-profile-modeling-using-boosting-algorithms/284084}.

\bibitem[Han et~al.(2022)Han, Li, Cai, and Li]{han_multi-aggregator_2022}
Jinkun Han, Wei Li, Zhipeng Cai, and Yingshu Li.
\newblock Multi-{Aggregator} {Time}-{Warping} {Heterogeneous} {Graph} {Neural}
  {Network} for {Personalized} {Micro}-{Video} {Recommendation}.
\newblock In \emph{Proceedings of the 31st {ACM} {International} {Conference}
  on {Information} \& {Knowledge} {Management}}, pages 676--685, Atlanta GA
  USA, October 2022. ACM.
\newblock ISBN 978-1-4503-9236-5.
\newblock \doi{10.1145/3511808.3557403}.
\newblock URL \url{https://dl.acm.org/doi/10.1145/3511808.3557403}.

\bibitem[Harvey et~al.(2011)Harvey, Carman, Ruthven, and
  Crestani]{harvey_bayesian_2011}
Morgan Harvey, Mark~J. Carman, Ian Ruthven, and Fabio Crestani.
\newblock Bayesian latent variable models for collaborative item rating
  prediction.
\newblock In \emph{Proceedings of the 20th {ACM} international conference on
  {Information} and knowledge management}, {CIKM} '11, pages 699--708, New
  York, NY, USA, 2011. Association for Computing Machinery.
\newblock ISBN 978-1-4503-0717-8.
\newblock \doi{10.1145/2063576.2063680}.
\newblock URL \url{https://dl.acm.org/doi/10.1145/2063576.2063680}.

\bibitem[Hase and Bansal(2020)]{hase_evaluating_2020}
Peter Hase and Mohit Bansal.
\newblock Evaluating {Explainable} {AI}: {Which} {Algorithmic} {Explanations}
  {Help} {Users} {Predict} {Model} {Behavior}?
\newblock In Dan Jurafsky, Joyce Chai, Natalie Schluter, and Joel Tetreault,
  editors, \emph{Proceedings of the 58th {Annual} {Meeting} of the
  {Association} for {Computational} {Linguistics}}, pages 5540--5552, Online,
  July 2020. Association for Computational Linguistics.
\newblock \doi{10.18653/v1/2020.acl-main.491}.
\newblock URL \url{https://aclanthology.org/2020.acl-main.491}.

\bibitem[Hassan et~al.(2021)Hassan, Edmison, Stelter, and
  McCrickard]{hassan_learning_2021}
Taha Hassan, Bob Edmison, Timothy Stelter, and D.~Scott McCrickard.
\newblock Learning to {Trust}: {Understanding} {Editorial} {Authority} and
  {Trust} in {Recommender} {Systems} for {Education}.
\newblock In \emph{Proceedings of the 29th {ACM} {Conference} on {User}
  {Modeling}, {Adaptation} and {Personalization}}, pages 24--32, Utrecht
  Netherlands, June 2021. ACM.
\newblock ISBN 978-1-4503-8366-0.
\newblock \doi{10.1145/3450613.3456811}.
\newblock URL \url{https://dl.acm.org/doi/10.1145/3450613.3456811}.

\bibitem[He et~al.(2023)He, Liu, Guo, Qin, Zhang, Hu, and Tang]{he_survey_2023}
Zhicheng He, Weiwen Liu, Wei Guo, Jiarui Qin, Yingxue Zhang, Yaochen Hu, and
  Ruiming Tang.
\newblock A {Survey} on {User} {Behavior} {Modeling} in {Recommender}
  {Systems}, February 2023.
\newblock URL \url{http://arxiv.org/abs/2302.11087}.

\bibitem[Heidari et~al.(2020)Heidari, Jones, and Uzuner]{heidari_deep_2020}
Maryam Heidari, James~H Jones, and Ozlem Uzuner.
\newblock Deep {Contextualized} {Word} {Embedding} for {Text}-based {Online}
  {User} {Profiling} to {Detect} {Social} {Bots} on {Twitter}.
\newblock In \emph{2020 {International} {Conference} on {Data} {Mining}
  {Workshops} ({ICDMW})}, pages 480--487, November 2020.
\newblock \doi{10.1109/ICDMW51313.2020.00071}.

\bibitem[Hijikata(2004)]{hijikata_implicit_2004}
Yoshinori Hijikata.
\newblock Implicit user profiling for on demand relevance feedback.
\newblock In \emph{Proceedings of the 9th international conference on
  {Intelligent} user interfaces}, {IUI} '04, pages 198--205, New York, NY, USA,
  January 2004. Association for Computing Machinery.
\newblock ISBN 978-1-58113-815-3.
\newblock \doi{10.1145/964442.964480}.
\newblock URL \url{https://dl.acm.org/doi/10.1145/964442.964480}.

\bibitem[Hof et~al.(1998)Hof, Green, and Himelstein]{hof_now_1998}
Robert~D Hof, Heather Green, and Linda Himelstein.
\newblock Now it's your web.
\newblock \emph{Business Week}, pages 68--74, 1998.

\bibitem[Horvitz et~al.(1998)Horvitz, Breese, Heckerman, Hovel, and
  Rommelse]{horvitz_lumiere_1998}
Eric Horvitz, Jack Breese, David Heckerman, David Hovel, and Koos Rommelse.
\newblock The lumière project: {Bayesian} user modeling for inferring the
  goals and needs of software users.
\newblock In \emph{Proceedings of the {Fourteenth} conference on {Uncertainty}
  in artificial intelligence}, {UAI}'98, pages 256--265, San Francisco, CA,
  USA, 1998. Morgan Kaufmann Publishers Inc.
\newblock ISBN 978-1-55860-555-8.

\bibitem[Hu et~al.(2017)Hu, Jin, Zhang, Wang, and Yang]{hu_user_2017}
Jianqiao Hu, Feng Jin, Guigang Zhang, Jian Wang, and Yi~Yang.
\newblock A {User} {Profile} {Modeling} {Method} {Based} on {Word2Vec}.
\newblock In \emph{2017 {IEEE} {International} {Conference} on {Software}
  {Quality}, {Reliability} and {Security} {Companion} ({QRS}-{C})}, pages
  410--414, July 2017.
\newblock \doi{10.1109/QRS-C.2017.74}.
\newblock URL
  \url{https://ieeexplore.ieee.org/abstract/document/8004351?casa_token=LBqpMSiajuYAAAAA:eDejrNq_k8xXgSujFmX9zOZzFk3S3O2oNpGrWOqu0IC-Zr3eXts_D4dtTZwxbYrQETOuiip6wQ}.

\bibitem[Hu et~al.(2020)Hu, Li, Shi, Yang, and Shao]{hu_graph_2020}
Linmei Hu, Chen Li, Chuan Shi, Cheng Yang, and Chao Shao.
\newblock Graph neural news recommendation with long-term and short-term
  interest modeling.
\newblock \emph{Information Processing \& Management}, 57\penalty0
  (2):\penalty0 102142, March 2020.
\newblock ISSN 0306-4573.
\newblock \doi{10.1016/j.ipm.2019.102142}.
\newblock URL
  \url{https://www.sciencedirect.com/science/article/pii/S0306457319307800}.

\bibitem[Huang et~al.(2019)Huang, Fang, Qian, Sang, Li, and
  Xu]{huang_explainable_2019}
Xiaowen Huang, Quan Fang, Shengsheng Qian, Jitao Sang, Yan Li, and Changsheng
  Xu.
\newblock Explainable {Interaction}-driven {User} {Modeling} over {Knowledge}
  {Graph} for {Sequential} {Recommendation}.
\newblock In \emph{Proceedings of the 27th {ACM} {International} {Conference}
  on {Multimedia}}, {MM} '19, pages 548--556, New York, NY, USA, 2019.
  Association for Computing Machinery.
\newblock ISBN 978-1-4503-6889-6.
\newblock \doi{10.1145/3343031.3350893}.
\newblock URL \url{https://dl.acm.org/doi/10.1145/3343031.3350893}.

\bibitem[Huertas-García et~al.(2021)Huertas-García, Huertas-Tato, Martín,
  and Camacho]{huertas-garcia_profiling_2021}
Alvaro Huertas-García, Javier Huertas-Tato, Alejandro Martín, and David
  Camacho.
\newblock Profiling {Hate} {Speech} {Spreaders} on {Twitter}: {Transformers}
  and mixed pooling.
\newblock In \emph{Proceedings of the {Working} {Notes} of {CLEF} 2021 -
  {Conference} and {Labs} of the {Evaluation} {Forum}}, volume 2936, pages
  1772--1789, Bucharest, Romania, 2021. CEUR Workshop Proceedings.

\bibitem[Humann et~al.(2023)Humann, Fletcher, and Gerdes]{humann_modeling_2023}
James Humann, TaLena Fletcher, and John Gerdes.
\newblock Modeling, simulation, and trade-off analysis for multirobot,
  multioperator surveillance.
\newblock \emph{Systems Engineering}, 26\penalty0 (5):\penalty0 627--640, 2023.
\newblock ISSN 1520-6858.
\newblock \doi{10.1002/sys.21685}.
\newblock URL \url{https://onlinelibrary.wiley.com/doi/abs/10.1002/sys.21685}.

\bibitem[Hämäläinen et~al.(2023)Hämäläinen, Çelikok, and
  Kaski]{hamalainen_differentiable_2023}
Alex Hämäläinen, Mustafa~Mert Çelikok, and Samuel Kaski.
\newblock Differentiable user models.
\newblock In \emph{Proceedings of the {Thirty}-{Ninth} {Conference} on
  {Uncertainty} in {Artificial} {Intelligence}}, pages 798--808. PMLR, July
  2023.
\newblock URL \url{https://proceedings.mlr.press/v216/hamalainen23a.html}.

\bibitem[Isaak and Hanna(2018)]{isaak_user_2018}
Jim Isaak and Mina~J. Hanna.
\newblock User {Data} {Privacy}: {Facebook}, {Cambridge} {Analytica}, and
  {Privacy} {Protection}.
\newblock \emph{Computer}, 51\penalty0 (8):\penalty0 56--59, August 2018.
\newblock ISSN 1558-0814.
\newblock \doi{10.1109/MC.2018.3191268}.
\newblock URL \url{https://ieeexplore.ieee.org/abstract/document/8436400}.

\bibitem[Ishitaki et~al.(2017)Ishitaki, Obukata, Oda, and
  Barolli]{ishitaki_application_2017}
Taro Ishitaki, Ryoichiro Obukata, Tetsuya Oda, and Leonard Barolli.
\newblock Application of {Deep} {Recurrent} {Neural} {Networks} for
  {Prediction} of {User} {Behavior} in {Tor} {Networks}.
\newblock In \emph{2017 31st {International} {Conference} on {Advanced}
  {Information} {Networking} and {Applications} {Workshops} ({WAINA})}, pages
  238--243, March 2017.
\newblock \doi{10.1109/WAINA.2017.63}.
\newblock URL
  \url{https://ieeexplore.ieee.org/abstract/document/7929684?casa_token=9Iq9cw7U8csAAAAA:Ie_jgGdZHnR-3WUz8eDK3QyRdEH0c9XmrAp3O7-CCt_ejaP8mTWHBfu0Pcab3ebzftpuZVa-wRVE}.

\bibitem[Ivanov et~al.(2011)Ivanov, Riccardi, Sporka, and
  Franc]{ivanov_recognition_2011}
Alexei~V Ivanov, Giuseppe Riccardi, Adam~J Sporka, and Jakub Franc.
\newblock Recognition of personality traits from human spoken conversations.
\newblock In \emph{Twelfth annual conference of the international speech
  communication association}, 2011.

\bibitem[Jaber and McMillan(2022)]{jaber_cross-modal_2022}
Razan Jaber and Donald McMillan.
\newblock Cross-{Modal} {Repair}: {Gaze} and {Speech} {Interaction} for {List}
  {Advancement}.
\newblock In \emph{Proceedings of the 4th {Conference} on {Conversational}
  {User} {Interfaces}}, {CUI} '22, pages 1--11, New York, NY, USA, 2022.
  Association for Computing Machinery.
\newblock ISBN 978-1-4503-9739-1.
\newblock \doi{10.1145/3543829.3543833}.
\newblock URL \url{https://dl.acm.org/doi/10.1145/3543829.3543833}.

\bibitem[Javed et~al.(2021)Javed, Shaukat, Hameed, Iqbal, Alam, and
  Luo]{javed_review_2021}
Umair Javed, Kamran Shaukat, Ibrahim~A. Hameed, Farhat Iqbal, Talha~Mahboob
  Alam, and Suhuai Luo.
\newblock A {Review} of {Content}-{Based} and {Context}-{Based}
  {Recommendation} {Systems}.
\newblock \emph{International Journal of Emerging Technologies in Learning
  (iJET)}, 16\penalty0 (03):\penalty0 274--306, February 2021.
\newblock ISSN 1863-0383.
\newblock \doi{10.3991/ijet.v16i03.18851}.
\newblock URL
  \url{https://online-journals.org/index.php/i-jet/article/view/18851}.

\bibitem[Jiamthapthaksin and Aung(2017)]{jiamthapthaksin_user_2017}
Rachsuda Jiamthapthaksin and Than~Htike Aung.
\newblock User preferences profiling based on user behaviors on {Facebook} page
  categories.
\newblock In \emph{2017 9th {International} {Conference} on {Knowledge} and
  {Smart} {Technology} ({KST})}, pages 248--253, February 2017.
\newblock \doi{10.1109/KST.2017.7886077}.
\newblock URL
  \url{https://ieeexplore.ieee.org/abstract/document/7886077?casa_token=RN8VjN1AFygAAAAA:F6a0e3SMgYc_k20YKeSoF1bskeADnQXWx3XBiT-3nbHbmqaEA93hoMyZwJX_XVpb3-KLLcAgBA&signout=success}.

\bibitem[Jin et~al.(2020)Jin, Gao, He, Jin, and Li]{jin_multi-behavior_2020}
Bowen Jin, Chen Gao, Xiangnan He, Depeng Jin, and Yong Li.
\newblock Multi-behavior {Recommendation} with {Graph} {Convolutional}
  {Networks}.
\newblock In \emph{Proceedings of the 43rd {International} {ACM} {SIGIR}
  {Conference} on {Research} and {Development} in {Information} {Retrieval}},
  {SIGIR} '20, pages 659--668, Virtual Event China, 2020. Association for
  Computing Machinery.
\newblock ISBN 978-1-4503-8016-4.
\newblock \doi{10.1145/3397271.3401072}.
\newblock URL \url{https://dl.acm.org/doi/10.1145/3397271.3401072}.

\bibitem[Jin(2023)]{jin_mooc_2023}
Cong Jin.
\newblock {MOOC} student dropout prediction model based on learning behavior
  features and parameter optimization.
\newblock \emph{Interactive Learning Environments}, 31\penalty0 (2):\penalty0
  714--732, February 2023.
\newblock ISSN 1049-4820.
\newblock \doi{10.1080/10494820.2020.1802300}.
\newblock URL \url{https://doi.org/10.1080/10494820.2020.1802300}.

\bibitem[Jin et~al.(2013)Jin, Chen, Wang, Hui, and
  Vasilakos]{jin_understanding_2013}
Long Jin, Yang Chen, Tianyi Wang, Pan Hui, and Athanasios~V. Vasilakos.
\newblock Understanding user behavior in online social networks: a survey.
\newblock \emph{IEEE Communications Magazine}, 51\penalty0 (9):\penalty0
  144--150, September 2013.
\newblock ISSN 1558-1896.
\newblock \doi{10.1109/MCOM.2013.6588663}.
\newblock URL
  \url{https://ieeexplore.ieee.org/abstract/document/6588663?casa_token=BLPT_wxLkJYAAAAA:EbmGps059bb8icyPPhAVSFS4c4AZROZ9G3JIoJzWNppvqeFhe_5rRzJpbmKk1AcuGOY8v-yl4g}.

\bibitem[Jégou and Chevaillier(2018)]{jegou_computational_2018}
Mathieu Jégou and Pierre Chevaillier.
\newblock A computational model for the emergence of turn-taking behaviors in
  user-agent interactions.
\newblock \emph{Journal on Multimodal User Interfaces}, 12\penalty0
  (3):\penalty0 199--223, September 2018.
\newblock ISSN 1783-8738.
\newblock \doi{10.1007/s12193-018-0265-3}.
\newblock URL \url{https://doi.org/10.1007/s12193-018-0265-3}.

\bibitem[Kanoje et~al.(2015)Kanoje, Girase, and Mukhopadhyay]{kanoje_user_2015}
Sumitkumar Kanoje, Sheetal Girase, and Debajyoti Mukhopadhyay.
\newblock User {Profiling} {Trends}, {Techniques} and {Applications}, March
  2015.
\newblock URL \url{http://arxiv.org/abs/1503.07474}.

\bibitem[Karatzoglou et~al.(2018)Karatzoglou, Schnell, and
  Beigl]{karatzoglou_convolutional_2018}
Antonios Karatzoglou, Nikolai Schnell, and Michael Beigl.
\newblock A {Convolutional} {Neural} {Network} {Approach} for {Modeling}
  {Semantic} {Trajectories} and {Predicting} {Future} {Locations}.
\newblock In Věra Kůrková, Yannis Manolopoulos, Barbara Hammer, Lazaros
  Iliadis, and Ilias Maglogiannis, editors, \emph{Artificial {Neural}
  {Networks} and {Machine} {Learning} – {ICANN} 2018}, Lecture {Notes} in
  {Computer} {Science}, pages 61--72, Cham, 2018. Springer International
  Publishing.
\newblock ISBN 978-3-030-01418-6.
\newblock \doi{10.1007/978-3-030-01418-6_7}.

\bibitem[Kasper et~al.(2017)Kasper, de~Siqueira~Braga, Martins, and
  Hellingrath]{kasper_user_2017}
Gerrit Kasper, Diego de~Siqueira~Braga, Denis Mayr~Lima Martins, and Bernd
  Hellingrath.
\newblock User profile acquisition: {A} comprehensive framework to support
  personal information agents.
\newblock In \emph{2017 {IEEE} {Latin} {American} {Conference} on
  {Computational} {Intelligence} ({LA}-{CCI})}, pages 1--6, November 2017.
\newblock \doi{10.1109/LA-CCI.2017.8285719}.
\newblock URL \url{https://ieeexplore.ieee.org/document/8285719}.

\bibitem[Kaur et~al.(2018)Kaur, Singh, and Kumar]{kaur_authcom_2018}
Ravneet Kaur, Sarbjeet Singh, and Harish Kumar.
\newblock {AuthCom}: {Authorship} verification and compromised account
  detection in online social networks using {AHP}-{TOPSIS} embedded profiling
  based technique.
\newblock \emph{Expert Systems with Applications}, 113:\penalty0 397--414,
  December 2018.
\newblock ISSN 0957-4174.
\newblock \doi{10.1016/j.eswa.2018.07.011}.
\newblock URL
  \url{https://www.sciencedirect.com/science/article/pii/S0957417418304275}.

\bibitem[Kaushal et~al.(2019)Kaushal, Ghose, and
  Kumaraguru]{kaushal_methods_2019}
Rishabh Kaushal, Vasundhara Ghose, and Ponnurangam Kumaraguru.
\newblock Methods for {User} {Profiling} across {Social} {Networks}.
\newblock In \emph{2019 {IEEE} {Intl} {Conf} on {Parallel} \& {Distributed}
  {Processing} with {Applications}, {Big} {Data} \& {Cloud} {Computing},
  {Sustainable} {Computing} \& {Communications}, {Social} {Computing} \&
  {Networking} ({ISPA}/{BDCloud}/{SocialCom}/{SustainCom})}, pages 1572--1579,
  Xiamen, China, December 2019. IEEE.
\newblock ISBN 978-1-72814-328-6.
\newblock \doi{10.1109/ISPA-BDCloud-SustainCom-SocialCom48970.2019.00231}.
\newblock URL \url{https://ieeexplore.ieee.org/document/9047322/}.

\bibitem[Kay(1995)]{kay_toolkit_1995}
Judy Kay.
\newblock The um toolkit for reusable, long term user models.
\newblock \emph{User Modeling and User-Adapted Interaction}, 4\penalty0
  (3):\penalty0 149--196, 1995.

\bibitem[Kay et~al.(2002)Kay, Kummerfeld, and Lauder]{kay_personis_2002}
Judy Kay, Bob Kummerfeld, and Piers Lauder.
\newblock Personis: {A} {Server} for {User} {Models}.
\newblock In Gerhard Goos, Juris Hartmanis, Jan Van~Leeuwen, Paul De~Bra, Peter
  Brusilovsky, and Ricardo Conejo, editors, \emph{Adaptive {Hypermedia} and
  {Adaptive} {Web}-{Based} {Systems}}, volume 2347, pages 203--212. Springer
  Berlin Heidelberg, Berlin, Heidelberg, 2002.
\newblock ISBN 978-3-540-43737-6 978-3-540-47952-9.
\newblock URL \url{http://link.springer.com/10.1007/3-540-47952-X_22}.

\bibitem[Kellner and Berthold(2012)]{kellner_i-know_2012}
Gudrun Kellner and Marcel Berthold.
\newblock I-know my users: user-centric profiling based on the perceptual
  preference questionnaire ({PPQ}).
\newblock In \emph{Proceedings of the 12th {International} {Conference} on
  {Knowledge} {Management} and {Knowledge} {Technologies}}, i-{KNOW} '12, pages
  1--4, New York, NY, USA, September 2012. Association for Computing Machinery.
\newblock ISBN 978-1-4503-1242-4.
\newblock \doi{10.1145/2362456.2362493}.
\newblock URL \url{https://dl.acm.org/doi/10.1145/2362456.2362493}.

\bibitem[Kern et~al.(2008)Kern, Harding, Storz, Davis, and
  Schmidt]{kern_shaping_2008}
Dagmar Kern, Michael Harding, Oliver Storz, Nigel Davis, and Albrecht Schmidt.
\newblock Shaping how advertisers see me: user views on implicit and explicit
  profile capture.
\newblock In \emph{{CHI} '08 {Extended} {Abstracts} on {Human} {Factors} in
  {Computing} {Systems}}, {CHI} {EA} '08, pages 3363--3368, New York, NY, USA,
  April 2008. Association for Computing Machinery.
\newblock ISBN 978-1-60558-012-8.
\newblock \doi{10.1145/1358628.1358858}.
\newblock URL \url{https://dl.acm.org/doi/10.1145/1358628.1358858}.

\bibitem[Keurulainen et~al.(2023)Keurulainen, Westerlund, Keurulainen, and
  Howes]{keurulainen_amortised_2023}
Antti Keurulainen, Isak~Rafael Westerlund, Oskar Keurulainen, and Andrew Howes.
\newblock Amortised {Experimental} {Design} and {Parameter} {Estimation} for
  {User} {Models} of {Pointing}.
\newblock In \emph{Proceedings of the 2023 {CHI} {Conference} on {Human}
  {Factors} in {Computing} {Systems}}, {CHI} '23, pages 1--17, New York, NY,
  USA, April 2023. Association for Computing Machinery.
\newblock ISBN 978-1-4503-9421-5.
\newblock \doi{10.1145/3544548.3581483}.
\newblock URL \url{https://dl.acm.org/doi/10.1145/3544548.3581483}.

\bibitem[Kim et~al.(2011)Kim, Alkhaldi, El~Saddik, and
  Jo]{kim_collaborative_2011}
Heung-Nam Kim, Abdulmajeed Alkhaldi, Abdulmotaleb El~Saddik, and Geun-Sik Jo.
\newblock Collaborative user modeling with user-generated tags for social
  recommender systems.
\newblock \emph{Expert Systems with Applications}, 38\penalty0 (7):\penalty0
  8488--8496, July 2011.
\newblock ISSN 09574174.
\newblock \doi{10.1016/j.eswa.2011.01.048}.
\newblock URL
  \url{https://linkinghub.elsevier.com/retrieve/pii/S0957417411000686}.

\bibitem[Kim et~al.(2013)Kim, Lee, and Ryu]{kim_personality_2013}
Jieun Kim, Ahreum Lee, and Hokyoung Ryu.
\newblock Personality and its effects on learning performance: {Design}
  guidelines for an adaptive e-learning system based on a user model.
\newblock \emph{International Journal of Industrial Ergonomics}, 43\penalty0
  (5):\penalty0 450--461, September 2013.
\newblock ISSN 0169-8141.
\newblock \doi{10.1016/j.ergon.2013.03.001}.
\newblock URL
  \url{https://www.sciencedirect.com/science/article/pii/S0169814113000334}.

\bibitem[Kim et~al.(2023)Kim, Lee, Kim, Yang, and Park]{kim_task_2023}
Sein Kim, Namkyeong Lee, Donghyun Kim, Minchul Yang, and Chanyoung Park.
\newblock Task {Relation}-aware {Continual} {User} {Representation} {Learning}.
\newblock In \emph{Proceedings of the 29th {ACM} {SIGKDD} {Conference} on
  {Knowledge} {Discovery} and {Data} {Mining}}, {KDD} '23, pages 1107--1119,
  New York, NY, USA, 2023. Association for Computing Machinery.
\newblock ISBN 9798400701030.
\newblock \doi{10.1145/3580305.3599516}.
\newblock URL \url{https://dl.acm.org/doi/10.1145/3580305.3599516}.

\bibitem[Kobsa(1990)]{kobsa_modeling_1990}
Alfred Kobsa.
\newblock Modeling the user's conceptual knowledge in {BGP}-{MS}, a user
  modeling shell system1.
\newblock \emph{Computational Intelligence}, 6\penalty0 (4):\penalty0 193--208,
  1990.
\newblock ISSN 1467-8640.
\newblock \doi{10.1111/j.1467-8640.1990.tb00295.x}.
\newblock URL
  \url{https://onlinelibrary.wiley.com/doi/abs/10.1111/j.1467-8640.1990.tb00295.x}.

\bibitem[Kobsa(2001)]{kobsa_generic_2001}
Alfred Kobsa.
\newblock Generic {User} {Modeling} {Systems}.
\newblock \emph{User Modeling and User-Adapted Interaction}, 11\penalty0
  (1):\penalty0 49--63, March 2001.
\newblock ISSN 1573-1391.
\newblock \doi{10.1023/A:1011187500863}.
\newblock URL \url{https://doi.org/10.1023/A:1011187500863}.

\bibitem[Kobsa and Fink(2006)]{kobsa_ldap-based_2006}
Alfred Kobsa and Josef Fink.
\newblock An {LDAP}-based {User} {Modeling} {Server} and its {Evaluation}.
\newblock \emph{User Modeling and User-Adapted Interaction}, 16\penalty0
  (2):\penalty0 129--169, May 2006.
\newblock ISSN 0924-1868, 1573-1391.
\newblock \doi{10.1007/s11257-006-9006-5}.
\newblock URL \url{https://link.springer.com/10.1007/s11257-006-9006-5}.

\bibitem[Kobsa and Pohl(1994)]{kobsa_user_1994}
Alfred Kobsa and Wolfgang Pohl.
\newblock The user modeling shell system {BGP}-{MS}.
\newblock \emph{User Modeling and User-Adapted Interaction}, 4\penalty0
  (2):\penalty0 59--106, June 1994.
\newblock ISSN 1573-1391.
\newblock \doi{10.1007/BF01099428}.
\newblock URL \url{https://doi.org/10.1007/BF01099428}.

\bibitem[Kobsa and Schreck(2003)]{kobsa_privacy_2003}
Alfred Kobsa and Jörg Schreck.
\newblock Privacy through pseudonymity in user-adaptive systems.
\newblock \emph{ACM Transactions on Internet Technology}, 3\penalty0
  (2):\penalty0 149--183, 2003.
\newblock ISSN 1533-5399.
\newblock \doi{10.1145/767193.767196}.
\newblock URL \url{https://dl.acm.org/doi/10.1145/767193.767196}.

\bibitem[Kobsa et~al.(2001)Kobsa, Koenemann, and Pohl]{kobsa_personalised_2001}
Alfred Kobsa, Jürgen Koenemann, and Wolfgang Pohl.
\newblock Personalised hypermedia presentation techniques for improving online
  customer relationships.
\newblock \emph{The Knowledge Engineering Review}, 16\penalty0 (2):\penalty0
  111--155, March 2001.
\newblock ISSN 1469-8005, 0269-8889.
\newblock \doi{10.1017/S0269888901000108}.
\newblock URL
  \url{https://www.cambridge.org/core/journals/knowledge-engineering-review/article/personalised-hypermedia-presentation-techniques-for-improving-online-customer-relationships/FBFF2DE0C1AA93E2796F7B9809FE01D2}.

\bibitem[Konstan and Riedl(2012)]{konstan_recommender_2012}
Joseph~A. Konstan and John Riedl.
\newblock Recommender systems: from algorithms to user experience.
\newblock \emph{User Modeling and User-Adapted Interaction}, 22\penalty0
  (1):\penalty0 101--123, April 2012.
\newblock ISSN 1573-1391.
\newblock \doi{10.1007/s11257-011-9112-x}.
\newblock URL \url{https://doi.org/10.1007/s11257-011-9112-x}.

\bibitem[Konstan et~al.(1997)Konstan, Miller, Maltz, Herlocker, Gordon, and
  Riedl]{konstan_grouplens_1997}
Joseph~A. Konstan, Bradley~N. Miller, David Maltz, Jonathan~L. Herlocker,
  Lee~R. Gordon, and John Riedl.
\newblock {GroupLens}: applying collaborative filtering to {Usenet} news.
\newblock \emph{Communications of the ACM}, 40\penalty0 (3):\penalty0 77--87,
  March 1997.
\newblock ISSN 0001-0782, 1557-7317.
\newblock \doi{10.1145/245108.245126}.
\newblock URL \url{https://dl.acm.org/doi/10.1145/245108.245126}.

\bibitem[Kostric et~al.(2021)Kostric, Balog, and
  Radlinski]{kostric_soliciting_2021}
Ivica Kostric, Krisztian Balog, and Filip Radlinski.
\newblock Soliciting {User} {Preferences} in {Conversational} {Recommender}
  {Systems} via {Usage}-related {Questions}.
\newblock In \emph{Fifteenth {ACM} {Conference} on {Recommender} {Systems}},
  pages 724--729, Amsterdam Netherlands, September 2021. ACM.
\newblock ISBN 978-1-4503-8458-2.
\newblock \doi{10.1145/3460231.3478861}.
\newblock URL \url{https://dl.acm.org/doi/10.1145/3460231.3478861}.

\bibitem[Kota et~al.(2021)Kota, Duppada, Jindal, and
  Wadhwa]{kota_understanding_2021}
Nagaraj Kota, Venkatesh Duppada, Ashvini Jindal, and Mohit Wadhwa.
\newblock Understanding {Job} {Seeker} {Funnel} for {Search} and {Discovery}
  {Personalization}.
\newblock In \emph{Proceedings of the 30th {ACM} {International} {Conference}
  on {Information} \& {Knowledge} {Management}}, pages 3888--3897, Virtual
  Event Queensland Australia, October 2021. ACM.
\newblock ISBN 978-1-4503-8446-9.
\newblock \doi{10.1145/3459637.3481959}.
\newblock URL \url{https://dl.acm.org/doi/10.1145/3459637.3481959}.

\bibitem[Krishnan and Kamath(2017)]{krishnan_dynamic_2017}
Gokul~S Krishnan and S~Sowmya Kamath.
\newblock Dynamic and temporal user profiling for personalized recommenders
  using heterogeneous data sources.
\newblock In \emph{2017 8th {International} {Conference} on {Computing},
  {Communication} and {Networking} {Technologies} ({ICCCNT})}, pages 1--7, July
  2017.
\newblock \doi{10.1109/ICCCNT.2017.8203963}.
\newblock URL \url{https://ieeexplore.ieee.org/document/8203963}.

\bibitem[Krulwich(1997)]{krulwich_lifestyle_1997}
Bruce Krulwich.
\newblock {LIFESTYLE} {FINDER}: {Intelligent} {User} {Profiling} {Using}
  {Large}-{Scale} {Demographic} {Data}.
\newblock \emph{AI Magazine}, 18\penalty0 (2):\penalty0 37--37, June 1997.
\newblock ISSN 2371-9621.
\newblock \doi{10.1609/aimag.v18i2.1292}.
\newblock URL
  \url{https://ojs.aaai.org/aimagazine/index.php/aimagazine/article/view/1292}.

\bibitem[Kuflik et~al.(2012)Kuflik, Kay, and
  Kummerfeld]{kuflik_challenges_2012}
Tsvi Kuflik, Judy Kay, and Bob Kummerfeld.
\newblock Challenges and {Solutions} of {Ubiquitous} {User} {Modeling}.
\newblock In Antonio Krüger and Tsvi Kuflik, editors, \emph{Ubiquitous
  {Display} {Environments}}, Cognitive {Technologies}, pages 7--30. Springer,
  Berlin, Heidelberg, 2012.
\newblock ISBN 978-3-642-27663-7.
\newblock \doi{10.1007/978-3-642-27663-7_2}.
\newblock URL \url{https://doi.org/10.1007/978-3-642-27663-7_2}.

\bibitem[Kulkarni et~al.(2019)Kulkarni, Kabra, and
  Shankarmani]{kulkarni_user_2019}
Tanay Kulkarni, Madhur Kabra, and Radha Shankarmani.
\newblock User {Profiling} {Based} {Recommendation} {System} for
  {E}-{Learning}.
\newblock In \emph{2019 {IEEE} 16th {India} {Council} {International}
  {Conference} ({INDICON})}, pages 1--4, December 2019.
\newblock \doi{10.1109/INDICON47234.2019.9028982}.
\newblock URL \url{https://ieeexplore.ieee.org/document/9028982}.

\bibitem[Kwon et~al.(2021)Kwon, Lee, and Jeong]{kwon_user_2021}
Hongkyun Kwon, Sangjin Lee, and Doowon Jeong.
\newblock User profiling via application usage pattern on digital devices for
  digital forensics.
\newblock \emph{Expert Systems with Applications}, 168:\penalty0 114488, April
  2021.
\newblock ISSN 09574174.
\newblock \doi{10.1016/j.eswa.2020.114488}.
\newblock URL
  \url{https://linkinghub.elsevier.com/retrieve/pii/S0957417420311349}.

\bibitem[Lakiotaki et~al.(2011)Lakiotaki, Matsatsinis, and
  Tsoukias]{lakiotaki_multicriteria_2011}
Kleanthi Lakiotaki, Nikolaos~F. Matsatsinis, and Alexis Tsoukias.
\newblock Multicriteria {User} {Modeling} in {Recommender} {Systems}.
\newblock \emph{IEEE Intelligent Systems}, 26\penalty0 (2):\penalty0 64--76,
  March 2011.
\newblock ISSN 1541-1672.
\newblock \doi{10.1109/MIS.2011.33}.
\newblock URL \url{http://ieeexplore.ieee.org/document/5751215/}.

\bibitem[Langley(1999)]{langley_user_1999}
Pat Langley.
\newblock User {Modeling} in {Adaptive} {Interfaces}.
\newblock In \emph{Proceedings of the {Seventh} {International} {Conference} on
  {User} {Modeling}}, 1999.
\newblock \doi{10.1007/978-3-7091-2490-1_48}.

\bibitem[Lashkari et~al.(2019)Lashkari, Chen, and
  Ghorbani]{lashkari_survey_2019}
Arash~Habibi Lashkari, Min Chen, and Ali~A. Ghorbani.
\newblock A {Survey} on {User} {Profiling} {Model} for {Anomaly} {Detection} in
  {Cyberspace}.
\newblock \emph{Journal of Cyber Security and Mobility}, pages 75--112, 2019.
\newblock ISSN 2245-4578.
\newblock \doi{10.13052/2245-1439.814}.
\newblock URL \url{https://journals.riverpublishers.com/index.php/JCSANDM}.

\bibitem[Li et~al.(2009)Li, Lau, and Dharmendran]{li_three-tier_2009}
Frederick W.~B. Li, Rynson W.~H. Lau, and Parthiban Dharmendran.
\newblock A {Three}-{Tier} {Profiling} {Framework} for {Adaptive} e-{Learning}.
\newblock In Marc Spaniol, Qing Li, Ralf Klamma, and Rynson W.~H. Lau, editors,
  \emph{Advances in {Web} {Based} {Learning} – {ICWL} 2009}, Lecture {Notes}
  in {Computer} {Science}, pages 235--244, Berlin, Heidelberg, 2009. Springer.
\newblock ISBN 978-3-642-03426-8.
\newblock \doi{10.1007/978-3-642-03426-8_30}.

\bibitem[Li et~al.(2022{\natexlab{a}})Li, Dong, Cheng, and
  Mo]{li_hierarchical_2022}
Hai Li, Xin Dong, Lei Cheng, and Linjian Mo.
\newblock A {Hierarchical} {User} {Behavior} {Modeling} {Framework} for
  {Cross}-{Domain} {Click}-{Through} {Rate} {Prediction}.
\newblock In \emph{Proceedings of the 31st {ACM} {International} {Conference}
  on {Information} \& {Knowledge} {Management}}, pages 4163--4167, Atlanta GA
  USA, October 2022{\natexlab{a}}. ACM.
\newblock ISBN 978-1-4503-9236-5.
\newblock \doi{10.1145/3511808.3557531}.
\newblock URL \url{https://dl.acm.org/doi/10.1145/3511808.3557531}.

\bibitem[Li et~al.(2023{\natexlab{a}})Li, Sun, Wang, Ma, Li, Zhang, Feng, and
  Xue]{li_intent-aware_2023}
Jiayu Li, Peijie Sun, Zhefan Wang, Weizhi Ma, Yangkun Li, Min Zhang, Zhoutian
  Feng, and Daiyue Xue.
\newblock Intent-aware {Ranking} {Ensemble} for {Personalized}
  {Recommendation}, April 2023{\natexlab{a}}.
\newblock URL \url{http://arxiv.org/abs/2304.07450}.

\bibitem[Li et~al.(2014)Li, Ritter, and Hovy]{li_weakly_2014}
Jiwei Li, Alan Ritter, and Eduard Hovy.
\newblock Weakly {Supervised} {User} {Profile} {Extraction} from {Twitter}.
\newblock In Kristina Toutanova and Hua Wu, editors, \emph{Proceedings of the
  52nd {Annual} {Meeting} of the {Association} for {Computational}
  {Linguistics} ({Volume} 1: {Long} {Papers})}, pages 165--174, Baltimore,
  Maryland, June 2014. Association for Computational Linguistics.
\newblock \doi{10.3115/v1/P14-1016}.
\newblock URL \url{https://aclanthology.org/P14-1016}.

\bibitem[Li et~al.(2007)Li, Yang, Wang, and Kitsuregawa]{li_dynamic_2007}
Lin Li, Zhenglu Yang, Botao Wang, and Masaru Kitsuregawa.
\newblock Dynamic {Adaptation} {Strategies} for {Long}-{Term} and
  {Short}-{Term} {User} {Profile} to {Personalize} {Search}.
\newblock In Guozhu Dong, Xuemin Lin, Wei Wang, Yun Yang, and Jeffrey~Xu Yu,
  editors, \emph{Advances in {Data} and {Web} {Management}}, Lecture {Notes} in
  {Computer} {Science}, pages 228--240, Berlin, Heidelberg, 2007. Springer.
\newblock ISBN 978-3-540-72524-4.
\newblock \doi{10.1007/978-3-540-72524-4_26}.

\bibitem[Li et~al.(2022{\natexlab{b}})Li, Han, Sheng, Ma, Kong, Liu, and
  Mao]{li_novel_2022}
Ming Li, Xingwang Han, Hua Sheng, Lin Ma, Hanzhang Kong, Weite Liu, and Bo~Mao.
\newblock A {Novel} {RNN} {Model} with {Enhanced} {Behavior} {Semantic} for
  {Network} {User} {Profile}.
\newblock In \emph{2022 {Tenth} {International} {Conference} on {Advanced}
  {Cloud} and {Big} {Data} ({CBD})}, pages 190--193, November
  2022{\natexlab{b}}.
\newblock \doi{10.1109/CBD58033.2022.00041}.
\newblock URL \url{https://ieeexplore.ieee.org/document/10024536}.

\bibitem[Li and Zhao(2020)]{li_survey_2020}
Sheng Li and Handong Zhao.
\newblock A {Survey} on {Representation} {Learning} for {User} {Modeling}.
\newblock In \emph{Proceedings of the {Twenty}-{Ninth} {International} {Joint}
  {Conference} on {Artificial} {Intelligence}}, pages 4997--5003, Yokohama,
  Japan, July 2020. International Joint Conferences on Artificial Intelligence
  Organization.
\newblock ISBN 978-0-9992411-6-5.
\newblock \doi{10.24963/ijcai.2020/695}.
\newblock URL \url{https://www.ijcai.org/proceedings/2020/695}.

\bibitem[Li et~al.(2022{\natexlab{c}})Li, Xie, Zhu, Ao, Zhuang, and
  He]{li_user-centric_2022}
Shuokai Li, Ruobing Xie, Yongchun Zhu, Xiang Ao, Fuzhen Zhuang, and Qing He.
\newblock User-{Centric} {Conversational} {Recommendation} with
  {Multi}-{Aspect} {User} {Modeling}.
\newblock In \emph{Proceedings of the 45th {International} {ACM} {SIGIR}
  {Conference} on {Research} and {Development} in {Information} {Retrieval}},
  pages 223--233, Madrid Spain, July 2022{\natexlab{c}}. ACM.
\newblock ISBN 978-1-4503-8732-3.
\newblock \doi{10.1145/3477495.3532074}.
\newblock URL \url{https://dl.acm.org/doi/10.1145/3477495.3532074}.

\bibitem[Li et~al.(2023{\natexlab{b}})Li, Zheng, Xiao, and Wang]{li_stan_2023}
Wanda Li, Wenhao Zheng, Xuanji Xiao, and Suhang Wang.
\newblock {STAN}: {Stage}-{Adaptive} {Network} for {Multi}-{Task}
  {Recommendation} by {Learning} {User} {Lifecycle}-{Based} {Representation},
  June 2023{\natexlab{b}}.
\newblock URL \url{http://arxiv.org/abs/2306.12232}.

\bibitem[Liang and De~Rijke(2016)]{liang_formal_2016}
Shangsong Liang and Maarten De~Rijke.
\newblock Formal language models for finding groups of experts.
\newblock \emph{Information Processing \& Management}, 52\penalty0
  (4):\penalty0 529--549, July 2016.
\newblock ISSN 03064573.
\newblock \doi{10.1016/j.ipm.2015.11.005}.
\newblock URL
  \url{https://linkinghub.elsevier.com/retrieve/pii/S0306457315001405}.

\bibitem[Lin et~al.(2019)Lin, Gao, and Li]{lin_cross_2019}
Tzu-Heng Lin, Chen Gao, and Yong Li.
\newblock {CROSS}: {Cross}-platform {Recommendation} for {Social}
  {E}-{Commerce}.
\newblock In \emph{Proceedings of the 42nd {International} {ACM} {SIGIR}
  {Conference} on {Research} and {Development} in {Information} {Retrieval}},
  {SIGIR}'19, pages 515--524, New York, NY, USA, 2019. Association for
  Computing Machinery.
\newblock ISBN 978-1-4503-6172-9.
\newblock \doi{10.1145/3331184.3331191}.
\newblock URL \url{https://dl.acm.org/doi/10.1145/3331184.3331191}.

\bibitem[Liu et~al.(2023{\natexlab{a}})Liu, Li, Gu, Lu, Zhang, Shang, and
  Gu]{liu_triple_2023}
Jiahao Liu, Dongsheng Li, Hansu Gu, Tun Lu, Peng Zhang, Li~Shang, and Ning Gu.
\newblock Triple {Structural} {Information} {Modelling} for {Accurate},
  {Explainable} and {Interactive} {Recommendation}, April 2023{\natexlab{a}}.
\newblock URL \url{http://arxiv.org/abs/2304.11528}.

\bibitem[Liu et~al.(2023{\natexlab{b}})Liu, Chen, Huang, Li, and
  Min]{liu_gnn-based_2023}
Jinbo Liu, Yunliang Chen, Xiaohui Huang, Jianxin Li, and Geyong Min.
\newblock {GNN}-based long and short term preference modeling for next-location
  prediction.
\newblock \emph{Information Sciences}, 629:\penalty0 1--14, June
  2023{\natexlab{b}}.
\newblock ISSN 00200255.
\newblock \doi{10.1016/j.ins.2023.01.131}.
\newblock URL
  \url{https://linkinghub.elsevier.com/retrieve/pii/S0020025523001433}.

\bibitem[Liu et~al.(2007)Liu, Chen, Bu, Chen, and Zhang]{liu_user_2007}
Kangmiao Liu, Wei Chen, Jiajun Bu, Chun Chen, and Lijun Zhang.
\newblock User {Modeling} for {Recommendation} in {Blogspace}.
\newblock In \emph{2007 {IEEE}/{WIC}/{ACM} {International} {Conferences} on
  {Web} {Intelligence} and {Intelligent} {Agent} {Technology} - {Workshops}},
  pages 79--82, November 2007.
\newblock \doi{10.1109/WI-IATW.2007.23}.
\newblock URL
  \url{https://ieeexplore.ieee.org/abstract/document/4427544?casa_token=CSJ-JvzE2IcAAAAA:fnGYlRgTPH2l-2A6BUyuYwmYlaUrV1OCcJjtJrNrJzjMqq8aQ6xSRuSqfSJR9T-agEhn_mOPhw}.

\bibitem[Liu et~al.(2023{\natexlab{c}})Liu, Wu, Huang, Wang, Ning, Chen, Chen,
  Yi, and Zhou]{liu_federated_2023}
Qi~Liu, Jinze Wu, Zhenya Huang, Hao Wang, Yuting Ning, Ming Chen, Enhong Chen,
  Jinfeng Yi, and Bowen Zhou.
\newblock Federated {User} {Modeling} from {Hierarchical} {Information}.
\newblock \emph{ACM Transactions on Information Systems}, 41\penalty0
  (2):\penalty0 46:1--46:33, April 2023{\natexlab{c}}.
\newblock ISSN 1046-8188.
\newblock \doi{10.1145/3560485}.
\newblock URL \url{https://dl.acm.org/doi/10.1145/3560485}.

\bibitem[Liu(2015)]{liu_modeling_2015}
Xin Liu.
\newblock Modeling users' dynamic preference for personalized recommendation.
\newblock In \emph{Proceedings of the 24th {International} {Conference} on
  {Artificial} {Intelligence}}, {IJCAI}'15, pages 1785--1791, Buenos Aires,
  Argentina, 2015. AAAI Press.
\newblock ISBN 978-1-57735-738-4.

\bibitem[Logesh et~al.(2019)Logesh, Subramaniyaswamy, Vijayakumar, and
  Li]{logesh_efficient_2019}
R.~Logesh, V.~Subramaniyaswamy, V.~Vijayakumar, and Xiong Li.
\newblock Efficient {User} {Profiling} {Based} {Intelligent} {Travel}
  {Recommender} {System} for {Individual} and {Group} of {Users}.
\newblock \emph{Mobile Networks and Applications}, 24\penalty0 (3):\penalty0
  1018--1033, June 2019.
\newblock ISSN 1572-8153.
\newblock \doi{10.1007/s11036-018-1059-2}.
\newblock URL \url{https://doi.org/10.1007/s11036-018-1059-2}.

\bibitem[Lu et~al.(2021)Lu, Ma, Zhang, De~Rijke, Liu, and Ma]{lu_standing_2021}
Hongyu Lu, Weizhi Ma, Min Zhang, Maarten De~Rijke, Yiqun Liu, and Shaoping Ma.
\newblock Standing in {Your} {Shoes}: {External} {Assessments} for
  {Personalized} {Recommender} {Systems}.
\newblock In \emph{Proceedings of the 44th {International} {ACM} {SIGIR}
  {Conference} on {Research} and {Development} in {Information} {Retrieval}},
  pages 1523--1533, Virtual Event Canada, July 2021. ACM.
\newblock ISBN 978-1-4503-8037-9.
\newblock \doi{10.1145/3404835.3462916}.
\newblock URL \url{https://dl.acm.org/doi/10.1145/3404835.3462916}.

\bibitem[Lugo et~al.(2021)Lugo, Moreno, and Hubert]{lugo_modeling_2021}
Luis Lugo, Jose~G. Moreno, and Gilles Hubert.
\newblock Modeling {User} {Search} {Tasks} with a {Language}-{Agnostic}
  {Unsupervised} {Approach}.
\newblock In Djoerd Hiemstra, Marie-Francine Moens, Josiane Mothe, Raffaele
  Perego, Martin Potthast, and Fabrizio Sebastiani, editors, \emph{Advances in
  {Information} {Retrieval}}, Lecture {Notes} in {Computer} {Science}, pages
  405--418, Cham, 2021. Springer International Publishing.
\newblock ISBN 978-3-030-72113-8.
\newblock \doi{10.1007/978-3-030-72113-8_27}.

\bibitem[Luo et~al.(2022)Luo, Xiao, and Song]{luo_personalized_2022}
Sichun Luo, Yuanzhang Xiao, and Linqi Song.
\newblock Personalized {Federated} {Recommendation} via {Joint}
  {Representation} {Learning}, {User} {Clustering}, and {Model} {Adaptation}.
\newblock In \emph{Proceedings of the 31st {ACM} {International} {Conference}
  on {Information} \& {Knowledge} {Management}}, pages 4289--4293, Atlanta GA
  USA, October 2022. ACM.
\newblock ISBN 978-1-4503-9236-5.
\newblock \doi{10.1145/3511808.3557668}.
\newblock URL \url{https://dl.acm.org/doi/10.1145/3511808.3557668}.

\bibitem[Luo et~al.(2014)Luo, Xu, Cai, and Bu]{luo_hybrid_2014}
Yang Luo, Boyi Xu, Hongming Cai, and Fenglin Bu.
\newblock A {Hybrid} {User} {Profile} {Model} for {Personalized} {Recommender}
  {System} with {Linked} {Open} {Data}.
\newblock In \emph{2014 {Enterprise} {Systems} {Conference}}, pages 243--248,
  August 2014.
\newblock \doi{10.1109/ES.2014.16}.
\newblock URL
  \url{https://ieeexplore.ieee.org/abstract/document/6997053?casa_token=_xtD5qA_nO4AAAAA:9HSZ6x8CnenAC0zBxTXKmvWHPlPgnlSiWSC8p0DWTZFEXfAcDpMYcJE3A6MB2plDw4FvITPtKA}.

\bibitem[Ma et~al.(2022{\natexlab{a}})Ma, Ren, Castells, and
  Sanderson]{ma_nest_2022}
Chenglong Ma, Yongli Ren, Pablo Castells, and Mark Sanderson.
\newblock {NEST}: {Simulating} {Pandemic}-like {Events} for {Collaborative}
  {Filtering} by {Modeling} {User} {Needs} {Evolution}.
\newblock In \emph{Proceedings of the 31st {ACM} {International} {Conference}
  on {Information} \& {Knowledge} {Management}}, pages 1430--1440, Atlanta GA
  USA, October 2022{\natexlab{a}}. ACM.
\newblock ISBN 978-1-4503-9236-5.
\newblock \doi{10.1145/3511808.3557407}.
\newblock URL \url{https://dl.acm.org/doi/10.1145/3511808.3557407}.

\bibitem[Ma et~al.(2022{\natexlab{b}})Ma, Liu, Yuan, Yang, and
  Zhang]{ma_caen_2022}
Rui Ma, Ning Liu, Jingsong Yuan, Huafeng Yang, and Jiandong Zhang.
\newblock {CAEN}: {A} {Hierarchically} {Attentive} {Evolution} {Network} for
  {Item}-{Attribute}-{Change}-{Aware} {Recommendation} in the {Growing}
  {E}-commerce {Environment}.
\newblock In \emph{Proceedings of the 16th {ACM} {Conference} on {Recommender}
  {Systems}}, pages 278--287, Seattle WA USA, September 2022{\natexlab{b}}.
  ACM.
\newblock ISBN 978-1-4503-9278-5.
\newblock \doi{10.1145/3523227.3546773}.
\newblock URL \url{https://dl.acm.org/doi/10.1145/3523227.3546773}.

\bibitem[Ma et~al.(2022{\natexlab{c}})Ma, Yang, Gao, and Xu]{ma_model_2022}
Xinhong Ma, Xiaoshan Yang, Junyu Gao, and Changsheng Xu.
\newblock The {Model} {May} {Fit} {You}: {User}-{Generalized} {Cross}-{Modal}
  {Retrieval}.
\newblock \emph{IEEE Transactions on Multimedia}, 24:\penalty0 2998--3012,
  2022{\natexlab{c}}.
\newblock ISSN 1941-0077.
\newblock \doi{10.1109/TMM.2021.3091888}.
\newblock URL \url{https://ieeexplore.ieee.org/document/9465686}.

\bibitem[Ma et~al.(2020)Ma, Dou, Bian, and Wen]{ma_pstie_2020}
Zhengyi Ma, Zhicheng Dou, Guanyue Bian, and Ji-Rong Wen.
\newblock {PSTIE}: {Time} {Information} {Enhanced} {Personalized} {Search}.
\newblock In \emph{Proceedings of the 29th {ACM} {International} {Conference}
  on {Information} \& {Knowledge} {Management}}, pages 1075--1084, Virtual
  Event Ireland, October 2020. ACM.
\newblock ISBN 978-1-4503-6859-9.
\newblock \doi{10.1145/3340531.3411877}.
\newblock URL \url{https://dl.acm.org/doi/10.1145/3340531.3411877}.

\bibitem[Ma et~al.(2021)Ma, Dou, Zhu, Zhong, and Wen]{ma_one_2021}
Zhengyi Ma, Zhicheng Dou, Yutao Zhu, Hanxun Zhong, and Ji-Rong Wen.
\newblock One {Chatbot} {Per} {Person}: {Creating} {Personalized} {Chatbots}
  based on {Implicit} {User} {Profiles}.
\newblock In \emph{Proceedings of the 44th {International} {ACM} {SIGIR}
  {Conference} on {Research} and {Development} in {Information} {Retrieval}},
  pages 555--564, Virtual Event Canada, July 2021. ACM.
\newblock ISBN 978-1-4503-8037-9.
\newblock \doi{10.1145/3404835.3462828}.
\newblock URL \url{https://dl.acm.org/doi/10.1145/3404835.3462828}.

\bibitem[Madsen et~al.(2019)Madsen, Bailey, Carrella, and
  Koralus]{madsen_analytic_2019}
Jens~Koed Madsen, Richard Bailey, Ernesto Carrella, and Philipp Koralus.
\newblock Analytic {Versus} {Computational} {Cognitive} {Models}:
  {Agent}-{Based} {Modeling} as a {Tool} in {Cognitive} {Sciences}.
\newblock \emph{Current Directions in Psychological Science}, 28\penalty0
  (3):\penalty0 299--305, June 2019.
\newblock ISSN 0963-7214.
\newblock \doi{10.1177/0963721419834547}.
\newblock URL \url{https://doi.org/10.1177/0963721419834547}.

\bibitem[Mairesse and Walker(2010)]{mairesse_towards_2010}
François Mairesse and Marilyn~A. Walker.
\newblock Towards personality-based user adaptation: psychologically informed
  stylistic language generation.
\newblock \emph{User Modeling and User-Adapted Interaction}, 20\penalty0
  (3):\penalty0 227--278, August 2010.
\newblock ISSN 0924-1868, 1573-1391.
\newblock \doi{10.1007/s11257-010-9076-2}.
\newblock URL \url{http://link.springer.com/10.1007/s11257-010-9076-2}.

\bibitem[Majumder et~al.(2019)Majumder, Li, Ni, and
  McAuley]{majumder_generating_2019}
Bodhisattwa~Prasad Majumder, Shuyang Li, Jianmo Ni, and Julian McAuley.
\newblock Generating {Personalized} {Recipes} from {Historical} {User}
  {Preferences}, August 2019.
\newblock URL \url{http://arxiv.org/abs/1909.00105}.

\bibitem[Malinowski and Zimányi(2006)]{malinowski_hierarchies_2006}
E.~Malinowski and E.~Zimányi.
\newblock Hierarchies in a multidimensional model: {From} conceptual modeling
  to logical representation.
\newblock \emph{Data \& Knowledge Engineering}, 59\penalty0 (2):\penalty0
  348--377, November 2006.
\newblock ISSN 0169-023X.
\newblock \doi{10.1016/j.datak.2005.08.003}.
\newblock URL
  \url{https://www.sciencedirect.com/science/article/pii/S0169023X0500145X}.

\bibitem[Martinez et~al.(2009)Martinez, Arias, Vilas, Garcia~Duque, and
  Lopez~Nores]{martinez_whats_2009}
Ana Belen~Barragans Martinez, Jose J.~Pazos Arias, Ana~Fernandez Vilas, Jorge
  Garcia~Duque, and Martin Lopez~Nores.
\newblock What's on {TV} tonight? {An} efficient and effective personalized
  recommender system of {TV} programs.
\newblock \emph{IEEE Transactions on Consumer Electronics}, 55\penalty0
  (1):\penalty0 286--294, February 2009.
\newblock ISSN 1558-4127.
\newblock \doi{10.1109/TCE.2009.4814447}.
\newblock URL
  \url{https://ieeexplore.ieee.org/abstract/document/4814447?casa_token=_cS7kDsAFe0AAAAA:fpgwqYT9CfzLb2s-GpOxdIUKIrERDJh8sxUJYd8NdBwvyU2eEQNfcuPiD9cPyI_QkCtJUsV8aw}.

\bibitem[Masthoff(2004)]{masthoff_group_2004}
Judith Masthoff.
\newblock Group {Modeling}: {Selecting} a {Sequence} of {Television} {Items} to
  {Suit} a {Group} of {Viewers}.
\newblock In \emph{Personalized {Digital} {Television}: {Targeting} {Programs}
  to individual {Viewers}}, Human-{Computer} {Interaction} {Series}, pages
  93--141. Springer Netherlands, Dordrecht, 2004.
\newblock ISBN 978-1-4020-2164-0.
\newblock URL \url{https://doi.org/10.1007/1-4020-2164-X_5}.

\bibitem[Masthoff(2011)]{masthoff_group_2011}
Judith Masthoff.
\newblock Group {Recommender} {Systems}: {Combining} {Individual} {Models}.
\newblock In Francesco Ricci, Lior Rokach, Bracha Shapira, and Paul~B. Kantor,
  editors, \emph{Recommender {Systems} {Handbook}}, pages 677--702. Springer
  US, Boston, MA, 2011.
\newblock ISBN 978-0-387-85820-3.
\newblock \doi{10.1007/978-0-387-85820-3_21}.
\newblock URL \url{https://doi.org/10.1007/978-0-387-85820-3_21}.

\bibitem[McCrae and John(1992)]{mccrae_introduction_1992}
Robert~R. McCrae and Oliver~P. John.
\newblock An {Introduction} to the {Five}-{Factor} {Model} and {Its}
  {Applications}.
\newblock \emph{Journal of Personality}, 60\penalty0 (2):\penalty0 175--215,
  1992.
\newblock ISSN 1467-6494.
\newblock \doi{10.1111/j.1467-6494.1992.tb00970.x}.
\newblock URL
  \url{https://onlinelibrary.wiley.com/doi/abs/10.1111/j.1467-6494.1992.tb00970.x}.

\bibitem[Medina-Medina and García-Cabrera(2016)]{medina-medina_taxonomy_2016}
Nuria Medina-Medina and Lina García-Cabrera.
\newblock A taxonomy for user models in adaptive systems: special
  considerations for learning environments.
\newblock \emph{The Knowledge Engineering Review}, 31\penalty0 (2):\penalty0
  124--141, March 2016.
\newblock ISSN 0269-8889, 1469-8005.
\newblock \doi{10.1017/S0269888916000035}.
\newblock URL
  \url{https://www.cambridge.org/core/journals/knowledge-engineering-review/article/taxonomy-for-user-models-in-adaptive-systems-special-considerations-for-learning-environments/B46A2FF2E88ACEC48BFD5D0A1CF7E4BB}.

\bibitem[Meftah et~al.(2012)Meftah, Le~Thanh, and
  Ben~Amar]{meftah_emotion_2012}
Imen~Tayari Meftah, Nhan Le~Thanh, and Chokri Ben~Amar.
\newblock Emotion {Recognition} {Using} {KNN} {Classification} for {User}
  {Modeling} and {Sharing} of {Affect} {States}.
\newblock In Tingwen Huang, Zhigang Zeng, Chuandong Li, and Chi~Sing Leung,
  editors, \emph{Neural {Information} {Processing}}, Lecture {Notes} in
  {Computer} {Science}, pages 234--242, Berlin, Heidelberg, 2012. Springer.
\newblock ISBN 978-3-642-34475-6.
\newblock \doi{10.1007/978-3-642-34475-6_29}.

\bibitem[Mehta et~al.(2005)Mehta, Niederee, Stewart, Degemmis, Lops, and
  Semeraro]{mehta_ontologically-enriched_2005}
Bhaskar Mehta, Claudia Niederee, Avare Stewart, Marco Degemmis, Pasquale Lops,
  and Giovanni Semeraro.
\newblock Ontologically-{Enriched} {Unified} {User} {Modeling} for
  {Cross}-{System} {Personalization}.
\newblock In Liliana Ardissono, Paul Brna, and Antonija Mitrovic, editors,
  \emph{User {Modeling} 2005}, Lecture {Notes} in {Computer} {Science}, pages
  119--123, Berlin, Heidelberg, 2005. Springer.
\newblock ISBN 978-3-540-31878-1.
\newblock \doi{10.1007/11527886_16}.

\bibitem[Mekruksavanich and
  Jitpattanakul(2021)]{mekruksavanich_convolutional_2021}
Sakorn Mekruksavanich and Anuchit Jitpattanakul.
\newblock Convolutional {Neural} {Network} and {Data} {Augmentation} for
  {Behavioral}-{Based} {Biometric} {User} {Identification}.
\newblock In Milan Tuba, Shyam Akashe, and Amit Joshi, editors, \emph{{ICT}
  {Systems} and {Sustainability}}, Advances in {Intelligent} {Systems} and
  {Computing}, pages 753--761, Singapore, 2021. Springer.
\newblock ISBN 9789811582899.
\newblock \doi{10.1007/978-981-15-8289-9_72}.

\bibitem[Mercado et~al.(2016)Mercado, Munaiah, and
  Meneely]{mercado_impact_2016}
Iván~Tactuk Mercado, Nuthan Munaiah, and Andrew Meneely.
\newblock The impact of cross-platform development approaches for mobile
  applications from the user's perspective.
\newblock In \emph{Proceedings of the {International} {Workshop} on {App}
  {Market} {Analytics}}, {WAMA} 2016, pages 43--49, New York, NY, USA, November
  2016. Association for Computing Machinery.
\newblock ISBN 978-1-4503-4398-5.
\newblock \doi{10.1145/2993259.2993268}.
\newblock URL \url{https://dl.acm.org/doi/10.1145/2993259.2993268}.

\bibitem[Mezghani et~al.(2012)Mezghani, Zayani, Amous, and
  Gargouri]{mezghani_user_2012}
Manel Mezghani, Corinne~Amel Zayani, Ikram Amous, and Faiez Gargouri.
\newblock A user profile modelling using social annotations: a survey.
\newblock In \emph{Proceedings of the 21st {International} {Conference} on
  {World} {Wide} {Web}}, {WWW} '12 {Companion}, pages 969--976, New York, NY,
  USA, April 2012. Association for Computing Machinery.
\newblock ISBN 978-1-4503-1230-1.
\newblock \doi{10.1145/2187980.2188230}.
\newblock URL \url{https://dl.acm.org/doi/10.1145/2187980.2188230}.

\bibitem[Middleton et~al.(2004)Middleton, Shadbolt, and
  De~Roure]{middleton_ontological_2004}
Stuart~E. Middleton, Nigel~R. Shadbolt, and David~C. De~Roure.
\newblock Ontological user profiling in recommender systems.
\newblock \emph{ACM Transactions on Information Systems}, 22\penalty0
  (1):\penalty0 54--88, 2004.
\newblock ISSN 1046-8188.
\newblock \doi{10.1145/963770.963773}.
\newblock URL \url{https://dl.acm.org/doi/10.1145/963770.963773}.

\bibitem[Minn et~al.(2022)Minn, Vie, Takeuchi, Kashima, and
  Zhu]{minn_interpretable_2022}
Sein Minn, Jill-Jênn Vie, Koh Takeuchi, Hisashi Kashima, and Feida Zhu.
\newblock Interpretable {Knowledge} {Tracing}: {Simple} and {Efficient}
  {Student} {Modeling} with {Causal} {Relations}.
\newblock \emph{Proceedings of the AAAI Conference on Artificial Intelligence},
  36\penalty0 (11):\penalty0 12810--12818, June 2022.
\newblock ISSN 2374-3468.
\newblock \doi{10.1609/aaai.v36i11.21560}.
\newblock URL \url{https://ojs.aaai.org/index.php/AAAI/article/view/21560}.

\bibitem[Mobasher(2007)]{mobasher_data_2007}
Bamshad Mobasher.
\newblock Data {Mining} for {Web} {Personalization}.
\newblock In Peter Brusilovsky, Alfred Kobsa, and Wolfgang Nejdl, editors,
  \emph{The {Adaptive} {Web}: {Methods} and {Strategies} of {Web}
  {Personalization}}, Lecture {Notes} in {Computer} {Science}, pages 90--135.
  Springer, Berlin, Heidelberg, 2007.
\newblock ISBN 978-3-540-72079-9.
\newblock \doi{10.1007/978-3-540-72079-9_3}.
\newblock URL \url{https://doi.org/10.1007/978-3-540-72079-9_3}.

\bibitem[Monti et~al.(2019)Monti, Frasca, Eynard, Mannion, and
  Bronstein]{monti_fake_2019}
Federico Monti, Fabrizio Frasca, Davide Eynard, Damon Mannion, and Michael~M.
  Bronstein.
\newblock Fake {News} {Detection} on {Social} {Media} using {Geometric} {Deep}
  {Learning}, February 2019.
\newblock URL \url{https://arxiv.org/abs/1902.06673v1}.

\bibitem[Morshed~Fahid et~al.(2021)Morshed~Fahid, Tian, Emerson, B.~Wiggins,
  Bounajim, Smith, Wiebe, Mott, Elizabeth~Boyer, and
  Lester]{morshed_fahid_progression_2021}
Fahmid Morshed~Fahid, Xiaoyi Tian, Andrew Emerson, Joseph B.~Wiggins, Dolly
  Bounajim, Andy Smith, Eric Wiebe, Bradford Mott, Kristy Elizabeth~Boyer, and
  James Lester.
\newblock Progression {Trajectory}-{Based} {Student} {Modeling} for {Novice}
  {Block}-{Based} {Programming}.
\newblock In \emph{Proceedings of the 29th {ACM} {Conference} on {User}
  {Modeling}, {Adaptation} and {Personalization}}, pages 189--200, Utrecht
  Netherlands, June 2021. ACM.
\newblock ISBN 978-1-4503-8366-0.
\newblock \doi{10.1145/3450613.3456833}.
\newblock URL \url{https://dl.acm.org/doi/10.1145/3450613.3456833}.

\bibitem[Musto et~al.(2018)Musto, Semeraro, Lovascio, de~Gemmis, and
  Lops]{musto_framework_2018}
Cataldo Musto, Giovanni Semeraro, Cosimo Lovascio, Marco de~Gemmis, and
  Pasquale Lops.
\newblock A {Framework} for {Holistic} {User} {Modeling} {Merging}
  {Heterogeneous} {Digital} {Footprints}.
\newblock In \emph{Adjunct {Publication} of the 26th {Conference} on {User}
  {Modeling}, {Adaptation} and {Personalization}}, {UMAP} '18, pages 97--101,
  New York, NY, USA, 2018. Association for Computing Machinery.
\newblock ISBN 978-1-4503-5784-5.
\newblock \doi{10.1145/3213586.3226218}.
\newblock URL \url{https://dl.acm.org/doi/10.1145/3213586.3226218}.

\bibitem[Musto et~al.(2020{\natexlab{a}})Musto, Narducci, Polignano, de~Gemmis,
  Lops, and Semeraro]{musto_towards_queryable_2020}
Cataldo Musto, Fedelucio Narducci, Marco Polignano, Marco de~Gemmis, Pasquale
  Lops, and Giovanni Semeraro.
\newblock Towards {Queryable} {User} {Profiles}: {Introducing} {Conversational}
  {Agents} in a {Platform} for {Holistic} {User} {Modeling}.
\newblock In \emph{Adjunct {Publication} of the 28th {ACM} {Conference} on
  {User} {Modeling}, {Adaptation} and {Personalization}}, {UMAP} '20 {Adjunct},
  pages 213--218, New York, NY, USA, 2020{\natexlab{a}}. Association for
  Computing Machinery.
\newblock ISBN 978-1-4503-7950-2.
\newblock \doi{10.1145/3386392.3399298}.
\newblock URL \url{https://dl.acm.org/doi/10.1145/3386392.3399298}.

\bibitem[Musto et~al.(2020{\natexlab{b}})Musto, Polignano, Semeraro, de~Gemmis,
  and Lops]{musto_myrror_2020}
Cataldo Musto, Marco Polignano, Giovanni Semeraro, Marco de~Gemmis, and
  Pasquale Lops.
\newblock Myrror: a platform for holistic user modeling.
\newblock \emph{User Modeling and User-Adapted Interaction}, 30\penalty0
  (3):\penalty0 477--511, July 2020{\natexlab{b}}.
\newblock ISSN 1573-1391.
\newblock \doi{10.1007/s11257-020-09272-6}.
\newblock URL \url{https://doi.org/10.1007/s11257-020-09272-6}.

\bibitem[Musto et~al.(2020{\natexlab{c}})Musto, Trattner, Starke, and
  Semeraro]{musto_towards_2020}
Cataldo Musto, Christoph Trattner, Alain Starke, and Giovanni Semeraro.
\newblock Towards a {Knowledge}-aware {Food} {Recommender} {System}
  {Exploiting} {Holistic} {User} {Models}.
\newblock In \emph{Proceedings of the 28th {ACM} {Conference} on {User}
  {Modeling}, {Adaptation} and {Personalization}}, {UMAP} '20, pages 333--337,
  New York, NY, USA, 2020{\natexlab{c}}. Association for Computing Machinery.
\newblock ISBN 978-1-4503-6861-2.
\newblock \doi{10.1145/3340631.3394880}.
\newblock URL \url{https://dl.acm.org/doi/10.1145/3340631.3394880}.

\bibitem[Musto et~al.(2021)Musto, Narducci, Polignano, de~Gemmis, Lops, and
  Semeraro]{musto_myrrorbot_2021}
Cataldo Musto, Fedelucio Narducci, Marco Polignano, Marco de~Gemmis, Pasquale
  Lops, and Giovanni Semeraro.
\newblock {MyrrorBot}: {A} {Digital} {Assistant} {Based} on {Holistic} {User}
  {Models} for {Personalized} {Access} to {Online} {Services}.
\newblock \emph{ACM Transactions on Information Systems}, 39\penalty0
  (4):\penalty0 46:1--46:34, 2021.
\newblock ISSN 1046-8188.
\newblock \doi{10.1145/3447679}.
\newblock URL \url{https://dl.acm.org/doi/10.1145/3447679}.

\bibitem[Nagaraj et~al.(2022)Nagaraj, Saiteja, Ram, Kanta, Aditya, and
  V]{nagaraj_university_2022}
P~Nagaraj, K~Saiteja, K~Kalyan Ram, K~Mani Kanta, S~Krishna Aditya, and
  Muneeswaran V.
\newblock University {Recommender} {System} based on {Student} {Profile} using
  {Feature} {Weighted} {Algorithm} and {KNN}.
\newblock In \emph{2022 {International} {Conference} on {Sustainable}
  {Computing} and {Data} {Communication} {Systems} ({ICSCDS})}, pages 479--484,
  April 2022.
\newblock \doi{10.1109/ICSCDS53736.2022.9760852}.
\newblock URL
  \url{https://ieeexplore.ieee.org/abstract/document/9760852?casa_token=8EDAS89mh7IAAAAA:2ewEa1q30-4U7jT_pGfCM9o3EBfWRT4tfJ6eq4BqMzPBztfFJkZxeaxLv74SlvJx1r8T0Or76-HZ}.

\bibitem[Nanas et~al.(2003)Nanas, Uren, and De~Roeck]{nanas_building_2003}
Nikolaos Nanas, Victoria Uren, and Anne De~Roeck.
\newblock Building and applying a concept hierarchy representation of a user
  profile.
\newblock In \emph{Proceedings of the 26th annual international {ACM} {SIGIR}
  conference on {Research} and development in informaion retrieval}, {SIGIR}
  '03, pages 198--204, New York, NY, USA, 2003. Association for Computing
  Machinery.
\newblock ISBN 978-1-58113-646-3.
\newblock \doi{10.1145/860435.860473}.
\newblock URL \url{https://dl.acm.org/doi/10.1145/860435.860473}.

\bibitem[Nasrudin et~al.(2023)Nasrudin, Ahmad, Salleh, and
  Azahari]{nasrudin_systematic_2023}
Nur~Hasni Nasrudin, Samsiah Ahmad, Khairulliza~Ahmad Salleh, and Lily~Adina
  Azahari.
\newblock A {Systematic} {Review} of {User} {Mental} {Models} on {Applications}
  {Sustainability}.
\newblock \emph{International Journal of Sustainable Construction Engineering
  and Technology}, 14\penalty0 (3):\penalty0 376--389, September 2023.
\newblock ISSN 2600-7959.
\newblock URL
  \url{https://publisher.uthm.edu.my/ojs/index.php/IJSCET/article/view/15283}.

\bibitem[Natarajan and Moh(2016)]{natarajan_recommending_2016}
Suraj Natarajan and Melody Moh.
\newblock Recommending {News} {Based} on {Hybrid} {User} {Profile},
  {Popularity}, {Trends}, and {Location}.
\newblock In \emph{2016 {International} {Conference} on {Collaboration}
  {Technologies} and {Systems} ({CTS})}, pages 204--211, October 2016.
\newblock \doi{10.1109/CTS.2016.0050}.
\newblock URL \url{https://ieeexplore.ieee.org/abstract/document/7870988}.

\bibitem[Ni et~al.(2018)Ni, Ou, Liu, Li, Ou, Zeng, and Si]{ni_perceive_2018}
Yabo Ni, Dan Ou, Shichen Liu, Xiang Li, Wenwu Ou, Anxiang Zeng, and Luo Si.
\newblock Perceive {Your} {Users} in {Depth}: {Learning} {Universal} {User}
  {Representations} from {Multiple} {E}-commerce {Tasks}.
\newblock In \emph{Proceedings of the 24th {ACM} {SIGKDD} {International}
  {Conference} on {Knowledge} {Discovery} \& {Data} {Mining}}, {KDD} '18, pages
  596--605, New York, NY, USA, 2018. Association for Computing Machinery.
\newblock ISBN 978-1-4503-5552-0.
\newblock \doi{10.1145/3219819.3219828}.
\newblock URL \url{https://dl.acm.org/doi/10.1145/3219819.3219828}.

\bibitem[Nigel~Shadbolt(2020)]{shadbolt_architectures_2020}
Sir Nigel~Shadbolt.
\newblock Architectures for {Autonomy}: {Towards} an {Equitable} {Web} of
  {Data} in the {Age} of {AI}.
\newblock In \emph{Proceedings of {The} {Web} {Conference} 2020}, {WWW} '20,
  pages 3141--3142, New York, NY, USA, April 2020. Association for Computing
  Machinery.
\newblock ISBN 978-1-4503-7023-3.
\newblock \doi{10.1145/3366423.3382668}.
\newblock URL \url{https://doi.org/10.1145/3366423.3382668}.

\bibitem[Nkambou et~al.(2023)Nkambou, Brisson, Tato, and
  Robert]{nkambou_learning_2023}
Roger Nkambou, Janie Brisson, Ange Tato, and Serge Robert.
\newblock Learning {Logical} {Reasoning} {Using} an {Intelligent} {Tutoring}
  {System}: {A} {Hybrid} {Approach} to {Student} {Modeling}.
\newblock \emph{Proceedings of the AAAI Conference on Artificial Intelligence},
  37\penalty0 (13):\penalty0 15930--15937, September 2023.
\newblock ISSN 2374-3468.
\newblock \doi{10.1609/aaai.v37i13.26891}.
\newblock URL \url{https://ojs.aaai.org/index.php/AAAI/article/view/26891}.

\bibitem[Olaleke et~al.(2021)Olaleke, Oseledets, and
  Frolov]{olaleke_dynamic_2021}
Oluwafemi Olaleke, Ivan Oseledets, and Evgeny Frolov.
\newblock Dynamic {Modeling} of {User} {Preferences} for {Stable}
  {Recommendations}.
\newblock In \emph{Proceedings of the 29th {ACM} {Conference} on {User}
  {Modeling}, {Adaptation} and {Personalization}}, pages 262--266, Utrecht
  Netherlands, June 2021. ACM.
\newblock ISBN 978-1-4503-8366-0.
\newblock \doi{10.1145/3450613.3456830}.
\newblock URL \url{https://dl.acm.org/doi/10.1145/3450613.3456830}.

\bibitem[Olufisayo~Dahunsi(2021)]{olufisayo_dahunsi_ontology-based_2021}
Bolanle Olufisayo~Dahunsi.
\newblock An {Ontology}-based {Knowledgebase} for {User} {Profile} and
  {Garment} {Features} in {Apparel} {Recommender} {Systems}.
\newblock In \emph{Fifteenth {ACM} {Conference} on {Recommender} {Systems}},
  pages 851--854, Amsterdam Netherlands, September 2021. ACM.
\newblock ISBN 978-1-4503-8458-2.
\newblock \doi{10.1145/3460231.3473901}.
\newblock URL \url{https://dl.acm.org/doi/10.1145/3460231.3473901}.

\bibitem[Oosterhuis and de~Rijke(2021)]{oosterhuis_robust_2021}
Harrie Oosterhuis and Maarten~de de~Rijke.
\newblock Robust {Generalization} and {Safe} {Query}-{Specializationin}
  {Counterfactual} {Learning} to {Rank}.
\newblock In \emph{Proceedings of the {Web} {Conference} 2021}, {WWW} '21,
  pages 158--170, New York, NY, USA, 2021. Association for Computing Machinery.
\newblock ISBN 978-1-4503-8312-7.
\newblock \doi{10.1145/3442381.3450018}.
\newblock URL \url{https://dl.acm.org/doi/10.1145/3442381.3450018}.

\bibitem[Ortony et~al.(1988)Ortony, Clore, and Collins]{ortony_cognitive_1988}
Andrew Ortony, Gerald~L Clore, and Allan Collins.
\newblock \emph{The cognitive structure of emotions}.
\newblock Cambridge University Press, 1988.

\bibitem[Orwant(1994)]{orwant_heterogeneous_1994}
Jon Orwant.
\newblock Heterogeneous learning in the {Doppelgänger} user modeling system.
\newblock \emph{User Modeling and User-Adapted Interaction}, 4\penalty0
  (2):\penalty0 107--130, June 1994.
\newblock ISSN 1573-1391.
\newblock \doi{10.1007/BF01099429}.
\newblock URL \url{https://doi.org/10.1007/BF01099429}.

\bibitem[Ouaftouh et~al.(2015)Ouaftouh, Zellou, and Idri]{ouaftouh_user_2015}
Sara Ouaftouh, Ahmed Zellou, and Ali Idri.
\newblock User profile model: {A} user dimension based classification.
\newblock In \emph{2015 10th {International} {Conference} on {Intelligent}
  {Systems}: {Theories} and {Applications} ({SITA})}, pages 1--5, Rabat,
  October 2015. IEEE.
\newblock ISBN 978-1-5090-0220-7.
\newblock \doi{10.1109/SITA.2015.7358378}.
\newblock URL \url{http://ieeexplore.ieee.org/document/7358378/}.

\bibitem[Ouaftouh et~al.(2019)Ouaftouh, Zellou, and Idri]{ouaftouh_social_2019}
Sara Ouaftouh, Ahmed Zellou, and Ali Idri.
\newblock Social recommendation: {A} user profile clustering-based approach.
\newblock \emph{Concurrency and Computation: Practice and Experience},
  31\penalty0 (20):\penalty0 e5330, 2019.
\newblock ISSN 1532-0634.
\newblock \doi{10.1002/cpe.5330}.
\newblock URL \url{https://onlinelibrary.wiley.com/doi/abs/10.1002/cpe.5330}.

\bibitem[Ouanaim et~al.(2010)Ouanaim, Harroud, Berrado, and
  Boulmalf]{ouanaim_dynamic_2010}
Mariam Ouanaim, Hamid Harroud, Aziz Berrado, and Mohammed Boulmalf.
\newblock Dynamic user profiling approach for services discovery in mobile
  environments.
\newblock In \emph{Proceedings of the 6th {International} {Wireless}
  {Communications} and {Mobile} {Computing} {Conference}}, {IWCMC} '10, pages
  550--554, New York, NY, USA, June 2010. Association for Computing Machinery.
\newblock ISBN 978-1-4503-0062-9.
\newblock \doi{10.1145/1815396.1815523}.
\newblock URL \url{https://dl.acm.org/doi/10.1145/1815396.1815523}.

\bibitem[Ouyang et~al.(2019)Ouyang, Zhang, Ren, Qi, Liu, and
  Du]{ouyang_representation_2019}
Wentao Ouyang, Xiuwu Zhang, Shukui Ren, Chao Qi, Zhaojie Liu, and Yanlong Du.
\newblock Representation {Learning}-{Assisted} {Click}-{Through} {Rate}
  {Prediction}.
\newblock In \emph{Proceedings of the {Twenty}-{Eighth} {International} {Joint}
  {Conference} on {Artificial} {Intelligence}}, pages 4561--4567, Macao, China,
  August 2019. International Joint Conferences on Artificial Intelligence
  Organization.
\newblock ISBN 978-0-9992411-4-1.
\newblock \doi{10.24963/ijcai.2019/634}.
\newblock URL \url{https://www.ijcai.org/proceedings/2019/634}.

\bibitem[Paiva and Self(1994)]{paiva_tagus_1994}
Ana Paiva and John Self.
\newblock {TAGUS} — {A} user and learner modeling workbench.
\newblock \emph{User Modeling and User-Adapted Interaction}, 4\penalty0
  (3):\penalty0 197--226, September 1994.
\newblock ISSN 1573-1391.
\newblock \doi{10.1007/BF01100244}.
\newblock URL \url{https://doi.org/10.1007/BF01100244}.

\bibitem[Pan et~al.(2020)Pan, He, and Yu]{pan_learning_2020}
Yiteng Pan, Fazhi He, and Haiping Yu.
\newblock Learning social representations with deep autoencoder for recommender
  system.
\newblock \emph{World Wide Web}, 23\penalty0 (4):\penalty0 2259--2279, July
  2020.
\newblock ISSN 1573-1413.
\newblock \doi{10.1007/s11280-020-00793-z}.
\newblock URL \url{https://doi.org/10.1007/s11280-020-00793-z}.

\bibitem[Pannu et~al.(2011)Pannu, Anane, Odetayo, and
  James]{pannu_explicit_2011}
Mandeep Pannu, Rachid Anane, Michael Odetayo, and Anne James.
\newblock Explicit user profiles in web search personalisation.
\newblock In \emph{Proceedings of the 2011 15th {International} {Conference} on
  {Computer} {Supported} {Cooperative} {Work} in {Design} ({CSCWD})}, pages
  416--421, June 2011.
\newblock \doi{10.1109/CSCWD.2011.5960107}.
\newblock URL
  \url{https://ieeexplore.ieee.org/abstract/document/5960107?casa_token=Bq9oEJKCwqwAAAAA:SfgXL9zUt1DP81HPlasEZi1CEmZm5wm4c8OwYkdTwRVtHNoj-luYxwaNEfpZLRjKFKgRbrEnAg}.

\bibitem[Pavan et~al.(2015)Pavan, Lee, and De~Luca]{pavan_semantic_2015}
Marco Pavan, Thebin Lee, and Ernesto~William De~Luca.
\newblock Semantic enrichment for adaptive expert search.
\newblock In \emph{Proceedings of the 15th {International} {Conference} on
  {Knowledge} {Technologies} and {Data}-driven {Business}}, i-{KNOW} '15, pages
  1--4, New York, NY, USA, 2015. Association for Computing Machinery.
\newblock ISBN 978-1-4503-3721-2.
\newblock \doi{10.1145/2809563.2809621}.
\newblock URL \url{https://dl.acm.org/doi/10.1145/2809563.2809621}.

\bibitem[Peng et~al.(2016)Peng, Choo, and Ashman]{peng_user_2016}
Jian Peng, Kim-Kwang~Raymond Choo, and Helen Ashman.
\newblock User profiling in intrusion detection: {A} review.
\newblock \emph{Journal of Network and Computer Applications}, 72:\penalty0
  14--27, September 2016.
\newblock ISSN 1084-8045.
\newblock \doi{10.1016/j.jnca.2016.06.012}.
\newblock URL
  \url{https://www.sciencedirect.com/science/article/pii/S1084804516301412}.

\bibitem[Peppers and Rogers(1993)]{peppers_one_1993}
Don Peppers and Martha Rogers.
\newblock \emph{The one to one future: Building relationships one customer at a
  time}.
\newblock Currency Doubleday, 1993.

\bibitem[Peppers and Rogers(1997)]{peppers_enterprise_1997}
Don Peppers and Martha Rogers.
\newblock Enterprise one to one: Tools for competing in the interactive age.
\newblock 1997.

\bibitem[Perrault et~al.(1978)Perrault, Allen, and Cohen]{perrault_speech_1978}
C.~Raymond Perrault, James~F. Allen, and Philip~R. Cohen.
\newblock Speech {Acts} as a {Basis} for {Understanding} {Dialogue}
  {Coherence}.
\newblock \emph{American Journal of Computational Linguistics}, pages 32--39,
  December 1978.
\newblock URL \url{https://aclanthology.org/J78-3024}.

\bibitem[Piao and Breslin(2016)]{piao_exploring_2016}
Guangyuan Piao and John~G. Breslin.
\newblock Exploring {Dynamics} and {Semantics} of {User} {Interests} for {User}
  {Modeling} on {Twitter} for {Link} {Recommendations}.
\newblock In \emph{Proceedings of the 12th {International} {Conference} on
  {Semantic} {Systems}}, {SEMANTiCS} 2016, pages 81--88, New York, NY, USA,
  2016. Association for Computing Machinery.
\newblock ISBN 978-1-4503-4752-5.
\newblock \doi{10.1145/2993318.2993332}.
\newblock URL \url{https://dl.acm.org/doi/10.1145/2993318.2993332}.

\bibitem[Piao and Breslin(2018)]{piao_inferring_2018}
Guangyuan Piao and John~G. Breslin.
\newblock Inferring user interests in microblogging social networks: a survey.
\newblock \emph{User Modeling and User-Adapted Interaction}, 28\penalty0
  (3):\penalty0 277--329, August 2018.
\newblock ISSN 0924-1868, 1573-1391.
\newblock \doi{10.1007/s11257-018-9207-8}.
\newblock URL \url{http://link.springer.com/10.1007/s11257-018-9207-8}.

\bibitem[Pierrakos et~al.(2003)Pierrakos, Paliouras, Papatheodorou, and
  Spyropoulos]{pierrakos_web_2003}
Dimitrios Pierrakos, Georgios Paliouras, Christos Papatheodorou, and
  Constantine~D. Spyropoulos.
\newblock Web {Usage} {Mining} as a {Tool} for {Personalization}: {A} {Survey}.
\newblock \emph{User Modeling and User-Adapted Interaction}, 13\penalty0
  (4):\penalty0 311--372, November 2003.
\newblock ISSN 1573-1391.
\newblock \doi{10.1023/A:1026238916441}.
\newblock URL \url{https://doi.org/10.1023/A:1026238916441}.

\bibitem[Plumbaum et~al.(2011)Plumbaum, Wu, De~Luca, and
  Albayrak]{plumbaum_user_2011}
Till Plumbaum, Songxuan Wu, Ernesto~William De~Luca, and Sahin Albayrak.
\newblock User modeling for the social semantic web.
\newblock In \emph{Proceedings of the {Second} {International} {Conference} on
  {Semantic} {Personalized} {Information} {Management}: {Retrieval} and
  {Recommendation} - {Volume} 781}, {SPIM}'11, pages 78--89, Aachen, DEU, 2011.
  CEUR-WS.org.

\bibitem[Plumbaum et~al.(2012)Plumbaum, Lommatzsch, De~Luca, and
  Albayrak]{plumbaum_serum_2012}
Till Plumbaum, Andreas Lommatzsch, Ernesto~William De~Luca, and Sahin Albayrak.
\newblock {SERUM}: {Collecting} {Semantic} {User} {Behavior} for {Improved}
  {News} {Recommendations}.
\newblock In Liliana Ardissono and Tsvi Kuflik, editors, \emph{Advances in
  {User} {Modeling}}, Lecture {Notes} in {Computer} {Science}, pages 402--405,
  Berlin, Heidelberg, 2012. Springer.
\newblock ISBN 978-3-642-28509-7.
\newblock \doi{10.1007/978-3-642-28509-7_37}.

\bibitem[Poo et~al.(2003)Poo, Chng, and Goh]{poo_hybrid_2003}
D.~Poo, B.~Chng, and Jie-Mein Goh.
\newblock A hybrid approach for user profiling.
\newblock In \emph{36th {Annual} {Hawaii} {International} {Conference} on
  {System} {Sciences}, 2003. {Proceedings} of the}, pages 9 pp.--, January
  2003.
\newblock \doi{10.1109/HICSS.2003.1174242}.

\bibitem[Pujahari and Sisodia(2022)]{pujahari_item_2022}
Abinash Pujahari and Dilip~Singh Sisodia.
\newblock Item feature refinement using matrix factorization and boosted
  learning based user profile generation for content-based recommender systems.
\newblock \emph{Expert Systems with Applications}, 206:\penalty0 117849,
  November 2022.
\newblock ISSN 09574174.
\newblock \doi{10.1016/j.eswa.2022.117849}.
\newblock URL
  \url{https://linkinghub.elsevier.com/retrieve/pii/S0957417422011046}.

\bibitem[Purificato and De~Luca(2023)]{purificato_what_2023}
Erasmo Purificato and Ernesto~William De~Luca.
\newblock What {Are} {We} {Missing} in {Algorithmic} {Fairness}? {Discussing}
  {Open} {Challenges} for {Fairness} {Analysis} in {User} {Profiling}
  with {Graph} {Neural} {Networks}.
\newblock In Ludovico Boratto, Stefano Faralli, Mirko Marras, and Giovanni
  Stilo, editors, \emph{Advances in {Bias} and {Fairness} in {Information}
  {Retrieval}}, Communications in {Computer} and {Information} {Science}, pages
  169--175, Cham, 2023. Springer Nature Switzerland.
\newblock ISBN 978-3-031-37249-0.
\newblock \doi{10.1007/978-3-031-37249-0_14}.

\bibitem[Purificato et~al.(2022)Purificato, Boratto, and
  De~Luca]{purificato_graph_2022}
Erasmo Purificato, Ludovico Boratto, and Ernesto~William De~Luca.
\newblock Do {Graph} {Neural} {Networks} {Build} {Fair} {User} {Models}?
  {Assessing} {Disparate} {Impact} and {Mistreatment} in {Behavioural} {User}
  {Profiling}.
\newblock In \emph{Proceedings of the 31st {ACM} {International} {Conference}
  on {Information} \& {Knowledge} {Management}}, {CIKM} '22, pages 4399--4403,
  New York, NY, USA, 2022. Association for Computing Machinery.
\newblock ISBN 978-1-4503-9236-5.
\newblock \doi{10.1145/3511808.3557584}.
\newblock URL \url{https://doi.org/10.1145/3511808.3557584}.

\bibitem[Qi et~al.(2022{\natexlab{a}})Qi, Wu, Wu, and Huang]{qi_fum_2022}
Tao Qi, Fangzhao Wu, Chuhan Wu, and Yongfeng Huang.
\newblock {FUM}: {Fine}-grained and {Fast} {User} {Modeling} for {News}
  {Recommendation}.
\newblock In \emph{Proceedings of the 45th {International} {ACM} {SIGIR}
  {Conference} on {Research} and {Development} in {Information} {Retrieval}},
  pages 1974--1978, Madrid Spain, July 2022{\natexlab{a}}. ACM.
\newblock ISBN 978-1-4503-8732-3.
\newblock \doi{10.1145/3477495.3531790}.
\newblock URL \url{https://dl.acm.org/doi/10.1145/3477495.3531790}.

\bibitem[Qi et~al.(2022{\natexlab{b}})Qi, Wu, Wu, and Huang]{qi_news_2022}
Tao Qi, Fangzhao Wu, Chuhan Wu, and Yongfeng Huang.
\newblock News {Recommendation} with {Candidate}-aware {User} {Modeling}.
\newblock In \emph{Proceedings of the 45th {International} {ACM} {SIGIR}
  {Conference} on {Research} and {Development} in {Information} {Retrieval}},
  pages 1917--1921, Madrid Spain, July 2022{\natexlab{b}}. ACM.
\newblock ISBN 978-1-4503-8732-3.
\newblock \doi{10.1145/3477495.3531778}.
\newblock URL \url{https://dl.acm.org/doi/10.1145/3477495.3531778}.

\bibitem[Qi(2010)]{qi_research_2010}
Xu~Qi.
\newblock Research on {User} {Profiling} {Technology} for {Personalized}
  {Demands}.
\newblock In \emph{2010 {International} {Conference} on {Intelligent}
  {Computation} {Technology} and {Automation}}, volume~3, pages 198--201, May
  2010.
\newblock \doi{10.1109/ICICTA.2010.252}.
\newblock URL \url{https://ieeexplore.ieee.org/document/5523259}.

\bibitem[Qi et~al.(2021)Qi, Hu, Zhang, Cheng, and Lei]{qi_trilateral_2021}
Yi~Qi, Ke~Hu, Bo~Zhang, Jia Cheng, and Jun Lei.
\newblock Trilateral {Spatiotemporal} {Attention} {Network} for {User}
  {Behavior} {Modeling} in {Location}-based {Search}.
\newblock In \emph{Proceedings of the 30th {ACM} {International} {Conference}
  on {Information} \& {Knowledge} {Management}}, pages 3373--3377, Virtual
  Event Queensland Australia, October 2021. ACM.
\newblock ISBN 978-1-4503-8446-9.
\newblock \doi{10.1145/3459637.3482206}.
\newblock URL \url{https://dl.acm.org/doi/10.1145/3459637.3482206}.

\bibitem[Qian et~al.(2022)Qian, Wu, Li, Wu, Zhang, Zhou, Gu, and
  Gu]{qian_fwseqblock_2022}
Hao Qian, Qintong Wu, MingHao Li, Zhengwei Wu, Zhiqiang Zhang, Jun Zhou, Lihong
  Gu, and Jinjie Gu.
\newblock {FwSeqBlock}: {A} {Field}-wise {Approach} for {Modeling} {Behavior}
  {Representation} in {Sequential} {Recommendation}.
\newblock In \emph{Proceedings of the 31st {ACM} {International} {Conference}
  on {Information} \& {Knowledge} {Management}}, pages 4404--4408, Atlanta GA
  USA, October 2022. ACM.
\newblock ISBN 978-1-4503-9236-5.
\newblock \doi{10.1145/3511808.3557601}.
\newblock URL \url{https://dl.acm.org/doi/10.1145/3511808.3557601}.

\bibitem[Qian et~al.(2021)Qian, Dou, Zhu, Ma, and Wen]{qian_learning_2021}
Hongjin Qian, Zhicheng Dou, Yutao Zhu, Yueyuan Ma, and Ji-Rong Wen.
\newblock Learning {Implicit} {User} {Profile} for {Personalized}
  {Retrieval}-{Based} {Chatbot}.
\newblock In \emph{Proceedings of the 30th {ACM} {International} {Conference}
  on {Information} \& {Knowledge} {Management}}, pages 1467--1477, Virtual
  Event Queensland Australia, October 2021. ACM.
\newblock ISBN 978-1-4503-8446-9.
\newblock \doi{10.1145/3459637.3482269}.
\newblock URL \url{https://dl.acm.org/doi/10.1145/3459637.3482269}.

\bibitem[Raber and Krüger(2022)]{raber_transferring_2022}
Frederic Raber and Antonio Krüger.
\newblock Transferring recommendations through privacy user models across
  domains.
\newblock \emph{User Modeling and User-Adapted Interaction}, 32\penalty0
  (1):\penalty0 25--90, April 2022.
\newblock ISSN 1573-1391.
\newblock \doi{10.1007/s11257-021-09307-6}.
\newblock URL \url{https://doi.org/10.1007/s11257-021-09307-6}.

\bibitem[Raghu et~al.(2001)Raghu, Kannan, Rao, and
  Whinston]{raghu_dynamic_2001}
T.~S. Raghu, P.~K. Kannan, H.~R. Rao, and A.~B. Whinston.
\newblock Dynamic profiling of consumers for customized offerings over the
  {Internet}: a model and analysis.
\newblock \emph{Decision Support Systems}, 32\penalty0 (2):\penalty0 117--134,
  December 2001.
\newblock ISSN 0167-9236.
\newblock \doi{10.1016/S0167-9236(01)00106-3}.
\newblock URL
  \url{https://www.sciencedirect.com/science/article/pii/S0167923601001063}.

\bibitem[Raghuram et~al.(2016)Raghuram, Akshay, and
  Chandrasekaran]{raghuram_efficient_2016}
M.~A. Raghuram, K.~Akshay, and K.~Chandrasekaran.
\newblock Efficient {User} {Profiling} in {Twitter} {Social} {Network} {Using}
  {Traditional} {Classifiers}.
\newblock In Stefano Berretti, Sabu~M. Thampi, and Soura Dasgupta, editors,
  \emph{Intelligent {Systems} {Technologies} and {Applications}}, Advances in
  {Intelligent} {Systems} and {Computing}, pages 399--411, Cham, 2016. Springer
  International Publishing.
\newblock ISBN 978-3-319-23258-4.
\newblock \doi{10.1007/978-3-319-23258-4_35}.

\bibitem[Rajashekar et~al.(2016)Rajashekar, Zincir-Heywood, and
  Heywood]{rajashekar_smart_2016}
Deepthi Rajashekar, A.~Nur Zincir-Heywood, and Malcolm~I. Heywood.
\newblock Smart {Phone} {User} {Behaviour} {Characterization} {Based} on
  {Autoencoders} and {Self} {Organizing} {Maps}.
\newblock In \emph{2016 {IEEE} 16th {International} {Conference} on {Data}
  {Mining} {Workshops} ({ICDMW})}, pages 319--326, December 2016.
\newblock \doi{10.1109/ICDMW.2016.0052}.
\newblock URL \url{https://ieeexplore.ieee.org/abstract/document/7836683}.
\newblock ISSN: 2375-9259.

\bibitem[Razmerita et~al.(2003)Razmerita, Angehrn, and
  Maedche]{razmerita_ontology-based_2003}
Liana Razmerita, Albert Angehrn, and Alexander Maedche.
\newblock Ontology-{Based} {User} {Modeling} for {Knowledge} {Management}
  {Systems}.
\newblock In Peter Brusilovsky, Albert Corbett, and Fiorella de~Rosis, editors,
  \emph{User {Modeling} 2003}, Lecture {Notes} in {Computer} {Science}, pages
  213--217, Berlin, Heidelberg, 2003. Springer.
\newblock ISBN 978-3-540-44963-8.
\newblock \doi{10.1007/3-540-44963-9_29}.

\bibitem[Ren et~al.(2019)Ren, Qin, Fang, Zhang, Zheng, Bian, Zhou, Xu, Yu, Zhu,
  and Gai]{ren_lifelong_2019}
Kan Ren, Jiarui Qin, Yuchen Fang, Weinan Zhang, Lei Zheng, Weijie Bian, Guorui
  Zhou, Jian Xu, Yong Yu, Xiaoqiang Zhu, and Kun Gai.
\newblock Lifelong {Sequential} {Modeling} with {Personalized} {Memorization}
  for {User} {Response} {Prediction}.
\newblock In \emph{Proceedings of the 42nd {International} {ACM} {SIGIR}
  {Conference} on {Research} and {Development} in {Information} {Retrieval}},
  pages 565--574, Paris France, July 2019. ACM.
\newblock ISBN 978-1-4503-6172-9.
\newblock \doi{10.1145/3331184.3331230}.
\newblock URL \url{https://dl.acm.org/doi/10.1145/3331184.3331230}.

\bibitem[Rich(1979)]{rich_user_1979}
Elaine Rich.
\newblock User {Modeling} via {Stereotypes}.
\newblock \emph{Cognitive Science}, 3\penalty0 (4):\penalty0 329--354, 1979.
\newblock ISSN 1551-6709.
\newblock \doi{10.1207/s15516709cog0304_3}.
\newblock URL
  \url{https://onlinelibrary.wiley.com/doi/abs/10.1207/s15516709cog0304_3}.

\bibitem[Rimitha et~al.(2018)Rimitha, Abburu, Kiranmai, and
  Chandrasekaran]{rimitha_ontologies_2018}
S.R. Rimitha, Vedasamhitha Abburu, Annem Kiranmai, and K.~Chandrasekaran.
\newblock Ontologies to {Model} {User} {Profiles} in {Personalized} {Job}
  {Recommendation}.
\newblock In \emph{2018 {IEEE} {Distributed} {Computing}, {VLSI}, {Electrical}
  {Circuits} and {Robotics} ({DISCOVER})}, pages 98--103, August 2018.
\newblock \doi{10.1109/DISCOVER.2018.8674084}.
\newblock URL \url{https://ieeexplore.ieee.org/document/8674084}.

\bibitem[Rizzo et~al.(2016)Rizzo, Dondio, Delany, and
  Longo]{rizzo_modeling_2016}
Lucas Rizzo, Pierpaolo Dondio, Sarah~Jane Delany, and Luca Longo.
\newblock Modeling {Mental} {Workload} {Via} {Rule}-{Based} {Expert} {System}:
  {A} {Comparison} with {NASA}-{TLX} and {Workload} {Profile}.
\newblock In Lazaros Iliadis and Ilias Maglogiannis, editors, \emph{Artificial
  {Intelligence} {Applications} and {Innovations}}, {IFIP} {Advances} in
  {Information} and {Communication} {Technology}, pages 215--229, Cham, 2016.
  Springer International Publishing.
\newblock ISBN 978-3-319-44944-9.
\newblock \doi{10.1007/978-3-319-44944-9_19}.

\bibitem[Romero and Ventura(2013)]{romero_data_2013}
Cristobal Romero and Sebastian Ventura.
\newblock Data mining in education.
\newblock \emph{WIREs Data Mining and Knowledge Discovery}, 3\penalty0
  (1):\penalty0 12--27, 2013.
\newblock ISSN 1942-4795.
\newblock \doi{10.1002/widm.1075}.
\newblock URL \url{https://onlinelibrary.wiley.com/doi/abs/10.1002/widm.1075}.

\bibitem[Rozen et~al.(2021)Rozen, Oren, and Raviv]{rozen_predicting_2021}
Ohad Rozen, Joel Oren, and Ariel Raviv.
\newblock Predicting {User} {Demography} and {Device} from {News} {Comments}.
\newblock In \emph{Proceedings of the 44th {International} {ACM} {SIGIR}
  {Conference} on {Research} and {Development} in {Information} {Retrieval}},
  {SIGIR} '21, pages 1995--1999, New York, NY, USA, 2021. Association for
  Computing Machinery.
\newblock ISBN 978-1-4503-8037-9.
\newblock \doi{10.1145/3404835.3463024}.
\newblock URL \url{https://dl.acm.org/doi/10.1145/3404835.3463024}.

\bibitem[Saevanee et~al.(2012)Saevanee, Clarke, and
  Furnell]{saevanee_multi-modal_2012}
Hataichanok Saevanee, Nathan~L. Clarke, and Steven~M. Furnell.
\newblock Multi-modal {Behavioural} {Biometric} {Authentication} for {Mobile}
  {Devices}.
\newblock In Dimitris Gritzalis, Steven Furnell, and Marianthi Theoharidou,
  editors, \emph{Information {Security} and {Privacy} {Research}}, {IFIP}
  {Advances} in {Information} and {Communication} {Technology}, pages 465--474,
  Berlin, Heidelberg, 2012. Springer.
\newblock ISBN 978-3-642-30436-1.
\newblock \doi{10.1007/978-3-642-30436-1_38}.

\bibitem[Saevanee et~al.(2015)Saevanee, Clarke, Furnell, and
  Biscione]{saevanee_continuous_2015}
Hataichanok Saevanee, Nathan Clarke, Steven Furnell, and Valerio Biscione.
\newblock Continuous user authentication using multi-modal biometrics.
\newblock \emph{Computers \& Security}, 53:\penalty0 234--246, September 2015.
\newblock ISSN 0167-4048.
\newblock \doi{10.1016/j.cose.2015.06.001}.
\newblock URL
  \url{https://www.sciencedirect.com/science/article/pii/S0167404815000875}.

\bibitem[Sahoo and Gupta(2021)]{sahoo_multiple_2021}
Somya~Ranjan Sahoo and B.~B. Gupta.
\newblock Multiple features based approach for automatic fake news detection on
  social networks using deep learning.
\newblock \emph{Applied Soft Computing}, 100:\penalty0 106983, March 2021.
\newblock ISSN 1568-4946.
\newblock \doi{10.1016/j.asoc.2020.106983}.
\newblock URL
  \url{https://www.sciencedirect.com/science/article/pii/S1568494620309224}.

\bibitem[Said et~al.(2013)Said, Berkovsky, and De~Luca]{said_movie_2013}
Alan Said, Shlomo Berkovsky, and Ernesto~W. De~Luca.
\newblock Movie recommendation in context.
\newblock \emph{ACM Transactions on Intelligent Systems and Technology},
  4\penalty0 (13):\penalty0 1--9, February 2013.
\newblock ISSN 2157-6904.
\newblock \doi{10.1145/2414425.2414438}.
\newblock URL \url{https://dl.acm.org/doi/10.1145/2414425.2414438}.

\bibitem[Sajib Al~Seraj(2018)]{sajib_al_seraj_survey_2018}
Mohammad Sajib Al~Seraj.
\newblock A {Survey} on {User} {Modeling} in {HCI}.
\newblock \emph{Computer Applications: An International Journal}, 5\penalty0
  (1):\penalty0 21--28, February 2018.
\newblock ISSN 23938455.
\newblock \doi{10.5121/caij.2018.5102}.
\newblock URL \url{http://airccse.com/caij/papers/5118caij02.pdf}.

\bibitem[Sanchez-Gordon et~al.(2021)Sanchez-Gordon, Aguilar-Mayanquer, and
  Calle-Jimenez]{sanchez-gordon_model_2021}
Sandra Sanchez-Gordon, Carmen Aguilar-Mayanquer, and Tania Calle-Jimenez.
\newblock Model for {Profiling} {Users} {With} {Disabilities} on e-{Learning}
  {Platforms}.
\newblock \emph{IEEE Access}, 9:\penalty0 74258--74274, 2021.
\newblock ISSN 2169-3536.
\newblock \doi{10.1109/ACCESS.2021.3081061}.
\newblock URL \url{https://ieeexplore.ieee.org/document/9432934}.

\bibitem[Santra and Jayasudha(2012)]{santra_classification_2012}
AK~Santra and S~Jayasudha.
\newblock Classification of web log data to identify interested users using
  na{\"\i}ve bayesian classification.
\newblock \emph{International Journal of Computer Science Issues (IJCSI)},
  9\penalty0 (1):\penalty0 381, 2012.

\bibitem[Sarker and Salim(2018)]{sarker_mining_2018}
Iqbal~H. Sarker and Flora~D. Salim.
\newblock Mining {User} {Behavioral} {Rules} from {Smartphone} {Data} {Through}
  {Association} {Analysis}.
\newblock In Dinh Phung, Vincent~S. Tseng, Geoffrey~I. Webb, Bao Ho, Mohadeseh
  Ganji, and Lida Rashidi, editors, \emph{Advances in {Knowledge} {Discovery}
  and {Data} {Mining}}, Lecture {Notes} in {Computer} {Science}, pages
  450--461, Cham, 2018. Springer International Publishing.
\newblock ISBN 978-3-319-93034-3.
\newblock \doi{10.1007/978-3-319-93034-3_36}.

\bibitem[Sarker et~al.(2020)Sarker, Colman, Han, Khan, Abushark, and
  Salah]{sarker_behavdt_2020}
Iqbal~H. Sarker, Alan Colman, Jun Han, Asif~Irshad Khan, Yoosef~B. Abushark,
  and Khaled Salah.
\newblock {BehavDT}: {A} {Behavioral} {Decision} {Tree} {Learning} to {Build}
  {User}-{Centric} {Context}-{Aware} {Predictive} {Model}.
\newblock \emph{Mobile Networks and Applications}, 25\penalty0 (3):\penalty0
  1151--1161, June 2020.
\newblock ISSN 1572-8153.
\newblock \doi{10.1007/s11036-019-01443-z}.
\newblock URL \url{https://doi.org/10.1007/s11036-019-01443-z}.

\bibitem[Schiaffino and Amandi(2009)]{schiaffino_intelligent_2009}
Silvia Schiaffino and Analía Amandi.
\newblock Intelligent {User} {Profiling}.
\newblock In Max Bramer, editor, \emph{Artificial {Intelligence} {An}
  {International} {Perspective}: {An} {International} {Perspective}}, Lecture
  {Notes} in {Computer} {Science}, pages 193--216. Springer, Berlin,
  Heidelberg, 2009.
\newblock ISBN 978-3-642-03226-4.
\newblock \doi{10.1007/978-3-642-03226-4_11}.
\newblock URL \url{https://doi.org/10.1007/978-3-642-03226-4_11}.

\bibitem[Schreck(2003)]{schreck_security_2003}
J.~Schreck.
\newblock \emph{Security and {Privacy} in {User} {Modeling}}.
\newblock Springer Science \& Business Media, January 2003.
\newblock ISBN 978-1-4020-1130-6.

\bibitem[Semeraro et~al.(2008)Semeraro, Andersen, Andersen, de~Gemmis, and
  Lops]{semeraro_user_2008}
Giovanni Semeraro, Verner Andersen, Hans H.~K. Andersen, Marco de~Gemmis, and
  Pasquale Lops.
\newblock User profiling and virtual agents: a case study on e-commerce
  services.
\newblock \emph{Universal Access in the Information Society}, 7\penalty0
  (3):\penalty0 179--194, September 2008.
\newblock ISSN 1615-5297.
\newblock \doi{10.1007/s10209-008-0116-1}.
\newblock URL \url{https://doi.org/10.1007/s10209-008-0116-1}.

\bibitem[Seroussi et~al.(2011)Seroussi, Bohnert, and
  Zukerman]{seroussi_personalised_2011}
Yanir Seroussi, Fabian Bohnert, and Ingrid Zukerman.
\newblock Personalised rating prediction for new users using latent factor
  models.
\newblock In \emph{Proceedings of the 22nd {ACM} conference on {Hypertext} and
  hypermedia}, {HT} '11, pages 47--56, New York, NY, USA, 2011. Association for
  Computing Machinery.
\newblock ISBN 978-1-4503-0256-2.
\newblock \doi{10.1145/1995966.1995976}.
\newblock URL \url{https://dl.acm.org/doi/10.1145/1995966.1995976}.

\bibitem[Sguerra et~al.(2023)Sguerra, Tran, and Hennequin]{sguerra_ex2vec_2023}
Bruno Sguerra, Viet-Anh Tran, and Romain Hennequin.
\newblock {Ex2Vec}: {Characterizing} {Users} and {Items} from the {Mere}
  {Exposure} {Effect}.
\newblock In \emph{Proceedings of the 17th {ACM} {Conference} on {Recommender}
  {Systems}}, pages 971--977, Singapore Singapore, September 2023. ACM.
\newblock ISBN 9798400702419.
\newblock \doi{10.1145/3604915.3608856}.
\newblock URL \url{https://dl.acm.org/doi/10.1145/3604915.3608856}.

\bibitem[Shahar(2021)]{shahar_ethical_2021}
Yuval Shahar.
\newblock The {Ethical} {Implications} of {Shared} {Medical} {Decision}
  {Making} without {Providing} {Adequate} {Computational} {Support} to the
  {Care} {Provider} and to the {Patient}, February 2021.
\newblock URL \url{http://arxiv.org/abs/2102.01811}.

\bibitem[Sharma et~al.(2020)Sharma, Pokharel, and Joshi]{sharma_user_2020}
Balaram Sharma, Prabhat Pokharel, and Basanta Joshi.
\newblock User {Behavior} {Analytics} for {Anomaly} {Detection} {Using} {LSTM}
  {Autoencoder} - {Insider} {Threat} {Detection}.
\newblock In \emph{Proceedings of the 11th {International} {Conference} on
  {Advances} in {Information} {Technology}}, {IAIT} '20, pages 1--9, New York,
  NY, USA, 2020. Association for Computing Machinery.
\newblock ISBN 978-1-4503-7759-1.
\newblock \doi{10.1145/3406601.3406610}.
\newblock URL \url{https://dl.acm.org/doi/10.1145/3406601.3406610}.

\bibitem[Shazad et~al.(2019)Shazad, Ullah~Khan, Rehman, Farooq, Mahmood,
  Mehmood, Rho, and Nam]{shazad_finding_2019}
Babar Shazad, Hikmat Ullah~Khan, Zahoor-ur Rehman, Muhammad Farooq, Ahsan
  Mahmood, Irfan Mehmood, Seungmin Rho, and Yunyoung Nam.
\newblock Finding {Temporal} {Influential} {Users} in {Social} {Media} {Using}
  {Association} {Rule} {Learning}.
\newblock \emph{Intelligent Automation and Soft Computing}, 2019.
\newblock ISSN 1079-8587, 2326-005X.
\newblock \doi{10.31209/2019.100000130}.
\newblock URL \url{http://autosoftjournal.net/paperShow.php?paper=100000130}.

\bibitem[Shen et~al.(2021{\natexlab{a}})Shen, Lederman, Cao, Tang, and
  Pentland]{shen_user_2021}
Jiaxing Shen, Oren Lederman, Jiannong Cao, Shaojie Tang, and Alex~‘Sandy’
  Pentland.
\newblock User {Profiling} based on {Nonlinguistic} {Audio} {Data}.
\newblock In \emph{2021 {IEEE} 37th {International} {Conference} on {Data}
  {Engineering} ({ICDE})}, pages 2303--2308, April 2021{\natexlab{a}}.
\newblock \doi{10.1109/ICDE51399.2021.00241}.
\newblock URL \url{https://ieeexplore.ieee.org/document/9458756}.

\bibitem[Shen et~al.(2021{\natexlab{b}})Shen, Tao, Zhang, Wen, Chen, and
  Lu]{shen_sar-net_2021}
Qijie Shen, Wanjie Tao, Jing Zhang, Hong Wen, Zulong Chen, and Quan Lu.
\newblock {SAR}-{Net}: {A} {Scenario}-{Aware} {Ranking} {Network} for
  {Personalized} {Fair} {Recommendation} in {Hundreds} of {Travel} {Scenarios}.
\newblock In \emph{Proceedings of the 30th {ACM} {International} {Conference}
  on {Information} \& {Knowledge} {Management}}, pages 4094--4103, Virtual
  Event Queensland Australia, October 2021{\natexlab{b}}. ACM.
\newblock ISBN 978-1-4503-8446-9.
\newblock \doi{10.1145/3459637.3481948}.
\newblock URL \url{https://dl.acm.org/doi/10.1145/3459637.3481948}.

\bibitem[Shen et~al.(2005)Shen, Tan, and Zhai]{shen_implicit_2005}
Xuehua Shen, Bin Tan, and ChengXiang Zhai.
\newblock Implicit user modeling for personalized search.
\newblock In \emph{Proceedings of the 14th {ACM} international conference on
  {Information} and knowledge management}, {CIKM} '05, pages 824--831, New
  York, NY, USA, 2005. Association for Computing Machinery.
\newblock ISBN 978-1-59593-140-5.
\newblock \doi{10.1145/1099554.1099747}.
\newblock URL \url{https://dl.acm.org/doi/10.1145/1099554.1099747}.

\bibitem[Shin(2016)]{shin_cross-platform_2016}
Dong-Hee Shin.
\newblock Cross-{Platform} {Users}’ {Experiences} {Toward} {Designing}
  {Interusable} {Systems}.
\newblock \emph{International Journal of Human–Computer Interaction},
  32\penalty0 (7):\penalty0 503--514, July 2016.
\newblock ISSN 1044-7318.
\newblock \doi{10.1080/10447318.2016.1177277}.
\newblock URL \url{https://doi.org/10.1080/10447318.2016.1177277}.

\bibitem[Shneiderman(2022)]{shneiderman_human_2022}
Ben Shneiderman.
\newblock \emph{Human-centered AI}.
\newblock Oxford University Press, 2022.

\bibitem[Shu et~al.(2018)Shu, Wang, and Liu]{shu_understanding_2018}
Kai Shu, Suhang Wang, and Huan Liu.
\newblock Understanding {User} {Profiles} on {Social} {Media} for {Fake} {News}
  {Detection}.
\newblock In \emph{2018 {IEEE} {Conference} on {Multimedia} {Information}
  {Processing} and {Retrieval} ({MIPR})}, pages 430--435, April 2018.
\newblock \doi{10.1109/MIPR.2018.00092}.
\newblock URL
  \url{https://ieeexplore.ieee.org/abstract/document/8397048?casa_token=yFzmKSodQYwAAAAA:gXCf7euQLFE1d9A_TYH9Pv0T7VlTJB6rRu6CdiOz8XZ6Yxs0s4rR1Lg8QOtNfznG92juwPHIsg}.

\bibitem[Shu et~al.(2019)Shu, Zhou, Wang, Zafarani, and Liu]{shu_role_2019}
Kai Shu, Xinyi Zhou, Suhang Wang, Reza Zafarani, and Huan Liu.
\newblock The role of user profiles for fake news detection.
\newblock In \emph{Proceedings of the 2019 {IEEE}/{ACM} {International}
  {Conference} on {Advances} in {Social} {Networks} {Analysis} and {Mining}},
  pages 436--439, Vancouver British Columbia Canada, August 2019. ACM.
\newblock ISBN 978-1-4503-6868-1.
\newblock \doi{10.1145/3341161.3342927}.
\newblock URL \url{https://dl.acm.org/doi/10.1145/3341161.3342927}.

\bibitem[Si et~al.(2019)Si, Zhou, Chen, Wan, Xiong, Zhang, and
  Vasilakos]{si_association_2019}
Huayou Si, Jiayong Zhou, Zhihui Chen, Jian Wan, Neal~N. Xiong, Wei Zhang, and
  Athanasios~V. Vasilakos.
\newblock Association {Rules} {Mining} {Among} {Interests} and {Applications}
  for {Users} on {Social} {Networks}.
\newblock \emph{IEEE Access}, 7:\penalty0 116014--116026, 2019.
\newblock ISSN 2169-3536.
\newblock \doi{10.1109/ACCESS.2019.2925819}.
\newblock URL \url{https://ieeexplore.ieee.org/abstract/document/8751963}.
\newblock Conference Name: IEEE Access.

\bibitem[Sieg et~al.(2007)Sieg, Mobasher, and Burke]{sieg_web_2007}
Ahu Sieg, Bamshad Mobasher, and Robin Burke.
\newblock Web search personalization with ontological user profiles.
\newblock In \emph{Proceedings of the sixteenth {ACM} conference on
  {Conference} on information and knowledge management}, pages 525--534, Lisbon
  Portugal, November 2007. ACM.
\newblock ISBN 978-1-59593-803-9.
\newblock \doi{10.1145/1321440.1321515}.
\newblock URL \url{https://dl.acm.org/doi/10.1145/1321440.1321515}.

\bibitem[Singh et~al.(2019)Singh, Mehtre, and Sangeetha]{singh_user_2019}
Malvika Singh, B.M. Mehtre, and S.~Sangeetha.
\newblock User {Behavior} {Profiling} using {Ensemble} {Approach} for {Insider}
  {Threat} {Detection}.
\newblock In \emph{2019 {IEEE} 5th {International} {Conference} on {Identity},
  {Security}, and {Behavior} {Analysis} ({ISBA})}, pages 1--8, January 2019.
\newblock \doi{10.1109/ISBA.2019.8778466}.
\newblock URL
  \url{https://ieeexplore.ieee.org/abstract/document/8778466?casa_token=-dP5bf2m0PkAAAAA:qUdg7W27ArLk0Rg6ahD_Pz9NjOhlsCNYAeRl8Xz-l6HCq1XH3WtDVXCfGiiax6YKovKcvwvWxWoI}.

\bibitem[Singh et~al.(2021)Singh, Othman, Ahmed, Mahmood, Dhahri, and
  Choudhury]{singh_optimized_2021}
Pradeep~Kumar Singh, Esam Othman, Rafeeq Ahmed, Awais Mahmood, Habib Dhahri,
  and Prasenjit Choudhury.
\newblock Optimized recommendations by user profiling using apriori algorithm.
\newblock \emph{Applied Soft Computing}, 106:\penalty0 107272, July 2021.
\newblock ISSN 15684946.
\newblock \doi{10.1016/j.asoc.2021.107272}.
\newblock URL
  \url{https://linkinghub.elsevier.com/retrieve/pii/S1568494621001952}.

\bibitem[Skillen et~al.(2012)Skillen, Chen, Nugent, Donnelly, Burns, and
  Solheim]{skillen_ontological_2012}
Kerry-Louise Skillen, Liming Chen, Chris~D. Nugent, Mark~P. Donnelly, William
  Burns, and Ivar Solheim.
\newblock Ontological {User} {Profile} {Modeling} for {Context}-{Aware}
  {Application} {Personalization}.
\newblock In José Bravo, Diego López-de Ipiña, and Francisco Moya, editors,
  \emph{Ubiquitous {Computing} and {Ambient} {Intelligence}}, Lecture {Notes}
  in {Computer} {Science}, pages 261--268, Berlin, Heidelberg, 2012. Springer.
\newblock ISBN 978-3-642-35377-2.
\newblock \doi{10.1007/978-3-642-35377-2_36}.

\bibitem[Sleeman(1985)]{sleeman_umfe_1985}
D.~Sleeman.
\newblock {UMFE}: {A} user modelling front-end subsystem.
\newblock \emph{International Journal of Man-Machine Studies}, 23\penalty0
  (1):\penalty0 71--88, July 1985.
\newblock ISSN 0020-7373.
\newblock \doi{10.1016/S0020-7373(85)80025-0}.
\newblock URL
  \url{https://www.sciencedirect.com/science/article/pii/S0020737385800250}.

\bibitem[Sola et~al.(2021)Sola, Meilicke, van~der Aa, and
  Stuckenschmidt]{sola_rule-based_2021}
Diana Sola, Christian Meilicke, Han van~der Aa, and Heiner Stuckenschmidt.
\newblock A {Rule}-{Based} {Recommendation} {Approach} for {Business} {Process}
  {Modeling}.
\newblock In Marcello La~Rosa, Shazia Sadiq, and Ernest Teniente, editors,
  \emph{Advanced {Information} {Systems} {Engineering}}, Lecture {Notes} in
  {Computer} {Science}, pages 328--343, Cham, 2021. Springer International
  Publishing.
\newblock ISBN 978-3-030-79382-1.
\newblock \doi{10.1007/978-3-030-79382-1_20}.

\bibitem[Solomon et~al.(2018)Solomon, Bar, Yanai, Shapira, and
  Rokach]{solomon_predict_2018}
Adir Solomon, Ariel Bar, Chen Yanai, Bracha Shapira, and Lior Rokach.
\newblock Predict {Demographic} {Information} {Using} {Word2vec} on {Spatial}
  {Trajectories}.
\newblock In \emph{Proceedings of the 26th {Conference} on {User} {Modeling},
  {Adaptation} and {Personalization}}, {UMAP} '18, pages 331--339, New York,
  NY, USA, 2018. Association for Computing Machinery.
\newblock ISBN 978-1-4503-5589-6.
\newblock \doi{10.1145/3209219.3209224}.
\newblock URL \url{https://dl.acm.org/doi/10.1145/3209219.3209224}.

\bibitem[Sosnovsky and Dicheva(2010)]{sosnovsky_ontological_2010}
Sergey Sosnovsky and Darina Dicheva.
\newblock Ontological technologies for user modelling.
\newblock \emph{International Journal of Metadata, Semantics and Ontologies},
  5\penalty0 (1):\penalty0 32, 2010.
\newblock ISSN 1744-2621, 1744-263X.
\newblock \doi{10.1504/IJMSO.2010.032649}.
\newblock URL \url{http://www.inderscience.com/link.php?id=32649}.

\bibitem[Stanescu et~al.(2013)Stanescu, Nagar, and
  Caragea]{stanescu_hybrid_2013}
Ana Stanescu, Swapnil Nagar, and Doina Caragea.
\newblock A {Hybrid} {Recommender} {System}: {User} {Profiling} from {Keywords}
  and {Ratings}.
\newblock In \emph{2013 {IEEE}/{WIC}/{ACM} {International} {Joint}
  {Conferences} on {Web} {Intelligence} ({WI}) and {Intelligent} {Agent}
  {Technologies} ({IAT})}, volume~1, pages 73--80, November 2013.
\newblock \doi{10.1109/WI-IAT.2013.11}.
\newblock URL \url{https://ieeexplore.ieee.org/abstract/document/6689996}.

\bibitem[Streeb et~al.(2022)Streeb, Metz, Schlegel, Schneider, El-Assady, Neth,
  Chen, and Keim]{streeb_task-based_2022}
Dirk Streeb, Yannick Metz, Udo Schlegel, Bruno Schneider, Mennatallah
  El-Assady, Hansjorg Neth, Min Chen, and Daniel~A. Keim.
\newblock Task-{Based} {Visual} {Interactive} {Modeling}: {Decision} {Trees}
  and {Rule}-{Based} {Classifiers}.
\newblock \emph{IEEE Transactions on Visualization and Computer Graphics},
  28\penalty0 (9):\penalty0 3307--3323, September 2022.
\newblock ISSN 1077-2626, 1941-0506, 2160-9306.
\newblock \doi{10.1109/TVCG.2020.3045560}.
\newblock URL \url{https://ieeexplore.ieee.org/document/9321557/}.

\bibitem[Streicher and Bauer(2024)]{streicher_cognitive_2024}
Alexander Streicher and Kolja Bauer.
\newblock Cognitive {User} {Modeling} for {Adaptivity} in {Serious} {Games}.
\newblock In \emph{Intelligent {Human} {Systems} {Integration} ({IHSI} 2024):
  {Integrating} {People} and {Intelligent} {Systems}}, volume 119. AHFE Open
  Acces, 2024.
\newblock ISBN 978-1-958651-95-7.
\newblock \doi{10.54941/ahfe1004472}.
\newblock URL
  \url{https://openaccess.cms-conferences.org/publications/book/978-1-958651-95-7/article/978-1-958651-95-7_8}.

\bibitem[Subramaniyaswamy and Logesh(2017)]{subramaniyaswamy_adaptive_2017}
V.~Subramaniyaswamy and R.~Logesh.
\newblock Adaptive {KNN} based {Recommender} {System} through {Mining} of
  {User} {Preferences}.
\newblock \emph{Wireless Personal Communications}, 97\penalty0 (2):\penalty0
  2229--2247, November 2017.
\newblock ISSN 1572-834X.
\newblock \doi{10.1007/s11277-017-4605-5}.
\newblock URL \url{https://doi.org/10.1007/s11277-017-4605-5}.

\bibitem[Sudhakar et~al.(2022)Sudhakar, Smail, Reddy, Shitharth, Tripathi, and
  Fahlevi]{sudhakar_web_2022}
K~Sudhakar, Boussaadi Smail, Tatireddy~Subba Reddy, S~Shitharth,
  Diwakar~Ramanuj Tripathi, and Mochammad Fahlevi.
\newblock Web {User} {Profile} {Generation} and {Discovery} {Analysis} using
  {LSTM} {Architecture}.
\newblock In \emph{2022 2nd {International} {Conference} on {Technological}
  {Advancements} in {Computational} {Sciences} ({ICTACS})}, pages 371--375,
  October 2022.
\newblock \doi{10.1109/ICTACS56270.2022.9988505}.
\newblock URL \url{https://ieeexplore.ieee.org/document/9988505}.

\bibitem[Sun et~al.(2020)Sun, Qian, Chen, Liang, Nguyen, and
  Yin]{sun_where_2020}
Ke~Sun, Tieyun Qian, Tong Chen, Yile Liang, Quoc Viet~Hung Nguyen, and Hongzhi
  Yin.
\newblock Where to {Go} {Next}: {Modeling} {Long}- and {Short}-{Term} {User}
  {Preferences} for {Point}-of-{Interest} {Recommendation}.
\newblock \emph{Proceedings of the AAAI Conference on Artificial Intelligence},
  34\penalty0 (01):\penalty0 214--221, April 2020.
\newblock ISSN 2374-3468.
\newblock \doi{10.1609/aaai.v34i01.5353}.
\newblock URL \url{https://ojs.aaai.org/index.php/AAAI/article/view/5353}.

\bibitem[Sunar et~al.(2020)Sunar, Abbasi, Davis, White, and
  Aljohani]{sunar_modelling_2020}
Ayse~Saliha Sunar, Rabeeh~Ayaz Abbasi, Hugh~C. Davis, Su~White, and Naif~R.
  Aljohani.
\newblock Modelling {MOOC} learners' social behaviours.
\newblock \emph{Computers in Human Behavior}, 107:\penalty0 105835, June 2020.
\newblock ISSN 0747-5632.
\newblock \doi{10.1016/j.chb.2018.12.013}.
\newblock URL
  \url{https://www.sciencedirect.com/science/article/pii/S0747563218305995}.

\bibitem[Sánchez and Bellogín(2019)]{sanchez_building_2019}
Pablo Sánchez and Alejandro Bellogín.
\newblock Building user profiles based on sequences for content and
  collaborative filtering.
\newblock \emph{Information Processing \& Management}, 56\penalty0
  (1):\penalty0 192--211, January 2019.
\newblock ISSN 0306-4573.
\newblock \doi{10.1016/j.ipm.2018.10.003}.
\newblock URL
  \url{https://www.sciencedirect.com/science/article/pii/S0306457318302735}.

\bibitem[Tan and Jiang(2023)]{tan_user_2023}
Zhaoxuan Tan and Meng Jiang.
\newblock User {Modeling} in the {Era} of {Large} {Language} {Models}:
  {Current} {Research} and {Future} {Directions}, December 2023.
\newblock URL \url{https://arxiv.org/abs/2312.11518v2}.

\bibitem[Tang et~al.(2015)Tang, Qin, Liu, and Yang]{tang_user_2015}
Duyu Tang, Bing Qin, Ting Liu, and Yuekui Yang.
\newblock User modeling with neural network for review rating prediction.
\newblock In \emph{Proceedings of the 24th {International} {Conference} on
  {Artificial} {Intelligence}}, {IJCAI}'15, pages 1340--1346, Buenos Aires,
  Argentina, 2015. AAAI Press.
\newblock ISBN 978-1-57735-738-4.

\bibitem[Tang et~al.(2010)Tang, Yao, Zhang, and Zhang]{tang_combination_2010}
Jie Tang, Limin Yao, Duo Zhang, and Jing Zhang.
\newblock A {Combination} {Approach} to {Web} {User} {Profiling}.
\newblock \emph{ACM Transactions on Knowledge Discovery from Data}, 5\penalty0
  (1):\penalty0 2:1--2:44, December 2010.
\newblock ISSN 1556-4681.
\newblock \doi{10.1145/1870096.1870098}.
\newblock URL \url{https://dl.acm.org/doi/10.1145/1870096.1870098}.

\bibitem[Tchuente(2022)]{tchuente_user_2022}
Dieudonne Tchuente.
\newblock User {Modeling} and {Profiling} in {Information} {Systems}: {A}
  {Bibliometric} {Study} and {Future} {Research} {Directions}.
\newblock \emph{Journal of Global Information Management (JGIM)}, 30\penalty0
  (1):\penalty0 1--25, January 2022.
\newblock ISSN 1062-7375.
\newblock \doi{10.4018/JGIM.307116}.
\newblock URL
  \url{https://www.igi-global.com/article/user-modeling-and-profiling-in-information-systems/www.igi-global.com/article/user-modeling-and-profiling-in-information-systems/307116}.

\bibitem[Temor et~al.(2022)Temor, Husain, and Coppin]{temor_cross-modal_2022}
Lucas Temor, Zainab Husain, and Peter Coppin.
\newblock A cross-modal {UX} design pedagogy for industrial design.
\newblock In \emph{Proceedings of the 17th {International} {Audio} {Mostly}
  {Conference}}, {AM} '22, pages 195--198, New York, NY, USA, 2022. Association
  for Computing Machinery.
\newblock ISBN 978-1-4503-9701-8.
\newblock \doi{10.1145/3561212.3561241}.
\newblock URL \url{https://doi.org/10.1145/3561212.3561241}.

\bibitem[Tenemaza(2022)]{tenemaza_user_2022}
Maritzol Tenemaza.
\newblock User models for recommendation systems.
\newblock In \emph{Human {Factors} and {Systems} {Interaction}}, volume~52.
  AHFE Open Acces, 2022.
\newblock ISBN 978-1-958651-28-5.
\newblock \doi{10.54941/ahfe1002188}.
\newblock URL
  \url{https://openaccess.cms-conferences.org/publications/book/978-1-958651-28-5/article/978-1-958651-28-5_56}.

\bibitem[Tomeo et~al.(2015)Tomeo, Noia, de~Gemmis, Lops, Semeraro, and
  Sciascio]{tomeo_exploiting_2015}
Paolo Tomeo, Tommaso~Di Noia, Marco de~Gemmis, Pasquale Lops, Giovanni
  Semeraro, and Eugenio~Di Sciascio.
\newblock Exploiting {Regression} {Trees} as {User} {Models} for
  {Intent}-{Aware} {Multi}-attribute {Diversity}.
\newblock In \emph{Proceedings of the 2nd {Workshop} on {New} {Trends} on
  {Content}-{Based} {Recommender} {Systems} co-located with 9th {ACM}
  {Conference} on {Recommender} {Systems} ({RecSys} 2015)}, pages 2--9, Vienna,
  Austria, 2015.

\bibitem[Tripathi et~al.(2018)Tripathi, T.S., and
  Guddeti]{tripathi_reinforcement_2018}
Abhishek Tripathi, Ashwin T.S., and Ram Mohana~Reddy Guddeti.
\newblock A {Reinforcement} {Learning} and {Recurrent} {Neural} {Network}
  {Based} {Dynamic} {User} {Modeling} {System}.
\newblock In \emph{2018 {IEEE} 18th {International} {Conference} on {Advanced}
  {Learning} {Technologies} ({ICALT})}, pages 411--415, July 2018.
\newblock \doi{10.1109/ICALT.2018.00103}.
\newblock URL
  \url{https://ieeexplore.ieee.org/abstract/document/8433551?casa_token=BpGaGAEPhu4AAAAA:x42YQHbh30JK5eSer5Mky4k8Y3pTxF-8YxZIZZ3qmN_7I2GYkMSuauIa9i2AJ1KP9Sj71CsT7KpL}.

\bibitem[Tripathi et~al.(2019)Tripathi, Yadav, and Rajan]{tripathi_naive_2019}
Akarshita Tripathi, Saumya Yadav, and Rajiv Rajan.
\newblock Naive {Bayes} {Classification} {Model} for the {Student}
  {Performance} {Prediction}.
\newblock In \emph{2019 2nd {International} {Conference} on {Intelligent}
  {Computing}, {Instrumentation} and {Control} {Technologies} ({ICICICT})},
  volume~1, pages 1548--1553, July 2019.
\newblock \doi{10.1109/ICICICT46008.2019.8993237}.
\newblock URL \url{https://ieeexplore.ieee.org/abstract/document/8993237}.

\bibitem[van Dam and van~de Velden(2015)]{van_dam_online_2015}
Jan-Willem van Dam and Michel van~de Velden.
\newblock Online profiling and clustering of {Facebook} users.
\newblock \emph{Decision Support Systems}, 70:\penalty0 60--72, February 2015.
\newblock ISSN 0167-9236.
\newblock \doi{10.1016/j.dss.2014.12.001}.
\newblock URL
  \url{https://www.sciencedirect.com/science/article/pii/S0167923614002796}.

\bibitem[Vardasbi et~al.(2020)Vardasbi, Oosterhuis, and
  de~Rijke]{vardasbi_when_2020}
Ali Vardasbi, Harrie Oosterhuis, and Maarten de~Rijke.
\newblock When {Inverse} {Propensity} {Scoring} does not {Work}: {Affine}
  {Corrections} for {Unbiased} {Learning} to {Rank}.
\newblock In \emph{Proceedings of the 29th {ACM} {International} {Conference}
  on {Information} \& {Knowledge} {Management}}, {CIKM} '20, pages 1475--1484,
  New York, NY, USA, 2020. Association for Computing Machinery.
\newblock ISBN 978-1-4503-6859-9.
\newblock \doi{10.1145/3340531.3412031}.
\newblock URL \url{https://dl.acm.org/doi/10.1145/3340531.3412031}.

\bibitem[Vassileva(2012)]{vassileva_motivating_2012}
Julita Vassileva.
\newblock Motivating participation in social computing applications: a user
  modeling perspective.
\newblock \emph{User Modeling and User-Adapted Interaction}, 22\penalty0
  (1):\penalty0 177--201, April 2012.
\newblock ISSN 1573-1391.
\newblock \doi{10.1007/s11257-011-9109-5}.
\newblock URL \url{https://doi.org/10.1007/s11257-011-9109-5}.

\bibitem[Verbert et~al.(2012)Verbert, Manouselis, Ochoa, Wolpers, Drachsler,
  Bosnic, and Duval]{verbert_context-aware_2012}
Katrien Verbert, Nikos Manouselis, Xavier Ochoa, Martin Wolpers, Hendrik
  Drachsler, Ivana Bosnic, and Erik Duval.
\newblock Context-{Aware} {Recommender} {Systems} for {Learning}: {A} {Survey}
  and {Future} {Challenges}.
\newblock \emph{IEEE Transactions on Learning Technologies}, 5\penalty0
  (4):\penalty0 318--335, October 2012.
\newblock ISSN 1939-1382, 2372-0050.
\newblock \doi{10.1109/TLT.2012.11}.
\newblock URL \url{https://ieeexplore.ieee.org/document/6189308/}.

\bibitem[Virós-i Martin and Selva(2022)]{viros-i-martin_improving_2022}
Antoni Virós-i Martin and Daniel Selva.
\newblock Improving {Designer} {Learning} in {Design} {Space} {Exploration} by
  {Adapting} to the {Designer}’s {Learning} {Goals}.
\newblock In \emph{International Design Engineering Technical Conferences \&
  Computers and Information in Engineering Conference}. American Society of
  Mechanical Engineers Digital Collection, November 2022.
\newblock \doi{10.1115/DETC2022-89207}.
\newblock URL \url{https://dx.doi.org/10.1115/DETC2022-89207}.

\bibitem[Viviani et~al.(2010)Viviani, Bennani, and
  Egyed-Zsigmond]{viviani_survey_2010}
Marco Viviani, Nadia Bennani, and Elod Egyed-Zsigmond.
\newblock A {Survey} on {User} {Modeling} in {Multi}-application
  {Environments}.
\newblock In \emph{2010 {Third} {International} {Conference} on {Advances} in
  {Human}-{Oriented} and {Personalized} {Mechanisms}, {Technologies} and
  {Services}}, pages 111--116, August 2010.
\newblock \doi{10.1109/CENTRIC.2010.30}.
\newblock URL \url{https://ieeexplore.ieee.org/abstract/document/5600339}.

\bibitem[Vo et~al.(2021)Vo, Karnjana, and Huynh]{vo_integrated_2021}
Duc-Vinh Vo, Jessada Karnjana, and Van-Nam Huynh.
\newblock An integrated framework of learning and evidential reasoning for user
  profiling using short texts.
\newblock \emph{Information Fusion}, 70:\penalty0 27--42, June 2021.
\newblock ISSN 15662535.
\newblock \doi{10.1016/j.inffus.2020.12.004}.
\newblock URL
  \url{https://linkinghub.elsevier.com/retrieve/pii/S1566253520304292}.

\bibitem[Vu et~al.(2015)Vu, Willis, Tran, and Song]{vu_temporal_2015}
Thanh Vu, Alistair Willis, Son~N. Tran, and Dawei Song.
\newblock Temporal {Latent} {Topic} {User} {Profiles} for {Search}
  {Personalisation}.
\newblock In Allan Hanbury, Gabriella Kazai, Andreas Rauber, and Norbert Fuhr,
  editors, \emph{Advances in {Information} {Retrieval}}, Lecture {Notes} in
  {Computer} {Science}, pages 605--616, Cham, 2015. Springer International
  Publishing.
\newblock ISBN 978-3-319-16354-3.
\newblock \doi{10.1007/978-3-319-16354-3_67}.

\bibitem[Wahlster and Kobsa(1989)]{wahlster_user_1989}
Wolfgang Wahlster and Alfred Kobsa.
\newblock User {Models} in {Dialog} {Systems}.
\newblock In Alfred Kobsa and Wolfgang Wahlster, editors, \emph{User {Models}
  in {Dialog} {Systems}}, Symbolic {Computation}, pages 4--34, Berlin,
  Heidelberg, 1989. Springer.
\newblock ISBN 978-3-642-83230-7.
\newblock \doi{10.1007/978-3-642-83230-7_1}.

\bibitem[Wanda and Jie(2020)]{wanda_deepprofile_2020}
Putra Wanda and Huang~Jin Jie.
\newblock {DeepProfile}: {Finding} fake profile in online social network using
  dynamic {CNN}.
\newblock \emph{Journal of Information Security and Applications}, 52:\penalty0
  102465, June 2020.
\newblock ISSN 22142126.
\newblock \doi{10.1016/j.jisa.2020.102465}.
\newblock URL
  \url{https://linkinghub.elsevier.com/retrieve/pii/S2214212619303801}.

\bibitem[Wang et~al.(2021{\natexlab{a}})Wang, Zhu, Liu, Ma, Zang, and
  Yu]{wang_enhancing_2021}
Chunyang Wang, Yanmin Zhu, Haobing Liu, Wenze Ma, Tianzi Zang, and Jiadi Yu.
\newblock Enhancing {User} {Interest} {Modeling} with {Knowledge}-{Enriched}
  {Itemsets} for {Sequential} {Recommendation}.
\newblock In \emph{Proceedings of the 30th {ACM} {International} {Conference}
  on {Information} \& {Knowledge} {Management}}, pages 1889--1898, Virtual
  Event Queensland Australia, October 2021{\natexlab{a}}. ACM.
\newblock ISBN 978-1-4503-8446-9.
\newblock \doi{10.1145/3459637.3482256}.
\newblock URL \url{https://dl.acm.org/doi/10.1145/3459637.3482256}.

\bibitem[Wang et~al.(2020{\natexlab{a}})Wang, Jiang, Syed, Conway, Juneja,
  Subramanian, and Chawla]{wang_calendar_2020}
Daheng Wang, Meng Jiang, Munira Syed, Oliver Conway, Vishal Juneja, Sriram
  Subramanian, and Nitesh~V. Chawla.
\newblock Calendar {Graph} {Neural} {Networks} for {Modeling} {Time}
  {Structures} in {Spatiotemporal} {User} {Behaviors}.
\newblock In \emph{Proceedings of the 26th {ACM} {SIGKDD} {International}
  {Conference} on {Knowledge} {Discovery} \& {Data} {Mining}}, {KDD} '20, pages
  2581--2589, New York, NY, USA, 2020{\natexlab{a}}. Association for Computing
  Machinery.
\newblock ISBN 978-1-4503-7998-4.
\newblock \doi{10.1145/3394486.3403308}.
\newblock URL \url{https://dl.acm.org/doi/10.1145/3394486.3403308}.

\bibitem[Wang et~al.(2019{\natexlab{a}})Wang, Yang, Abdul, and
  Lim]{wang_designing_2019}
Danding Wang, Qian Yang, Ashraf Abdul, and Brian~Y. Lim.
\newblock Designing {Theory}-{Driven} {User}-{Centric} {Explainable} {AI}.
\newblock In \emph{Proceedings of the 2019 {CHI} {Conference} on {Human}
  {Factors} in {Computing} {Systems}}, {CHI} '19, pages 1--15, New York, NY,
  USA, 2019{\natexlab{a}}. Association for Computing Machinery.
\newblock ISBN 978-1-4503-5970-2.
\newblock \doi{10.1145/3290605.3300831}.
\newblock URL \url{https://dl.acm.org/doi/10.1145/3290605.3300831}.

\bibitem[Wang et~al.(2021{\natexlab{b}})Wang, Wang, Liu, Zhou, Hughes, and
  Fu]{wang_reinforced_2021}
Dongjie Wang, Pengyang Wang, Kunpeng Liu, Yuanchun Zhou, Charles~E. Hughes, and
  Yanjie Fu.
\newblock Reinforced {Imitative} {Graph} {Representation} {Learning} for
  {Mobile} {User} {Profiling}: {An} {Adversarial} {Training} {Perspective}.
\newblock \emph{Proceedings of the AAAI Conference on Artificial Intelligence},
  35\penalty0 (5):\penalty0 4410--4417, May 2021{\natexlab{b}}.
\newblock ISSN 2374-3468.
\newblock \doi{10.1609/aaai.v35i5.16567}.
\newblock URL \url{https://ojs.aaai.org/index.php/AAAI/article/view/16567}.

\bibitem[Wang et~al.(2023)Wang, Jiang, Li, and Zhao]{wang_improving_2023}
Jingkun Wang, Yongtao Jiang, Haochen Li, and Wen Zhao.
\newblock Improving {News} {Recommendation} with {Channel}-{Wise} {Dynamic}
  {Representations} and {Contrastive} {User} {Modeling}.
\newblock In \emph{Proceedings of the {Sixteenth} {ACM} {International}
  {Conference} on {Web} {Search} and {Data} {Mining}}, pages 562--570,
  Singapore Singapore, February 2023. ACM.
\newblock ISBN 978-1-4503-9407-9.
\newblock \doi{10.1145/3539597.3570447}.
\newblock URL \url{https://dl.acm.org/doi/10.1145/3539597.3570447}.

\bibitem[Wang et~al.(2019{\natexlab{b}})Wang, Fu, Xiong, and
  Li]{wang_adversarial_2019}
Pengyang Wang, Yanjie Fu, Hui Xiong, and Xiaolin Li.
\newblock Adversarial {Substructured} {Representation} {Learning} for {Mobile}
  {User} {Profiling}.
\newblock In \emph{Proceedings of the 25th {ACM} {SIGKDD} {International}
  {Conference} on {Knowledge} {Discovery} \& {Data} {Mining}}, {KDD} '19, pages
  130--138, New York, NY, USA, 2019{\natexlab{b}}. Association for Computing
  Machinery.
\newblock ISBN 978-1-4503-6201-6.
\newblock \doi{10.1145/3292500.3330869}.
\newblock URL \url{https://dl.acm.org/doi/10.1145/3292500.3330869}.

\bibitem[Wang et~al.(2020{\natexlab{b}})Wang, Liu, Jiang, Li, and
  Fu]{wang_incremental_2020}
Pengyang Wang, Kunpeng Liu, Lu~Jiang, Xiaolin Li, and Yanjie Fu.
\newblock Incremental {Mobile} {User} {Profiling}: {Reinforcement} {Learning}
  with {Spatial} {Knowledge} {Graph} for {Modeling} {Event} {Streams}.
\newblock In \emph{Proceedings of the 26th {ACM} {SIGKDD} {International}
  {Conference} on {Knowledge} {Discovery} \& {Data} {Mining}}, {KDD} '20, pages
  853--861, New York, NY, USA, August 2020{\natexlab{b}}. Association for
  Computing Machinery.
\newblock ISBN 978-1-4503-7998-4.
\newblock \doi{10.1145/3394486.3403128}.
\newblock URL \url{https://dl.acm.org/doi/10.1145/3394486.3403128}.

\bibitem[Wang et~al.(2020{\natexlab{c}})Wang, Zhu, Wang, Chen, Gao, and
  Xin]{wang_user_2020}
Ruiheng Wang, Hongliang Zhu, Lu~Wang, Zhaoyun Chen, Mingcheng Gao, and Yang
  Xin.
\newblock User {Identity} {Linkage} {Across} {Social} {Networks} by
  {Heterogeneous} {Graph} {Attention} {Network} {Modeling}.
\newblock \emph{Applied Sciences}, 10\penalty0 (16):\penalty0 5478, January
  2020{\natexlab{c}}.
\newblock ISSN 2076-3417.
\newblock \doi{10.3390/app10165478}.
\newblock URL \url{https://www.mdpi.com/2076-3417/10/16/5478}.

\bibitem[Wang et~al.(2022{\natexlab{a}})Wang, Wu, Lou, and
  Jiang]{wang_attention-based_2022}
Ruiqin Wang, Zongda Wu, Jungang Lou, and Yunliang Jiang.
\newblock Attention-based dynamic user modeling and {Deep} {Collaborative}
  filtering recommendation.
\newblock \emph{Expert Systems with Applications}, 188:\penalty0 116036,
  February 2022{\natexlab{a}}.
\newblock ISSN 09574174.
\newblock \doi{10.1016/j.eswa.2021.116036}.
\newblock URL
  \url{https://linkinghub.elsevier.com/retrieve/pii/S0957417421013816}.

\bibitem[Wang et~al.(2022{\natexlab{b}})Wang, Sun, Fang, Yang, and
  Wang]{wang_modeling_2022}
Xiaolin Wang, Guohao Sun, Xiu Fang, Jian Yang, and Shoujin Wang.
\newblock Modeling {Spatio}-temporal {Neighbourhood} for {Personalized}
  {Point}-of-interest {Recommendation}.
\newblock In \emph{Proceedings of the {Thirty}-{First} {International} {Joint}
  {Conference} on {Artificial} {Intelligence}}, pages 3530--3536, Vienna,
  Austria, July 2022{\natexlab{b}}. International Joint Conferences on
  Artificial Intelligence Organization.
\newblock ISBN 978-1-956792-00-3.
\newblock \doi{10.24963/ijcai.2022/490}.
\newblock URL \url{https://www.ijcai.org/proceedings/2022/490}.

\bibitem[Wei et~al.(2017)Wei, Zhang, Yuan, Cao, Fu, Xie, Rui, and
  Ma]{wei_beyond_2017}
Honghao Wei, Fuzheng Zhang, Nicholas~Jing Yuan, Chuan Cao, Hao Fu, Xing Xie,
  Yong Rui, and Wei-Ying Ma.
\newblock Beyond the {Words}: {Predicting} {User} {Personality} from
  {Heterogeneous} {Information}.
\newblock In \emph{Proceedings of the {Tenth} {ACM} {International}
  {Conference} on {Web} {Search} and {Data} {Mining}}, {WSDM} '17, pages
  305--314, New York, NY, USA, February 2017. Association for Computing
  Machinery.
\newblock ISBN 978-1-4503-4675-7.
\newblock \doi{10.1145/3018661.3018717}.
\newblock URL \url{https://dl.acm.org/doi/10.1145/3018661.3018717}.

\bibitem[Wei et~al.(2022)Wei, Wang, He, Nie, Rui, and
  Chua]{wei_hierarchical_2022}
Yinwei Wei, Xiang Wang, Xiangnan He, Liqiang Nie, Yong Rui, and Tat-Seng Chua.
\newblock Hierarchical {User} {Intent} {Graph} {Network} for {Multimedia}
  {Recommendation}.
\newblock \emph{IEEE Transactions on Multimedia}, 24:\penalty0 2701--2712,
  2022.
\newblock ISSN 1520-9210, 1941-0077.
\newblock \doi{10.1109/TMM.2021.3088307}.
\newblock URL \url{https://ieeexplore.ieee.org/document/9453189/}.

\bibitem[Wen et~al.(2021)Wen, Zhang, Lv, Bao, Wang, and
  Chen]{wen_hierarchically_2021}
Hong Wen, Jing Zhang, Fuyu Lv, Wentian Bao, Tianyi Wang, and Zulong Chen.
\newblock Hierarchically {Modeling} {Micro} and {Macro} {Behaviors} via
  {Multi}-{Task} {Learning} for {Conversion} {Rate} {Prediction}.
\newblock In \emph{Proceedings of the 44th {International} {ACM} {SIGIR}
  {Conference} on {Research} and {Development} in {Information} {Retrieval}},
  pages 2187--2191, Virtual Event Canada, July 2021. ACM.
\newblock ISBN 978-1-4503-8037-9.
\newblock \doi{10.1145/3404835.3463053}.
\newblock URL \url{https://dl.acm.org/doi/10.1145/3404835.3463053}.

\bibitem[Wu et~al.(2019{\natexlab{a}})Wu, Wu, Liu, and
  Huang]{wu_hierarchical_2019}
Chuhan Wu, Fangzhao Wu, Junxin Liu, and Yongfeng Huang.
\newblock Hierarchical {User} and {Item} {Representation} with {Three}-{Tier}
  {Attention} for {Recommendation}.
\newblock In Jill Burstein, Christy Doran, and Thamar Solorio, editors,
  \emph{Proceedings of the 2019 {Conference} of the {North} {American}
  {Chapter} of the {Association} for {Computational} {Linguistics}: {Human}
  {Language} {Technologies}, {Volume} 1 ({Long} and {Short} {Papers})}, pages
  1818--1826, Minneapolis, Minnesota, June 2019{\natexlab{a}}. Association for
  Computational Linguistics.
\newblock \doi{10.18653/v1/N19-1180}.
\newblock URL \url{https://aclanthology.org/N19-1180}.

\bibitem[Wu et~al.(2020{\natexlab{a}})Wu, Wu, Qi, and Huang]{wu_user_2020}
Chuhan Wu, Fangzhao Wu, Tao Qi, and Yongfeng Huang.
\newblock User {Modeling} with {Click} {Preference} and {Reading}
  {Satisfaction} for {News} {Recommendation}.
\newblock In \emph{Proceedings of the {Twenty}-{Ninth} {International} {Joint}
  {Conference} on {Artificial} {Intelligence}}, pages 3023--3029, Yokohama,
  Japan, July 2020{\natexlab{a}}. International Joint Conferences on Artificial
  Intelligence Organization.
\newblock ISBN 978-0-9992411-6-5.
\newblock \doi{10.24963/ijcai.2020/418}.
\newblock URL \url{https://www.ijcai.org/proceedings/2020/418}.

\bibitem[Wu et~al.(2020{\natexlab{b}})Wu, Wu, Qi, Lian, Huang, and
  Xie]{wu_ptum_2020}
Chuhan Wu, Fangzhao Wu, Tao Qi, Jianxun Lian, Yongfeng Huang, and Xing Xie.
\newblock {PTUM}: {Pre}-training {User} {Model} from {Unlabeled} {User}
  {Behaviors} via {Self}-supervision.
\newblock In Trevor Cohn, Yulan He, and Yang Liu, editors, \emph{Findings of
  the {Association} for {Computational} {Linguistics}: {EMNLP} 2020}, pages
  1939--1944, Online, November 2020{\natexlab{b}}. Association for
  Computational Linguistics.
\newblock \doi{10.18653/v1/2020.findings-emnlp.174}.
\newblock URL \url{https://aclanthology.org/2020.findings-emnlp.174}.

\bibitem[Wu et~al.(2021{\natexlab{a}})Wu, Wu, Huang, and
  Xie]{wu_user-as-graph_2021}
Chuhan Wu, Fangzhao Wu, Yongfeng Huang, and Xing Xie.
\newblock User-as-{Graph}: {User} {Modeling} with {Heterogeneous} {Graph}
  {Pooling} for {News} {Recommendation}.
\newblock In \emph{Proceedings of the {Thirtieth} {International} {Joint}
  {Conference} on {Artificial} {Intelligence}}, pages 1624--1630, Montreal,
  Canada, August 2021{\natexlab{a}}. International Joint Conferences on
  Artificial Intelligence Organization.
\newblock ISBN 978-0-9992411-9-6.
\newblock \doi{10.24963/ijcai.2021/224}.
\newblock URL \url{https://www.ijcai.org/proceedings/2021/224}.

\bibitem[Wu et~al.(2022)Wu, Wu, Qi, and Huang]{wu_userbert_2022}
Chuhan Wu, Fangzhao Wu, Tao Qi, and Yongfeng Huang.
\newblock {UserBERT}: {Pre}-training {User} {Model} with {Contrastive}
  {Self}-supervision.
\newblock In \emph{Proceedings of the 45th {International} {ACM} {SIGIR}
  {Conference} on {Research} and {Development} in {Information} {Retrieval}},
  {SIGIR} '22, pages 2087--2092, New York, NY, USA, 2022. Association for
  Computing Machinery.
\newblock ISBN 978-1-4503-8732-3.
\newblock \doi{10.1145/3477495.3531810}.
\newblock URL \url{https://dl.acm.org/doi/10.1145/3477495.3531810}.

\bibitem[Wu et~al.(2021{\natexlab{b}})Wu, Liu, Huang, Ning, Wang, Chen, Yi, and
  Zhou]{wu_hierarchical_2021}
Jinze Wu, Qi~Liu, Zhenya Huang, Yuting Ning, Hao Wang, Enhong Chen, Jinfeng Yi,
  and Bowen Zhou.
\newblock Hierarchical {Personalized} {Federated} {Learning} for {User}
  {Modeling}.
\newblock In \emph{Proceedings of the {Web} {Conference} 2021}, {WWW} '21,
  pages 957--968, New York, NY, USA, 2021{\natexlab{b}}. Association for
  Computing Machinery.
\newblock ISBN 978-1-4503-8312-7.
\newblock \doi{10.1145/3442381.3449926}.
\newblock URL \url{https://dl.acm.org/doi/10.1145/3442381.3449926}.

\bibitem[Wu et~al.(2019{\natexlab{b}})Wu, Li, Zhao, and Qian]{wu_long-_2019}
Yuxia Wu, Ke~Li, Guoshuai Zhao, and Xueming Qian.
\newblock Long- and {Short}-term {Preference} {Learning} for {Next} {POI}
  {Recommendation}.
\newblock In \emph{Proceedings of the 28th {ACM} {International} {Conference}
  on {Information} and {Knowledge} {Management}}, {CIKM} '19, pages 2301--2304,
  New York, NY, USA, November 2019{\natexlab{b}}. Association for Computing
  Machinery.
\newblock ISBN 978-1-4503-6976-3.
\newblock \doi{10.1145/3357384.3358171}.
\newblock URL \url{https://dl.acm.org/doi/10.1145/3357384.3358171}.

\bibitem[Xia et~al.(2021{\natexlab{a}})Xia, Huang, Xu, Dai, Zhang, Yang, Pei,
  and Bo]{xia_knowledge-enhanced_2021}
Lianghao Xia, Chao Huang, Yong Xu, Peng Dai, Xiyue Zhang, Hongsheng Yang, Jian
  Pei, and Liefeng Bo.
\newblock Knowledge-{Enhanced} {Hierarchical} {Graph} {Transformer} {Network}
  for {Multi}-{Behavior} {Recommendation}.
\newblock \emph{Proceedings of the AAAI Conference on Artificial Intelligence},
  35\penalty0 (5):\penalty0 4486--4493, May 2021{\natexlab{a}}.
\newblock ISSN 2374-3468.
\newblock \doi{10.1609/aaai.v35i5.16576}.
\newblock URL \url{https://ojs.aaai.org/index.php/AAAI/article/view/16576}.

\bibitem[Xia et~al.(2021{\natexlab{b}})Xia, Xu, Huang, Dai, and
  Bo]{xia_graph_2021}
Lianghao Xia, Yong Xu, Chao Huang, Peng Dai, and Liefeng Bo.
\newblock Graph {Meta} {Network} for {Multi}-{Behavior} {Recommendation}.
\newblock In \emph{Proceedings of the 44th {International} {ACM} {SIGIR}
  {Conference} on {Research} and {Development} in {Information} {Retrieval}},
  {SIGIR} '21, pages 757--766, New York, NY, USA, 2021{\natexlab{b}}.
  Association for Computing Machinery.
\newblock ISBN 978-1-4503-8037-9.
\newblock \doi{10.1145/3404835.3462972}.
\newblock URL \url{https://dl.acm.org/doi/10.1145/3404835.3462972}.

\bibitem[Xian et~al.(2021)Xian, Zhao, Li, Chan, Kan, Ma, Dong, Faloutsos,
  Karypis, Muthukrishnan, and Zhang]{xian_ex3_2021}
Yikun Xian, Tong Zhao, Jin Li, Jim Chan, Andrey Kan, Jun Ma, Xin~Luna Dong,
  Christos Faloutsos, George Karypis, S.~Muthukrishnan, and Yongfeng Zhang.
\newblock {EX3}: {Explainable} {Attribute}-aware {Item}-set {Recommendations}.
\newblock In \emph{Fifteenth {ACM} {Conference} on {Recommender} {Systems}},
  pages 484--494, Amsterdam Netherlands, September 2021. ACM.
\newblock ISBN 978-1-4503-8458-2.
\newblock \doi{10.1145/3460231.3474240}.
\newblock URL \url{https://dl.acm.org/doi/10.1145/3460231.3474240}.

\bibitem[Xie and Yu(2009)]{xie_large-scale_2009}
Yi~Xie and Shun-Zheng Yu.
\newblock A {Large}-{Scale} {Hidden} {Semi}-{Markov} {Model} for {Anomaly}
  {Detection} on {User} {Browsing} {Behaviors}.
\newblock \emph{IEEE/ACM Transactions on Networking}, 17\penalty0 (1):\penalty0
  54--65, February 2009.
\newblock ISSN 1558-2566.
\newblock \doi{10.1109/TNET.2008.923716}.
\newblock URL
  \url{https://ieeexplore.ieee.org/abstract/document/4515888?casa_token=ZOYglc4HMlIAAAAA:MBeNUBQELp00Pr41tN22uSBK3BB_zJUbsXvHR2bceHVLS-NN6NvK6W5awUZlSCBSH1Bu1I45z2Wr}.
\newblock Conference Name: IEEE/ACM Transactions on Networking.

\bibitem[Xu et~al.(2022)Xu, Zhai, and Rosenberg]{xu_rethinking_2022}
Jiajing Xu, Andrew Zhai, and Charles Rosenberg.
\newblock Rethinking {Personalized} {Ranking} at {Pinterest}: {An}
  {End}-to-{End} {Approach}.
\newblock In \emph{Proceedings of the 16th {ACM} {Conference} on {Recommender}
  {Systems}}, pages 502--505, Seattle WA USA, September 2022. ACM.
\newblock ISBN 978-1-4503-9278-5.
\newblock \doi{10.1145/3523227.3547394}.
\newblock URL \url{https://dl.acm.org/doi/10.1145/3523227.3547394}.

\bibitem[Xuan et~al.(2023)Xuan, Liu, Li, and Yin]{xuan_knowledge_2023}
Hongrui Xuan, Yi~Liu, Bohan Li, and Hongzhi Yin.
\newblock Knowledge {Enhancement} for {Contrastive} {Multi}-{Behavior}
  {Recommendation}.
\newblock In \emph{Proceedings of the {Sixteenth} {ACM} {International}
  {Conference} on {Web} {Search} and {Data} {Mining}}, pages 195--203,
  Singapore Singapore, February 2023. ACM.
\newblock ISBN 978-1-4503-9407-9.
\newblock \doi{10.1145/3539597.3570386}.
\newblock URL \url{https://dl.acm.org/doi/10.1145/3539597.3570386}.

\bibitem[Xue et~al.(2022)Xue, Yang, and Xiao]{xue_factorial_2022}
Lyuxin Xue, Deqing Yang, and Yanghua Xiao.
\newblock Factorial {User} {Modeling} with {Hierarchical} {Graph} {Neural}
  {Network} for {Enhanced} {Sequential} {Recommendation}.
\newblock In \emph{2022 {IEEE} {International} {Conference} on {Multimedia} and
  {Expo} ({ICME})}, pages 01--06, July 2022.
\newblock \doi{10.1109/ICME52920.2022.9859593}.
\newblock URL
  \url{https://ieeexplore.ieee.org/abstract/document/9859593?casa_token=PAT2krzHwr4AAAAA:omo9Z2VMKd-NUqf84e9NK2JJCk8yYsivZV_VJljR-DOrRYlLzyXfYNaJS3CKG3h-yZo1oz-w6g}.

\bibitem[Yadav and Selvakumar(2015)]{yadav_detection_2015}
Satyajit Yadav and S.~Selvakumar.
\newblock Detection of application layer {DDoS} attack by modeling user
  behavior using logistic regression.
\newblock In \emph{2015 4th {International} {Conference} on {Reliability},
  {Infocom} {Technologies} and {Optimization} ({ICRITO}) ({Trends} and {Future}
  {Directions})}, pages 1--6, September 2015.
\newblock \doi{10.1109/ICRITO.2015.7359289}.
\newblock URL \url{https://ieeexplore.ieee.org/abstract/document/7359289}.

\bibitem[Yan et~al.(2021)Yan, Zhang, Liu, Wu, and
  Wang]{yan_relation-aware_2021}
Qilong Yan, Yufeng Zhang, Qiang Liu, Shu Wu, and Liang Wang.
\newblock Relation-aware {Heterogeneous} {Graph} for {User} {Profiling}.
\newblock In \emph{Proceedings of the 30th {ACM} {International} {Conference}
  on {Information} \& {Knowledge} {Management}}, {CIKM} '21, pages 3573--3577,
  New York, NY, USA, 2021. Association for Computing Machinery.
\newblock ISBN 978-1-4503-8446-9.
\newblock \doi{10.1145/3459637.3482170}.
\newblock URL \url{https://doi.org/10.1145/3459637.3482170}.

\bibitem[Yan et~al.(2022)Yan, Zhao, and Deng]{yan_interaction-aware_2022}
Shaojie Yan, Tao Zhao, and Jinsheng Deng.
\newblock Interaction-aware {Hypergraph} {Neural} {Networks} for {User}
  {Profiling}.
\newblock In \emph{2022 {IEEE} 9th {International} {Conference} on {Data}
  {Science} and {Advanced} {Analytics} ({DSAA})}, pages 1--10, Shenzhen, China,
  October 2022. IEEE.
\newblock ISBN 978-1-66547-330-9.
\newblock \doi{10.1109/DSAA54385.2022.10032374}.
\newblock URL \url{https://ieeexplore.ieee.org/document/10032374/}.

\bibitem[Yang et~al.(2023)Yang, Liu, Suzumura, Dong, and Li]{yang_going_2023}
Boming Yang, Dairui Liu, Toyotaro Suzumura, Ruihai Dong, and Irene Li.
\newblock Going {Beyond} {Local}: {Global} {Graph}-{Enhanced} {Personalized}
  {News} {Recommendations}.
\newblock In \emph{Proceedings of the 17th {ACM} {Conference} on {Recommender}
  {Systems}}, pages 24--34, Singapore Singapore, September 2023. ACM.
\newblock ISBN 9798400702419.
\newblock \doi{10.1145/3604915.3608801}.
\newblock URL \url{https://dl.acm.org/doi/10.1145/3604915.3608801}.

\bibitem[Yang et~al.(2022)Yang, Hou, Song, Zhang, Wen, and
  Zhao]{yang_modeling_2022}
Chen Yang, Yupeng Hou, Yang Song, Tao Zhang, Ji-Rong Wen, and Wayne~Xin Zhao.
\newblock Modeling {Two}-{Way} {Selection} {Preference} for {Person}-{Job}
  {Fit}.
\newblock In \emph{Proceedings of the 16th {ACM} {Conference} on {Recommender}
  {Systems}}, pages 102--112, Seattle WA USA, September 2022. ACM.
\newblock ISBN 978-1-4503-9278-5.
\newblock \doi{10.1145/3523227.3546752}.
\newblock URL \url{https://dl.acm.org/doi/10.1145/3523227.3546752}.

\bibitem[Yang et~al.(2021)Yang, Schnabel, Bennett, and Dumais]{yang_local_2021}
Longqi Yang, Tobias Schnabel, Paul~N. Bennett, and Susan Dumais.
\newblock Local {Factor} {Models} for {Large}-{Scale} {Inductive}
  {Recommendation}.
\newblock In \emph{Fifteenth {ACM} {Conference} on {Recommender} {Systems}},
  pages 252--262, Amsterdam Netherlands, September 2021. ACM.
\newblock ISBN 978-1-4503-8458-2.
\newblock \doi{10.1145/3460231.3474276}.
\newblock URL \url{https://dl.acm.org/doi/10.1145/3460231.3474276}.

\bibitem[Yilma et~al.(2021)Yilma, Naudet, and
  Panetto]{yilma_personalisation_2021}
Bereket~Abera Yilma, Yannick Naudet, and Hervé Panetto.
\newblock Personalisation in {Cyber}-{Physical}-{Social} {Systems}: {A}
  {Multi}-stakeholder aware {Recommendation} and {Guidance}.
\newblock In \emph{Proceedings of the 29th {ACM} {Conference} on {User}
  {Modeling}, {Adaptation} and {Personalization}}, pages 251--255, Utrecht
  Netherlands, June 2021. ACM.
\newblock ISBN 978-1-4503-8366-0.
\newblock \doi{10.1145/3450613.3456847}.
\newblock URL \url{https://dl.acm.org/doi/10.1145/3450613.3456847}.

\bibitem[Yu et~al.(2019)Yu, Lian, Mahmoody, Liu, and Xie]{yu_adaptive_2019}
Zeping Yu, Jianxun Lian, Ahmad Mahmoody, Gongshen Liu, and Xing Xie.
\newblock Adaptive {User} {Modeling} with {Long} and {Short}-{Term}
  {Preferences} for {Personalized} {Recommendation}.
\newblock In \emph{Proceedings of the {Twenty}-{Eighth} {International} {Joint}
  {Conference} on {Artificial} {Intelligence}}, pages 4213--4219, Macao, China,
  August 2019. International Joint Conferences on Artificial Intelligence
  Organization.
\newblock ISBN 978-0-9992411-4-1.
\newblock \doi{10.24963/ijcai.2019/585}.
\newblock URL \url{https://www.ijcai.org/proceedings/2019/585}.

\bibitem[Yuan et~al.(2020{\natexlab{a}})Yuan, He, Karatzoglou, and
  Zhang]{yuan_parameter-efficient_2020}
Fajie Yuan, Xiangnan He, Alexandros Karatzoglou, and Liguang Zhang.
\newblock Parameter-{Efficient} {Transfer} from {Sequential} {Behaviors} for
  {User} {Modeling} and {Recommendation}.
\newblock In \emph{Proceedings of the 43rd {International} {ACM} {SIGIR}
  {Conference} on {Research} and {Development} in {Information} {Retrieval}},
  pages 1469--1478, Virtual Event China, July 2020{\natexlab{a}}. ACM.
\newblock ISBN 978-1-4503-8016-4.
\newblock \doi{10.1145/3397271.3401156}.
\newblock URL \url{https://dl.acm.org/doi/10.1145/3397271.3401156}.

\bibitem[Yuan et~al.(2021)Yuan, Zhang, Karatzoglou, Jose, Kong, and
  Li]{yuan_one_2021}
Fajie Yuan, Guoxiao Zhang, Alexandros Karatzoglou, Joemon Jose, Beibei Kong,
  and Yudong Li.
\newblock One {Person}, {One} {Model}, {One} {World}: {Learning} {Continual}
  {User} {Representation} without {Forgetting}.
\newblock In \emph{Proceedings of the 44th {International} {ACM} {SIGIR}
  {Conference} on {Research} and {Development} in {Information} {Retrieval}},
  pages 696--705, Virtual Event Canada, July 2021. ACM.
\newblock ISBN 978-1-4503-8037-9.
\newblock \doi{10.1145/3404835.3462884}.
\newblock URL \url{https://dl.acm.org/doi/10.1145/3404835.3462884}.

\bibitem[Yuan et~al.(2020{\natexlab{b}})Yuan, Luo, Shang, and
  Wu]{yuan_generalized_2020}
Ye~Yuan, Xin Luo, Mingsheng Shang, and Di~Wu.
\newblock A {Generalized} and {Fast}-converging {Non}-negative {Latent}
  {Factor} {Model} for {Predicting} {User} {Preferences} in {Recommender}
  {Systems}.
\newblock In \emph{Proceedings of {The} {Web} {Conference} 2020}, {WWW} '20,
  pages 498--507, New York, NY, USA, April 2020{\natexlab{b}}. Association for
  Computing Machinery.
\newblock ISBN 978-1-4503-7023-3.
\newblock \doi{10.1145/3366423.3380133}.
\newblock URL \url{https://dl.acm.org/doi/10.1145/3366423.3380133}.

\bibitem[Zafarani and Liu(2013)]{zafarani_connecting_2013}
Reza Zafarani and Huan Liu.
\newblock Connecting users across social media sites: a behavioral-modeling
  approach.
\newblock In \emph{Proceedings of the 19th {ACM} {SIGKDD} international
  conference on {Knowledge} discovery and data mining}, pages 41--49, Chicago
  Illinois USA, August 2013. ACM.
\newblock ISBN 978-1-4503-2174-7.
\newblock \doi{10.1145/2487575.2487648}.
\newblock URL \url{https://dl.acm.org/doi/10.1145/2487575.2487648}.

\bibitem[Zeng(2021)]{zeng_how_2021}
Yilei Zeng.
\newblock How {Human} {Centered} {AI} {Will} {Contribute} {Towards}
  {Intelligent} {Gaming} {Systems}.
\newblock \emph{Proceedings of the AAAI Conference on Artificial Intelligence},
  35\penalty0 (18):\penalty0 15742--15743, May 2021.
\newblock ISSN 2374-3468.
\newblock \doi{10.1609/aaai.v35i18.17868}.
\newblock URL \url{https://ojs.aaai.org/index.php/AAAI/article/view/17868}.

\bibitem[Zerhoudi et~al.(2022)Zerhoudi, Granitzer, Seifert, and
  Schloetterer]{zerhoudi_evaluating_2022}
Saber Zerhoudi, Michael Granitzer, Christin Seifert, and Joerg Schloetterer.
\newblock Evaluating {Simulated} {User} {Interaction} and {Search} {Behaviour}.
\newblock In Matthias Hagen, Suzan Verberne, Craig Macdonald, Christin Seifert,
  Krisztian Balog, Kjetil Nørvåg, and Vinay Setty, editors, \emph{Advances in
  {Information} {Retrieval}}, Lecture {Notes} in {Computer} {Science}, pages
  240--247, Cham, 2022. Springer International Publishing.
\newblock ISBN 978-3-030-99739-7.
\newblock \doi{10.1007/978-3-030-99739-7_28}.

\bibitem[Zghal~Rebaï et~al.(2013)Zghal~Rebaï, Ghorbel, Zayani, and
  Amous]{zghal_rebai_adaptive_2013}
Rim Zghal~Rebaï, Leila Ghorbel, Corinne~Amel Zayani, and Ikram Amous.
\newblock An {Adaptive} {Method} for {User} {Profile} {Learning}.
\newblock In Barbara Catania, Giovanna Guerrini, and Jaroslav Pokorný,
  editors, \emph{Advances in {Databases} and {Information} {Systems}}, Lecture
  {Notes} in {Computer} {Science}, pages 126--134, Berlin, Heidelberg, 2013.
  Springer.
\newblock ISBN 978-3-642-40683-6.
\newblock \doi{10.1007/978-3-642-40683-6_10}.

\bibitem[Zhang et~al.(2023{\natexlab{a}})Zhang, Long, Zhou, Yan, Zhang, Zhang,
  and Yang]{zhang_dual_2023}
Chunxu Zhang, Guodong Long, Tianyi Zhou, Peng Yan, Zijian Zhang, Chengqi Zhang,
  and Bo~Yang.
\newblock Dual {Personalization} on {Federated} {Recommendation}, May
  2023{\natexlab{a}}.
\newblock URL \url{http://arxiv.org/abs/2301.08143}.

\bibitem[Zhang et~al.(2020)Zhang, Yan, Liu, Zhang, and Lv]{zhang_daily_2020}
Jing Zhang, Jie Yan, Yongqian Liu, Haoran Zhang, and Guoliang Lv.
\newblock Daily electric vehicle charging load profiles considering
  demographics of vehicle users.
\newblock \emph{Applied Energy}, 274:\penalty0 115063, September 2020.
\newblock ISSN 03062619.
\newblock \doi{10.1016/j.apenergy.2020.115063}.
\newblock URL
  \url{https://linkinghub.elsevier.com/retrieve/pii/S0306261920305754}.

\bibitem[Zhang et~al.(2022)Zhang, Liao, Liu, Xu, and Zheng]{zhang_leaving_2022}
Qianqian Zhang, Xinru Liao, Quan Liu, Jian Xu, and Bo~Zheng.
\newblock Leaving {No} {One} {Behind}: {A} {Multi}-{Scenario} {Multi}-{Task}
  {Meta} {Learning} {Approach} for {Advertiser} {Modeling}.
\newblock In \emph{Proceedings of the {Fifteenth} {ACM} {International}
  {Conference} on {Web} {Search} and {Data} {Mining}}, pages 1368--1376,
  Virtual Event AZ USA, February 2022. ACM.
\newblock ISBN 978-1-4503-9132-0.
\newblock \doi{10.1145/3488560.3498479}.
\newblock URL \url{https://dl.acm.org/doi/10.1145/3488560.3498479}.

\bibitem[Zhang and Challis(2021)]{zhang_learning_2021}
Wei Zhang and Chris Challis.
\newblock Learning {User} {Preferences} {Without} {Feedbacks}.
\newblock In \emph{2021 {IEEE} 8th {International} {Conference} on {Data}
  {Science} and {Advanced} {Analytics} ({DSAA})}, pages 1--2, Porto, Portugal,
  October 2021. IEEE.
\newblock ISBN 978-1-66542-099-0.
\newblock \doi{10.1109/DSAA53316.2021.9564131}.
\newblock URL \url{https://ieeexplore.ieee.org/document/9564131/}.

\bibitem[Zhang and Koren(2007)]{zhang_efficient_2007}
Yi~Zhang and Jonathan Koren.
\newblock Efficient bayesian hierarchical user modeling for recommendation
  system.
\newblock In \emph{Proceedings of the 30th annual international {ACM} {SIGIR}
  conference on {Research} and development in information retrieval}, {SIGIR}
  '07, pages 47--54, New York, NY, USA, 2007. Association for Computing
  Machinery.
\newblock ISBN 978-1-59593-597-7.
\newblock \doi{10.1145/1277741.1277752}.
\newblock URL \url{https://dl.acm.org/doi/10.1145/1277741.1277752}.

\bibitem[Zhang et~al.(2023{\natexlab{b}})Zhang, Liu, Jiang, Wang, Zhuang, Wu,
  Gao, and Chen]{zhang_fairlisa_2023}
Zheng Zhang, Qi~Liu, Hao Jiang, Fei Wang, Yan Zhuang, Le~Wu, Weibo Gao, and
  Enhong Chen.
\newblock {FairLISA}: {Fair} {User} {Modeling} with {Limited} {Sensitive}
  {Attributes} {Information}.
\newblock In \emph{Thirty-seventh Conference on Neural Information Processing
  Systems}, November 2023{\natexlab{b}}.
\newblock URL \url{https://openreview.net/forum?id=uFpjPJMkv6}.

\bibitem[Zhao et~al.(2022)Zhao, Xie, and Wang]{zhao_co-learning_2022}
Hongyu Zhao, Jiazhi Xie, and Hongbin Wang.
\newblock Co-learning {Graph} {Convolution} {Network} for {Mobile} {User}
  {Profiling}.
\newblock \emph{Neural Processing Letters}, 54\penalty0 (6):\penalty0
  5299--5316, December 2022.
\newblock ISSN 1573-773X.
\newblock \doi{10.1007/s11063-022-10862-1}.
\newblock URL \url{https://doi.org/10.1007/s11063-022-10862-1}.

\bibitem[Zhao et~al.(2021)Zhao, Xiao, Zhang, Bian, and Yan]{zhao_ameir_2021}
Pengyu Zhao, Kecheng Xiao, Yuanxing Zhang, Kaigui Bian, and Wei Yan.
\newblock {AMEIR}: {Automatic} {Behavior} {Modeling}, {Interaction}
  {Exploration} and {MLP} {Investigation} in the {Recommender} {System}.
\newblock In \emph{Proceedings of the {Thirtieth} {International} {Joint}
  {Conference} on {Artificial} {Intelligence}}, pages 2104--2110, August 2021.
\newblock \doi{10.24963/ijcai.2021/290}.
\newblock URL \url{http://arxiv.org/abs/2006.05933}.

\bibitem[Zhao et~al.(2019{\natexlab{a}})Zhao, Willemsen, Adomavicius, Harper,
  and Konstan]{zhao_preference_2019}
Qian Zhao, Martijn~C. Willemsen, Gediminas Adomavicius, F.~Maxwell Harper, and
  Joseph~A. Konstan.
\newblock From preference into decision making: modeling user interactions in
  recommender systems.
\newblock In \emph{Proceedings of the 13th {ACM} {Conference} on {Recommender}
  {Systems}}, {RecSys} '19, pages 29--33, New York, NY, USA,
  2019{\natexlab{a}}. Association for Computing Machinery.
\newblock ISBN 978-1-4503-6243-6.
\newblock \doi{10.1145/3298689.3347065}.
\newblock URL \url{https://dl.acm.org/doi/10.1145/3298689.3347065}.

\bibitem[Zhao et~al.(2019{\natexlab{b}})Zhao, Li, Ramos, Luo, Jiang, Dey, and
  Pan]{zhao_user_2019}
Sha Zhao, Shijian Li, Julian Ramos, Zhiling Luo, Ziwen Jiang, Anind~K. Dey, and
  Gang Pan.
\newblock User profiling from their use of smartphone applications: {A} survey.
\newblock \emph{Pervasive and Mobile Computing}, 59:\penalty0 101052, October
  2019{\natexlab{b}}.
\newblock ISSN 15741192.
\newblock \doi{10.1016/j.pmcj.2019.101052}.
\newblock URL
  \url{https://linkinghub.elsevier.com/retrieve/pii/S1574119219300124}.

\bibitem[Zheng et~al.(2022{\natexlab{a}})Zheng, Zhao, Zhu, and
  Qian]{zheng_perd_2022}
Xuanzhi Zheng, Guoshuai Zhao, Li~Zhu, and Xueming Qian.
\newblock {PERD}: {Personalized} {Emoji} {Recommendation} with {Dynamic} {User}
  {Preference}.
\newblock In \emph{Proceedings of the 45th {International} {ACM} {SIGIR}
  {Conference} on {Research} and {Development} in {Information} {Retrieval}},
  pages 1922--1926, Madrid Spain, July 2022{\natexlab{a}}. ACM.
\newblock ISBN 978-1-4503-8732-3.
\newblock \doi{10.1145/3477495.3531779}.
\newblock URL \url{https://dl.acm.org/doi/10.1145/3477495.3531779}.

\bibitem[Zheng et~al.(2022{\natexlab{b}})Zheng, Qiu, Xu, Wu, Zhao, Chen, and
  Xiong]{zheng_cbr_2022}
Zhi Zheng, Zhaopeng Qiu, Tong Xu, Xian Wu, Xiangyu Zhao, Enhong Chen, and Hui
  Xiong.
\newblock {CBR}: {Context} {Bias} aware {Recommendation} for {Debiasing} {User}
  {Modeling} and {Click} {Prediction}.
\newblock In \emph{Proceedings of the {ACM} {Web} {Conference} 2022}, {WWW}
  '22, pages 2268--2276, New York, NY, USA, April 2022{\natexlab{b}}.
  Association for Computing Machinery.
\newblock ISBN 978-1-4503-9096-5.
\newblock \doi{10.1145/3485447.3512099}.
\newblock URL \url{https://dl.acm.org/doi/10.1145/3485447.3512099}.

\bibitem[Zhong et~al.(2012)Zhong, Fan, Wang, Xiao, and Li]{zhong_comsoc_2012}
Erheng Zhong, Wei Fan, Junwei Wang, Lei Xiao, and Yong Li.
\newblock {ComSoc}: adaptive transfer of user behaviors over composite social
  network.
\newblock In \emph{Proceedings of the 18th {ACM} {SIGKDD} international
  conference on {Knowledge} discovery and data mining}, {KDD} '12, pages
  696--704, New York, NY, USA, 2012. Association for Computing Machinery.
\newblock ISBN 978-1-4503-1462-6.
\newblock \doi{10.1145/2339530.2339641}.
\newblock URL \url{https://dl.acm.org/doi/10.1145/2339530.2339641}.

\bibitem[Zhou et~al.(2018)Zhou, Zhu, Song, Fan, Zhu, Ma, Yan, Jin, Li, and
  Gai]{zhou_deep_2018}
Guorui Zhou, Xiaoqiang Zhu, Chenru Song, Ying Fan, Han Zhu, Xiao Ma, Yanghui
  Yan, Junqi Jin, Han Li, and Kun Gai.
\newblock Deep {Interest} {Network} for {Click}-{Through} {Rate} {Prediction}.
\newblock In \emph{Proceedings of the 24th {ACM} {SIGKDD} {International}
  {Conference} on {Knowledge} {Discovery} \& {Data} {Mining}}, {KDD} '18, pages
  1059--1068, New York, NY, USA, 2018. Association for Computing Machinery.
\newblock ISBN 978-1-4503-5552-0.
\newblock \doi{10.1145/3219819.3219823}.
\newblock URL \url{https://dl.acm.org/doi/10.1145/3219819.3219823}.

\bibitem[Zhou et~al.(2023)Zhou, Gao, Xie, Ye, Hua, Kim, Wang, and
  Kim]{zhou_equivariant_2023}
Peilin Zhou, Jingqi Gao, Yueqi Xie, Qichen Ye, Yining Hua, Jaeboum Kim, Shoujin
  Wang, and Sunghun Kim.
\newblock Equivariant {Contrastive} {Learning} for {Sequential}
  {Recommendation}.
\newblock In \emph{Proceedings of the 17th {ACM} {Conference} on {Recommender}
  {Systems}}, pages 129--140, Singapore Singapore, September 2023. ACM.
\newblock ISBN 9798400702419.
\newblock \doi{10.1145/3604915.3608786}.
\newblock URL \url{https://dl.acm.org/doi/10.1145/3604915.3608786}.

\bibitem[Zhou and Conati(2003)]{zhou_inferring_2003}
Xiaoming Zhou and Cristina Conati.
\newblock Inferring user goals from personality and behavior in a causal model
  of user affect.
\newblock In \emph{Proceedings of the 8th international conference on
  {Intelligent} user interfaces}, {IUI} '03, pages 211--218, New York, NY, USA,
  2003. Association for Computing Machinery.
\newblock ISBN 978-1-58113-586-2.
\newblock \doi{10.1145/604045.604078}.
\newblock URL \url{https://dl.acm.org/doi/10.1145/604045.604078}.

\bibitem[Zhou et~al.(2012)Zhou, Xu, Li, Josang, and Cox]{zhou_state---art_2012}
Xujuan Zhou, Yue Xu, Yuefeng Li, Audun Josang, and Clive Cox.
\newblock The state-of-the-art in personalized recommender systems for social
  networking.
\newblock \emph{Artificial Intelligence Review}, 37\penalty0 (2):\penalty0
  119--132, February 2012.
\newblock ISSN 0269-2821, 1573-7462.
\newblock \doi{10.1007/s10462-011-9222-1}.
\newblock URL \url{http://link.springer.com/10.1007/s10462-011-9222-1}.

\bibitem[Zhou et~al.(2020{\natexlab{a}})Zhou, Dou, and Wen]{zhou_encoding_2020}
Yujia Zhou, Zhicheng Dou, and Ji-Rong Wen.
\newblock Encoding {History} with {Context}-aware {Representation} {Learning}
  for {Personalized} {Search}.
\newblock In \emph{Proceedings of the 43rd {International} {ACM} {SIGIR}
  {Conference} on {Research} and {Development} in {Information} {Retrieval}},
  pages 1111--1120, Virtual Event China, July 2020{\natexlab{a}}. ACM.
\newblock ISBN 978-1-4503-8016-4.
\newblock \doi{10.1145/3397271.3401175}.
\newblock URL \url{https://dl.acm.org/doi/10.1145/3397271.3401175}.

\bibitem[Zhou et~al.(2020{\natexlab{b}})Zhou, Dou, and
  Wen]{zhou_enhancing_2020}
Yujia Zhou, Zhicheng Dou, and Ji-Rong Wen.
\newblock Enhancing {Re}-finding {Behavior} with {External} {Memories} for
  {Personalized} {Search}.
\newblock In \emph{Proceedings of the 13th {International} {Conference} on
  {Web} {Search} and {Data} {Mining}}, {WSDM} '20, pages 789--797, New York,
  NY, USA, 2020{\natexlab{b}}. Association for Computing Machinery.
\newblock ISBN 978-1-4503-6822-3.
\newblock \doi{10.1145/3336191.3371794}.
\newblock URL \url{https://dl.acm.org/doi/10.1145/3336191.3371794}.

\bibitem[Zhou et~al.(2021{\natexlab{a}})Zhou, Dou, Wei, Xie, and
  Wen]{zhou_group_2021}
Yujia Zhou, Zhicheng Dou, Bingzheng Wei, Ruobing Xie, and Ji-Rong Wen.
\newblock Group based {Personalized} {Search} by {Integrating} {Search}
  {Behaviour} and {Friend} {Network}.
\newblock In \emph{Proceedings of the 44th {International} {ACM} {SIGIR}
  {Conference} on {Research} and {Development} in {Information} {Retrieval}},
  pages 92--101, Virtual Event Canada, July 2021{\natexlab{a}}. ACM.
\newblock ISBN 978-1-4503-8037-9.
\newblock \doi{10.1145/3404835.3462918}.
\newblock URL \url{https://dl.acm.org/doi/10.1145/3404835.3462918}.

\bibitem[Zhou et~al.(2021{\natexlab{b}})Zhou, Dou, Zhu, and
  Wen]{zhou_pssl_2021}
Yujia Zhou, Zhicheng Dou, Yutao Zhu, and Ji-Rong Wen.
\newblock {PSSL}: {Self}-supervised {Learning} for {Personalized} {Search} with
  {Contrastive} {Sampling}.
\newblock In \emph{Proceedings of the 30th {ACM} {International} {Conference}
  on {Information} \& {Knowledge} {Management}}, pages 2749--2758, Virtual
  Event Queensland Australia, October 2021{\natexlab{b}}. ACM.
\newblock ISBN 978-1-4503-8446-9.
\newblock \doi{10.1145/3459637.3482379}.
\newblock URL \url{https://dl.acm.org/doi/10.1145/3459637.3482379}.

\bibitem[Zhu et~al.(2017)Zhu, Li, Liao, Wang, Guan, Liu, and
  Cai]{zhu_what_2017}
Yu~Zhu, Hao Li, Yikang Liao, Beidou Wang, Ziyu Guan, Haifeng Liu, and Deng Cai.
\newblock What to do next: modeling user behaviors by time-{LSTM}.
\newblock In \emph{Proceedings of the 26th {International} {Joint} {Conference}
  on {Artificial} {Intelligence}}, {IJCAI}'17, pages 3602--3608, Melbourne,
  Australia, 2017. AAAI Press.
\newblock ISBN 978-0-9992411-0-3.

\bibitem[Zhu et~al.(2021)Zhu, Nie, Dou, Ma, Zhang, Du, Zuo, and
  Jiang]{zhu_contrastive_2021}
Yutao Zhu, Jian-Yun Nie, Zhicheng Dou, Zhengyi Ma, Xinyu Zhang, Pan Du,
  Xiaochen Zuo, and Hao Jiang.
\newblock Contrastive {Learning} of {User} {Behavior} {Sequence} for
  {Context}-{Aware} {Document} {Ranking}.
\newblock In \emph{Proceedings of the 30th {ACM} {International} {Conference}
  on {Information} \& {Knowledge} {Management}}, pages 2780--2791, Virtual
  Event Queensland Australia, October 2021. ACM.
\newblock ISBN 978-1-4503-8446-9.
\newblock \doi{10.1145/3459637.3482243}.
\newblock URL \url{https://dl.acm.org/doi/10.1145/3459637.3482243}.

\bibitem[Zigoris and Zhang(2006)]{zigoris_bayesian_2006}
Philip Zigoris and Yi~Zhang.
\newblock Bayesian adaptive user profiling with explicit \& implicit feedback.
\newblock In \emph{Proceedings of the 15th {ACM} international conference on
  {Information} and knowledge management}, {CIKM} '06, pages 397--404, New
  York, NY, USA, November 2006. Association for Computing Machinery.
\newblock ISBN 978-1-59593-433-8.
\newblock \doi{10.1145/1183614.1183672}.
\newblock URL \url{https://dl.acm.org/doi/10.1145/1183614.1183672}.

\bibitem[Zimmerman and Kurapati(2002)]{zimmerman_exposing_2002}
John Zimmerman and Kaushal Kurapati.
\newblock Exposing profiles to build trust in a recommender.
\newblock In \emph{{CHI} '02 {Extended} {Abstracts} on {Human} {Factors} in
  {Computing} {Systems}}, {CHI} {EA} '02, pages 608--609, New York, NY, USA,
  April 2002. Association for Computing Machinery.
\newblock ISBN 978-1-58113-454-4.
\newblock \doi{10.1145/506443.506507}.
\newblock URL \url{https://dl.acm.org/doi/10.1145/506443.506507}.

\bibitem[Zimmerman et~al.(2004)Zimmerman, Kauapati, Buczak, Schaffer, Gutta,
  and Martino]{zimmerman_tv_2004}
John Zimmerman, Kaushal Kauapati, Anna~L. Buczak, Dave Schaffer, Srinivas
  Gutta, and Jacquelyn Martino.
\newblock {TV} {Personalization} {System}.
\newblock In \emph{Personalized {Digital} {Television}: {Targeting} {Programs}
  to individual {Viewers}}, Human-{Computer} {Interaction} {Series}, pages
  27--51. Springer Netherlands, Dordrecht, 2004.
\newblock ISBN 978-1-4020-2164-0.
\newblock \doi{10.1007/1-4020-2164-X_2}.
\newblock URL \url{https://doi.org/10.1007/1-4020-2164-X_2}.

\bibitem[Zukerman and Albrecht(2001)]{zukerman_predictive_2001}
Ingrid Zukerman and David~W. Albrecht.
\newblock Predictive {Statistical} {Models} for {User} {Modeling}.
\newblock \emph{User Modeling and User-Adapted Interaction}, 11\penalty0
  (1):\penalty0 5--18, March 2001.
\newblock ISSN 1573-1391.
\newblock \doi{10.1023/A:1011175525451}.
\newblock URL \url{https://doi.org/10.1023/A:1011175525451}.

\bibitem[Çelikok et~al.(2023)Çelikok, Murena, and
  Kaski]{celikok_modeling_2023}
Mustafa~Mert Çelikok, Pierre-Alexandre Murena, and Samuel Kaski.
\newblock Modeling needs user modeling.
\newblock \emph{Frontiers in Artificial Intelligence}, 6, 2023.
\newblock ISSN 2624-8212.
\newblock \doi{10.3389/frai.2023.1097891}.
\newblock URL
  \url{https://www.frontiersin.org/articles/10.3389/frai.2023.1097891}.

\bibitem[Żołna and Romański(2017)]{zolna_user_2017}
Konrad Żołna and Bartłomiej Romański.
\newblock User {Modeling} {Using} {LSTM} {Networks}.
\newblock \emph{Proceedings of the AAAI Conference on Artificial Intelligence},
  31\penalty0 (1), February 2017.
\newblock ISSN 2374-3468.
\newblock \doi{10.1609/aaai.v31i1.11068}.
\newblock URL \url{https://ojs.aaai.org/index.php/AAAI/article/view/11068}.

\end{thebibliography}

\end{document}